%% file: paper.tex
\documentclass[]{xiaomiev}
\usepackage[toc,page,header]{appendix}
\usepackage{minitoc,bbding,solarized-light,amsfonts,booktabs,multirow,graphicx,array,siunitx}
\usepackage[table]{xcolor}
\usepackage{amssymb}
\usepackage{makecell,framed,amsthm,amsmath,placeins,longtable,hhline,fancyvrb,fvextra,CJKutf8,multicol,cleveref}
\usepackage{tablefootnote,threeparttable,tabularx,mdframed,subcaption,stackengine}
\usepackage[numbers,sort&compress]{natbib}
\usepackage{enumitem,float,wrapfig,caption,pifont,xspace}

\newcommand{\commentout}[1]{}
\renewcommand{\paragraph}[1]{\par\noindent\textcolor{black}{\textbf{#1}}\hspace*{1em}}
\newcommand{\cmark}{\ding{51}}
\newcommand{\xmark}{\ding{55}}

\newcommand{\XFM}{DriveFoundry}
\newcommand{\Model}{Xiaomi-Drive-o1}

\title{\XFM{}: Co-Evolving Data and Model for Efficient Closed-Loop Iteration in Autonomous Driving}
\author{Xiaomi EV}

\contribution{See the \protect\hyperlink{contributions-heading}{\textbf{Contributions and Acknowledgments}} section for a list of contributors.}

\abstract{
A driving foundation model must learn reusable decision knowledge
from rare critical events and apply it under strict runtime
constraints. We introduce \textbf{\XFM{}}, a closed-loop framework designed
to improve iteration efficiency under data and computation
budgets by coupling structured causal data refinement with
adaptive inference.
The Data Engine builds traceable scene--candidate records that
link scene evidence and factor--decision relations to candidate
trajectories and outcome assessments. Candidate comparisons and counterfactual annotations clarify how scene factors affect behavioral feasibility and preference. Verified model failures guide VLM-assisted annotation and expert correction, refining both the training data and the annotation system for subsequent learning.
The model engine, \textbf{\Model{}}, unifies Perception,
Causation, Action, World, and Value within a shared
vision--language backbone, enabling cross-task knowledge transfer through joint training.
Block-wise latent refill passes candidate-specific evidence and
predicted visual consequences between task blocks, connecting scene understanding, planning, and outcome evaluation. A learned
value-of-computation head selects additional tasks and allocates
latent tokens according to their expected benefit and incremental cost, adapting inference to the demands of each decision.

On PhysicalAI, the 9B model achieves the lowest ADE@1 across both evaluated splits and horizons among the tested models. In AlpaSim, it outperforms Alpamayo 1.5 on all reported metrics and achieves lower close-encounter and off-road rates than the 34B Alpamayo 2 Super, but trails it in progress and AlpaSim scores. For the 0.8B model, joint training reduces 5~s minADE@6 by approximately 54\% without changing the inference path or token budget. Adaptive inference mode further balances the performance of full reasoning while using 28\% of its tokens. 
Real-vehicle tests additionally demonstrate runtime feasibility.
These results support a unified approach to learning from
critical experience and adapting computation to the demands
of each driving decision.

\par\vspace{5pt}

\par\vspace{10pt}
\noindent\textbf{Date:} \today
\par
\noindent\textbf{Project page:}
\href{https://Xiaomi-EV.github.io/Xiaomi-Drive-o1}
     {https://Xiaomi-EV.github.io/Xiaomi-Drive-o1}
}

\begin{document}
\maketitle

\clearpage
\onecolumn

\tableofcontents
\newpage

\input{sections/introduction}
\input{sections/problem_formulation}
\input{sections/methodV2}
\input{sections/evaluation}
\input{sections/conclusion}
\clearpage
\bibliographystyle{plainnat}
\bibliography{main}
\clearpage
\input{sections/contribution}
\clearpage
\appendix
\input{sections/appendix}
\end{document}

%% file: sections/introduction.tex
\section{Introduction}
Progress in physical intelligence hinges on converting physical experience into generalizable behavior under acceptable budgets~\cite{firoozi2025foundation,gupta2021embodied,chen2024end}.
This is particularly acute in production-level autonomous driving (SAE L2/L3)~\cite{on2021taxonomy}, where vehicles act in a partially observed world and their decisions shape subsequent interactions~\cite{wang2022social}.
Each control cycle therefore requires identifying decision-relevant factors from observation histories, generating executable motion plans, predicting their consequences and selecting among alternatives under constraints.
Fleet expert clips offer broad coverage of normal driving, yet failures that expose the system limits are condensed in critical events~\cite{feng2023dense,liu2026r2se,liu2024curse}. Improving in this regime requires a costly cycle of discovery and learning. The central learning objective for autonomous driving should
therefore be closed-loop iteration efficiency for critical
scenarios~\cite{karkus2025beyond}, learning transferable decision
relationships from rare experiences at an acceptable cost and
reusing them in subsequent driving.


\begin{figure}[ht]
\centering
\includegraphics[width=\linewidth]{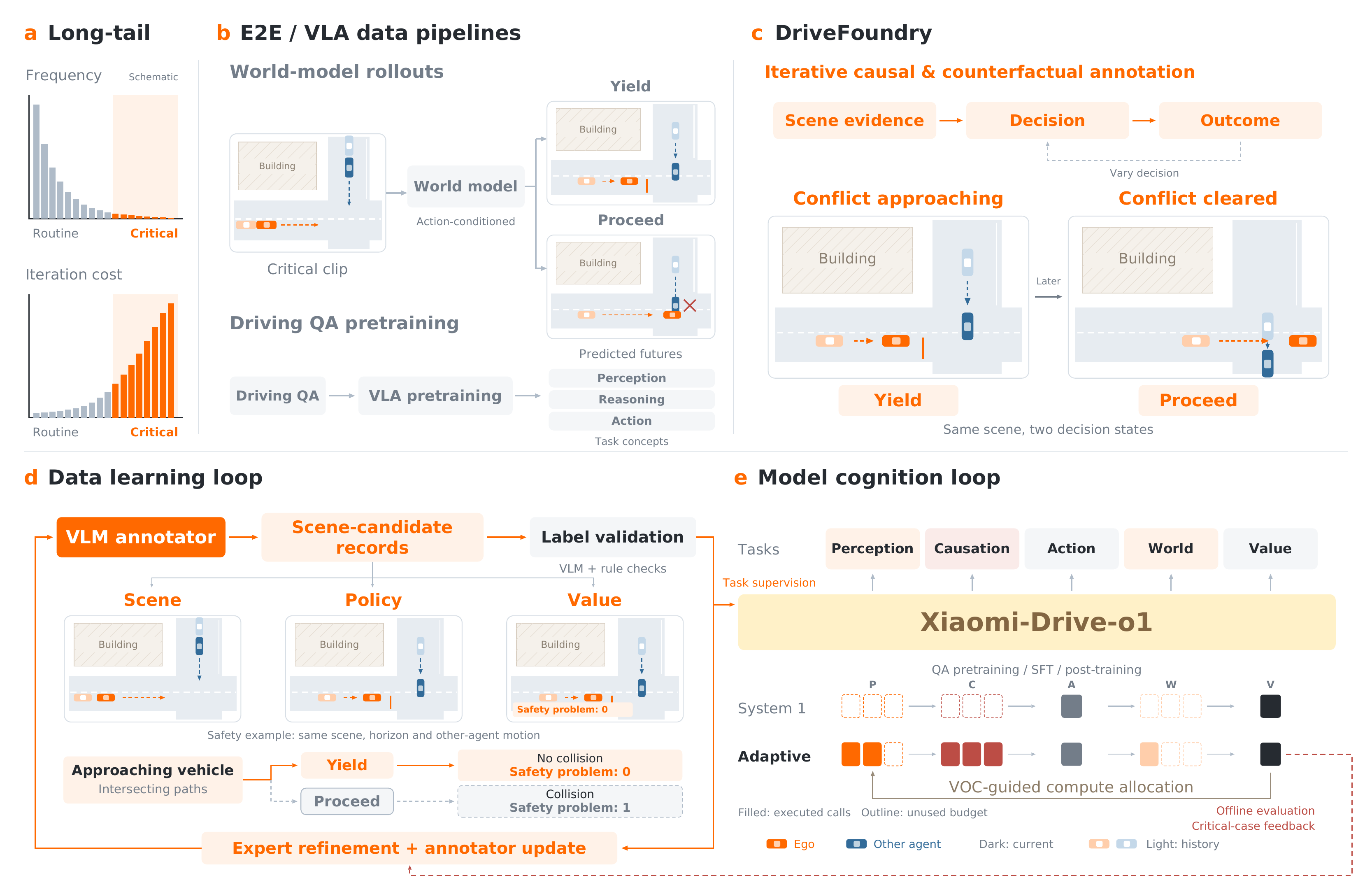}
\caption{\textbf{\XFM{}}, a closed-loop engine coupling a structured causal data-learning loop with an adaptive model-cognition loop that enhances the closed-loop iteration efficiency for critical cases.}
\label{fig:teaser}
\end{figure}

General-purpose vision-language models (VLMs) provide visual and semantic representations~\cite{team2026gemma,team2026qwen3,coreteam2025mimovltechnicalreport,hwang2024emma}, but driving expertise further requires precise geometric and motion modeling, decision-relevant reasoning, candidate comparison and real-time control.
This motivates unified learning across perception, decision and value, beyond direct observation--action mapping.
Driving VLMs and vision-language-action models (VLAs) extend supervision to scene understanding, explanations and action prediction~\cite{marcu2024lingoqa,sima2024drivelm,wang2024omnidrive,xu2024drivegpt4,li2025drivevla,huang2026mindvla,team2026one}, while vision-action models (VAs) learn direct mappings through multitask pretraining~\cite{liu2026driveva,wang2026drive,jing2026simvta}.
Explicit language reasoning captures scene relationships and decision rationales, but introduces information bottlenecks and sequential decoding costs.
World-action models (WAMs) predict action-conditioned futures~\cite{kim2026cosmos,wang2026latentwam,shi2026drivewam,yuan2026fast,yao2026discrete}, yet prediction alone does not explain which factors justify an action or favor it over counterfactual alternatives.
Latent reasoning offers compact task representations~\cite{lu2026xiaomi,li2025enhancing,lastvla}, but fixed-capacity, one-step interfaces constrain adaptation to scene demands.
These developments motivate shared latent reasoning that links scene evidence, action choices and outcome evaluation, with adaptive computation for efficient closed-loop iteration under limited data and compute budgets.

For a data engine, closed-loop iteration efficiency depends on coverage of critical experience and supervision density, supporting compositional generalization and reusable causal abstraction~\cite{battaglia2018relational}.
Data flywheels improve coverage and label quality through failure mining, targeted annotation and continual learning~\cite{yang2026human,feng2026breaking,cao2023continuous}, while generative world models expand experience through controllable synthesis and action-conditioned prediction~\cite{agarwal2025cosmos,hu2023gaia,gao2024vista,tian2025simscale}. World Engine integrates these capabilities with behavior modeling and reinforcement post-training, expanding rare critical scenarios for policy improvement~\cite{li2026world}. Still, rollouts alone do not identify which evidence, causal relationships or behaviors require correction.
Structured language annotations, including graph-structured QA, Chain of Causation (CoC) and structured chains of thought~\cite{wang2025alpamayo,nvidia2026alpamayo15,simcot,sima2024drivelm,team2026xcot}, complement experience expansion by linking grounded evidence, interactions, decision rationales and outcomes. Effective abstraction further requires supervision of how causal factors determine the suitability and consequences of alternative behaviors.
A data engine should therefore connect evidence, candidate decisions and consequences through structured causal annotations, using failure diagnosis and feedback to refine supervision so that corrections transfer across cases.

We introduce \textbf{\XFM{}} (\Cref{fig:teaser}), a framework for learning reusable driving behavior from critical experience under constrained data and computation budgets. Its common interface is a scene--candidate record linking scene facts, a candidate's semantic intent and continuous trajectory, and the evidence used to assess its outcome. Multiple candidates expose alternatives to the recorded behavior, while partial records allow annotation effort to focus on the distinctions that matter to the decision.

The model engine, \textbf{\Model{}}, connects Perception, Causation, Action, World, and Value tasks through a shared vision--language backbone. Shared parameters support learning across tasks and scenes;
block-wise latent refill passes intermediate results within the
current decision. Inference starts with a fast Action--Value path.
A value-of-computation (VOC) head guides further computation to
expand candidate coverage or improve candidate evaluation when
the expected benefit justifies the additional cost.

The two feedback processes operate at different timescales. Online inference allocates computation to the current decision. Offline iteration traces failures to gaps in evidence, candidate coverage, or outcome assessment, then revises the corresponding supervision. Shared scene and candidate identifiers connect these revisions to both the driving model and the annotation system.

Our contributions are threefold:
\begin{itemize}

    \item \textbf{A closed-loop framework and structured causal
    data engine.}
    We introduce \XFM{}, which connects data refinement,
    model learning, and adaptive inference under data and
    computation budgets. Its scene--candidate records combine
    temporal factor--decision relations, candidate comparisons,
    and counterfactual assessments into reusable supervision.
    Verified failures guide targeted annotation and subsequent
    updates to both the driving model and the annotation system.

    \item \textbf{A unified driving foundation model with latent
    refill.}
    We develop \Model{}, integrating Perception, Causation,
    Action, World, and Value through a shared vision--language
    backbone. Shared parameters allow cross-task learning to
    strengthen fast Action--Value inference, while block-wise
    latent refill connects intermediate results and preserves
    candidate correspondence across planning, future prediction,
    and evaluation. Staged training connects semantic and visual
    pretraining to continuous planning, with task supervision
    retained during reinforcement post-training.

    \item \textbf{Learned allocation of inference computation.}
    A dedicated value-of-computation head estimates the benefit
    of further reasoning after accounting for its incremental
    execution cost. It learns from SFT rollouts using
    candidate-evaluation supervision shared with Value.
    At inference, it uses the already computed state to guide
    additional task execution and latent-token allocations under compute, latency, and memory constraints.

\end{itemize}

%% file: sections/problem_formulation.tex
\section{Problem Formulation}

Let $t$ index driving decisions, $k$ index learning iterations,
and $\Theta$ denote model parameters.
Given training data $\mathcal D$ and a configuration
$m=(\mathcal M,\beta,\rho)$ specifying the architecture,
training strategy, and inference configuration.
The closed-loop driving return is:
\begin{equation}
J_{\mathrm{drive}}(\pi)
=
\mathbb E_{\tau\sim(\mu_0,P_{\mathrm{env}},Z,\pi)}
\left[
\sum_{t=0}^{T-1}\gamma^t r(s_t,a_t;g)
\right],
\end{equation}
Here, $\pi$ is the driving policy, and $\tau$ is a closed-loop
rollout generated from the initial-state distribution $\mu_0$,
environment transition model $P_{\mathrm{env}}$, and observation
model $Z$ under $\pi$.
At decision step $t$, $s_t$ is the underlying environment state,
$a_t$ is the selected action, and $r(s_t,a_t;g)$ is its reward
under navigation goal $g$.
$T$ is the evaluation horizon in decision steps, and $\gamma$
is the discount factor. The overall optimization objective is:
\begin{equation}
\max_{\mathcal D,m}\;J_{\mathrm{drive}}(\pi_{\mathcal D,m})
\quad\mathrm{s.t.}\quad
\mathbf C_{\mathrm{res}}(\mathcal D,m)\preceq\mathbf B_{\mathrm{res}},\;
\pi_{\mathcal D,m}\in\Pi_{\mathrm{feas}}.
\end{equation}
where $\mathbf C_{\mathrm{res}}$ and $\mathbf B_{\mathrm{res}}$ denote resource consumption and budgets, respectively, and $\preceq$ applies component-wise. The model loop improves driving capability for given data, while the data loop improves training utility for the current learning mechanism.
Here, $\Pi_{\mathrm{feas}}$ enforces applicable behavioral constraints ($\chi_\ell=1\Rightarrow g_\ell=1$), where $\chi_\ell$ and $g_\ell$ indicate applicability and satisfaction; always-active constraints use $\chi_\ell\equiv1$, while trade-offs among feasible behaviors enter the reward.
Learning starts from pretrained and newly initialized parameters $\Theta_0=(\theta_{\mathrm{VLM}},\phi_0)$, with $\beta(k)$ controlling supervision, task weights, and parameter updates using annotated experience.
Resource budgets bound data and training costs $(C_D,C_T)$ and inference compute, latency, and memory $(C_I,T_I,M_I)$.
Inference compute is evaluated under the deployment distribution $d_\pi$, and the latency budget reserves $\tau_{\mathrm{reserve}}$ for subsequent control execution.

\paragraph{Constrained driving and candidate evaluation.}
Driving is partially observed. The policy therefore uses past
observations, past actions, and the navigation goal to make
each decision. We represent this information and the resulting
uncertainty about the current state as
\begin{equation}
h_t=(o_{0:t},a_{0:t-1},g),\qquad
b_t(s)=p(s_t=s\mid h_t).
\end{equation}
Here, $o_t$, $a_t$, and $s_t$ are the observation, action,
and underlying state at step $t$, and $g$ is the navigation goal.
The belief $b_t(s)$ gives the probability that the current
state is $s$, based on the available history.

Hard driving constraints determine which actions are allowed.
Let $\mathcal A_{\mathrm{feas}}(b_t,g)$ denote these actions.
A feasible policy must select from this set at every reachable
history:
\begin{equation}
\Pi_{\mathrm{feas}}
=
\left\{
\pi:
\pi\!\left(\mathcal A_{\mathrm{feas}}(b_t,g)\mid h_t\right)=1
\text{ for every reachable }h_t
\right\}.
\end{equation}
Here, $\pi(\mathcal A_{\mathrm{feas}}(b_t,g)\mid h_t)$
is the probability of selecting an admissible action given
history $h_t$. Requiring this probability to equal one
means that the policy selects only admissible actions.

To compare allowed candidates, we evaluate the expected driving
return from each choice. For candidate $a$, we assume $a_t=a$
and that subsequent actions follow policy $\pi$:
\begin{equation}
Q_t^\pi(h_t,a)
=
\mathbb E_{\pi}
\left[
\sum_{u=t}^{T-1}\gamma^{u-t}r_u
\,\middle|\,
h_t,a_t=a
\right].
\end{equation}
Here, $r_u=r(s_u,a_u;g)$.
The expectation accounts for uncertainty in the current state,
future actions, and environment evolution.
The overall driving return is the expected value of the
initial decision:
\begin{equation}
J_{\mathrm{drive}}(\pi)
=
\mathbb E_{h_0,\,a_0\sim\pi(\cdot\mid h_0)}
\left[Q_0^\pi(h_0,a_0)\right].
\end{equation}
The ideal reference policy maximizes this return while
satisfying the driving constraints:
\begin{equation}
\pi^\star\in
\arg\max_{\pi\in\Pi_{\mathrm{feas}}}
J_{\mathrm{drive}}(\pi).
\end{equation}

In practice, we approximate candidate returns using rule-based
rewards and learned Value assessments. The resulting score
$\widehat Q_t^\pi$ ranks candidates using the information
available on the executed inference path.

\paragraph{Causal data refinement and continual learning.}
Each learning iteration uses evaluation results to correct
the training data. Driving failures and errors in decision
reasoning guide case selection. For the selected cases,
we add missing supervision or revise incorrect annotations.
The model is then trained on these corrections together with
a sample of previous data, and evaluated again to identify
the next set of problems.

For the cases $\mathcal E_k$ identified at iteration $k$,
the data and model updates are:
\begin{equation}
\begin{aligned}
\Delta\mathcal D_k
&=
\operatorname{Annotate}^{\mathrm{causal}}
\left(\operatorname{Select}(\mathcal E_k)\right),\\
\mathcal D_{k+1}
&=
\operatorname{Merge}(\mathcal D_k,\Delta\mathcal D_k),\\
\Theta_{k+1}
&=
\operatorname{Update}_{\beta(k)}
\left(
\Theta_k;\,
\Delta\mathcal D_k
\cup
\operatorname{Replay}(\mathcal D_k)
\right).
\end{aligned}
\end{equation}
Here, $\mathcal D_k$ is the current training dataset,
$\Theta_k$ the model parameters, and $\Delta\mathcal D_k$
the new or revised supervision. The model update follows
the training strategy $\beta(k)$.

$\operatorname{Merge}$ adds new records and replaces revised
annotations. $\operatorname{Replay}$ samples previous records,
excluding annotations superseded by $\Delta\mathcal D_k$.
Including this previous data helps preserve existing capabilities
while the model learns from the corrections.

\paragraph{From closed-loop return to local decision gaps.}
A decision can lose driving return because the candidate set
misses better behaviors or because the policy chooses a worse
candidate from that set. We connect these two sources of loss
to the overall closed-loop return.
Assume that a feasible policy $\pi^\star$ is optimal from every
reachable history.
Let $Q_t^\star=Q_t^{\pi^\star}$ and
$J_{\mathrm{drive}}^\star=J_{\mathrm{drive}}(\pi^\star)$.
At each history, $V_t^\star$ is the value of the best feasible
action, and $\delta_t$ is the loss from choosing $a_t$:
\[
\begin{aligned}
V_t^\star(h_t)
&=
\max_{a\in\mathcal A_{\mathrm{feas}}(b_t,g)}
Q_t^\star(h_t,a),\\[3pt]
\delta_t
&=
V_t^\star(h_t)-Q_t^\star(h_t,a_t),
\end{aligned}
\]
The evaluation ends at step $T$. No further rewards are
counted beyond this point, so $V_T^\star=0$.

For any feasible policy $\pi$, the expected discounted sum
of these decision gaps equals its overall return gap:
\begin{equation}
J_{\mathrm{drive}}^\star-J_{\mathrm{drive}}(\pi)
=
\mathbb E_\pi
\left[
\sum_{t=0}^{T-1}\gamma^t\delta_t
\right].
\label{eq:return_gap}
\end{equation}
The overall return gap equals the expected discounted sum
of these local gaps along rollouts induced by $\pi$.
For a finite, nonempty candidate set
$\mathcal B_t\subseteq\mathcal A_{\mathrm{feas}}(b_t,g)$
and a selected action $a_t\in\mathcal B_t$, this gap decomposes as
\begin{equation}
\delta_t
=
\underbrace{
V_t^\star(h_t)
-\max_{a\in\mathcal B_t}Q_t^\star(h_t,a)
}_{\varepsilon_{\mathrm{cover},t}}
+
\underbrace{
\max_{a\in\mathcal B_t}Q_t^\star(h_t,a)
-Q_t^\star(h_t,a_t)
}_{\varepsilon_{\mathrm{select},t}}.
\label{eq:candidate_loss}
\end{equation}

The coverage term measures the value gap between the best feasible
action and the best generated candidate. The selection term
measures the additional loss from choosing a worse candidate
within that set.

These two gaps guide both data refinement and inference.
Across learning iterations, observed failures guide corrections
to the supervision for candidate generation and evaluation.
Within a decision, additional task execution and latent
computation can improve the candidates or their evaluation.
The VOC head estimates whether the expected improvement
justifies the additional cost.
The following method connects these processes through structured
scene--candidate records and a shared driving model.

%% file: sections/methodV2.tex
\section{Method}
\label{sec:method}

\subsection{Overview}
\label{sec:overview}

\textbf{\XFM{}} links data annotation, model learning and inference to improve
closed-loop iteration efficiency. Its common unit is a scene--candidate
record connecting scene evidence, candidate behaviors and outcome
assessments. These records provide complementary supervision for a
shared driving model, while failures guide subsequent annotation
and training updates. The two loops operate at different timescales:
offline iteration improves the learned capabilities and online
inference allocates computation to the current decision.

\paragraph{Data annotation loop.}
The Data Engine organizes experience into scene facts and relations,
semantic decisions and trajectories, and behavioral assessments. Multiple candidates expose alternatives
to the recorded behavior, while partial records focus annotation on
the information needed to resolve a failure. VLM-assisted annotation,
human correction and verification produce training data and improve
the annotation system itself.

\begin{figure}[t]
\centering
\includegraphics[width=\linewidth]{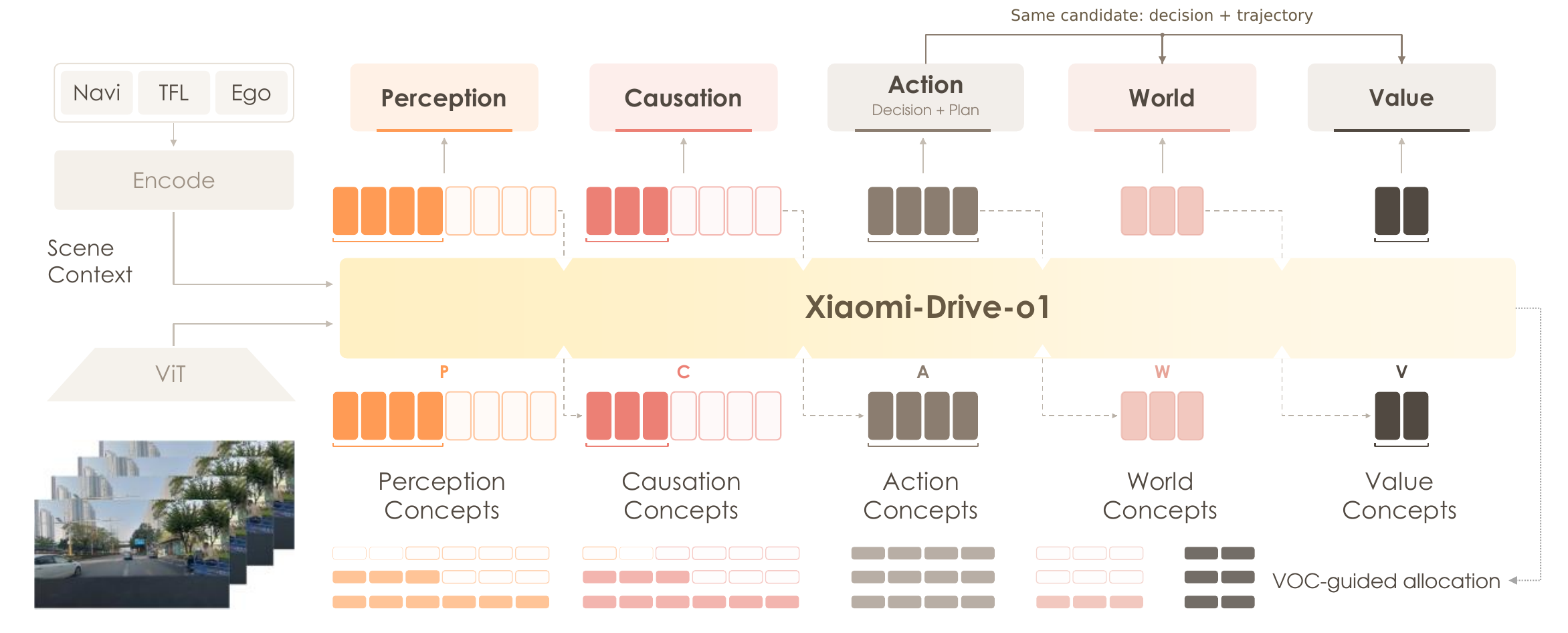}
\caption{Shared backbone and task interfaces in \Model{}.
The five tasks use a common vision--language backbone with
task-specific output heads. Block-wise latent refill passes
intermediate representations to downstream tasks, while the
Value interface supports computation scheduling.}
\label{fig:training_frame}
\end{figure}

\paragraph{Model cognition loop.}
\Model{} connects Perception, Causation, Action, World, and Value
through a shared vision--language backbone, detailed in
\Cref{fig:training_frame}. Shared parameters support learning across
tasks and scenes. Latent refill passes intermediate information
within a decision. The scheduler selects additional tasks and their
latent-token budgets when further computation is useful. The following sections
describe these model interfaces, their supervision, and the training
and feedback processes that connect the two loops.

\subsection{\Model{}}

\subsubsection{Practical Formulation}

At each decision time, the model encodes the available history, proposes
behaviors $a=(d,\xi)$ with a semantic decision $d$ and a trajectory $\xi$,
and selects among the admissible candidates:
\begin{equation}
\begin{aligned}
x_t &= f_{\mathrm{enc}}(h_t),\\
\mathcal A_t &= \mathcal G(x_t, c_{t,\rho}^{\mathrm{pre}};\rho),\\
\mathcal A_{t,\mathrm{adm}}
&=\{a\in\mathcal A_t:\operatorname{Check}(h_t,a)=1\},\\
\hat a_t
&\in\arg\max_{a\in\mathcal A_{t,\mathrm{adm}}}
\widehat Q_t^\pi\left(x_t,a,c_{t,\rho}(a)\right).
\end{aligned}
\end{equation}
Here, $f_{\mathrm{enc}}$ encodes history into $x_t$,
$\mathcal G$ is the Action candidate generator, and
$\operatorname{Check}$ is a separate rule-based feasibility module.
$\rho$ denotes the inference configuration controlling
candidate generation, task execution, and per-task latent-token budgets.
Before candidate generation, $c_{t,\rho}^{\mathrm{pre}}$ contains
the available Perception and Causation latent information and is
empty on the fast path. Action uses this context to generate candidates.
For each resulting candidate $a$, the evaluation context $c_{t,\rho}(a)$
retains the upstream information and incorporates the Action-side latents
associated with that candidate. When the World block is executed,
its candidate-conditioned prediction is added to this context.
Value then uses $c_{t,\rho}(a)$ to evaluate $a$. Rule-based rewards and learned Value assessments jointly provide
the engineering approximation $\widehat Q_t^\pi$ used to
rank candidates.

The five learned tasks support this process at different points. Perception
extracts scene evidence; Causation identifies decision-relevant factors and
relations; Action proposes candidate behaviors; World predicts their visual
consequences; and Value estimates the action values used to score candidates.
Together, these components support reducing the coverage and selection terms
in Eq.~\eqref{eq:candidate_loss} within the available computation budget.

\subsubsection{Shared Backbone and Task Interfaces}

\begin{figure}[htp]
    \centering
    \includegraphics[
        width=\linewidth,
        clip
    ]{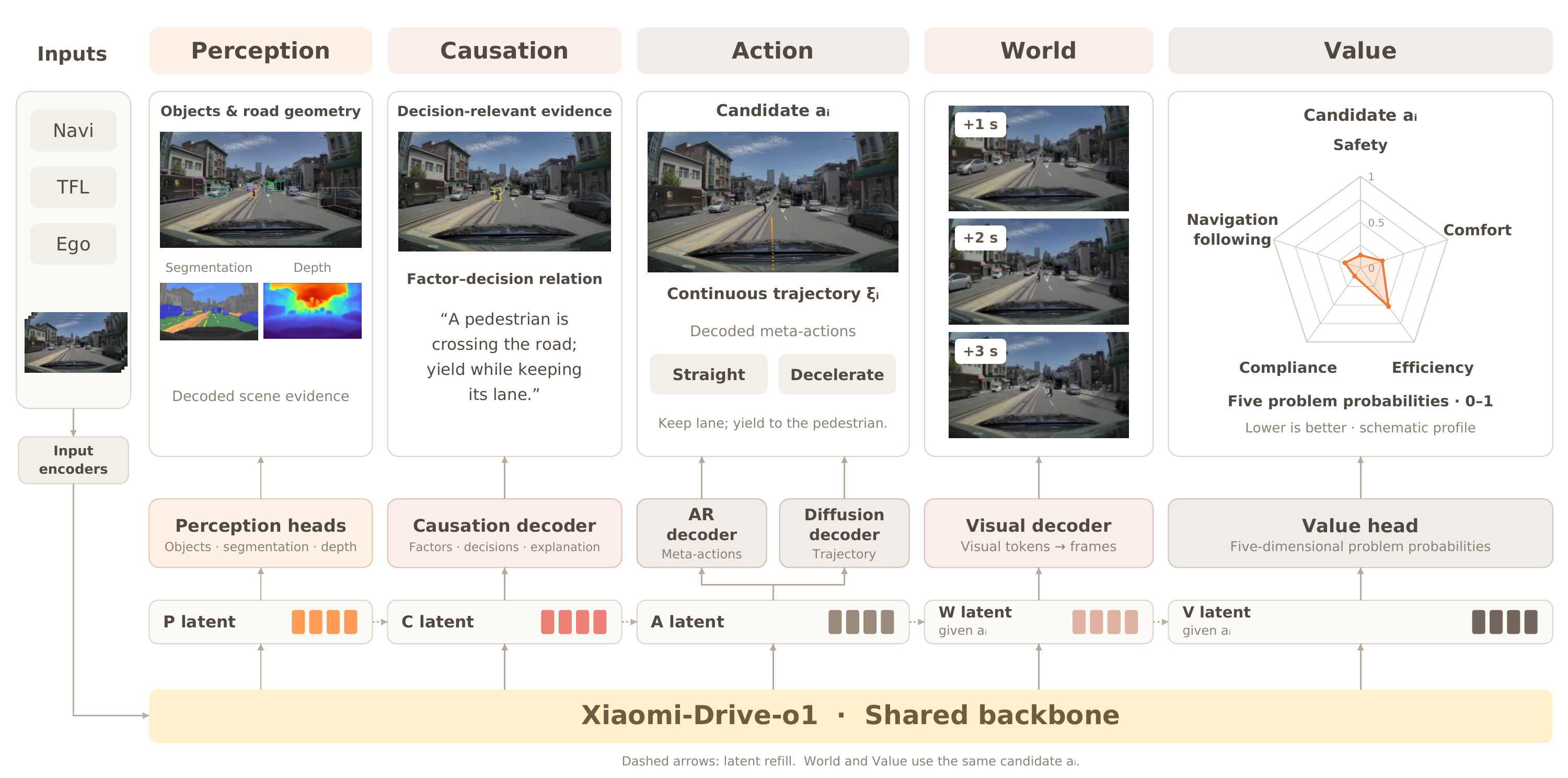}
    \vspace{-1.5em}
    \caption{Qualitative visualization of multi-task inference results from \Model{}, showing the ego vehicle decelerating to yield to a crossing pedestrian.}
    \label{fig:model_output}
\end{figure}

Xiaomi-Drive-o1 encodes driving history with a shared vision--language backbone and organizes five tasks through task queries, adapters, and output heads, detailed in \Cref{fig:model_output}:
\begin{equation}
\Theta=(\theta,\{\phi_j\}_{j\in\mathcal J}),\qquad
\mathcal J=\{P,C,A,W,V\},\qquad
H_0=E_\theta(h).
\end{equation}
Here, $\theta$ denotes shared parameters, while $\phi_j$ contains the queries, adapters, and heads for task $j$.
$E_\theta$ encodes the current history $h$ into the initial
scene context $H_0$, with the decision-time index omitted.
Action plans motion over 6.4\,s. World uses the first 3\,s of that plan, sampled at 2\,Hz, to predict visual evidence at future horizons of 1, 2, and 3\,s.
The two tasks thus model motion and its near-term consequences at different horizons.
\begin{table}[ht]
\centering
\small
\vspace{-1em}
\caption{Task interfaces and outputs of \Model{}.}
\vspace{-1em}
\begin{tabularx}{\linewidth}{@{}lX@{}}
\toprule
\rowcolor[HTML]{FFE0CC}
\textbf{Interface} & \textbf{Outputs} \\
\midrule
\textbf{Perception (P)} &
Objects, road geometry, segmentation, and depth. \\
\textbf{Causation (C)} &
Key factors, factor--decision relations, roles, and temporal validity. \\
\textbf{Action (A)} &
Semantic decisions, meta-actions, and candidate trajectories. \\
\textbf{World (W)} &
Candidate-conditioned future visual tokens. \\
\textbf{Value (V)} &
Five-dimensional driving-quality ratings and inference-budget recommendations. \\
\bottomrule
\end{tabularx}
\vspace{-1em}
\end{table}

\begin{figure}[t]
\centering
\includegraphics[width=\linewidth]{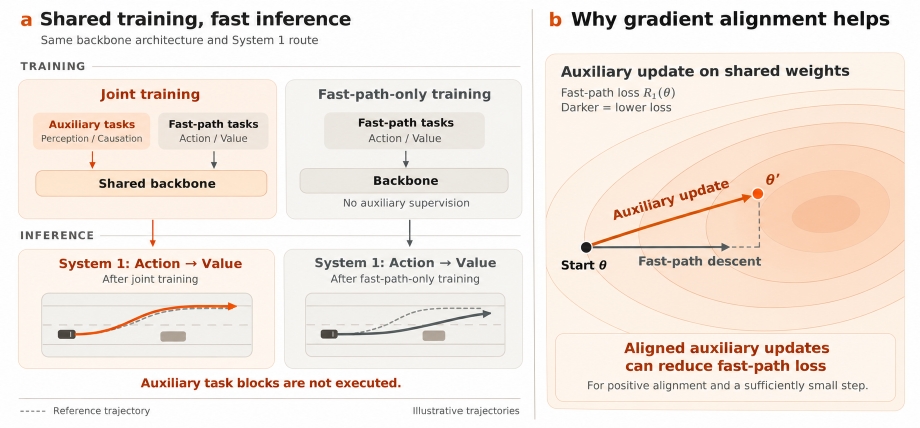}
\caption{Knowledge transfer through shared weights and the role of gradient alignment in joint training.}
\label{fig:gradient-alignment-insight}
\end{figure}

\paragraph{Shared training for fast-path inference.}
The five tasks use distinct interfaces but update a common backbone.
Driving Visual Question Answering (VQA) pretraining connects visual observations with the concepts
used to describe scene evidence and decisions. Joint training then
allows Perception, Causation, and the meta-action and trajectory objectives within Action to shape the representation used for planning.

A first-order expansion explains how an auxiliary update can benefit
the fast path. Let $R_1(\theta)$ be a smooth surrogate loss for the
fixed System~1 execution path, and let $\mathcal L_2(\theta)$ be an
auxiliary-task loss. With step size $\eta>0$, updating the shared parameters gives
\begin{equation}
\begin{aligned}
\theta' &= \theta-\eta\nabla_\theta\mathcal L_2(\theta),\\
R_1(\theta')-R_1(\theta)
&=-\eta\left\langle
\nabla_\theta R_1(\theta),
\nabla_\theta\mathcal L_2(\theta)
\right\rangle+O(\eta^2).
\end{aligned}
\end{equation}
A positive gradient inner product means that the auxiliary update
locally reduces the fast-path loss for a sufficiently small step size.
Alignment can vary across tasks, layers, and scenes.

System~1 uses the resulting shared backbone without executing all
the auxiliary tasks that contributed to its training.
Figure~\ref{fig:gradient-alignment-insight} illustrates this transfer
mechanism.

\subsubsection{From Joint Relations to Latent Refill}

Shared parameters transfer learning across tasks, while latent refill
passes intermediate results within the current decision. 

For an active route of $L$ tasks $j_1,\ldots,j_L$,
the block $B_{\theta,\phi_{j_\ell}}$ uses task queries
$\mathbf q^{j_\ell}$ to compute a latent representation,
produce a prediction $\widehat y^{j_\ell}$,
and update the context:
\begin{equation}
\begin{aligned}
U_\ell
&=B_{\theta,\phi_{j_\ell}}
  (H_{\ell-1},\mathbf q^{j_\ell};\kappa_\ell),\\
\widehat y^{j_\ell}
&=\operatorname{Head}_{\phi_{j_\ell}}(U_\ell),\\
H_\ell
&=\operatorname{Refill}(H_{\ell-1},U_\ell).
\end{aligned}
\end{equation}
Here, $U_\ell$ is the task latent, and $H_\ell$ retains the initial
scene context and information from completed blocks.
For Perception, Causation, and World, the length of $U_\ell$
depends on the selected computation configuration.
Skipped tasks are omitted from the active route. For World and
Value, $\kappa_\ell=a=(d,\xi)$ specifies the realized candidate being
evaluated; for Perception, Causation, and Action, this additional
candidate condition is empty.

Causation therefore reads the context updated by Perception.
After Action generates candidates, subsequent computation follows
a separate branch for each candidate. World predicts the consequence
of that candidate, and Value reads the same candidate together with
the refilled World information when available. This correspondence
is maintained even when several trajectories are sampled from the
same Action latent.

The full route follows
$P\rightarrow C\rightarrow A\rightarrow W\rightarrow V$.
Let $Y^j$ denote the random output of task $j$, and let $p$
be the reference joint distribution of histories and task outputs.
For a fixed route, the conditional dependence follows the chain rule:
\begin{equation}
p(Y^{j_1},\ldots,Y^{j_L}\mid h)
=
\prod_{\ell=1}^{L}
p(Y^{j_\ell}\mid h,Y^{j_1},\ldots,Y^{j_{\ell-1}}).
\end{equation}
Latent refill provides a compact interface for carrying information
needed by these conditional predictions.

Queries within a block can run in parallel, and candidate branches
can be evaluated in parallel once their conditions are fixed.
The autoregressive meta-action and diffusion decoders inside Action
operate within a task block; refill organizes information flow
between task blocks.


\paragraph{Reduced Structural Information Gap.}
Independent task predictions can miss dependencies between outputs.
The following idealized comparison measures how much of this dependence
can be retained by passing upstream information through refill.

Write $Y_\ell=Y^{j_\ell}$ and
$Y_{<\ell}=(Y_1,\ldots,Y_{\ell-1})$ along the fixed route
and candidate branch.
Let $\mathsf H$ denote driving history with realization $h$.
Independent composition ignores the conditional dependence
among $\mathbf Y=(Y_1,\ldots,Y_L)$:
\begin{equation}
\mathcal L_{\mathrm{ind}}
=
\mathbb E_h
D_{\mathrm{KL}}\!\left(
p(\mathbf Y\mid h)
\,\middle\|\,
\prod_{\ell=1}^{L}p(Y_\ell\mid h)
\right)
=
\operatorname{TC}_p(\mathbf Y\mid\mathsf H).
\end{equation}
Here, $D_{\mathrm{KL}}$ denotes KL divergence,
$\operatorname{TC}_p$ conditional total correlation, and
$I_p$ below conditional mutual information under $p$.
For an idealized comparison, let
$S_{\ell-1}=s_{\ell-1}(h,Y_{<\ell})$
summarize the upstream information available to task $\ell$,
including the realized candidate for World and Value.
This summary represents the information supplied through both
the refill state $H_{\ell-1}$ and the candidate condition
$\kappa_\ell$.

Using the exact conditional distributions given these summaries
defines
\begin{equation}
q_{\mathrm{ref}}(\mathbf Y\mid h)
=
\prod_{\ell=1}^{L}
p(Y_\ell\mid h,S_{\ell-1}).
\end{equation}
The remaining structural mismatch is
\begin{equation}
\begin{aligned}
\mathcal L_{\mathrm{res}}
&=
\mathbb E_h D_{\mathrm{KL}}\!\left(
p(\mathbf Y\mid h)
\,\middle\|\,
q_{\mathrm{ref}}(\mathbf Y\mid h)
\right)\\
&=
\sum_{\ell=1}^{L}
I_p(Y_\ell;Y_{<\ell}\mid\mathsf H,S_{\ell-1})\\
&\leq \mathcal L_{\mathrm{ind}}.
\end{aligned}
\end{equation}
The reduction depends on how much downstream-relevant information
the summaries retain from realized upstream outputs. This comparison
motivates preserving candidate-specific results across tasks.
The implemented model follows this information flow through
candidate conditioning and learned latent refill; its practical
benefit depends on the learned representations and conditional
prediction accuracy.


\subsubsection{Execution Paths Within One Model}
Table~\ref{tab:exc_path} summarizes representative task routes within the same model. System~1 generates and evaluates candidates
directly. System~2 adds Perception and Causation before candidate
generation and evaluation. Full reasoning additionally invokes World
to predict candidate-specific visual consequences. The System~1 and
System~2 terminology provides a functional analogy to fast and
deliberative processing.

All paths use the same backbone and task interfaces.
The model supports multiple configurations of task-specific
latent-token budgets for Perception, Causation, and World.
The scheduler selects which tasks to execute and how much
latent computation to allocate to each, according to the expected
benefit, incremental cost, and remaining resource budget.
Latent refill carries the resulting intermediate information
to downstream tasks.
\begin{table}[!ht]
\centering
\small
\caption{Task routes specify execution order; latent-token allocations
depend on the selected computation configuration.}
\label{tab:exc_path}
\begin{tabularx}{\linewidth}{@{}l l X@{}}
\toprule
\rowcolor[HTML]{FFE0CC}
\textbf{Path} & \textbf{Task route} & \textbf{Purpose} \\
\midrule
\textbf{System 1} &
$A\rightarrow V$ &
Generates and evaluates candidates directly using the shared backbone. \\
\midrule
\textbf{System 2} &
$P\rightarrow C\rightarrow A\rightarrow V$ &
Uses explicit scene evidence and factor--decision relations
to generate and evaluate candidates. \\
\midrule
\textbf{Full reasoning} &
$P\rightarrow C\rightarrow A\rightarrow W\rightarrow V$ &
Additionally predicts candidate-conditioned visual consequences
for evaluation. \\
\bottomrule
\end{tabularx}
\end{table}
\subsection{Structured Data Annotation}

\subsubsection{Structured Causal Modeling}

Structured causal annotation organizes scene factors that affect action feasibility or the relative value of candidates. A key object identifies an entity, whereas a key factor describes a state or relation relevant to the current decision. For example, ``pedestrian'' identifies an object; ``the pedestrian is entering the ego vehicle's path'' identifies a factor that affects whether to proceed.
We represent this at time $t$ as a directed bipartite graph:
\begin{equation}
\mathcal G_t^C=(\mathcal V_t^C,\mathcal E_t^C),\qquad
\mathcal V_t^C=\mathcal F_t\mathbin{\dot\cup}\mathcal D_t^{\mathrm{dec}},
\qquad
\mathcal E_t^C\subseteq\mathcal F_t\times\mathcal D_t^{\mathrm{dec}}.
\end{equation}
Here, $\mathcal F_t=\{f_i\}$ contains factor nodes and
$\mathcal D_t^{\mathrm{dec}}=\{d_j\}$ contains decision nodes. A decision node specifies a lateral or longitudinal semantic behavior and, when needed, its associated object. A candidate action $a=(d,\xi)$ also specifies a trajectory, so multiple trajectories may realize the same decision.
An edge $(f_i,d_j)\in\mathcal E_t^C$ indicates that factor $f_i$ affects the feasibility or preference of decision $d_j$. Nodes retain stable identities across time. Each active edge has a role, a set of valid times, and an evidence attribute:
\begin{equation}
\varnothing\neq\ell_{ij}(t)
\subseteq\{\mathrm{MOTIVE},\mathrm{CONSTRAINT}\},\qquad
I_{ij}=\{t:(f_i,d_j)\in\mathcal E_t^C\}.
\end{equation}
Here, $\ell_{ij}(t)$ specifies the active role or roles of the
relation, and $I_{ij}$ records when the relation is valid.
Each relation also stores links to supporting observations,
scene labels, or verification records.

\begin{table}[!htp]
\centering
\small
\caption{Roles of causal relations in structured annotation.}
\begin{tabularx}{\linewidth}{@{}lXX@{}}
\toprule
\rowcolor[HTML]{FFE0CC}
\textbf{Role} & \textbf{Meaning} & \textbf{Relation to the driving objective} \\
\midrule
\textbf{MOTIVE} & Supports or sustains a decision through the navigation goal or expected benefit. & Provides evidence for preferring a candidate. \\\midrule
\textbf{CONSTRAINT} & Restricts a behavior through conditions, risks, or other costs. & Defines a feasibility boundary or a candidate cost. \\
\bottomrule
\end{tabularx}
\end{table}

A factor may both motivate a decision and constrain its execution.
Each relation has its own validity interval, allowing lateral and
longitudinal decisions to overlap or change asynchronously.
Mandatory conditions restrict feasibility, while negotiable costs
enter candidate evaluation.

Counterfactual comparison asks two questions: does changing a factor
alter whether an action is feasible, and does it alter the relative
value of two actions? Let $\widetilde h$ be a comparable, physically
consistent scene in which factor $f$ is changed to $\widetilde f$.
Using the same feasible reference policy $\bar\pi$ in both scenes,
define
\begin{equation}
\begin{aligned}
\Delta_{f\to\widetilde f}^{\mathrm{feas}}(a)
&=
\mathbf 1[a\in\mathcal A_{\mathrm{feas}}(h)]
-
\mathbf 1[a\in\mathcal A_{\mathrm{feas}}(\widetilde h)],\\
\Delta_{f\to\widetilde f}^{\mathrm{value}}(a,a')
&=
\bigl[Q_t^{\bar\pi}(h,a)-Q_t^{\bar\pi}(h,a')\bigr]
-
\bigl[Q_t^{\bar\pi}(\widetilde h,a)
      -Q_t^{\bar\pi}(\widetilde h,a')\bigr].
\end{aligned}
\end{equation}
Here, $\mathcal A_{\mathrm{feas}}(h)$ abbreviates the admissible
action set under the belief induced by $h$ and its navigation goal.
The first quantity tests whether feasibility changes. The second compares the relative values of candidates that are feasible and semantically comparable in both scenes. A nonzero change for candidates associated with $d_j$ indicates that $f$ contains information relevant to that decision; the preferred candidate need not change.

Counterfactual supervision therefore compares both altered scene factors and alternative behaviors within the same scene and planning horizon. Observations, rule checks, simulation, and human review turn these comparisons into feasibility labels, outcome assessments, and preference pairs. The explicit graph $\mathcal G_t^C$ supports annotation and interpretation, while task latents support online computation. Text, structured labels, and behavior feedback provide different training interfaces to the same relations.

\subsubsection{Data Annotation Loop}

Each scene--candidate record combines scene evidence, candidate behavior, and quality assessments. Semantic labels capture decision-relevant distinctions, while physical measures such as clearance, time margins, smoothness, and progress ground candidate
assessments in physical outcomes. More details can be found in Appendix.~\Cref{sec:annotation}.

Five-dimensional assessments cover safety, comfort, efficiency,
compliance, and navigation following. Their textual analyses provide VQA supervision during pretraining; the associated labels and physical measures support candidate evaluation, post-training, and allocation of inference computation.

We organize experience into scene--candidate records:
\begin{equation}
\mathcal R_i=
\left(
h_i,y_i^P,\mathcal G_i^C,
\{(a_{in},y_{in}^W,\boldsymbol v_{in},\nu_{in})\}_{n=1}^{N_i},
\omega_i,\zeta_i
\right).
\end{equation}
Here, $i$ indexes a scene and decision time, and $n$ indexes
one of the $N_i$ candidates in the record.
$y_i^P$ denotes scene labels, $y_{in}^W$ a future visual target,
$\boldsymbol v_{in}\in\{0,1\}^5$ the five-dimensional assessment,
with $1$ indicating a problem in the corresponding dimension,
and $\nu_{in}$ a feasibility label. The mask $\omega_i$ indicates which supervision items are valid; $\zeta_i$ stores provenance, time range, confidence, and version information. Causation labels are derived from the graph's factors, decisions, roles, and valid times.

\begin{table}[t]
\centering
\small
\caption{Supervision sources and their roles in driving computation.}
\begin{tabularx}{\linewidth}{@{}p{0.20\linewidth}XX@{}}
\toprule
\rowcolor[HTML]{FFE0CC}
\textbf{Computation} & \textbf{Main supervision} & \textbf{Purpose} \\
\midrule
\textbf{Perception and Causation} & Scene facts, key objects, and factor--decision relations. & Retain information needed to generate and compare candidates. \\
\midrule
\textbf{Action} & Decisions, meta-actions, and continuous trajectories. & Improve candidate coverage. \\
\midrule
\textbf{World} & Executed actions and observed future scenes. & Improve consequence prediction. \\
\midrule
\textbf{Value} & Five-dimensional ratings, outcome evidence, and candidate preferences. & Improve candidate selection. \\
\midrule
\textbf{Feasibility check} & Physical quantities, rules, violations, and supporting evidence. & Determine whether behavioral constraints are met. \\
\bottomrule
\end{tabularx}
\end{table}

\begin{figure}[t]
\centering
\includegraphics[width=\linewidth]{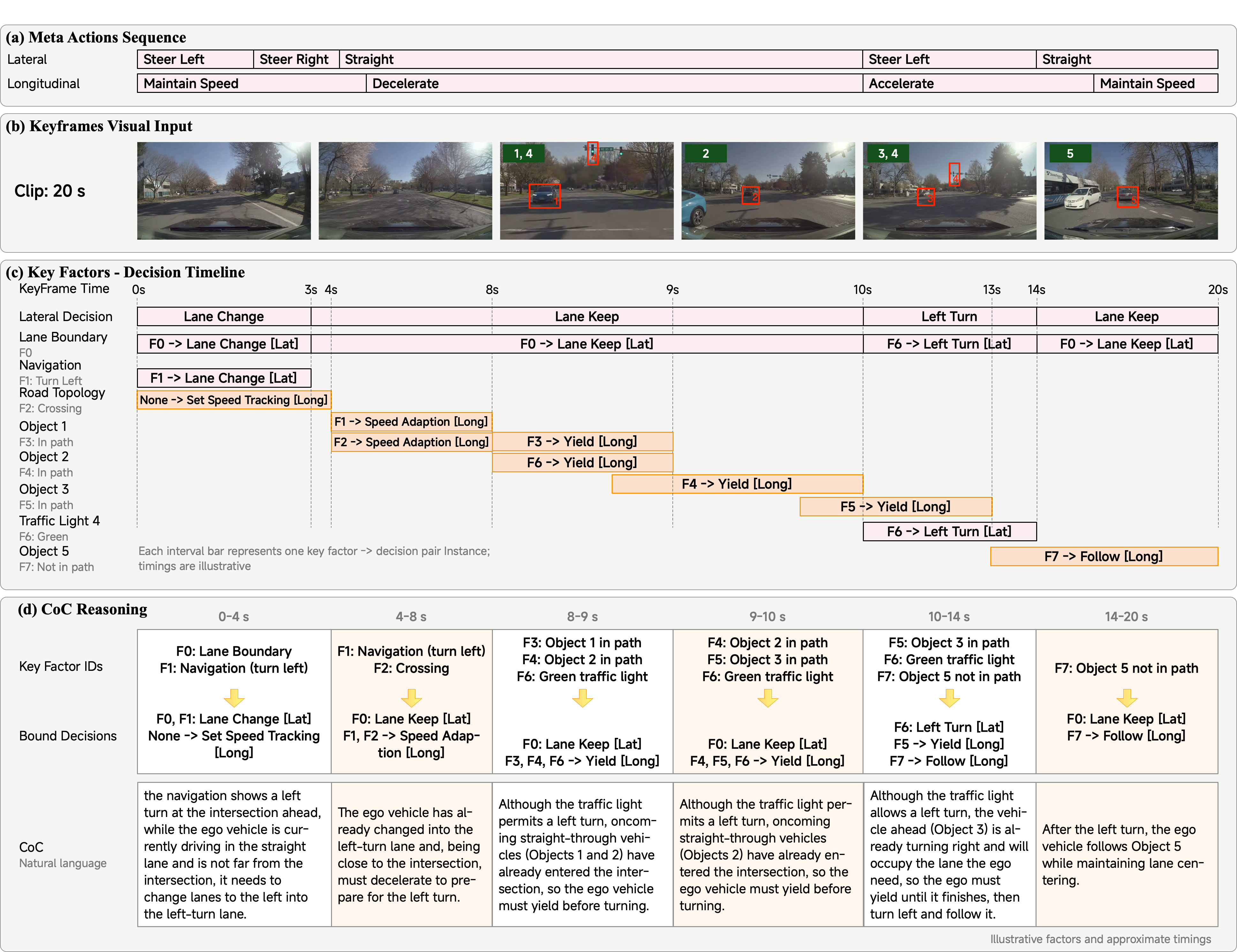}
\caption{Annotation pipeline for key factor--decision pairs and CoC, illustrated using a 20-second driving clip. The pipeline bridges perception and reasoning through four stages: 
(a) atomic Meta Actions sequence, categorized into lateral and longitudinal maneuvers; (b) visual inputs at selected keyframes, with detected objects annotated; (c) a timeline of lateral and longitudinal decisions associated with key factors; and (d) CoC generation, linking key factor IDs to their associated decisions and natural-language explanations. Timings and factor states are illustrative.}
\label{fig:overview_coc_autolabel_pipeline}
\end{figure}

The record binds supervision to the same scene, time range, and candidate identity. Observed futures supervise World predictions for executed actions. Consequences and preferences for unexecuted candidates come from rules, simulation, or human review, with their sources retained. Decision inputs contain only information available at decision time; later observations are stored separately as outcome supervision.

Causal annotation starts from action and semantic key moments to define decision segments. Within each segment, annotators identify Key factors, link them to lateral or longitudinal decisions, and record roles, valid times, and evidence. Each factor--decision relation forms a decision instance; instances may overlap or change asynchronously. Language explanations are generated from currently valid instances, while structured labels directly supervise objects, decisions, and relations.

\begin{table}[!htbp]
\centering
\caption{Categories and names of all candidate decisions.}
\label{tab:decision_classification_en}
\renewcommand{\arraystretch}{1.2} 
\begin{tabular}{@{}l@{\hspace{1em}}p{0.75\textwidth}@{}}
\hline
\textbf{Category} & \textbf{Name} \\
\hline
Lateral & Lane Keep, Lane Change, Nudge, Merge/Split, Turn \\
\hline
Longitudinal & Gap Search, Passing with Lateral Movement, Proceed with Priority, Speed Adaptation, Yield, Start up, Stop for Static Constraints, Lead Obstacle Following, Set Speed Tracking \\
\hline
\end{tabular}
\end{table}

\paragraph{Decision Definition}
The names of all decision candidates are listed in Table~\ref{tab:decision_classification_en}, and the specific definition of each decision is given in Table~\ref{tab:appendix_decision_classification_en} in the Appendix. 

\paragraph{CoC Autolabeling Pipeline.}
Inspired by AR1~\cite{wang2025alpamayo}, the pipeline combines
rule-based annotation, VLM prediction, and human correction to
produce factor--decision annotations and grounded explanations.
Its inputs include multi-view and historical images, navigation,
ego and surrounding-vehicle trajectories, lane geometry, and
traffic-control information. Figure~\ref{fig:overview_coc_autolabel_pipeline}
illustrates the autolabeling pipeline, which consists of four interconnected stages:

\begin{enumerate}

    \item \textbf{Annotate meta-actions.} Ego trajectories provide atomic meta-action labels. Lateral meta-actions include Steer Left, Steer Right, and Straight, while longitudinal meta-actions include Accelerate, Decelerate, and Maintain Speed.

    \item \textbf{Select keyframes.} Changes in meta-actions identify keyframes for subsequent decision analysis. Representative multi-view images are extracted at these times.

    \item \textbf{Generate rule-based factors and decisions.} Lane topology, cross-line events, and yaw changes support lateral
    annotation. For longitudinal annotation, the rules use time headway derived
    from the space--time (ST) representation, together with traffic-signal states
    and speed limits.
    The rules also identify relevant objects and road elements.

    \item \textbf{Refine annotations with a VLM.} Given rule-based labels and keyframes, a VLM predicts key factors and decisions. It generates natural-language CoC reasoning by linking key factor IDs to decisions (e.g., yielding to an in-path object). Human review corrects these outputs and refines trajectory assessments for safety, comfort, efficiency, compliance, and navigation following.
\end{enumerate}

The final outputs link key factors to their associated decisions and explanations. Prompts organize the available evidence, priorities, and scenario-specific conditions. Verified human corrections iteratively improve both the annotation model and the data for downstream driving-model training.

The record collection $\mathcal R=\{\mathcal R_i\}$ yields
pretraining, SFT, and feedback datasets:
$\mathcal R
\longrightarrow
\{\mathcal D_{\mathrm{pre}},
  \mathcal D_{\mathrm{FT}},
  \mathcal D_{\mathrm{feedback}}\}.$
Data cost depends on which annotation operations are performed:
$C_D=\sum_{i,r}\kappa_{ir}c_{ir}.$
Here, $r$ indexes an annotation operation, $\kappa_{ir}\in\{0,1\}$ indicates whether it is performed, and $c_{ir}$ includes automatic production, verification, and rework. The validity mask $\omega_i$ is recorded separately. A verified fact or relation can support multiple training targets.

\subsection{Training Strategy}
\begin{figure}[ht]
\centering
\includegraphics[width=\linewidth]{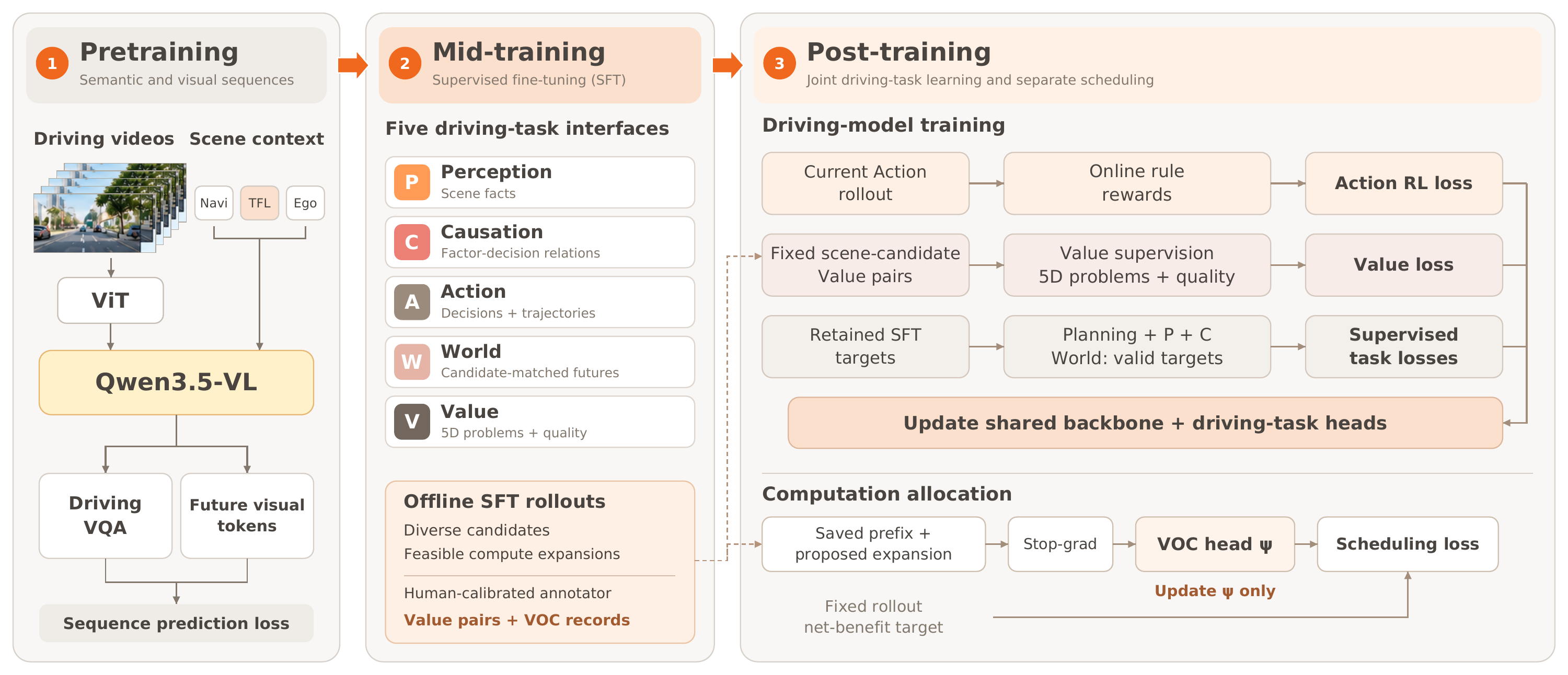}
\caption{Training pipeline of \Model{}, comprising pretraining,
mid-training (SFT), and post-training.}
\label{fig:architecture_pipeline}
\end{figure}

\Model{} is built on Qwen-3.5-VL and trained in three stages.
Pretraining establishes a shared semantic and visual sequence
interface through VQA and visual-token prediction.
Mid-training performs supervised fine-tuning (SFT), training the
Action decoder to generate continuous trajectories and grounding
the task interfaces in their corresponding supervision.
Post-training adds policy optimization for Action using rule-based rewards
while retaining the supervised objectives from mid-training.
It continues Value supervision, applies World supervision on
valid executed paths, and trains inference-budget allocation.

\subsubsection{Pretraining}
\begin{figure}[!htbp]
\centering
\includegraphics[width=\linewidth]{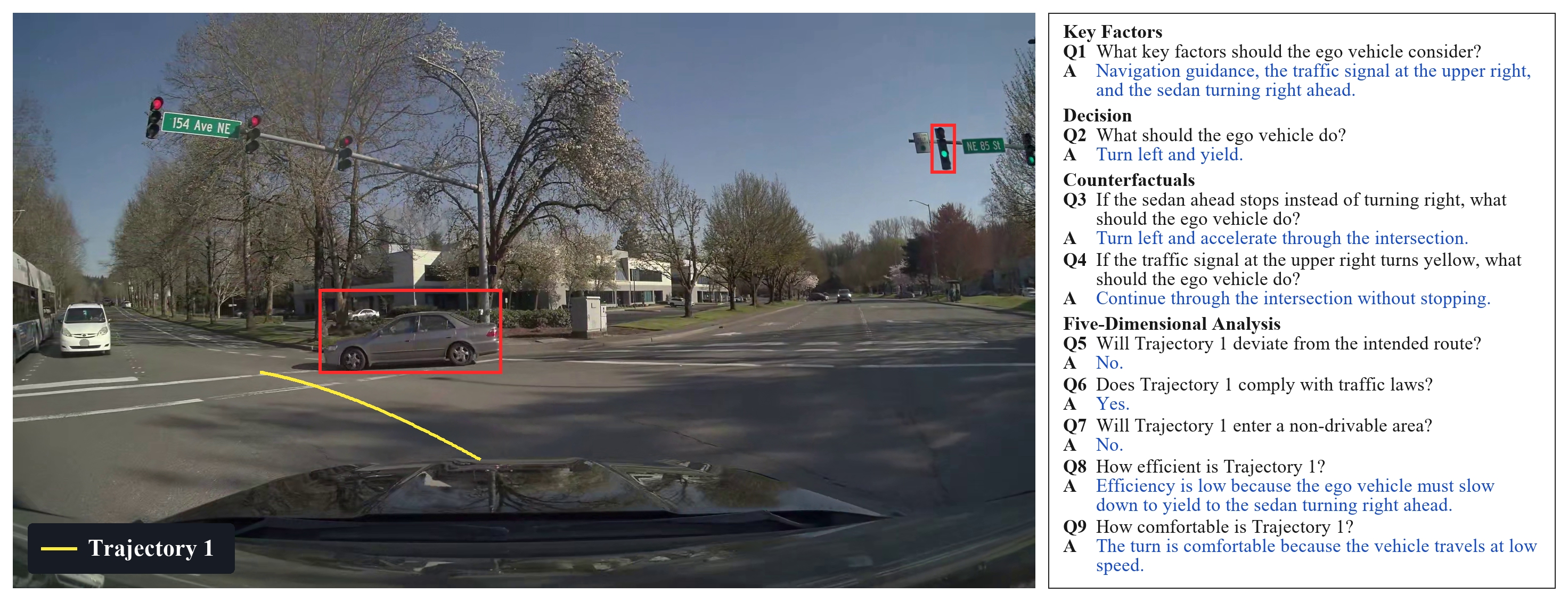}
\caption{Example of a training sample in the pretraining stage.}
\label{fig:example_pretrain_dataset}
\end{figure}
Pretraining establishes a shared sequence interface for driving
understanding. VQA targets cover key factor identification,
longitudinal and lateral decision prediction, counterfactual analysis,
and five-dimensional trajectory analysis. A visual tokenizer represents
future visual targets as discrete tokens. Text tokens, visual tokens,
and task markers form a common vocabulary.

Each training example pairs an input $X$ with a target sequence $Y$.
$X$ contains observations, task conditions, and the question prompt;
candidate evaluation and future prediction additionally specify
the corresponding action. The model learns to predict each target
token from the input and preceding target tokens:
\begin{equation}
\mathcal L_{\mathrm{pre}}
=
-\mathbb E_{(X,Y)}
\sum_{q=1}^{|Y|}
\log p_{\Theta_{\mathrm{pre}}}(Y_q\mid X,Y_{<q}).
\end{equation}
Here, $p_{\Theta_{\mathrm{pre}}}$ is the token-prediction distribution
of the pretraining model. $Y_q$ is the target token at position $q$,
$Y_{<q}$ contains the preceding tokens, and $|Y|$ is the sequence length.
The expectation is over pretraining examples.
This stage connects scene evidence, decision concepts, and outcome
assessments in the shared backbone. The continuous Action decoder
is trained later during SFT.

Figure~\ref{fig:example_pretrain_dataset} shows the front-camera view and associated QA targets of a training sample. Red boxes highlight key factors; the yellow line represents a candidate future trajectory of the ego vehicle; and the right side shows the question--answer pairs for each task.
The language-model loss is computed on the target text shown
in blue.


\subsubsection{Mid-Training}

Mid-training performs SFT on the multi-task interfaces. The Action decoder
learns continuous trajectory generation from planning supervision.
Candidate search expands its proposals into alternative plans,
which human experts and foundation models assess to provide
supervision beyond the recorded behavior. Perception and Causation
learn scene evidence and factor--decision relations, while World
and Value learn from available future visual targets and candidate
assessments. The objectives are:
\begin{equation}
\mathcal L_{\mathrm{Mid}}
=
\mathcal L_{\mathrm{plan}}
+\lambda_P\mathcal L_P
+\lambda_C\mathcal L_C
+\lambda_W\mathcal L_W
+\lambda_V\mathcal L_V.
\end{equation}
Here, $\mathcal L_{\mathrm{plan}}$ supervises Action,
while $\mathcal L_j$ and $\lambda_j$ denote the supervised
loss and its weight for task $j\in\{P,C,W,V\}$.
Each term uses only valid supervision.
The objectives can be optimized in separate updates; World and
Value training does not require gradients through candidate search
or a simultaneous planner update.




\subsubsection{Post-Training}
Post-training combines policy optimization for Action using rule-based rewards
with continued supervision of the previously learned tasks.
The supervised objectives and their loss weights are retained
to limit degradation of existing capabilities as the shared
backbone changes. Value continues to learn from annotated
scene--candidate pairs, while World receives supervision only
on enabled paths with valid candidate-matched future targets.
Although rollout annotations for post-training can also be
collected online, we prepare them offline from rollouts of the SFT model
before this stage to avoid the substantial generation and
annotation overhead during training.
\paragraph{Preparing Value supervision.}
Automatic annotations of obstacles, roads, traffic signals, and
motion states support five-dimensional assessment of recorded
driving trajectories. 
Human-refined judgments train and calibrate the annotation model,
which then evaluates diverse candidates generated by the SFT model.
Recorded trajectories with human-refined assessments and
annotated candidates generated by the SFT model form the
supervision set
\begin{equation}
\mathcal D_V
=
\{(h_i,a_i,\mathbf y_i,\mathbf m_i)\},
\end{equation}
where $a_i=(d_i,\xi_i)$ specifies the candidate decision and
trajectory. The binary target $\mathbf y_i\in\{0,1\}^5$ indicates
problems in safety, comfort, efficiency, compliance, and navigation following, with $y_{ij}=1$ denoting a problem.
The mask $m_{ij}=1$ identifies a valid and applicable judgment.
This dataset remains fixed within a post-training phase and is
refreshed in subsequent data iterations.

\paragraph{Joint Action and Value training.}
Alongside continued supervision of the previously learned tasks,
each training iteration combines two branches. In the Action branch,
the current policy generates trajectories, and online rules compute
rewards from scene geometry, traffic constraints, and trajectory kinematics.
Policy optimization updates Action using these rewards.
In the Value branch, fixed scene--candidate pairs are sampled from
$\mathcal D_V$ for supervised prediction of the five problem
probabilities. Supervised batches for planning, Perception, and Causation remain
in the training mixture to preserve their existing capabilities.

Training covers sampled task paths and latent-length configurations
with budgets $\mathbf b=(b_P,b_C,b_A,b_W)$.
The components $b_P,b_C,b_W$ specify the latent-token allocations
for Perception, Causation, and World, respectively, and vary
across supported inference configurations.
The Action latent-token budget $b_A$ remains fixed.
For a Value sample $(h,a)$, the context $c_{\mathbf b}(h,a)$ is
computed for that specific candidate. When World is enabled, it
conditions on the same $a$ and supplies future information to Value.
World loss is computed only when the block is executed and a valid
future visual target for that candidate is available.

The joint task-training objective retains the supervised losses
from mid-training and adds policy optimization:
\begin{equation}
\begin{aligned}
\mathcal L_{\mathrm{joint}}^{\mathrm{post}}
={}&
\mathcal L_A^{\mathrm{RL}}
+\mathcal L_{\mathrm{plan}}
+\lambda_P\mathcal L_P
+\lambda_C\mathcal L_C\\
&+\lambda_V\mathcal L_V
+\lambda_W\mathcal L_W^{\mathrm{valid}}.
\end{aligned}
\end{equation}
The supervised terms retain their mid-training definitions and
loss weights. Each term uses valid supervision on the corresponding
executed path.

Value retains its binary and continuous supervision. For the
binary judgments in $\mathcal D_V$, its loss component on a
sampled path is
\begin{equation}
\mathcal L_V^{\mathrm{bin}}
=
\mathbb E_{(h,a,\mathbf y,\mathbf m)\sim\mathcal D_V}
\left[
\sum_{j=1}^{5}
m_j\operatorname{BCE}
\left(
p_j(h,a,c_{\mathbf b}(h,a)),y_j
\right)
\right].
\end{equation}
Here, $\operatorname{BCE}$ is binary cross-entropy, and $p_j$
is the predicted probability of a problem in dimension $j$. The continuous component of
$\mathcal L_V$ supervises candidate-quality scores derived from
clearance, time margins, smoothness, and progress, using the
retained supervised data.


\paragraph{Gradient flow.}
All five tasks share the backbone while retaining their task
interfaces and output heads. The RL loss updates Action and
the shared backbone. The retained supervised losses continue
to train their corresponding tasks. Gradients also propagate
through upstream modules connected by continuous latent refill,
so the shared backbone receives the weighted contributions
of the joint objective.

Online rule rewards, offline candidate trajectories, and annotation
targets are treated as fixed within each update.
The gradient connections described above apply to the driving-task
losses. The value-of-computation (VOC) loss uses detached inputs
and targets and updates only the dedicated scheduling head.

\paragraph{Learning computation allocation.}
A dedicated VOC head within the Value interface predicts whether
additional computation is worth its cost. Let $\sigma_0$ denote
the current inference state, containing the latents of executed
blocks, candidate trajectories, their five-dimensional Value
outputs, the executed-task mask, the current budget $\mathbf b_0$,
and the remaining compute, time, and memory budgets.
For a proposed feasible expansion $\mathbf b\succeq\mathbf b_0$,
the head predicts
\begin{equation}
\begin{aligned}
x_{\mathrm{sched}}
&=\operatorname{sg}(\sigma_0),\\
\widehat{\operatorname{VOC}}_\psi
(\mathbf b\mid\sigma_0)
&=f_\psi(x_{\mathrm{sched}},\mathbf b),
\end{aligned}
\end{equation}
Here, $\psi$ denotes the parameters of the dedicated scheduling
head within $\phi_V$, and $\operatorname{sg}$ is the
stop-gradient operator. Unexecuted blocks are marked
by the mask and contribute no latent information.
Proposed expansions include activating skipped tasks and allocating
additional latent computation to tasks already in use.
The head scores each expansion without first executing
the proposed computation.

Scheduling supervision is collected using an SFT checkpoint.
Alternative feasible expansions are executed from the same
saved inference prefix, and their resulting candidates are
evaluated through human annotation and automatic scoring.
These evaluations provide shared supervision for Value and VOC:
Value learns to assess individual candidates, whereas VOC learns
the improvement in the selected candidate relative to
stopping at the current prefix, after accounting for additional
execution cost. The shared evaluations define the return
estimates $\widehat R$ used below.
The target and scheduling loss are
\begin{equation}
y_{\mathrm{VOC}}
=
\operatorname{sg}\!\left[
\widehat R(\sigma_0,\mathbf b)
-\widehat R(\sigma_0,\mathbf b_0)
-\lambda_I\Delta C_I(\sigma_0,\mathbf b\mid\mathbf b_0)
\right],
\end{equation}
\begin{equation}
\mathcal L_{\mathrm{sched}}
=
\mathbb E\!\left[
\left(
f_\psi(x_{\mathrm{sched}},\mathbf b)
-y_{\mathrm{VOC}}
\right)^2
\right].
\end{equation}
Here, $\widehat R$ estimates the driving return of the finally
selected candidate. Both returns use the same evaluation setting,
horizon, and return criterion.
$\Delta C_I$ counts the additional execution cost, including
required downstream recomputation, and $\lambda_I\geq0$
weights compute cost relative to driving return.
The expectation in $\mathcal L_{\mathrm{sched}}$ is over
the offline scheduling records.

At the system level, computation allocation aims to improve driving
return while keeping average inference compute within budget.
The corresponding Lagrangian objective is
\begin{equation}
\begin{gathered}
\max_{\Theta,\rho}\ \min_{\lambda_I\geq 0}
\mathcal F_{\mathrm{budget}}(\Theta,\rho;\lambda_I),\\
\mathcal F_{\mathrm{budget}}(\Theta,\rho;\lambda_I)
=
\widehat J_{\mathrm{drive}}(\pi_{\mathcal M,\Theta,\rho})
-\lambda_I\!\left(
\mathbb E_{h\sim d_\pi}[C_I(h,\rho)]-B_I
\right).
\end{gathered}
\end{equation}
Here, $\pi_{\mathcal M,\Theta,\rho}$ is the policy specified by
the architecture, parameters, and inference configuration, and
$\widehat J_{\mathrm{drive}}$ estimates its closed-loop return.
$C_I(h,\rho)$ is inference compute for history $h$;
$d_\pi$ is the distribution of histories encountered under the
deployed policy, and $B_I$ is the budget for average inference compute.
This expresses the design objective; the practical updates use the
task and scheduling losses above. Per-decision latency and memory
limits remain runtime constraints.

Training cost sums the resource use of all training iterations:
\[
C_T=
\sum_{k=0}^{K_{\mathrm{train}}-1}
c_{\mathrm{step}}
\bigl(\mathcal M,\beta(k),\mathcal D_k\bigr).
\]
Here, $K_{\mathrm{train}}$ is the total number of training
iterations across all stages, and $c_{\mathrm{step}}$ is
the resource cost of one iteration.

\subsection{Inference Scheduling}

\begin{figure}[t]
\centering
\includegraphics[width=\linewidth]{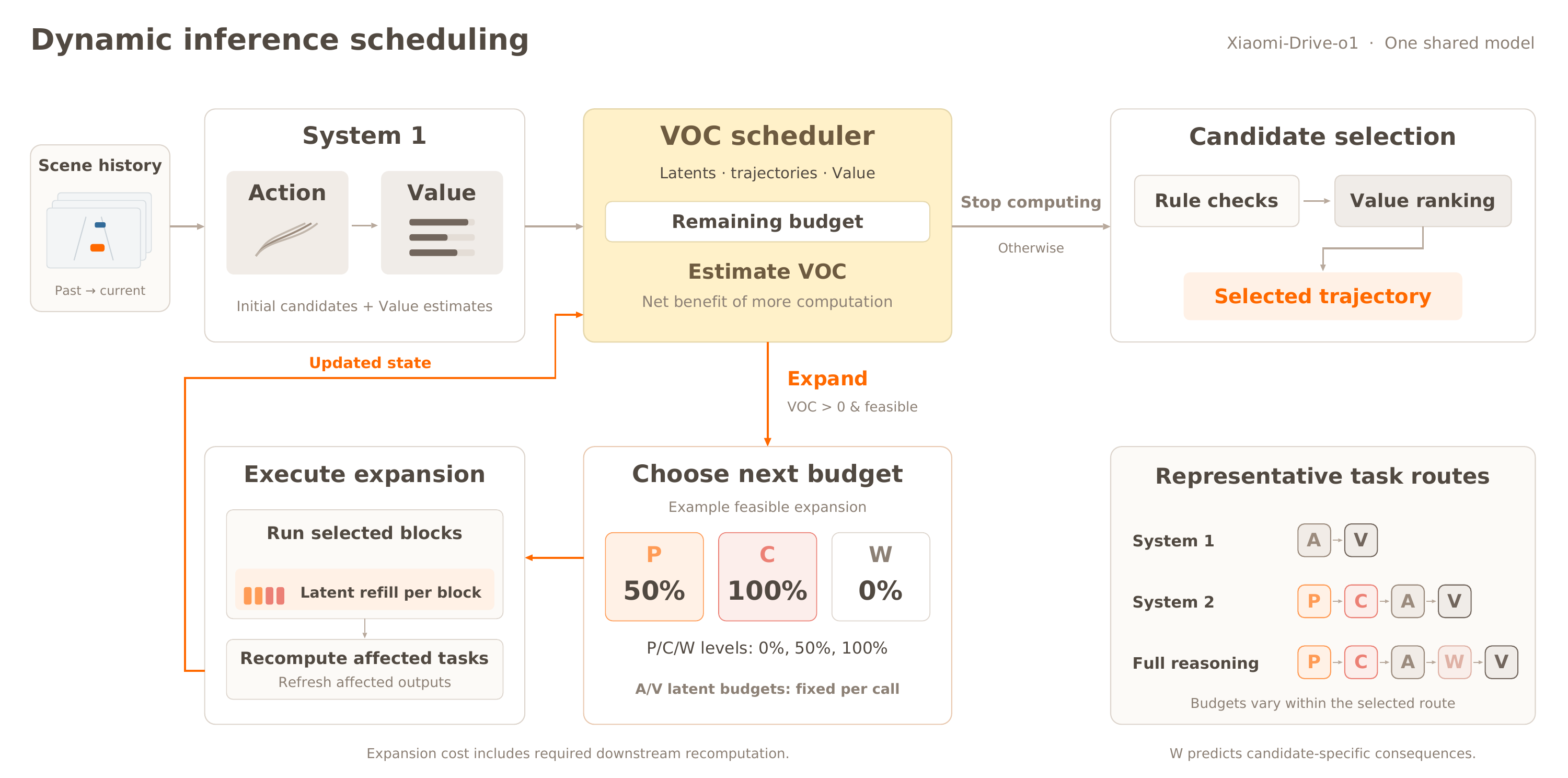}
\caption{Dynamic inference scheduling. The Action--Value path
first generates and evaluates candidates and the scheduler uses
value-of-computation (VOC) estimates to select additional tasks
and latent-token budgets within the remaining resources.
Perception and Causation refine candidate generation, while World
supplies candidate-conditioned future evidence for Value.
Rule-based checks and Value assessments support candidate selection;
a predefined fallback is invoked when no acceptable candidate
can be obtained within the runtime budget.}
\label{fig:infer_frame}
\end{figure}
Inference scheduling adapts the task path and latent-token budgets
to the needs of the current decision, as illustrated in
\Cref{fig:infer_frame}. Each decision begins with the Action--Value
path: Action generates candidate trajectories, Value predicts their
problem probabilities across the five evaluation dimensions, and
Check applies rule-based feasibility tests. Rule-based rewards and
learned Value assessments jointly rank the admissible candidates.
If the selected candidate meets the acceptance criteria, inference
terminates without invoking additional tasks.
When the initial result is insufficient, the scheduler uses the trained VOC head to compare feasible computation expansions.
These expansions can activate Perception, Causation, or World,
or increase the latent-token budgets of these tasks; the Action
latent-token budget remains fixed. Value assesses the current
candidates, whereas VOC estimates how much an expansion could
improve the final choice relative to stopping, after accounting
for its additional cost. The scheduler combines these estimates
with candidate acceptance criteria and the remaining resources
to decide whether and how to continue computation.

Each selected expansion follows the task dependencies.
Perception and Causation provide additional scene evidence and
factor--decision relations for Action to generate updated candidates.
World instead conditions on a generated candidate and predicts
its visual consequences, supplying future evidence for Value.
Latent refill carries the realized intermediate outputs to the
downstream blocks, keeping candidate generation and evaluation
aligned with the information actually computed. A skipped block
contributes no new result, and any downstream recomputation
required by an expansion is included in its incremental cost.

After an expansion, the updated candidates are re-evaluated,
problematic candidates are filtered, and the remaining candidates
are re-ranked. The highest-ranked candidate is returned if it meets
the acceptance criteria; otherwise, the system generates further
candidates or expands reasoning while resources permit.
When no acceptable candidate is available
and the remaining budget cannot support further computation,
the predefined fallback procedure is invoked.
Because online rule rewards can miss problematic Action proposals,
on-vehicle validation checks the candidates and their predicted
consequences to intercept identifiable problems before execution.

Runtime constraints bound every computation expansion.
We represent an execution as a directed acyclic graph
$G_\rho(h)=(\mathcal N_\rho(h),\mathcal E_\rho(h))$,
with computation calls as nodes and task dependencies and device
scheduling order as edges. Total compute, critical-path latency,
and peak memory usage are
\begin{equation}
\begin{aligned}
C_I(h,\rho)
&=\sum_{v\in\mathcal N_\rho(h)}c_v,\\
T_I(h,\rho)
&=T_{\mathrm{fixed}}
+\max_{p\in\operatorname{Paths}(G_\rho(h))}
  \sum_{v\in p}\Delta_v,\\
M_I(h,\rho)
&=M_{\mathrm{weights}}
+\max_{u}M_{\mathrm{active}}(u).
\end{aligned}
\end{equation}
$c_v$ and $\Delta_v$ are the compute and duration of call $v$;
$\operatorname{Paths}(G_\rho(h))$ contains the directed execution
paths, and $T_{\mathrm{fixed}}$ covers overhead outside the graph.
$M_{\mathrm{weights}}$ is resident model-weight memory, and
$M_{\mathrm{active}}(u)$ is additional active memory at execution
time $u$, including caches and workspaces.
Repeated calls are counted separately.
The scheduler admits additional calls only when time and memory remain available, including the reserve for control execution.

\subsection{Closed-Loop Data and Model Iteration}


Offline iteration uses driving failures to identify gaps in supervision.
Diagnosis examines the generated candidates, their supporting evidence,
and their evaluation. Missing behaviors motivate additional decision
or trajectory targets. Incorrect candidate rankings motivate review
of consequence and Value assessments. Feasibility errors motivate
review of scene facts, factor relations, and checks.

Incorrect rule judgments, problems missed by Value, and World
prediction errors identified during training or deployment
are fed back into the data loop.
Human review identifies which scene--candidate fields require
correction and updates the annotation model, labeled candidate records, or rules accordingly. Newly generated candidates are included
when constructing $\mathcal D_V$ for subsequent training phases.
Evaluation of the updated model checks improvement on related
cases and supplies the next set of failures.

Let $e$ be an experience segment, $\ell$ an annotation addition or revision, and $\Theta^+(e,\ell)$ the parameters after training with that supervision. Its ideal expected learning gain is
\[
G(e,\ell)
=
\mathbb E\!\left[
J_{\mathrm{drive}}\bigl(\pi_{\Theta^+(e,\ell)}\bigr)
-
J_{\mathrm{drive}}(\pi_\Theta)
\right].
\]
The comparison holds the architecture and inference configuration fixed. In practice, failure severity, scene coverage, model disagreement, and annotation cost guide prioritization; subsequent closed-loop evaluation measures the realized gain.

Online scheduling allocates computation to the current decision;
offline iteration allocates annotation and training effort to
improving future decisions. Shared records connect both processes
through the same scene evidence, candidates, and outcome assessments.


%% file: sections/evaluation.tex
\section{Experiments}
\label{sec:evaluation}
The evaluation focuses on two questions: whether structured causal supervision improves driving understanding and planning, and whether Value-guided dynamic reasoning allocates computation effectively. Using Qwen3.5-VL as the pretrained vision--language backbone, we compare \Model{} with Alpamayo 1.5~\cite{nvidia2026alpamayo15} Alpamayo 2 Super~\cite{ivanovic2026alpamayo2} and Qwen-Drive-1.0~\cite{zhou2026qwendrive10initialstepvisionlanguage} on NVIDIA's PhysicalAI-Autonomous-Vehicles~\cite{nvidia2026physicalai} dataset using a shared evaluation protocol that combines semantic causal assessment with geometric planning assessment. 

\subsection{Evaluation Protocol and Metrics}
\label{sec:eval_protocol}
We evaluate \Model{} on NVIDIA's PhysicalAI--Autonomous--Vehicles (PAI-AV) benchmark~\cite{nvidia2026physicalai}. The benchmark provides the scenes and recorded trajectories used to construct the linked annotation records in Section~\ref{sec:annotation}; each evaluation unit therefore retains the correspondence among a scene, candidate behavior, expected consequence, and value label.
Our high-quality test subset is drawn from clips excluded from
our training and validation data.
Candidate-selection rules and inference thresholds are calibrated on the validation split. Comparisons include Alpamayo 1.5 (AR1.5)~\cite{nvidia2026alpamayo15}, Alpamayo 2 Super (AR2)~\cite{ivanovic2026alpamayo2}, and Qwen-Drive-1.0~\cite{zhou2026qwendrive10initialstepvisionlanguage}. For every run, we record the released checkpoint, input configuration, output format, candidate count, and inference budget. The protocol scores observable outputs under a common scorer and does not assume that an AR baseline exposes the dedicated Perception, World, or Value heads used by \Model{}.

\paragraph{Open-loop PhysicalAI evaluation.}
PAI-AV requires each method to predict six candidate trajectories for every scene. We evaluate the selected trajectory with ADE@1 and the candidate set with minADE@6, where ADE@1 measures the geometric accuracy of the directly executed prediction and minADE@6 measures whether at least one candidate covers the recorded motion. Both metrics are reported at 3~s and 5~s horizons, with lower values indicating better trajectory fidelity. The standard 644-example Physical-AI NV split overlaps publicly available training data by its official construction. To make the comparison leakage-aware, we additionally report results on a high-quality subset curated exclusively from held-out test clips;

\paragraph{Closed-loop AlpaSim evaluation.}
AlpaSim evaluates behavior after the ego vehicle interacts with the simulated scene, exposing errors that accumulate beyond the recorded trajectory. We run 1,607 scenarios with PAI-AV-NuRec version 26.04. Following the evaluation protocol used by Alpamayo~\cite{wang2025alpamayo}, we report close-encounter rate for all and at-fault events, off-road rate, progress, and AlpaSim score for all and at-fault events. Lower values are preferred for encounter and off-road rates, whereas higher values are preferred for progress and score. The at-fault variants exclude events not caused by the ego vehicle; together, these metrics measure close-proximity events, roadway departures, scenario progress, and distance traveled between events.

\subsection{Reasoning Results}
\begin{figure}[!htp]
  \centering
  \begin{subfigure}[b]{0.49\textwidth}
    \centering
    \includegraphics[height=6.6cm]{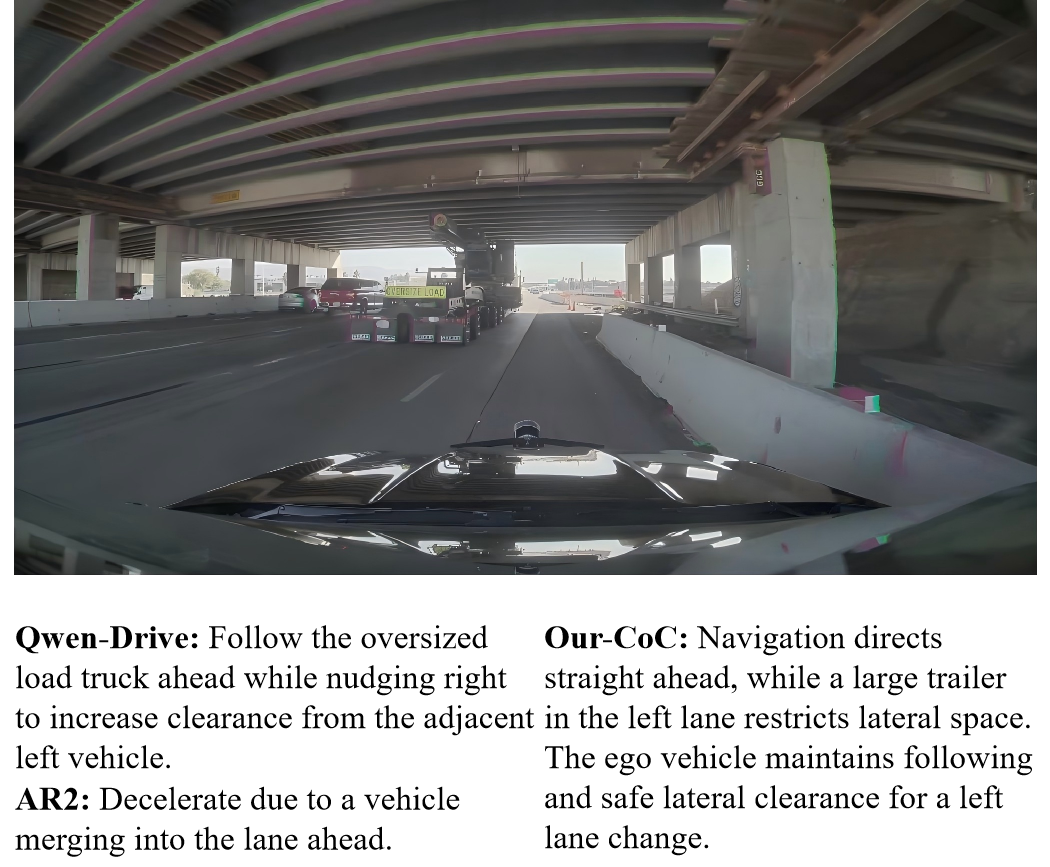}
  \end{subfigure}\hfill
  \begin{subfigure}[b]{0.49\textwidth}
    \centering
    \includegraphics[height=6.6cm]{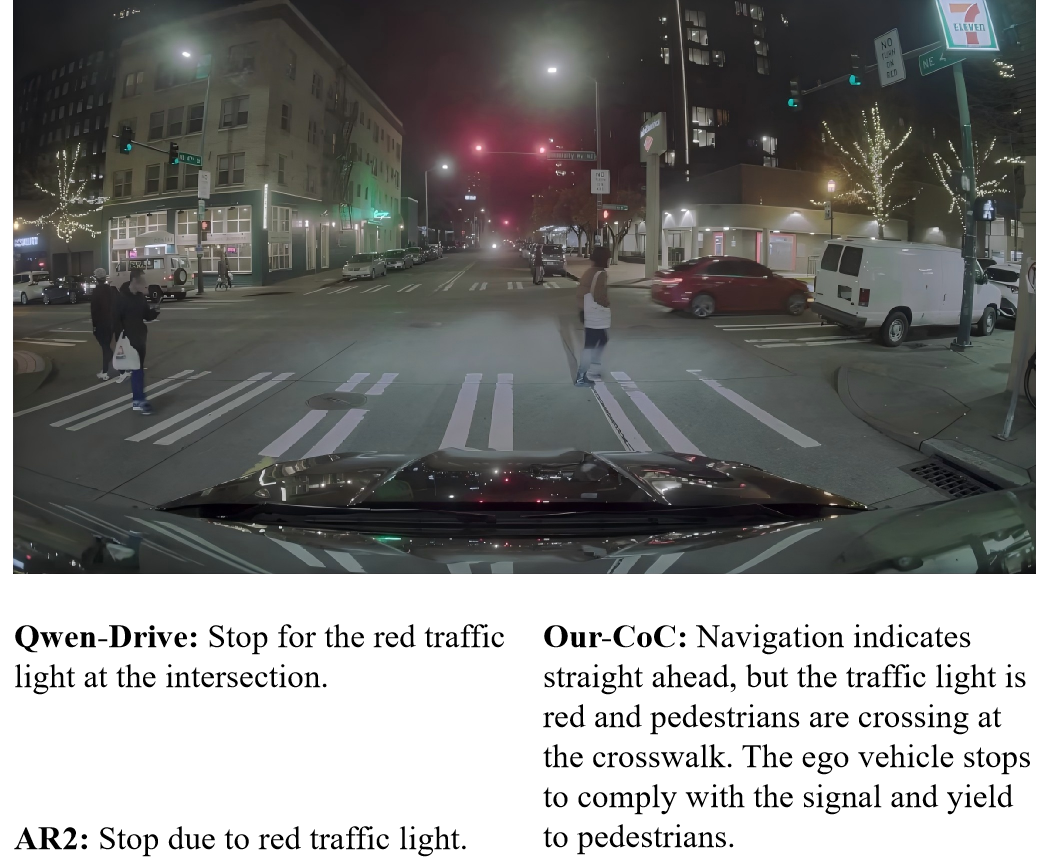}
  \end{subfigure}
  \caption{Two examples comparing the CoC outputs generated by the Qwen-Drive-1.0-SFT model, the Alpamayo 2 Super model, and our proposed method.}
  \label{fig:coc_three_compare_example}
\end{figure}

\paragraph{CoC output quality.}
To validate the quality of the CoC outputs produced by our proposed method, we manually curated 298 cases from the test split of the NVIDIA PhysicalAI-Autonomous-Vehicles dataset, all of which are critical scenarios involving interaction with the ego vehicle. We then generated CoC outputs using the Qwen-Drive-1.0-SFT model, the Alpamayo 2 Super model, and our proposed method, after which human annotators were asked to vote without knowing which model produced each CoC. The voting rule was that a CoC statement could only receive a vote if it simultaneously satisfied the evaluation criteria across the following four dimensions: correct identification of key factors, correct recall and correctness of decisions, correct attribution of decisions, and absence of hallucination. The CoC pass rates are reported in Table~\ref{tab:coc_comparison_three}, and two examples are shown in Figure~\ref{fig:coc_three_compare_example}.

\begin{table}[htbp]
\centering
\small
\caption{Human evaluation comparison of CoC outputs generated by the Qwen-Drive-1.0-SFT model, the Alpamayo 2 Super model, and our proposed method.}
\label{tab:coc_comparison_three}
\setlength{\tabcolsep}{8pt}
\renewcommand{\arraystretch}{1.15}
\begin{tabular}{c|cc>{\columncolor[HTML]{FFE0CC}}c}
\toprule
\textbf{Model} & \textbf{Qwen-Drive-1.0-SFT} & \textbf{Alpamayo 2 Super} & \textbf{\Model{}} \\
\midrule
\textbf{CoC pass rate}  & 68.1\% & 78.5\% & \textbf{84.2\%} \\
\bottomrule
\end{tabular}
\end{table}

The structured CoC output reaches a CoC pass rate of 84.2\%, compared with 78.5\% for Alpamayo 2 Super and 68.1\% for Qwen-Drive-1.0-SFT. This gap indicates that explicitly representing factors, decisions, and their causal attribution improves the completeness and factual consistency of decision explanations.

\begin{figure}[!htp]
  \centering
  \begin{subfigure}[b]{1.0\textwidth}
    \centering
    \includegraphics[height=4cm]{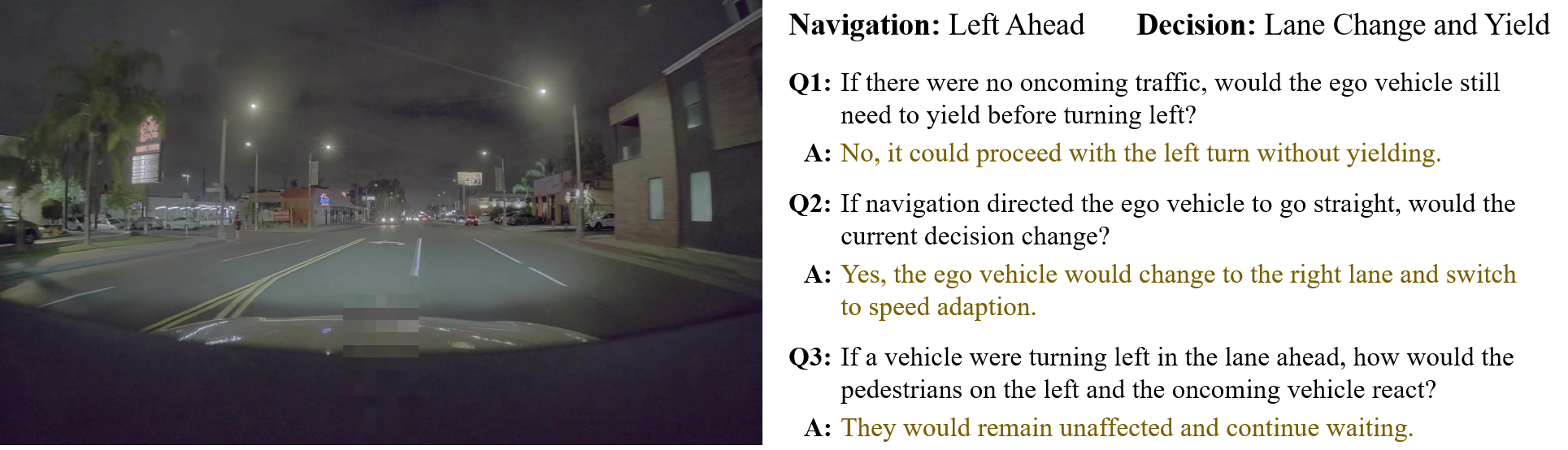}
    \vspace{1em}
  \end{subfigure}
  \begin{subfigure}[b]{1.0\textwidth}
    \centering
    \includegraphics[height=4cm]{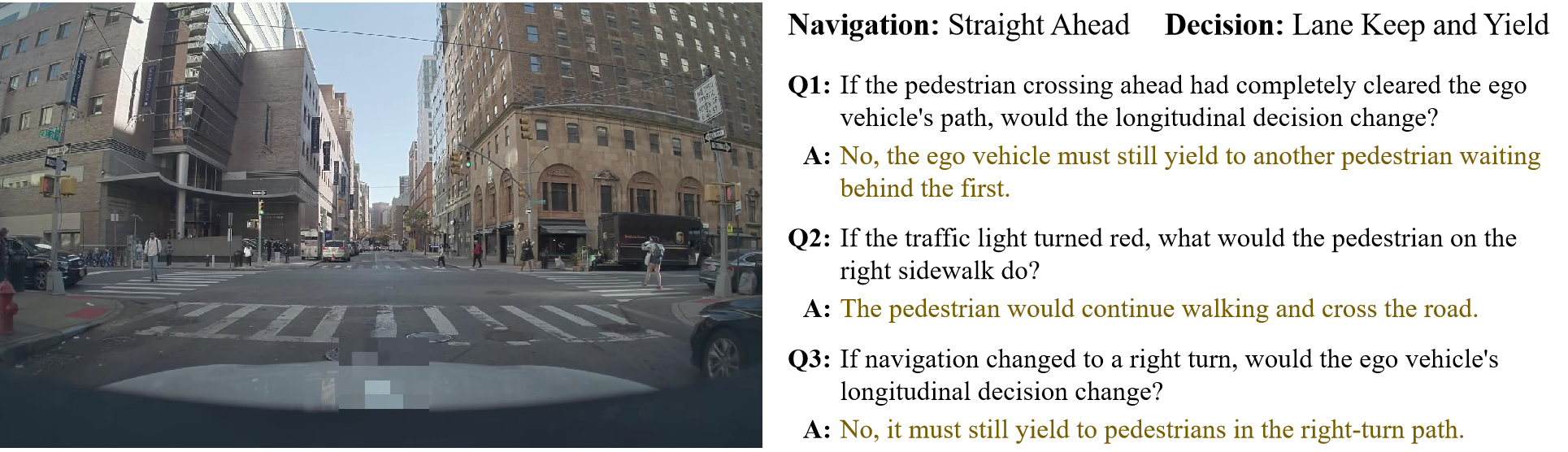}
  \end{subfigure}
  \caption{Two examples of model predictions for counterfactual reasoning.}
    \label{fig:two_examples_counterfactual_qa}
\end{figure}

\paragraph{Counterfactual reasoning} We also evaluate counterfactual reasoning on keyframes from
the same test split. For each keyframe, we first use Qwen3.8-27B to generate counterfactual QA pairs, and then manually correct erroneous questions and answers to obtain a counterfactual QA test set. We then evaluate our model's counterfactual reasoning capability using the following pipeline: for each counterfactual question, the model predicts a corresponding answer; the predicted answer and the human-corrected ground-truth answer are then fed together into Qwen3.8-27B for scoring. \Model{} obtains a judge score of 61.1 out of 100.~\Cref{fig:two_examples_counterfactual_qa} presents two examples of counterfactual reasoning results.

\begin{table}[htbp]
\centering
\small
\caption{Evaluation metrics of our proposed model for identifying key factors and producing longitudinal and lateral decisions on keyframes.}
\label{tab:f1_key_factor_decision}
\setlength{\tabcolsep}{8pt}
\renewcommand{\arraystretch}{1.15}
\begin{tabular}{c|cc}
\toprule
\rowcolor[HTML]{FFE0CC}
\textbf{Causation} & \textbf{Accuracy} & \textbf{F1} \\
\midrule
\textbf{Key factors}  & 75.2\% & 72.5\% \\
\textbf{Longitudinal decision}  & 81.5\% & 73.6\% \\
\textbf{Lateral decision}  & 72.1\% & 71.8\% \\
\bottomrule
\end{tabular}
\end{table}

\begin{figure}[!htp]
  \centering
  \begin{subfigure}[b]{0.5\textwidth}
    \centering
    \includegraphics[height=5.0cm]{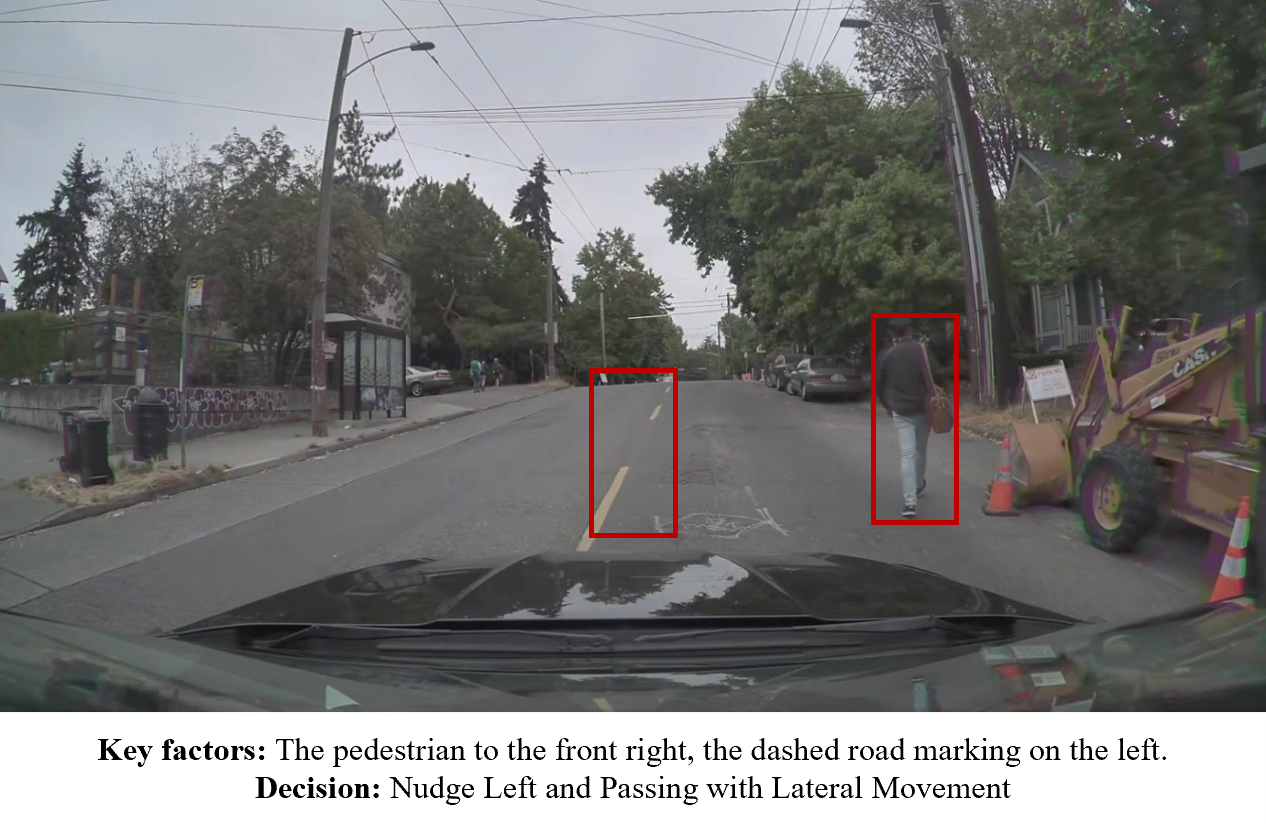}
  \end{subfigure}\hfill
  \begin{subfigure}[b]{0.5\textwidth}
    \centering
    \includegraphics[height=5.0cm]{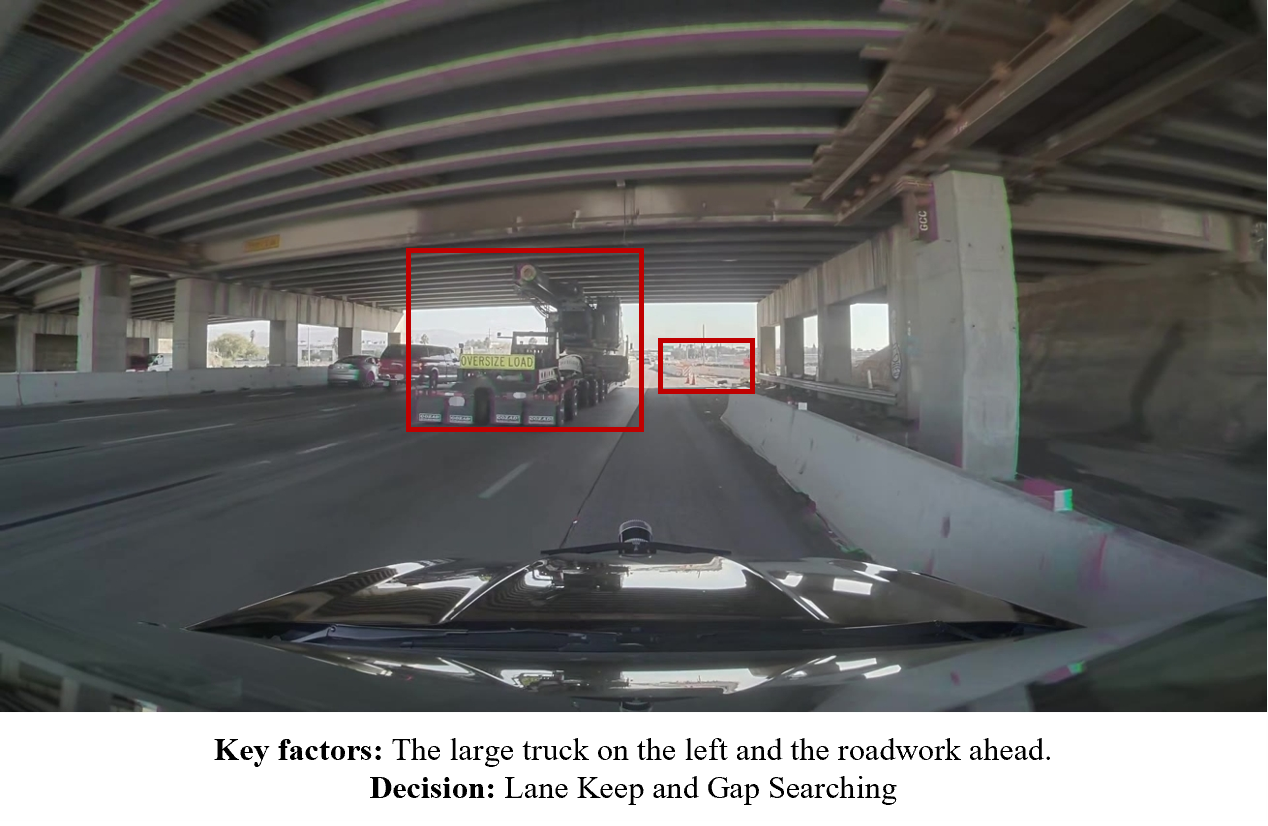}
  \end{subfigure}
  \caption{Two examples of model predictions for key factors and decisions.}
  \label{fig:two_examples_key_factor_decsion}
\end{figure}

\paragraph{Key factors, decisions, and trajectories.}
We evaluate whether the predicted evidence is translated into the correct discrete decisions and then into executable trajectories. We select 1,000 high-quality keyframes from the test split of NVIDIA's PhysicalAI-Autonomous-Vehicles dataset and manually annotate their key factors and longitudinal/lateral decisions. Table~\ref{tab:f1_key_factor_decision} reports the corresponding accuracy and F1 scores. Key-factor identification obtains 75.2\% accuracy and 72.5\% F1, while longitudinal decisions reach 81.5\% accuracy and 73.6\% F1. Lateral decisions are lower at 72.1\% accuracy and 71.8\% F1, indicating that lateral interactions remain the more difficult semantic decision class. Two cases are shown in Figure~\ref{fig:two_examples_key_factor_decsion}.
Trajectory prediction is evaluated separately using ADE@1 and
minADE@6 at 3~s and 5~s in
Table~\ref{tab:physicalai_comparison}. Closed-loop driving
performance is reported in
Table~\ref{tab:physicalai_closed_loop}, while per-class
meta-action recall is analyzed in
Figure~\ref{fig:meta_action_ablation}.


\subsection{Main Results}
Tables~\ref{tab:physicalai_comparison} and~\ref{tab:physicalai_closed_loop} report the open-loop trajectory and closed-loop driving results on PhysicalAI. The semantic decision and causal-reasoning measurements are reported separately above, allowing the geometric and closed-loop effects of the shared representation to be examined independently.

\begin{table*}[ht]
\centering
\small
\caption{Open-loop trajectory prediction results on PhysicalAI--Autonomous--Vehicles.
The high-quality test set is curated exclusively from the PhysicalAI test split, with high-quality, challenging cases selected manually through rigorous screening for evaluation.}
\label{tab:physicalai_comparison}
\resizebox{\textwidth}{!}{%
\begin{tabular}{l|c|llllllll}
\toprule
\textbf{Method}
& \makecell{\textbf{Params.}\\\textbf{(B)}}
& \multicolumn{4}{c}{\textbf{Physical-AI NV split}}
& \multicolumn{4}{c}{\textbf{High-quality test set}} \\
\cmidrule(lr){3-6}
\cmidrule(lr){7-10}
& &
  \multicolumn{2}{c}{\textbf{minADE@6$\,\downarrow$}}
& \multicolumn{2}{c}{\textbf{ADE@1$\,\downarrow$}}
& \multicolumn{2}{c}{\textbf{minADE@6$\,\downarrow$}}
& \multicolumn{2}{c}{\textbf{ADE@1$\,\downarrow$}} \\
\cmidrule(lr){3-4}
\cmidrule(lr){5-6}
\cmidrule(lr){7-8}
\cmidrule(lr){9-10}
& &
  3 s & 5 s
& 3 s & 5 s
& 3 s & 5 s
& 3 s & 5 s \\
\midrule
Qwen-Drive-1.0-SFT w / o reasoning
& 5.0
& 0.62 & 1.72
& 0.65 & 1.82
& 0.64 & 1.78
& 0.67 & 1.86 \\
Qwen-Drive-1.0-SFT w / reasoning
& 5.0
& 0.61 & 1.70
& 0.64 & 1.80
& 0.64 & 1.77
& 0.66 & 1.84 \\
Qwen-Drive-1.0-SFT-RL
& 5.0
& 0.64 & 1.67
& 0.68 & 1.77
& 0.67 & 1.80
& 0.71 & 1.89 \\
Alpamayo 1.5
& 9.8
& \underline{0.20} & \textbf{0.55}
& \underline{0.51} & \underline{1.34}
& \underline{0.24} & \underline{0.64}
& 0.62 & 1.60 \\
Alpamayo 2 Super
& 34.0
& 0.21 & \textbf{0.55}
& 0.52 & 1.35
& \textbf{0.23} & 0.65
& 0.60 & 1.51 \\
\midrule
\rowcolor[HTML]{FFE0CC}
\textbf{\Model{}-0.8B}
& 0.8
& 0.21 & 0.60
& 0.54 & 1.40
& 0.30 & 0.80
& \underline{0.53} & \underline{1.47} \\
\rowcolor[HTML]{FFE0CC}
\textbf{\Model{}-9B}
& 9.0
& \textbf{0.18} & \underline{0.56}
& \textbf{0.29} & \textbf{0.71}
& 0.25 & \textbf{0.43}
& \textbf{0.36} & \textbf{0.83} \\
\bottomrule
\end{tabular}%
}
\end{table*}

\paragraph{Open-loop planning.}
Table~\ref{tab:physicalai_comparison} evaluates open-loop planning on the standard Physical-AI NV split and the leakage-free high-quality split. We report minADE@6 for coverage of the six predicted candidates and ADE@1 for the selected trajectory, both at 3~s and 5~s. The two splits separate performance on the benchmark construction from performance on held-out, challenging clips.

The \Model{}-9B model delivers the strongest overall open-loop performance. It achieves the lowest ADE@1 across both splits and horizons and leads minADE@6 in most settings, including 0.18~m at 3~s on the NV split and 0.43~m at 5~s on the high-quality split. AR models retain only narrow advantages in the two remaining minADE@6 cases, so the main gain is consistent accuracy of the selected trajectory together with strong long-horizon candidate coverage rather than uniform dominance at every horizon. This result is obtained with 9B parameters, compared with 34B for AR2. The 0.8B variant remains competitive on ADE@1 but is weaker on minADE@6, showing that single-trajectory accuracy alone does not guarantee reliable candidate coverage.

\paragraph{Closed-loop driving.}
Table~\ref{tab:physicalai_closed_loop} evaluates the same models in 1,607 AlpaSim scenarios. The 9B model obtains the lowest close-encounter rate (19.1\%), at-fault rate (5.4\%), and off-road rate (9.8\%) among the listed methods, improving all three safety measures over AR1.5 and AR2. Its progress (62.6\%) and AlpaSim scores (0.38/0.68) are below AR2 (64.3\% and 0.44/0.75) but above AR1.5 (59.0\% and 0.32/0.50). Thus, the 9B configuration trades a small amount of progress and score for a lower observed safety risk, while remaining more favorable than AR1.5 on every reported closed-loop measure.

\begin{table*}[t]
\centering
\small
\caption{Closed-loop comparison on PhysicalAI--Autonomous--Vehicles.
Reference results use 1607 AlpaSim scenarios with PAI-AV-NuRec
version 26.04. Params. excludes LLM token embeddings and is
reported in billions.}
\label{tab:physicalai_closed_loop}
\resizebox{\textwidth}{!}{

\begin{tabular}{l|c|cccccc}
\toprule
\textbf{Method}
& \makecell{\textbf{Params.}\\\textbf{(B)}}
& \multicolumn{2}{c}{\textbf{Close Encounter Rate (\%)$\,\downarrow$}}
& \makecell{\textbf{Off-Road}\\\textbf{Rate (\%)$\,\downarrow$}}
& \makecell{\textbf{Progress}\\\textbf{(\%)$\,\uparrow$}}
& \multicolumn{2}{c}{\textbf{AlpaSim Score (km)$\,\uparrow$}} \\
\cmidrule{3-4}
\cmidrule{7-8}
& & All & At-fault & & & All & At-fault \\
\midrule
Qwen-Drive-1.0-SFT w/ reasoning
& 5.0 & 38.0 & 12.0 & 24.0 & 54.0 & 0.16 & 0.27 \\
Qwen-Drive-1.0-RL
& 5.0 & 41.0 & 11.0 & 12.0 & 48.0 & 0.16 & 0.37 \\
Alpamayo 1.5
& 9.8 & \underline{20.7} & 5.9 & 19.9 & 59.0 & 0.32 & 0.50 \\
Alpamayo 2 Super
& 34.0 & 21.9 & \underline{5.7} & \underline{10.3} & \textbf{64.3} & \textbf{0.44} & \textbf{0.75} \\
\midrule
\rowcolor[HTML]{FFE0CC}
\textbf{\Model{}-0.8B}
& 0.8 & 23.7 & 6.1 & 12.7 & 58.0 & 0.28 & 0.49 \\
\rowcolor[HTML]{FFE0CC}
\textbf{\Model{}-9B}
& 9.0 &\textbf{ 19.1 }& \textbf{5.4} & \textbf{9.8} & \underline{62.6} & \underline{0.38} & \underline{0.68} \\
\bottomrule
\end{tabular}
}
\end{table*}

\subsection{Ablations and Dynamic Reasoning Analysis}
We evaluate the contribution of each cognitive component and the behavior of the adaptive inference policy through controlled ablations.
Keeping the shared backbone architecture, data splits, and
Action decoder architecture fixed, we progressively add
Perception, Causation, Value, and World modules to the Action
baseline. Each added module is trained with its corresponding
objective and participates in inference. We evaluate these
module configurations and compare System~1 with adaptive
inference under a common evaluation protocol.

\paragraph{Adaptive computation ablation.}
Table~\ref{tab:dynamic_template} isolates the effect of joint training and adaptive mode under the same runtime and candidate-scoring configuration. The System~1 baseline uses the Action--Value path described in
Table~\ref{tab:exc_path}.
Adaptive inference selects among supported task and latent-budget
configurations for Perception, Causation, and World according
to the current decision.
With the same System~1 inference path and token consumption,
joint training reduces 3/5~s minADE@6 from 0.66/1.93~m to
0.35/0.89~m. Adaptive inference further reduces these errors
to 0.30/0.80~m and the close-encounter rate from 30.8\% to
23.7\%. It approaches the performance of the evaluated
full-reasoning configuration while using 28\% of its token
consumption.

\begin{table}[ht]
\centering
\small
\caption{PhysicalAI quality--compute ablation within \Model{}-0.8B. Metrics follow the evaluation protocol described in Section~\ref{sec:eval_protocol}. All policies use the same planning scorer.}
\resizebox{\columnwidth}{!}{%
\begin{tabular}{cccccccc}
\toprule
\multicolumn{3}{c}{} & \multicolumn{2}{c}{\textbf{Open-Loop}} & \multicolumn{3}{c}{\textbf{Closed-Loop (AlpaSim)}} \\
\cmidrule(lr){4-5}\cmidrule(lr){6-8}
\makecell{\textbf{Model}\\\textbf{configuration}} & \makecell{\textbf{Inference}\\\textbf{mode}} & \makecell{\textbf{Token}\\\textbf{Consumption}} & \makecell{\textbf{minADE@6}$\downarrow$ \\3s} & \makecell{\textbf{minADE@6}$\downarrow$ \\5s} & \makecell{\textbf{Close Encounter Rate}\\$\downarrow$ (\%)} & \makecell{\textbf{Off-Road Rate}\\$\downarrow$ (\%)} & \makecell{\textbf{AlpaSim Score}\\(km)$\uparrow$} \\
\midrule
System~1 & -- & 13\% & 0.66 & 1.93 & 43.6 & 23.9 & 0.10 \\
Joint training & System1 & 13\% & 0.35 & 0.89 & 30.8 & 19.7 & 0.18 \\
Joint training & Full reasoning & 100\% & \textbf{0.29} & 0.83 & \textbf{23.5} & \textbf{12.5} & \textbf{0.30} \\
\rowcolor[HTML]{FFE0CC}
\textbf{Joint training }& \textbf{Adaptive} & 28\% & 0.30 & \textbf{0.80} & 23.7 & 12.7 & 0.28 \\
\bottomrule
\end{tabular}
}
\label{tab:dynamic_template}
\end{table}


\begin{table}[ht]
\centering
\small
\caption{Modular ablation on PhysicalAI--Autonomous--Vehicles. Modules are added cumulatively to the Action baseline; \cmark{} and \xmark{} indicate enabled and disabled components. Planning is evaluated on the high-quality test set (2574 examples) using ADE@1 (m) at 3~s and 5~s and Collision Rate.}
\label{tab:modular_abl}
\setlength{\tabcolsep}{14pt}{
\begin{tabular}{cccccccc}
\toprule
\multicolumn{5}{c}{\textbf{Task components}} & \multicolumn{2}{c}{\makecell{\textbf{ADE@1 (m) $\downarrow$}\\\textbf{(High-quality Test)}}} & \makecell{\textbf{Collision Rate}\\(\%) $\downarrow$} \\
\cmidrule(lr){1-5}\cmidrule(lr){6-7}
Action & Perception & CoC & Value & World & 3 s & 5 s & \\
\midrule
\cmark & \xmark & \xmark & \xmark & \xmark & 0.66& 1.93 & 17.05  \\
\cmark & \cmark & \xmark & \xmark & \xmark & 0.60 & 1.61  & 16.38  \\
\cmark & \cmark & \cmark & \xmark & \xmark & 0.55 & 1.57  &16.07  \\
\cmark & \cmark & \cmark & \cmark & \xmark & \textbf{0.53} & \textbf{1.46} & 13.17  \\
\rowcolor[HTML]{FFE0CC}
\cmark & \cmark & \cmark & \cmark & \cmark & \textbf{0.53} & 1.47 & \textbf{11.40} \\
\bottomrule
\end{tabular}
}
\end{table}

\paragraph{Modular component ablation.}
Table~\ref{tab:modular_abl} cumulatively adds Perception,
Causation, Value, and World to the Action baseline.
ADE@1 decreases from 0.66/1.93~m to 0.53/1.47~m at 3/5~s,
while collision rate falls from 17.05\% to 11.40\%.
Adding Value produces the largest incremental reduction
in collision rate in this sequence, from 16.07\% to 13.17\%.
Adding World further lowers collision rate to 11.40\%,
with unchanged 3~s ADE@1 and a small increase in 5~s ADE@1.

\begin{table*}[ht]
  \centering
  \small
  \setlength{\tabcolsep}{0pt}
  \renewcommand{\arraystretch}{1.15}
  \caption{Model evaluation on two 200-clip adversarial scenarios. Each variant uses 20k new training samples with a 1:1 positive/negative ratio. We report decision F1 and clip pass rate for each scenario. A, P, and C denote Action, Perception, and Chain-of-Causation reasoning, respectively.}
  \label{table:model}
  \begin{tabularx}{\textwidth}{@{}>{\raggedright\arraybackslash}p{0.22\textwidth}>{\raggedright\arraybackslash}p{0.18\textwidth}*{4}{>{\centering\arraybackslash}X}@{}}
    \toprule
    \multirow{2}{*}{\makecell[l]{\textbf{Model}\\\textbf{variant}}} &
    \multirow{2}{*}{\makecell[l]{\textbf{Pipeline}\\\textbf{architecture}}} &
    \multicolumn{2}{c}{\makecell{\textbf{Scenario A: Advance lane }\\\textbf{change for roadwork}}} &
    \multicolumn{2}{c}{\makecell{\textbf{Scenario B: Stationary vehicle}\\\textbf{detection and detour}}} \\
    \cmidrule(lr){3-4}\cmidrule(lr){5-6}
    & & \makecell{\textbf{Decision}\\\textbf{F1}} & \makecell{\textbf{Clip pass}\\\textbf{rate}} & \makecell{\textbf{Decision}\\\textbf{F1}} & \makecell{\textbf{Clip pass}\\\textbf{rate}} \\
    \midrule
    \makecell[l]{Perception + Action\\(Baseline)} &
    \makecell[l]{P + A\\(without CoC or\\adaptive reasoning)} &
    68.5\% & 60.5\% & 57.0\% & 45.5\% \\
    \addlinespace
    \rowcolor[HTML]{FFE0CC}
    \makecell[l]{Perception + CoC\\+ Action (Ours)} &
    \makecell[l]{P $\rightarrow$ C $\rightarrow$ A\\(with causal chain)} &
    \textbf{89.0\%} & \textbf{84.5\%} & \textbf{82.5\%} & \textbf{73.5\%} \\
    \bottomrule
  \end{tabularx}
\end{table*}

\begin{figure}[!htp]
  \centering
  \begin{subfigure}[b]{0.5\textwidth}
    \centering
    \includegraphics[height=6cm]{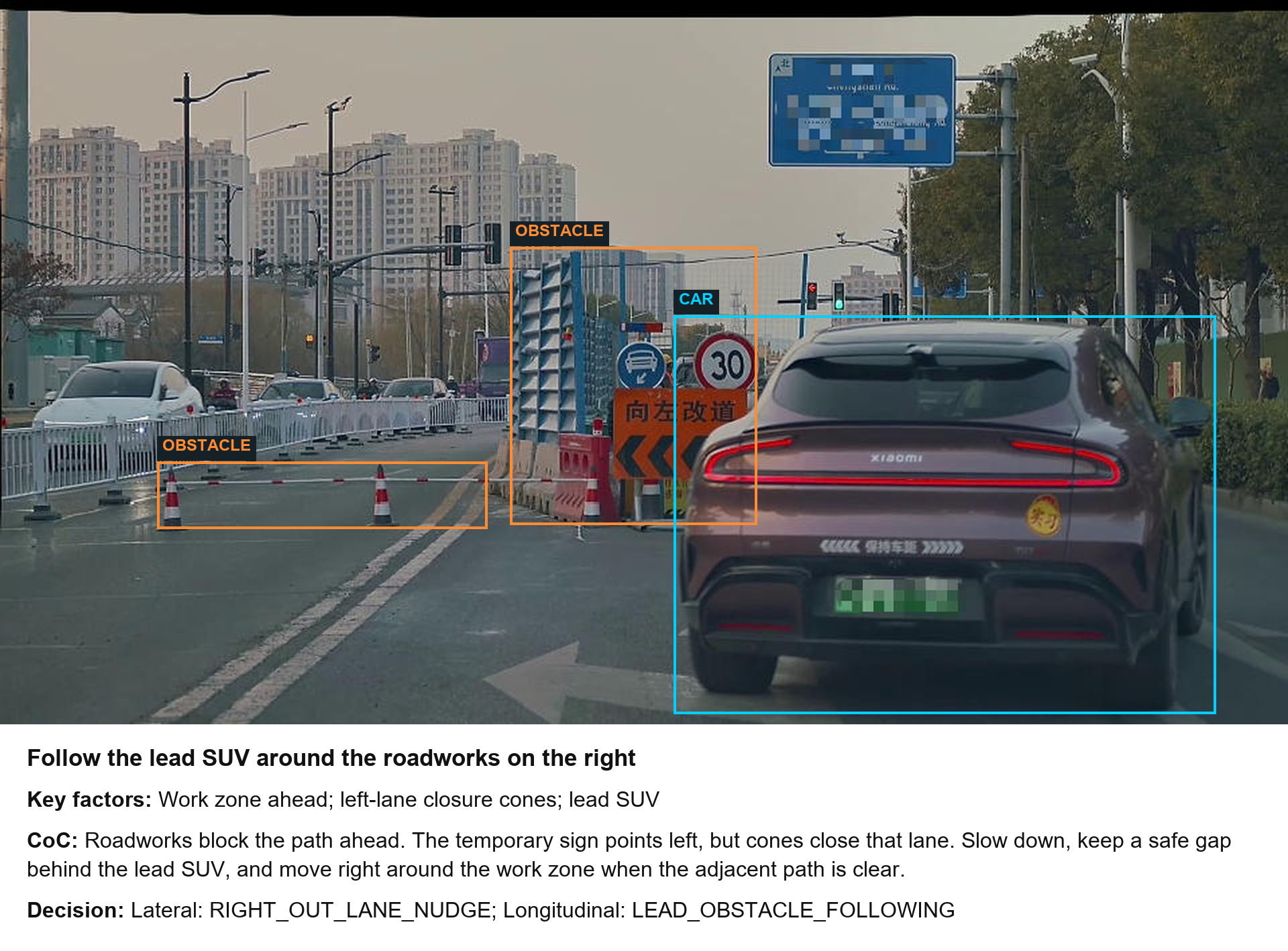}
    \caption{Positive case in a roadwork scenario.}
  \end{subfigure}\hfill
  \begin{subfigure}[b]{0.5\textwidth}
    \centering
    \includegraphics[height=6cm]{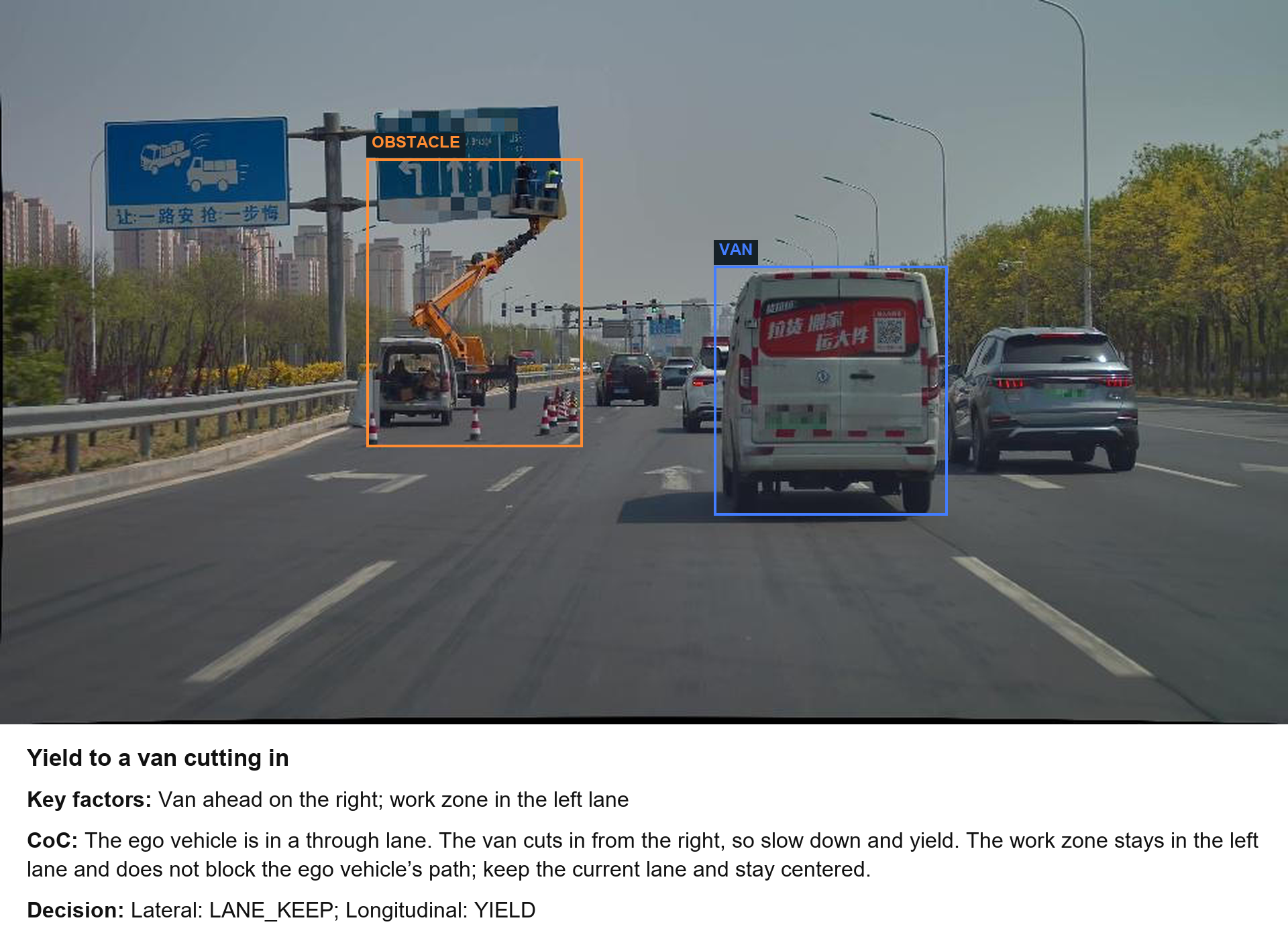}
    \caption{Similar negative case in a roadwork scenario.}
  \end{subfigure}

  \vspace{4pt}
  \begin{subfigure}[b]{0.5\textwidth}
    \centering
    \includegraphics[height=6cm]{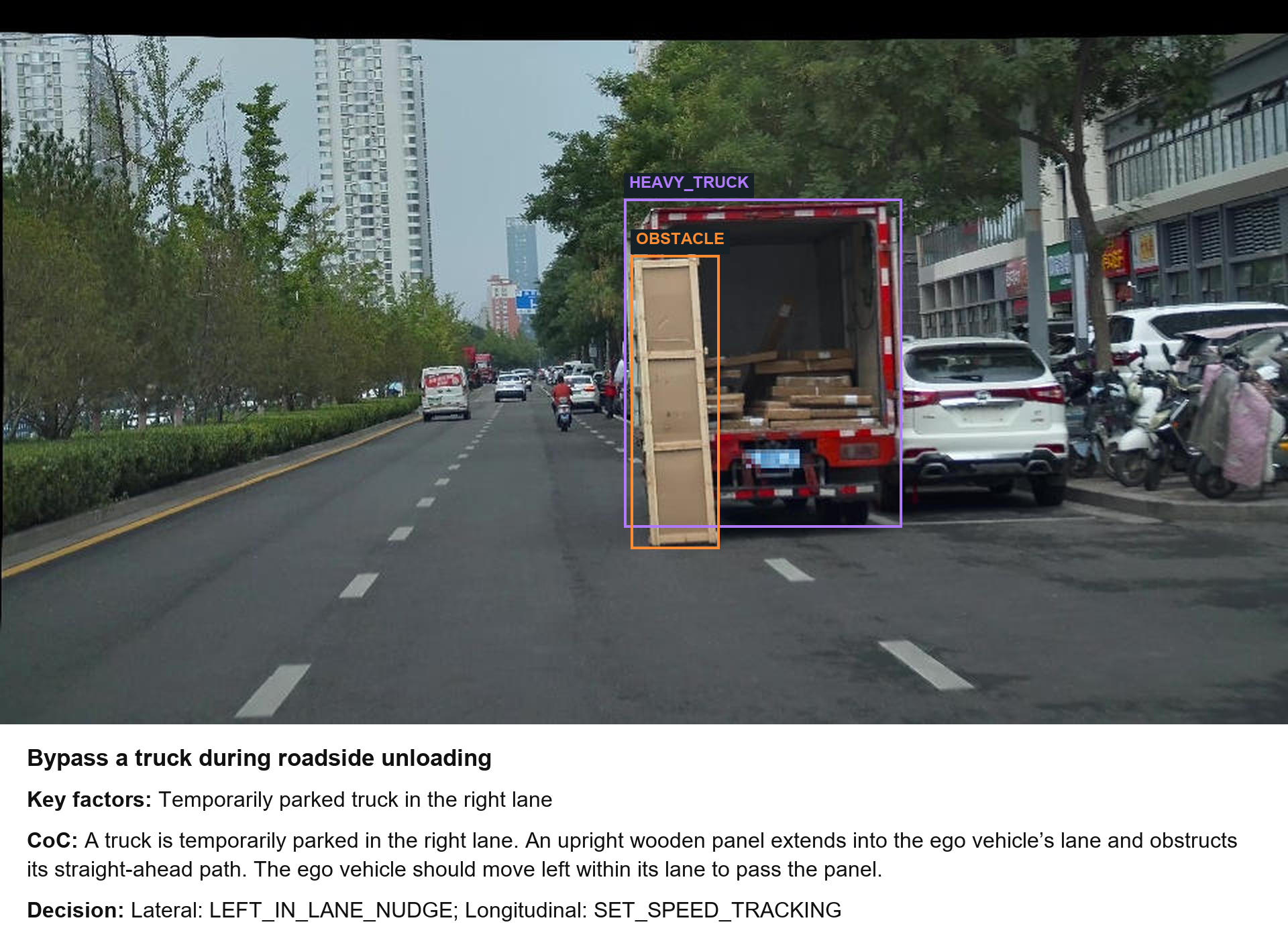}
    \caption{Positive case in a temporary-parking scenario.}
  \end{subfigure}\hfill
  \begin{subfigure}[b]{0.5\textwidth}
    \centering
    \includegraphics[height=6cm]{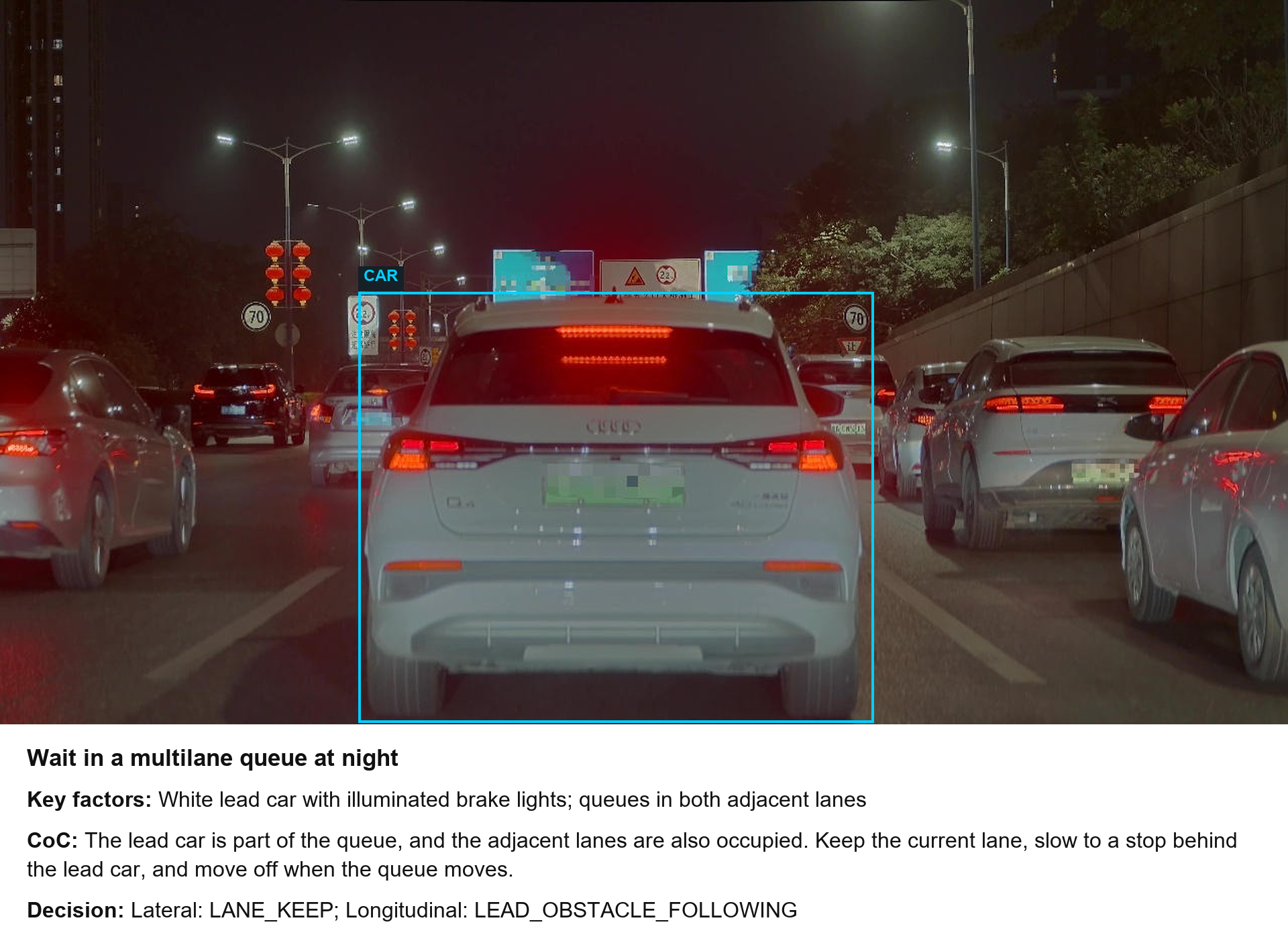}
    \caption{Similar negative case in a temporary-parking scenario.}
  \end{subfigure}
  \caption{Examples of positive and similar negative cases in construction and temporary-parking scenarios. For each case, the key factors, CoC, and decisions are shown.}
  \label{fig:example_pos_neg_roadwork_parking}
\end{figure}

\paragraph{Adversarial causal disambiguation.}
Table~\ref{table:model} isolates explicit causal reasoning with two 200-clip scenarios under a 1:1 mixture of visually similar positive--negative pairs, with examples shown in \Cref{fig:example_pos_neg_roadwork_parking}. The comparison tests whether the causal chain can identify lane-topology intrusion and the cause of a stationary obstacle before action selection, using decision F1 and clip pass rate as complementary measures.

\begin{itemize}
    \item \textbf{Causal disambiguation.}  Given only four frames over two seconds, the $A+P$ baseline often confuses a roadside barrier with lane-blocking roadwork or queued traffic with a disabled vehicle, yielding decision $F_1$ scores of 68.5\% on Scenario A and 57.0\% on Scenario B. The $P\rightarrow C\rightarrow A$ path uses Causation latents to represent whether an obstacle intrudes into the drivable lane
    and why it is stationary, then passes this information to Action through latent refill. Decision $F_1$ rises to 89.0\% and 82.5\% (+20.5 and +25.5 percentage points), while clip pass rates increase from 60.5\% to 84.5\% on Scenario A and from 45.5\% to 73.5\% on Scenario B.
    \item \textbf{Early detour commitment.} The CoC-conditioned path supports earlier detour commitment and helps avoid near-obstacle failures. The baseline often waits until the ego vehicle is close to a cone or stopped vehicle before changing lanes; on visually similar positive--negative pairs, this late response produces unstable decisions. In stationary-vehicle detours, the delayed maneuver can leave the end of the 6-second open-loop evaluation trajectory in a safety-risk region, so clip pass rate trails decision $F_1$. By identifying the obstacle's effect on lane feasibility earlier, CoC enables a timely maneuver.    
\end{itemize}

\begin{figure}[!htp]
\centering
\includegraphics[width=\linewidth]{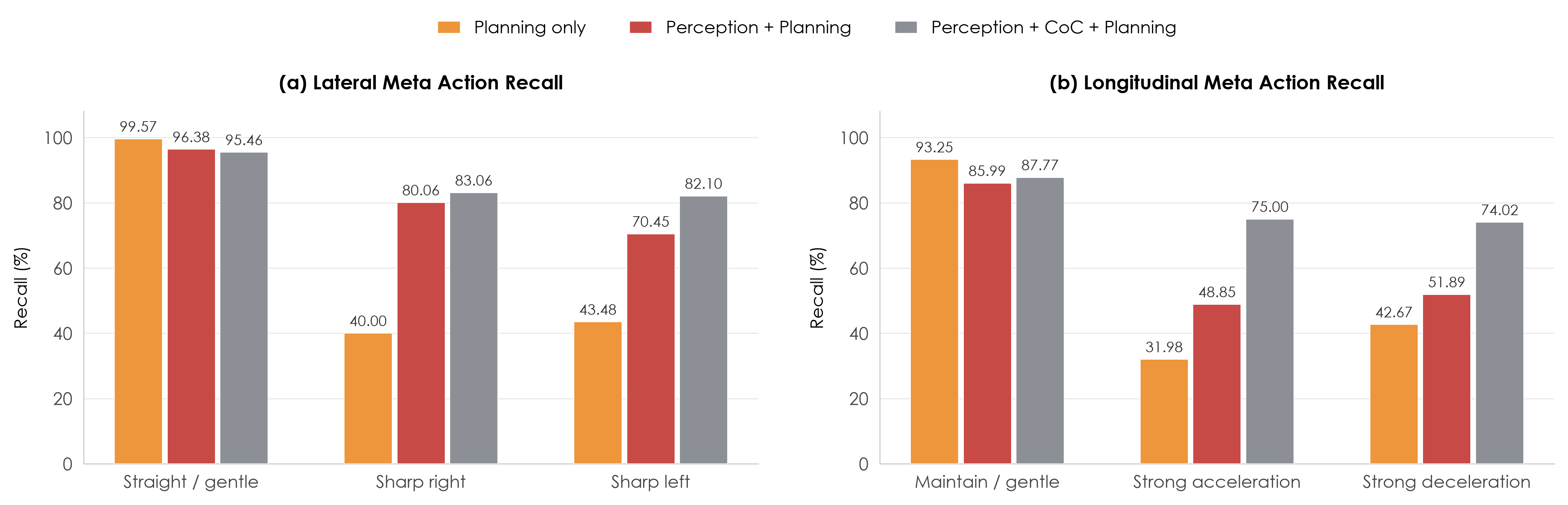}
\caption{meta-action recall in the \Model{}-0.8B ablation. We compare Action only, Perception + Action, and Perception + CoC + Action for lateral actions (a) and longitudinal actions (b); bars report per-class recall (\%).}
\label{fig:meta_action_ablation}
\end{figure}

\paragraph{meta-action ablation.}
We isolate the contribution of the Perception and Causation modules to discrete action prediction with the \Model{}-0.8B variant. Figure~\ref{fig:meta_action_ablation} compares an Action-only baseline with Perception + Action and Perception + CoC + Action, reporting per-class recall for lateral and longitudinal meta-actions.

The gains are concentrated on difficult maneuvers. For lateral actions, adding Perception raises recall for sharp-right decisions from 40.0\% to 80.1\%, while adding CoC further improves sharp-left recall from 70.5\% to 82.1\%. The same pattern appears for longitudinal actions: CoC increases recall for strong acceleration from 48.9\% to 75.0\% and for strong deceleration from 51.9\% to 74.0\%. Recall on straight or gentle actions changes little, indicating that causal supervision mainly resolves ambiguous maneuver conditions rather than improving already easy cases.

\begin{figure}[!htp]
  \centering
  \begin{subfigure}[b]{0.25\textwidth}
    \centering
    \includegraphics[height=3.5cm]{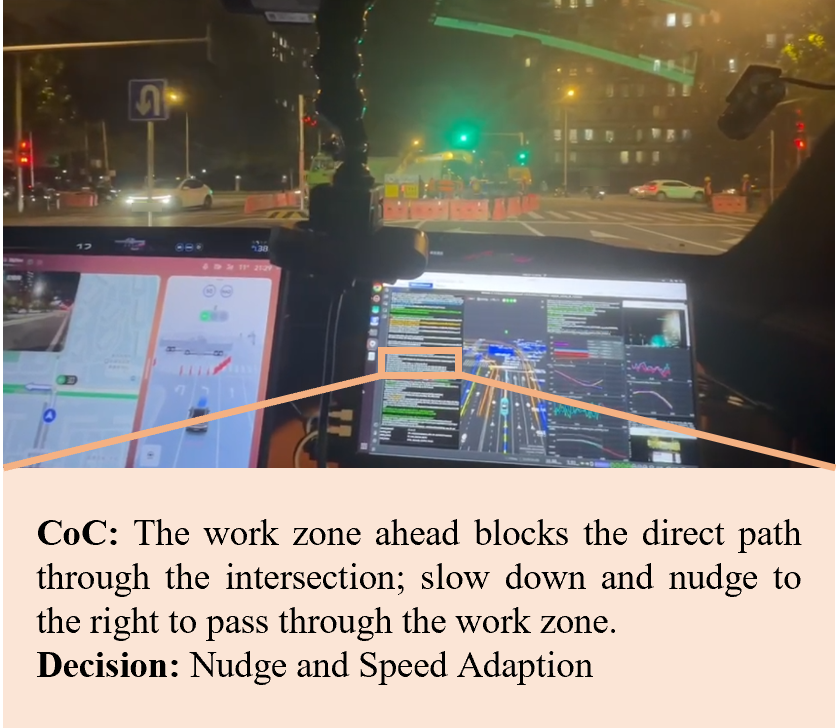}
  \end{subfigure}\hfill
  \begin{subfigure}[b]{0.25\textwidth}
    \centering
    \includegraphics[height=3.5cm]{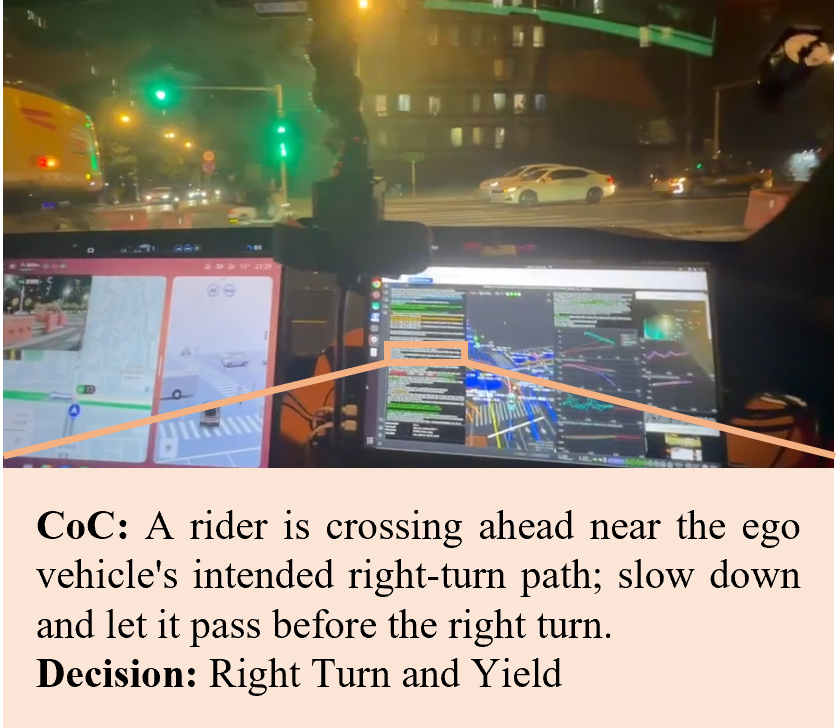}
  \end{subfigure}\hfill
  \begin{subfigure}[b]{0.25\textwidth}
    \centering
    \includegraphics[height=3.5cm]{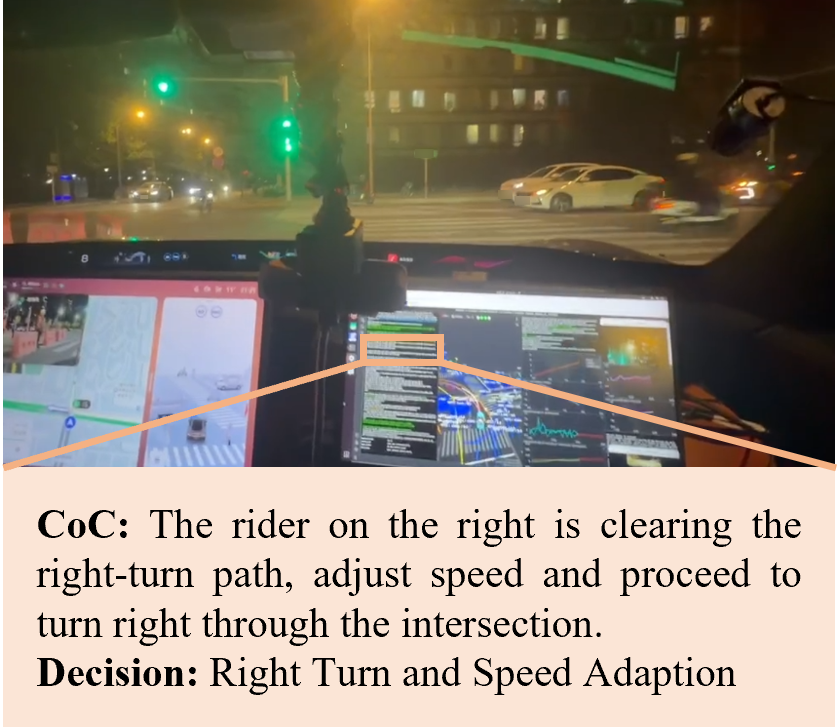}
  \end{subfigure}\hfill
  \begin{subfigure}[b]{0.25\textwidth}
    \centering
    \includegraphics[height=3.5cm]{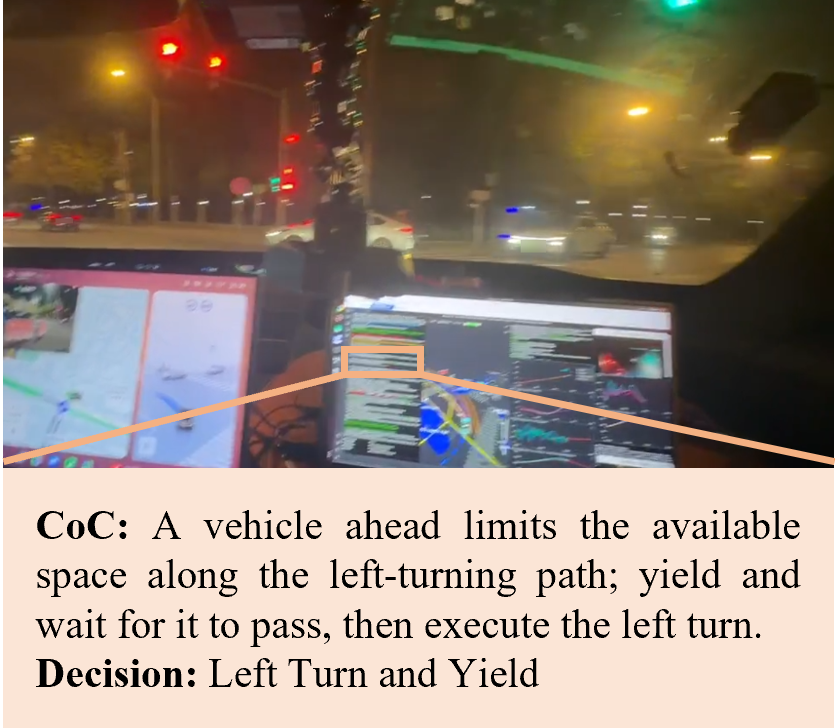}
  \end{subfigure}

  \vspace{4pt}
  \begin{subfigure}[b]{0.25\textwidth}
    \centering
    \includegraphics[height=3.5cm]{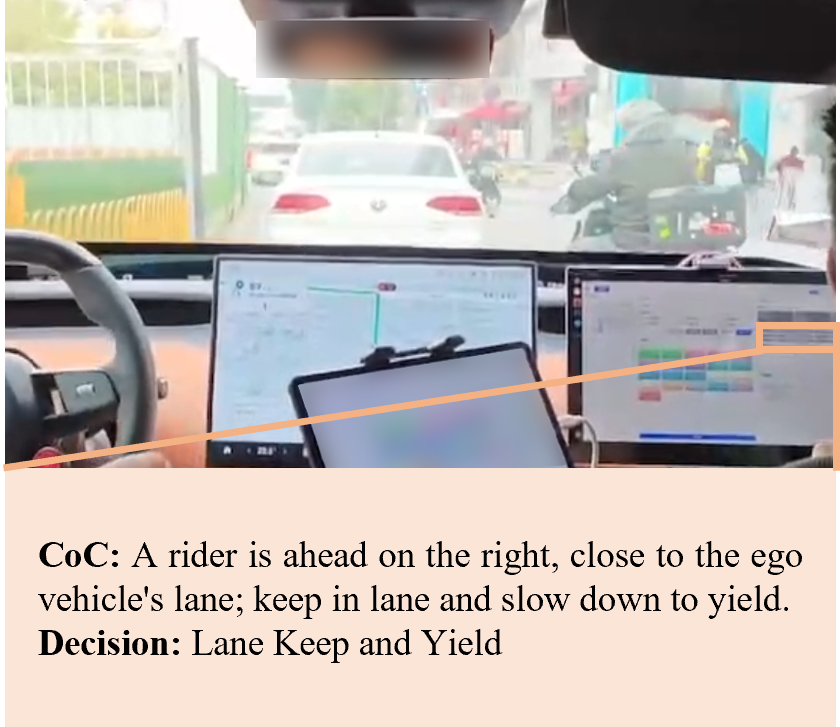}
  \end{subfigure}\hfill
  \begin{subfigure}[b]{0.25\textwidth}
    \centering
    \includegraphics[height=3.5cm]{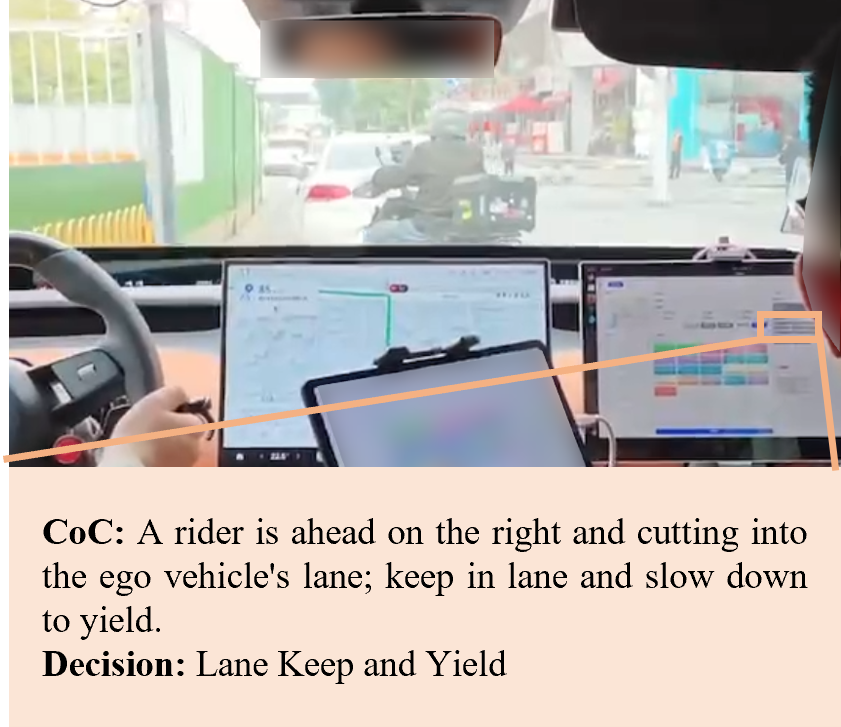}
  \end{subfigure}\hfill
  \begin{subfigure}[b]{0.25\textwidth}
    \centering
    \includegraphics[height=3.5cm]{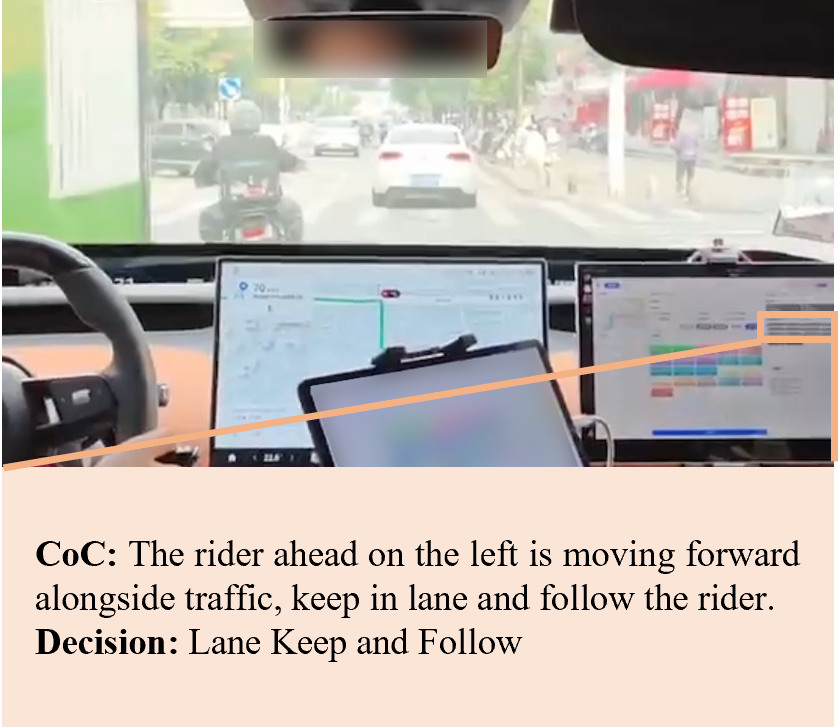}
  \end{subfigure}\hfill
  \begin{subfigure}[b]{0.25\textwidth}
    \centering
    \includegraphics[height=3.5cm]{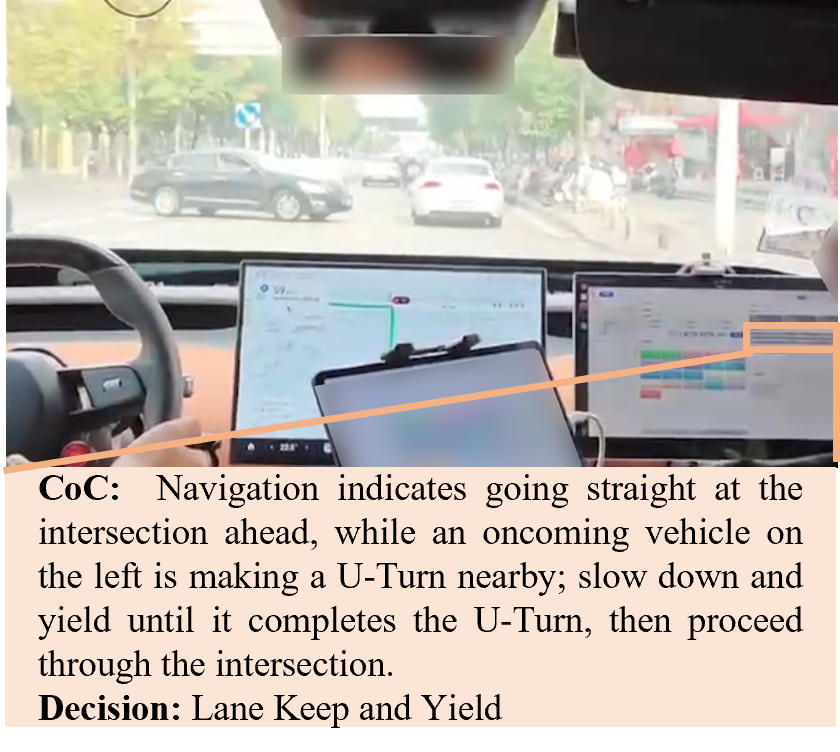}
  \end{subfigure}
  \caption{8 CoC examples from real-vehicle road test.}
  \label{fig:example_coc_in_real_vehicle}
\end{figure}

\subsection{Field Test}
\label{sec:roadtest}
We assess runtime feasibility through real-vehicle road tests
using the NVIDIA RTX 4090. The deployed system receives
synchronized multi-camera observations and ego state and
executes the task path selected by the inference scheduler.
The inference pipeline operates at 10~Hz in the tested
configuration. The in-vehicle CoC visualization results are shown in Figure~\ref{fig:example_coc_in_real_vehicle}.

%% file: sections/conclusion.tex
\section{Conclusion}
\label{sec:conclusion}

We introduced \textbf{\XFM{}}, a framework that connects learning
from critical driving experience with adaptive computation at
decision time. Its Data Engine organizes scene evidence,
factor--decision relations, candidate trajectories, and outcome
assessments into traceable scene--candidate records. Its model
engine, \textbf{\Model{}}, integrates Perception, Causation, Action,
World, and Value through a shared vision--language backbone.
Shared weights transfer learning across tasks and support
fast-path inference; candidate-conditioned latent refill carries
intermediate results between tasks; and a dedicated
value-of-computation head guides further reasoning. Verified
failures feed back into both annotation and driving-model updates,
connecting inference-time decisions with subsequent learning.
Public-benchmark comparisons and targeted ablations demonstrate
strong trajectory prediction and closed-loop driving performance,
with benefits from additional task modules, joint training, and
adaptive inference. Real-vehicle tests further demonstrate
on-vehicle runtime feasibility. These results support the
integration of scene understanding, candidate generation,
consequence prediction, and evaluation within a shared driving
model.

The broader contribution is an integrated technical foundation
for Xiaomi spanning data production, model learning, adaptive
inference, and vehicle validation. Its central principle is to
make critical experience reusable: failures can be examined in
terms of missing evidence, omitted candidate behaviors, or
inaccurate outcome assessments, allowing verified corrections
to inform subsequent annotation and training. This establishes
a concrete path toward learning more transferable decision
knowledge from each critical event and directing additional
computation to decisions where it is expected to improve the
final choice.

\paragraph{Limitations.}
Structured causal supervision depends on scene coverage and annotation  quality. The CoC and counterfactual evaluations measure agreement with
reviewed labels; controlled changes to scene factors are still needed
to assess whether the learned relationships support appropriate
closed-loop behavior under unfamiliar conditions. Learning from this
supervision also depends on the balance among task objectives. The
gradient-alignment analysis describes a local condition that can vary
across tasks, scenes, and training stages, leaving the trade-off between
planning improvement and auxiliary-task retention to be evaluated over
repeated training cycles.

At inference, candidate selection is limited by both behavioral coverage
and evaluation reliability. World receives visual supervision at 1, 2, and 3~s, whereas Action plans over 6.4~s, leaving longer-horizon consequences without direct World supervision. Shared representations may also correlate World and Value errors, while verification cannot
recover unproposed behaviors or correct every scene-interpretation error.
These limitations also affect computation allocation. Adaptive inference
operates over supported task and latent-budget configurations, and its
VOC estimates, learned from offline expansions of an SFT checkpoint,
may become inaccurate as the model, scene distribution, or execution
costs change. Detached scheduling inputs do not remove this shift, so scheduling records and allocation quality require reassessment across checkpoints and deployment platforms.

%% file: sections/contribution.tex
\newcommand{\envelopeicon}{\ding{41}}

\clearpage
\section*{\hypertarget{contributions-heading}{Contributions and Acknowledgments}}
\phantomsection
\addcontentsline{toc}{section}{Contributions and Acknowledgments}
\label{sec:contributions}

\noindent
\begin{minipage}[t]{0.47\linewidth}
  \subsubsection*{Core Contributors}
  \begin{itemize}[leftmargin=*, itemsep=0.15em, topsep=0.25em]
    \item Ziyang Yao
    \item Yuncheng Jiang
    \item Fang Li
    \item Wen Zhang
    \item Xiangyang Chu
    \item Zibin Guo
    \item Haochen Liu\textsuperscript{\envelopeicon}
    \item Shaoqing Xu\textsuperscript{\envelopeicon}    
  \end{itemize}
  \vspace{-1em}
\subsubsection*{Project Leader}
  \begin{itemize}[leftmargin=*, itemsep=0.15em, topsep=0.25em]
    \item Jingwei Zhao
  \end{itemize}
  \vspace{-1em}
    \subsubsection*{Project Advisors}
  \begin{itemize}[leftmargin=*, itemsep=0.15em, topsep=0.25em]
    \item Guang Chen
    \item Hangjun Ye
  \end{itemize}
\end{minipage}%
\hfill
\begin{minipage}[t]{0.47\linewidth}
  \subsubsection*{Contributors}
  \begin{itemize}[leftmargin=*, itemsep=0.15em, topsep=0.25em]
    \item Jiale Hou
    \item Chuang Wan
    \item Jia Hua
    \item Gengzhao Xiang
    \item Xiaoran Luo
    \item Chong He
    \item Yuechen Luo
    \item Junzhe Wang
    \item Zhiyu Sun
    \item Jun Zhang
    \item Wei Li
    \item Lei Li
    \item Hongwei Xie
    \item Bing Wang
    \item Yuannan Shen
    \item Jianwei Cui
  \end{itemize}
\end{minipage}

\vspace*{0.5em}
Technical details: 
\{~\underline{liuhaochen1}, ~\underline{xushaoqing}\}@xiaomi.com

\subsubsection*{Acknowledgments}
\noindent
We thank the Alpamayo team for their insightful advice on data processing and model deployment. We also express our gratitude to the development teams behind Qwen and the other open-source models, as well as the creators of the PhysicalAI-Autonomous-Vehicles datasets, benchmarks, and simulation environments utilized in this work. These resources were used exclusively for research and evaluation purposes.

%% file: sections/appendix.tex
\section*{\hypertarget{contributions-heading}{Appendix}}
\phantomsection
\addcontentsline{toc}{section}{Appendix}

\input{sections/related_work}
\input{sections/annotationV2}
\section{Supplementary Evaluation Details}
\label{sec:appendix}

\paragraph{CoC evaluation.}
Following the evaluation protocol in the main text, a CoC output
passes only if it correctly identifies key factors, covers the
relevant decisions correctly and completely, attributes those
decisions correctly, and contains no hallucinated claims.
Trajectory quality is evaluated separately from explanation quality.

\paragraph{Prediction horizons.}
Action generates a 6.4-second trajectory at 10~Hz.
World conditions on the first 3\,s of that trajectory, sampled
at 2\,Hz, and predicts visual tokens at future horizons of
1, 2, and 3\,s.
Its prediction therefore covers the near-term consequences
of the longer Action plan.

\section{Additional Qualitative Results}
\begin{figure}[!htp]
    \centering
    \includegraphics[width=\linewidth]{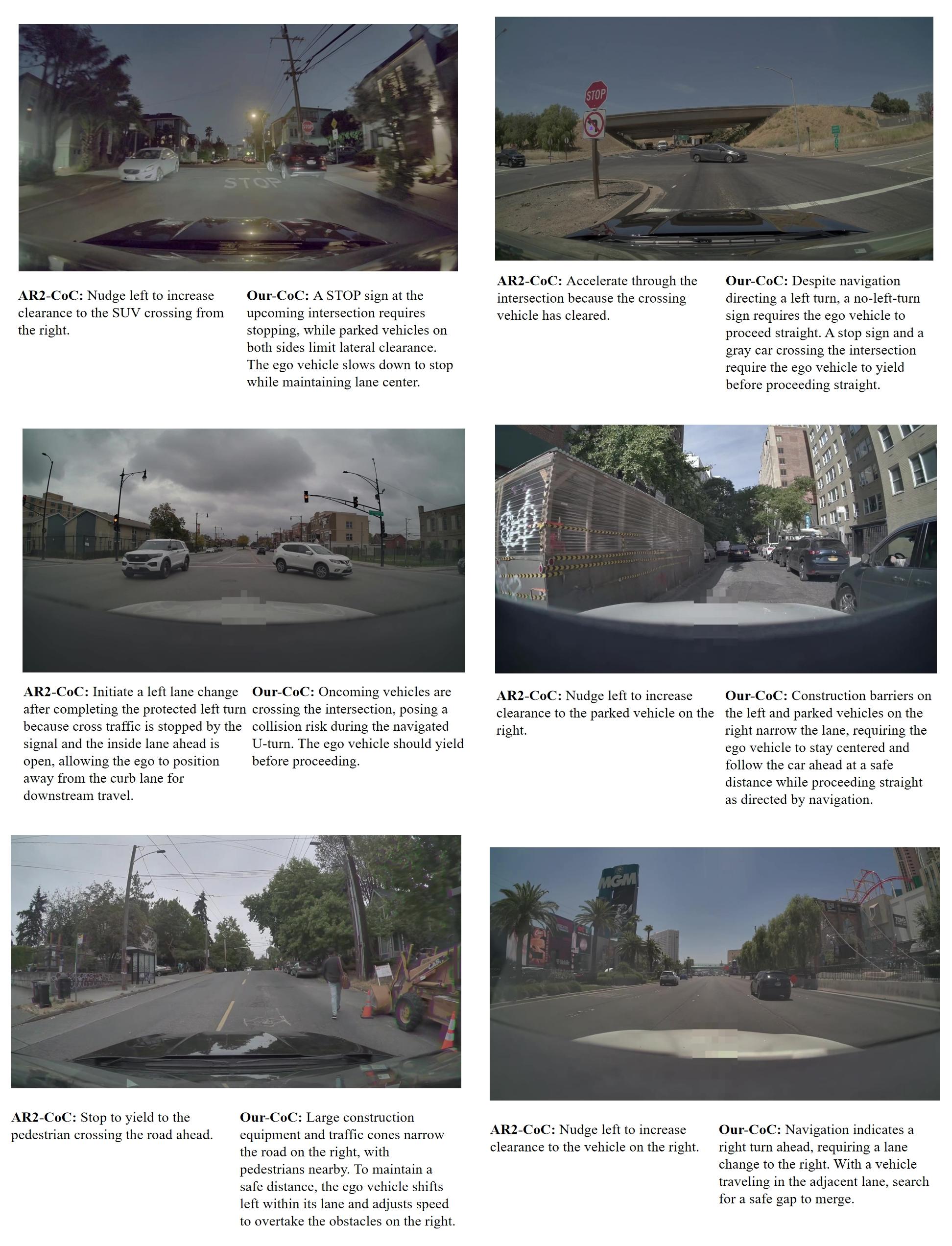}
    \caption{Cases of Our CoC compared with AR2's CoC}
    \label{fig:coc_case_study_0}
\end{figure}

  \begin{figure}[htbp]
    \centering
    \begin{minipage}{0.48\textwidth}
      \centering
      \includegraphics[width=\linewidth]{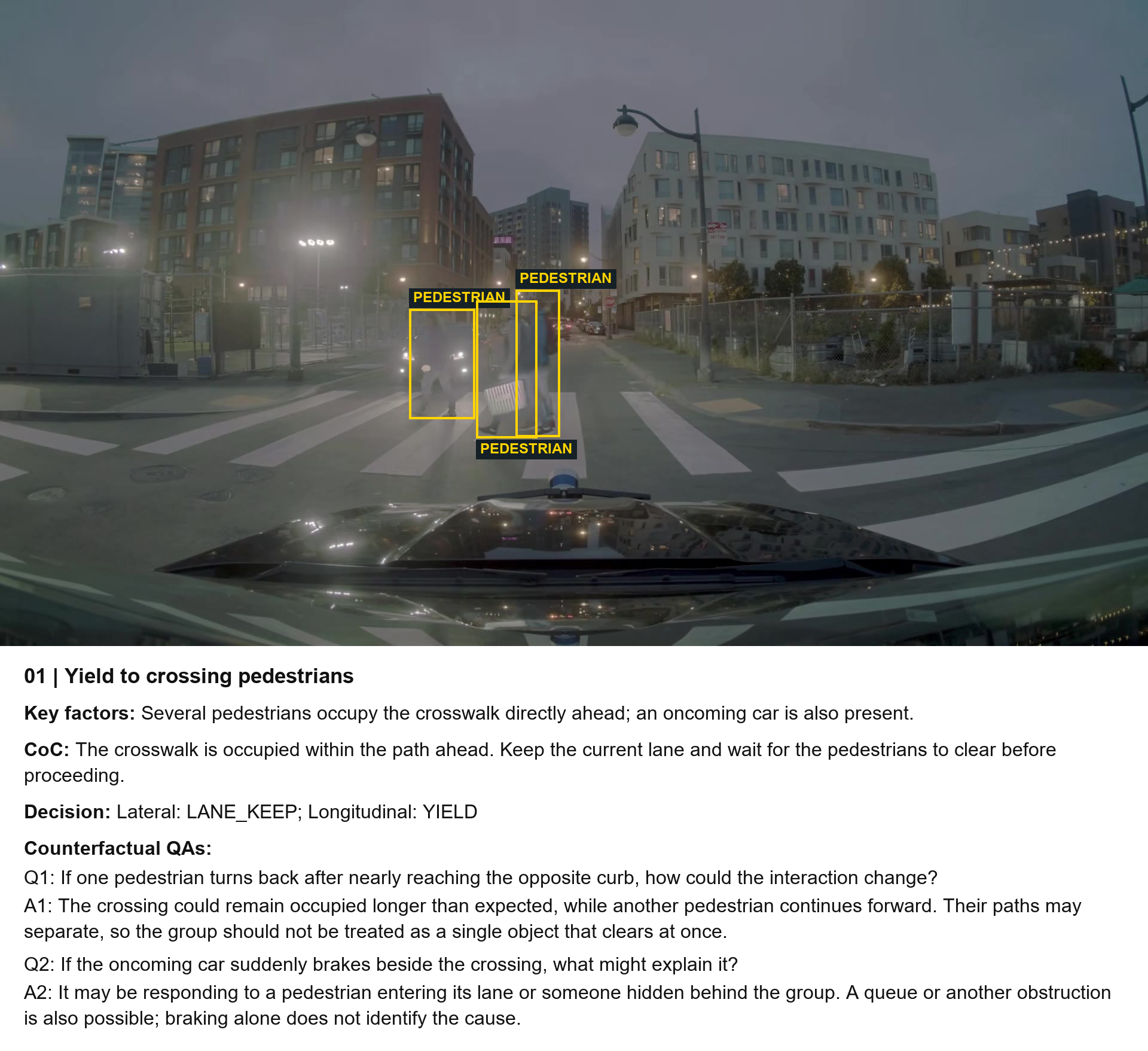}
    \end{minipage}\hfill
    \begin{minipage}{0.48\textwidth}
      \centering
      \includegraphics[width=\linewidth]{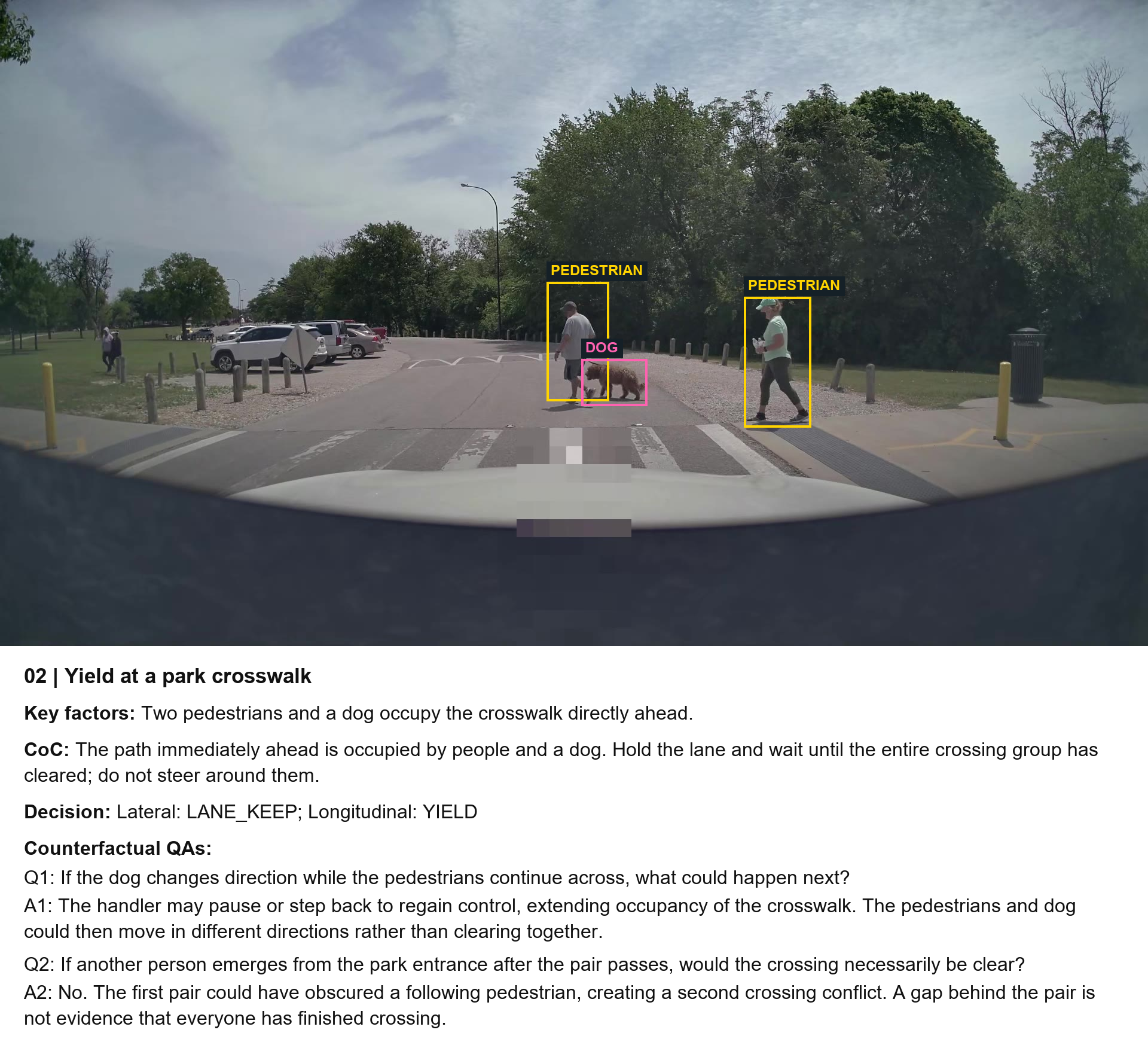}
    \end{minipage}\hfill
    \begin{minipage}{0.48\textwidth}
      \centering
      \includegraphics[width=\linewidth]{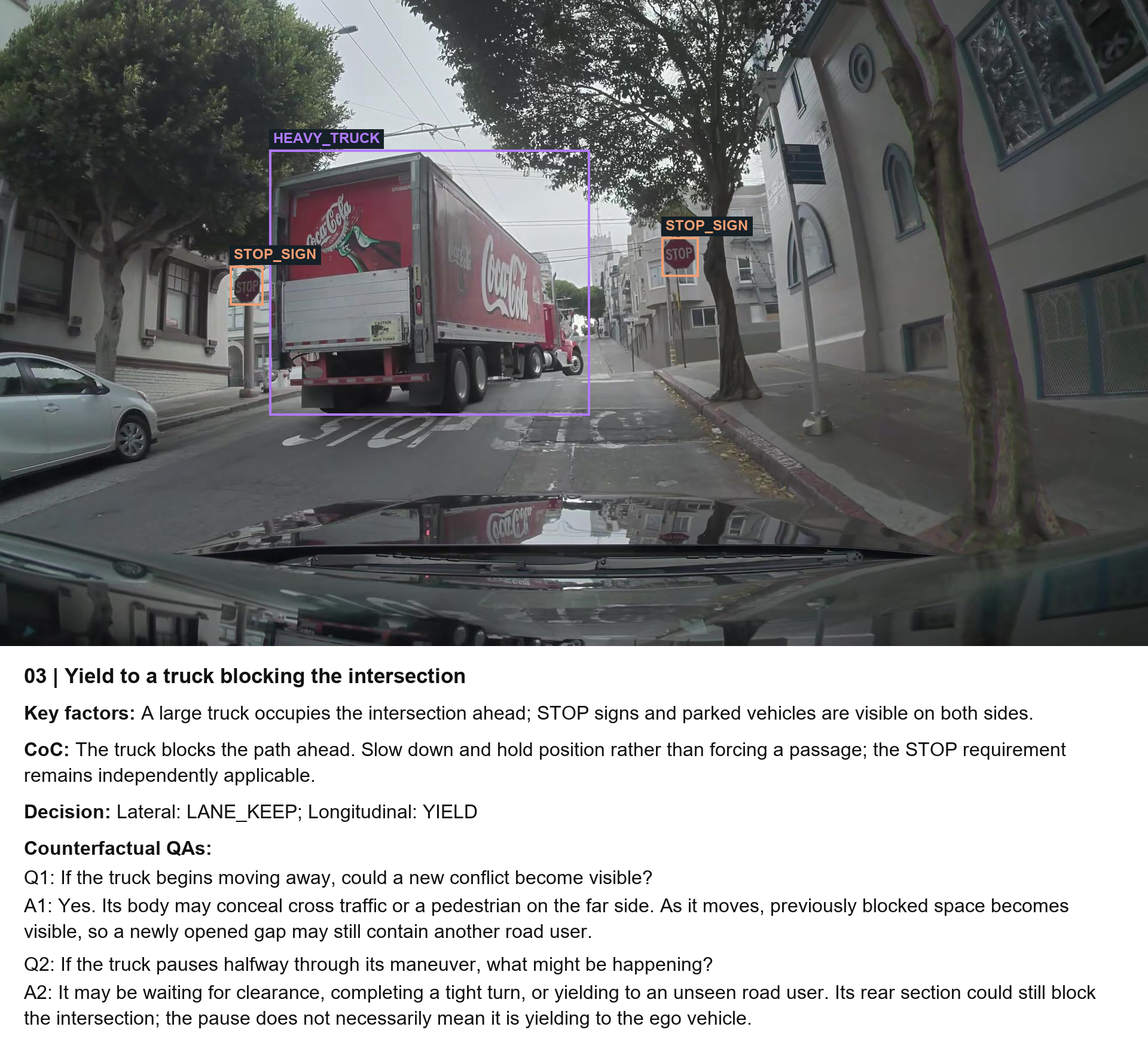}
    \end{minipage}\hfill
    \begin{minipage}{0.48\textwidth}
      \centering
      \includegraphics[width=\linewidth]{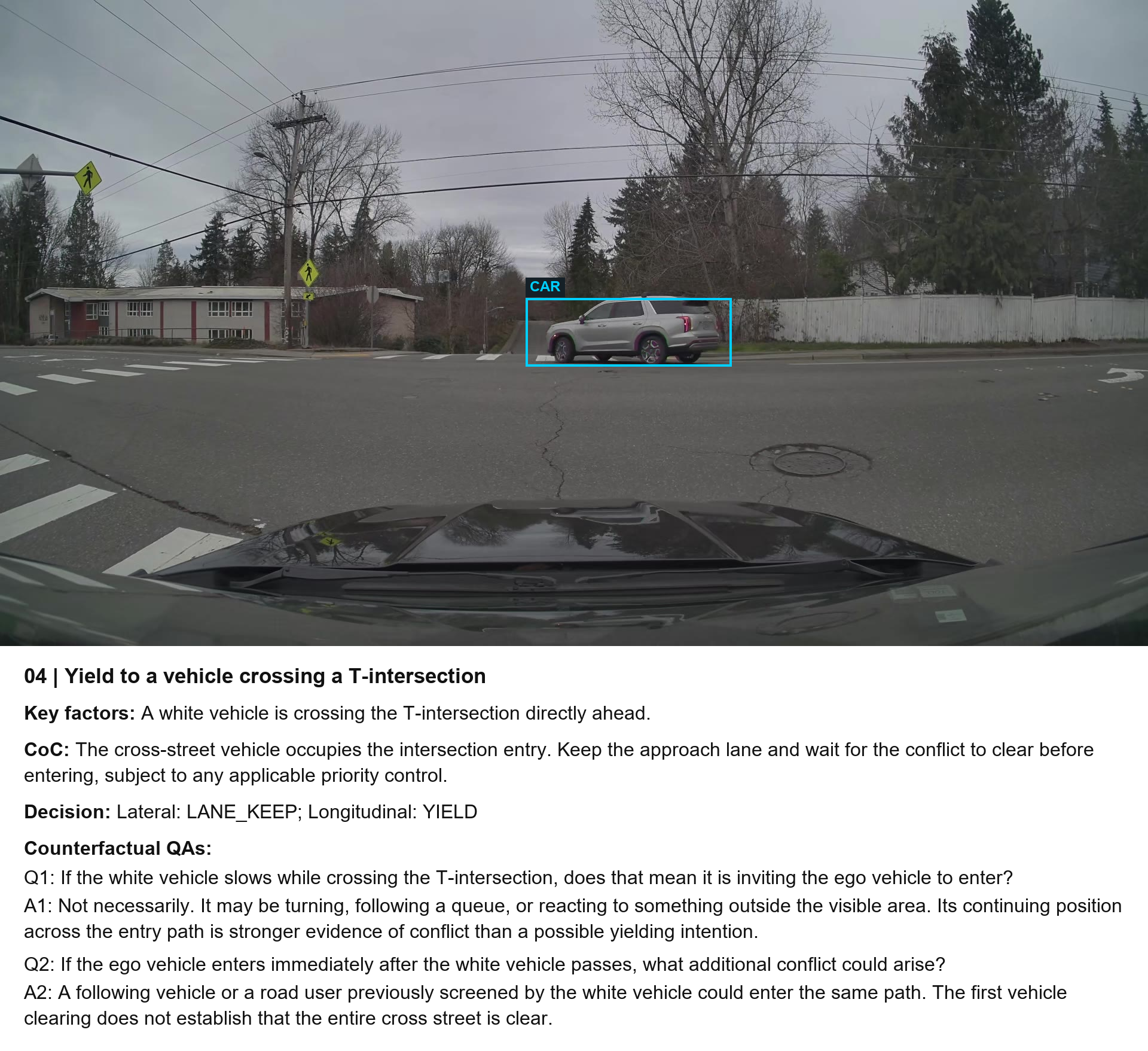}
    \end{minipage}\hfill
    \begin{minipage}{0.48\textwidth}
      \centering
      \includegraphics[width=\linewidth]{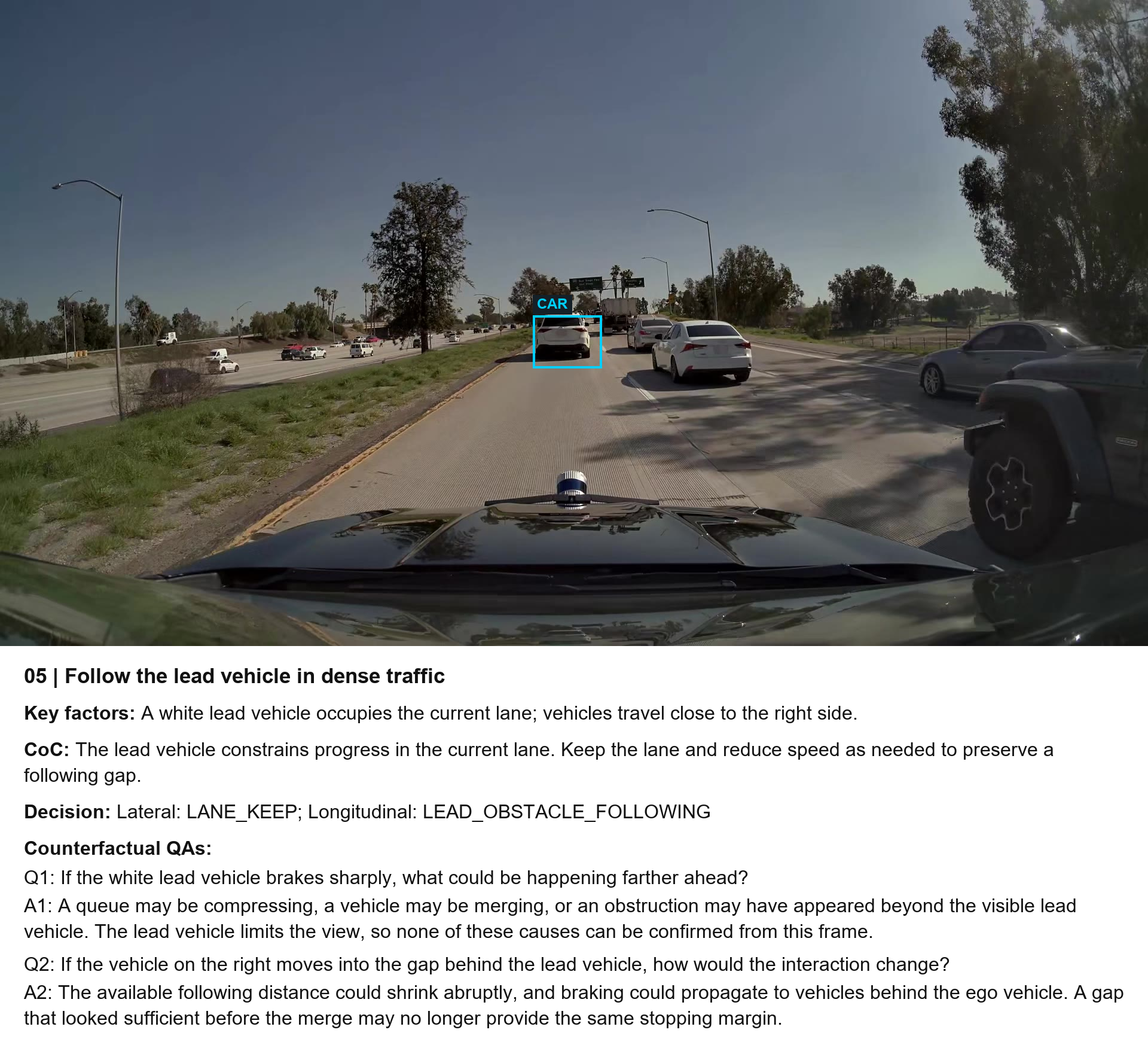}
    \end{minipage}\hfill
    \begin{minipage}{0.48\textwidth}
      \centering
      \includegraphics[width=\linewidth]{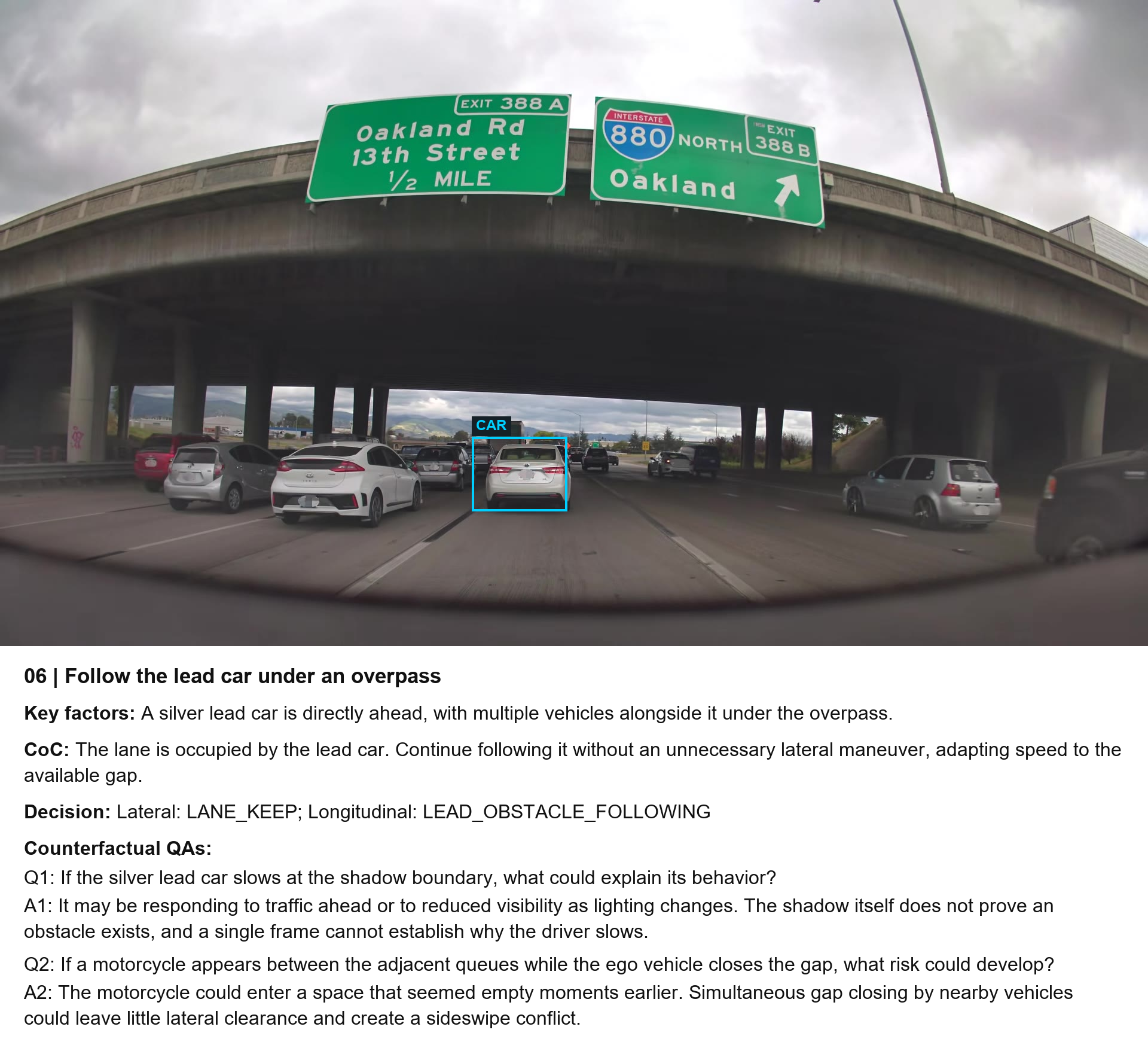}
    \end{minipage}\hfill
    \caption{Qualitative case study on 6 sign-dense scenes (1--6). Each case presents the key factors that the ego vehicle must attend to in the current frame, the CoC, the lateral and longitudinal decisions the ego vehicle should take, and the counterfactual QA pairs.}
    \label{fig:coc_case_study_1}
  \end{figure}

  \begin{figure}[htbp]
    \centering
    \begin{minipage}{0.48\textwidth}
      \centering
      \includegraphics[width=\linewidth]{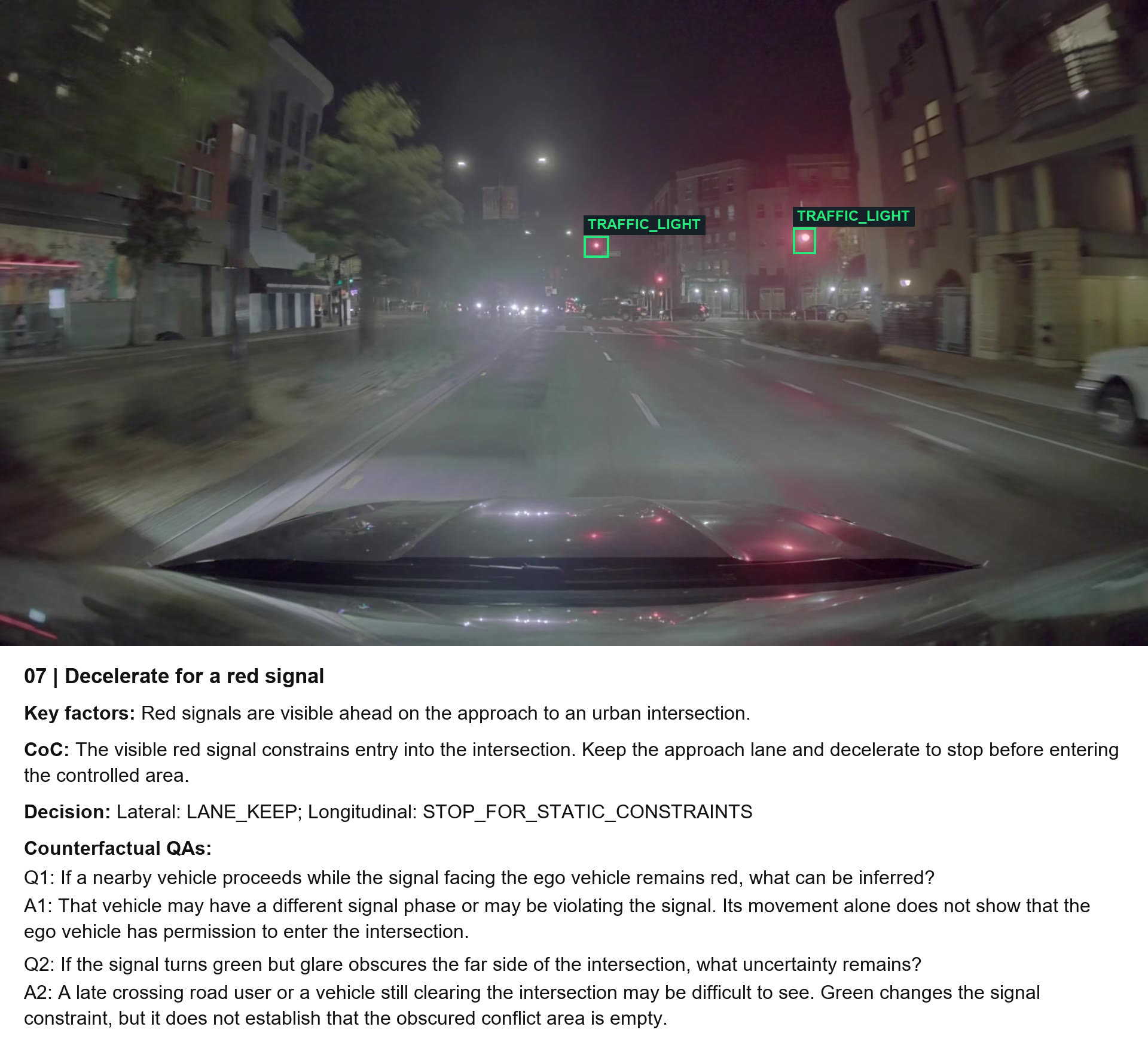}
    \end{minipage}\hfill
    \begin{minipage}{0.48\textwidth}
      \centering
      \includegraphics[width=\linewidth]{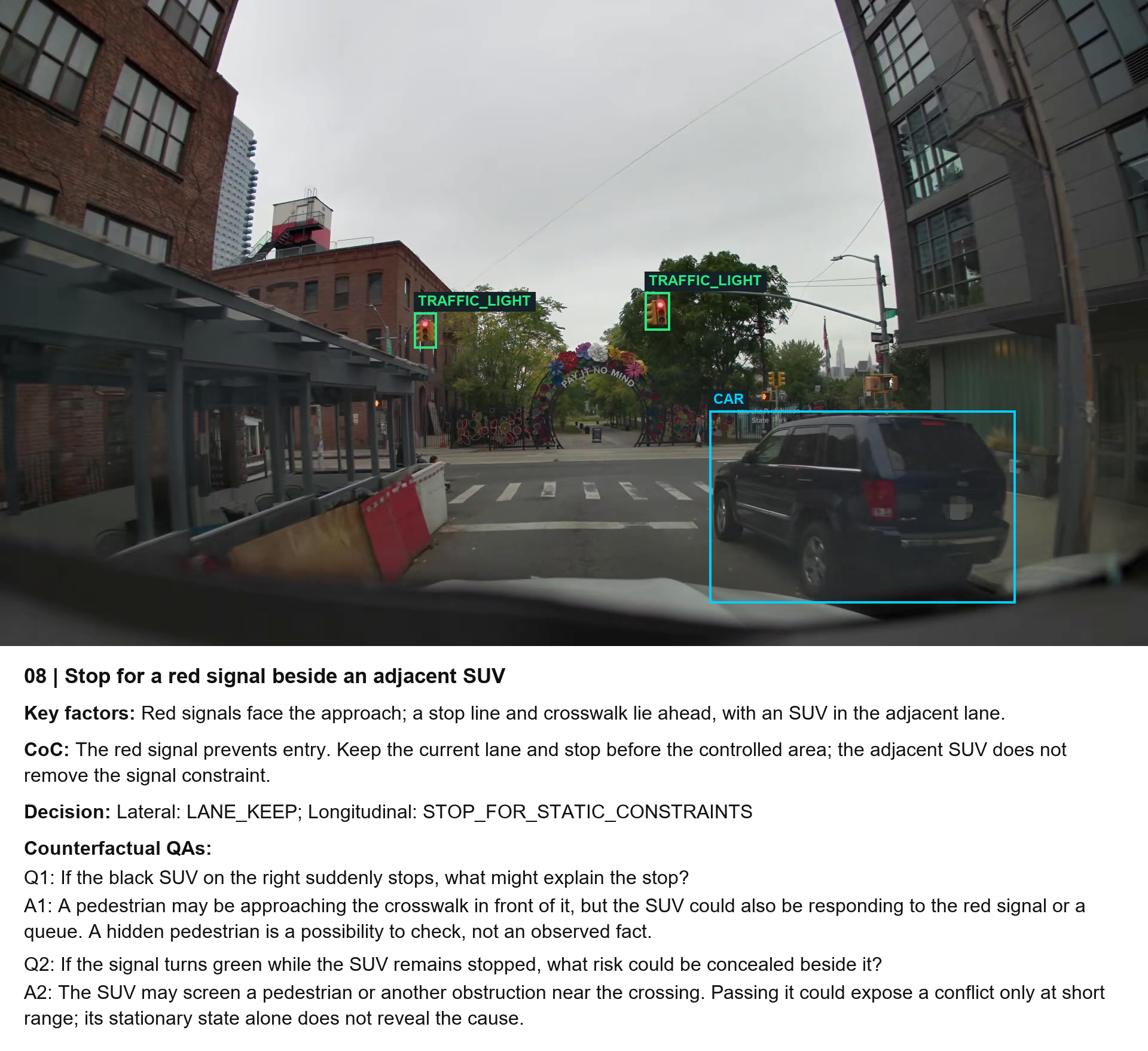}
    \end{minipage}\hfill
    \begin{minipage}{0.48\textwidth}
      \centering
      \includegraphics[width=\linewidth]{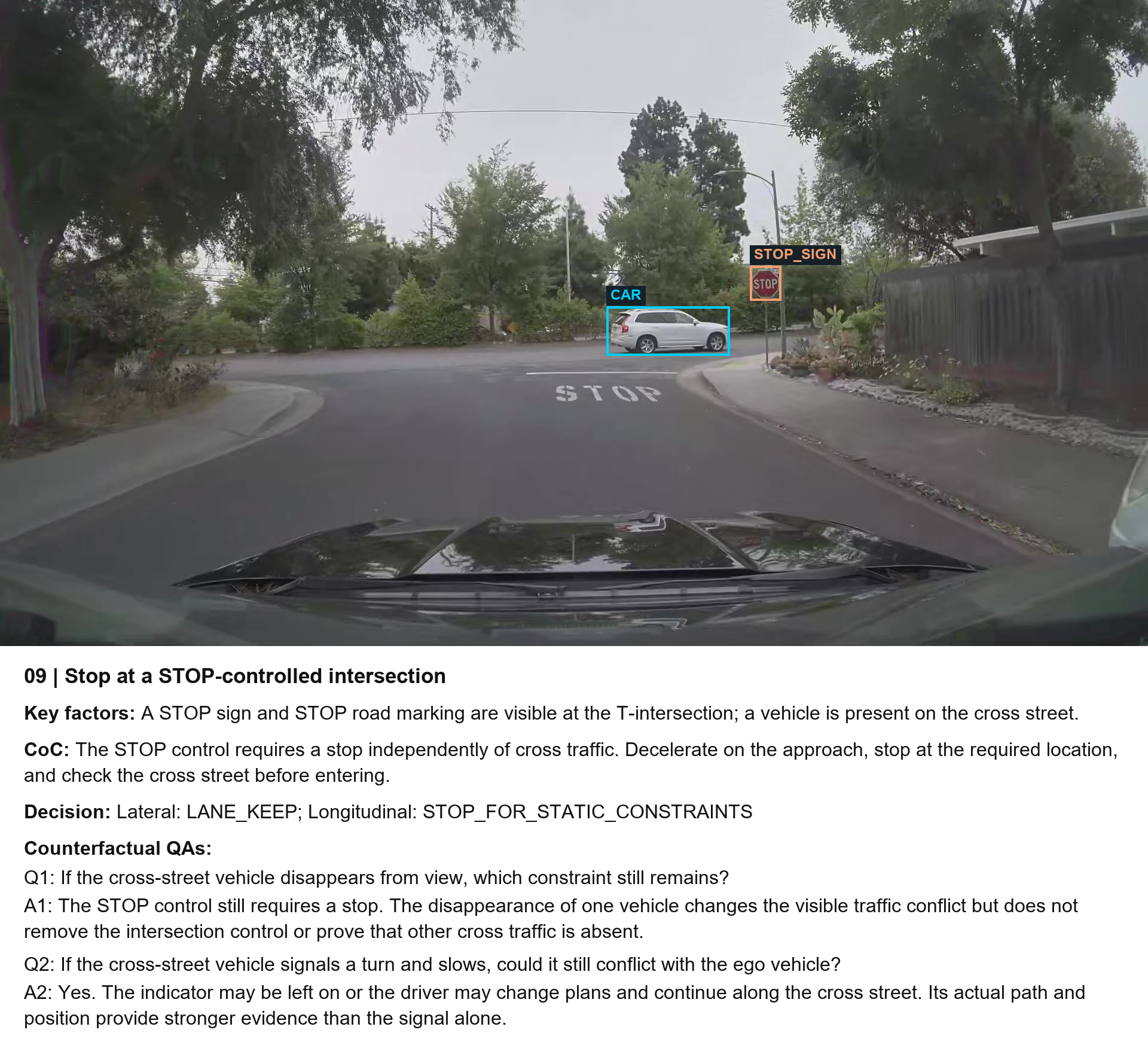}
    \end{minipage}\hfill
    \begin{minipage}{0.48\textwidth}
      \centering
      \includegraphics[width=\linewidth]{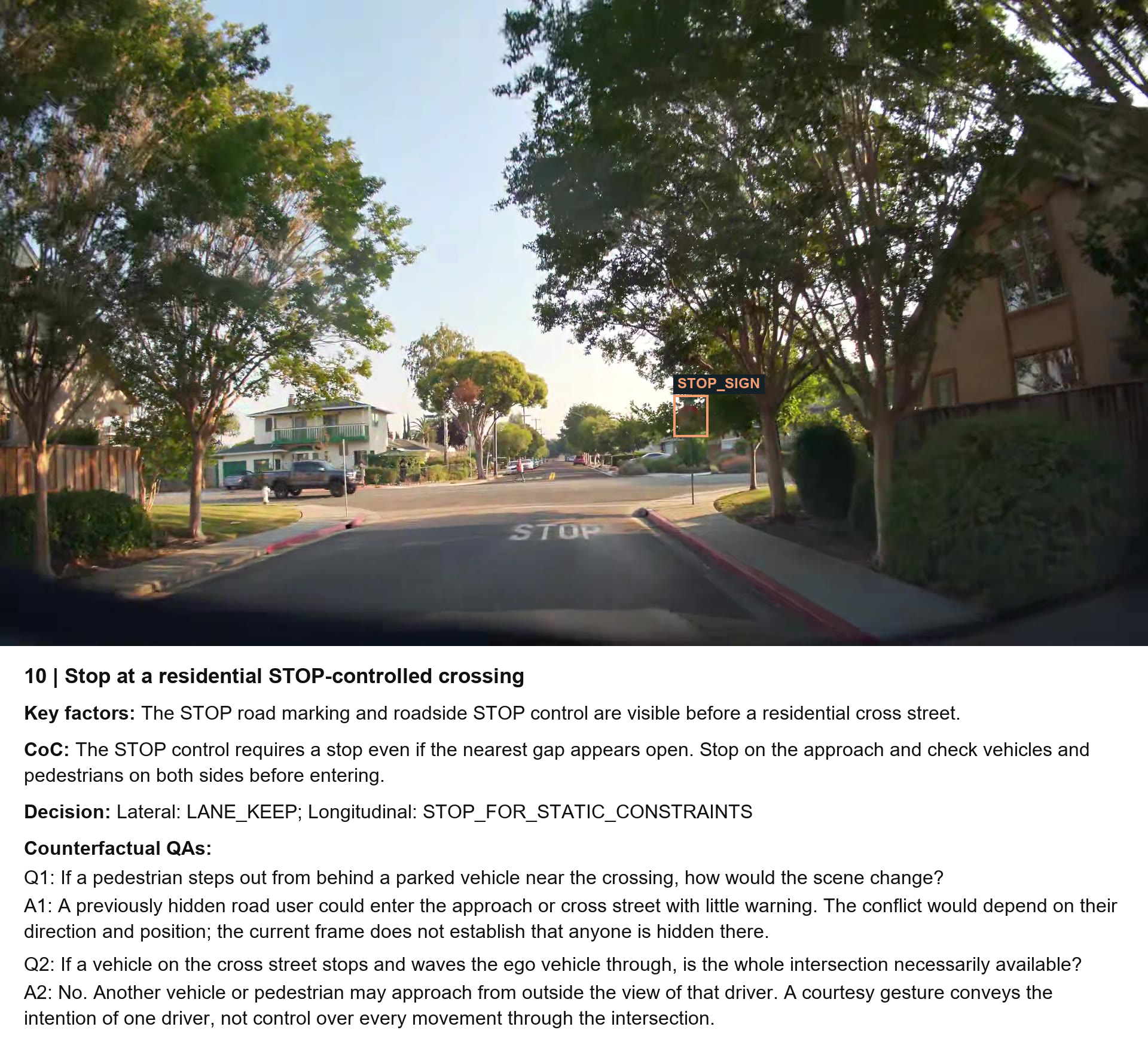}
    \end{minipage}\hfill
    \begin{minipage}{0.48\textwidth}
      \centering
      \includegraphics[width=\linewidth]{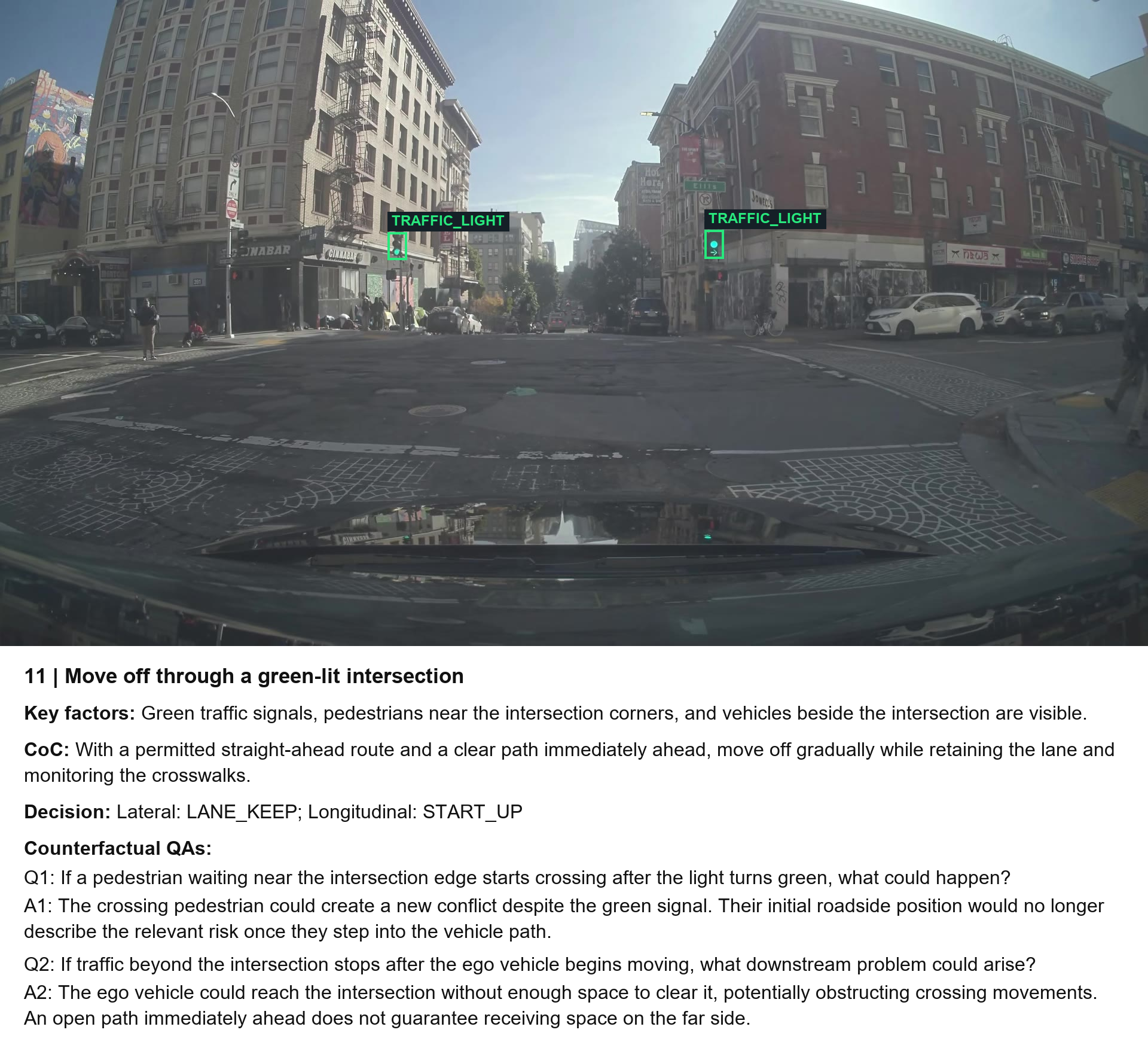}
    \end{minipage}\hfill
    \begin{minipage}{0.48\textwidth}
      \centering
      \includegraphics[width=\linewidth]{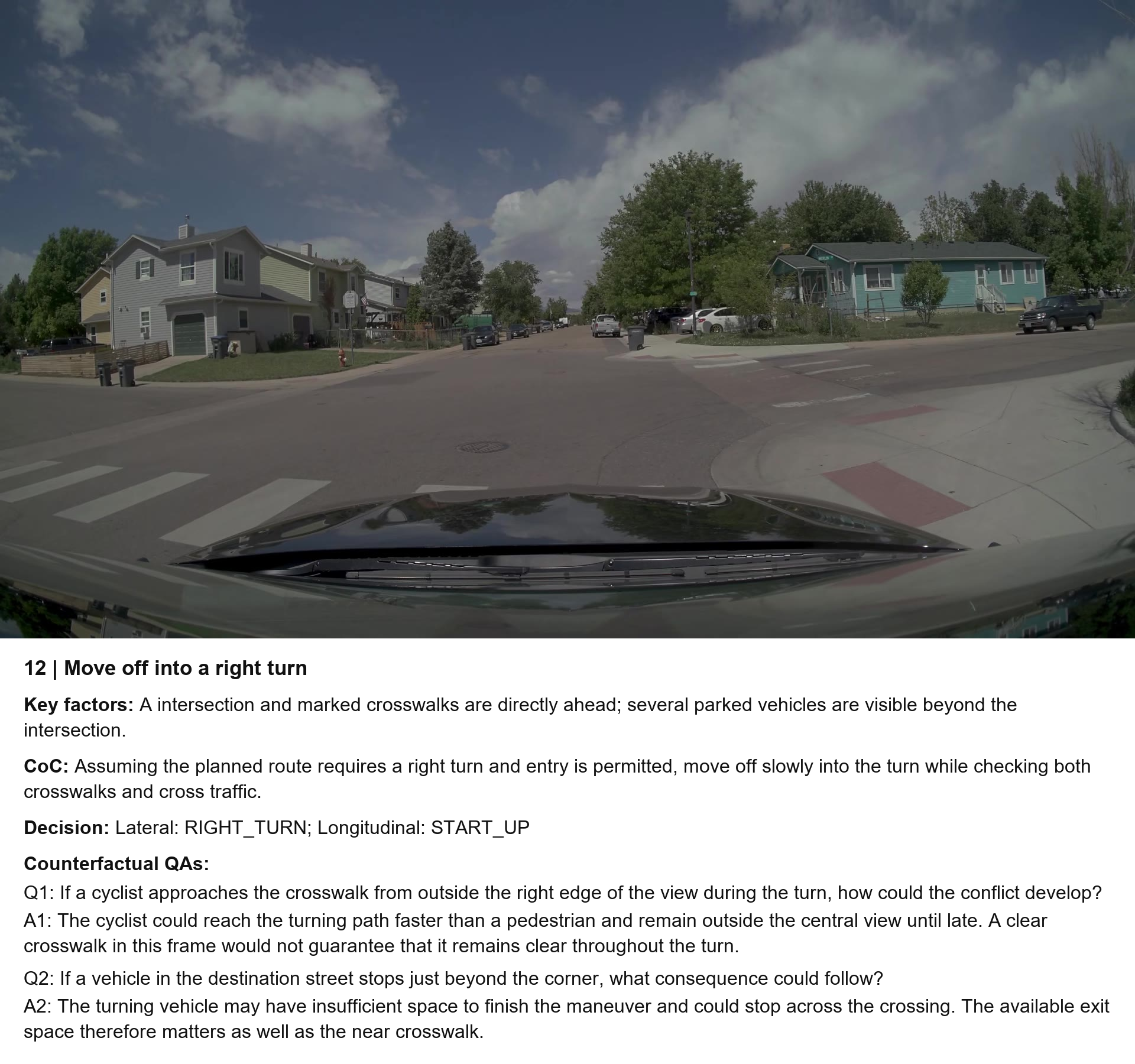}
    \end{minipage}\hfill
    \caption{Qualitative case study on 6 sign-dense scenes (7--12). Each case presents the key factors that the ego vehicle must attend to in the current frame, the CoC, the lateral and longitudinal decisions the ego vehicle should take, and the counterfactual QA pairs.}
    \label{fig:coc_case_study_2}
  \end{figure}

  \begin{figure}[htbp]
    \centering
    \begin{minipage}{0.48\textwidth}
      \centering
      \includegraphics[width=\linewidth]{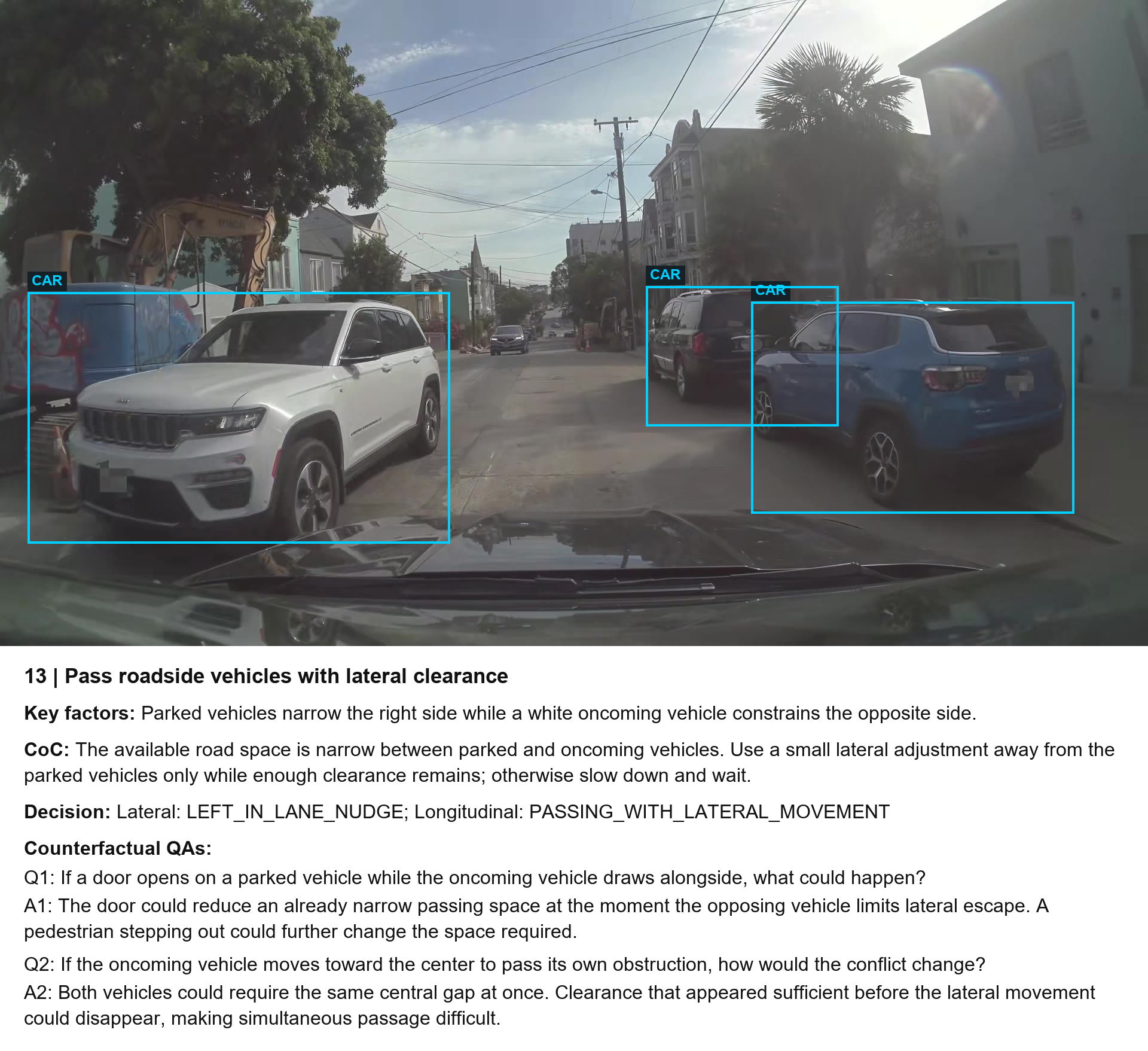}
    \end{minipage}\hfill
    \begin{minipage}{0.48\textwidth}
      \centering
      \includegraphics[width=\linewidth]{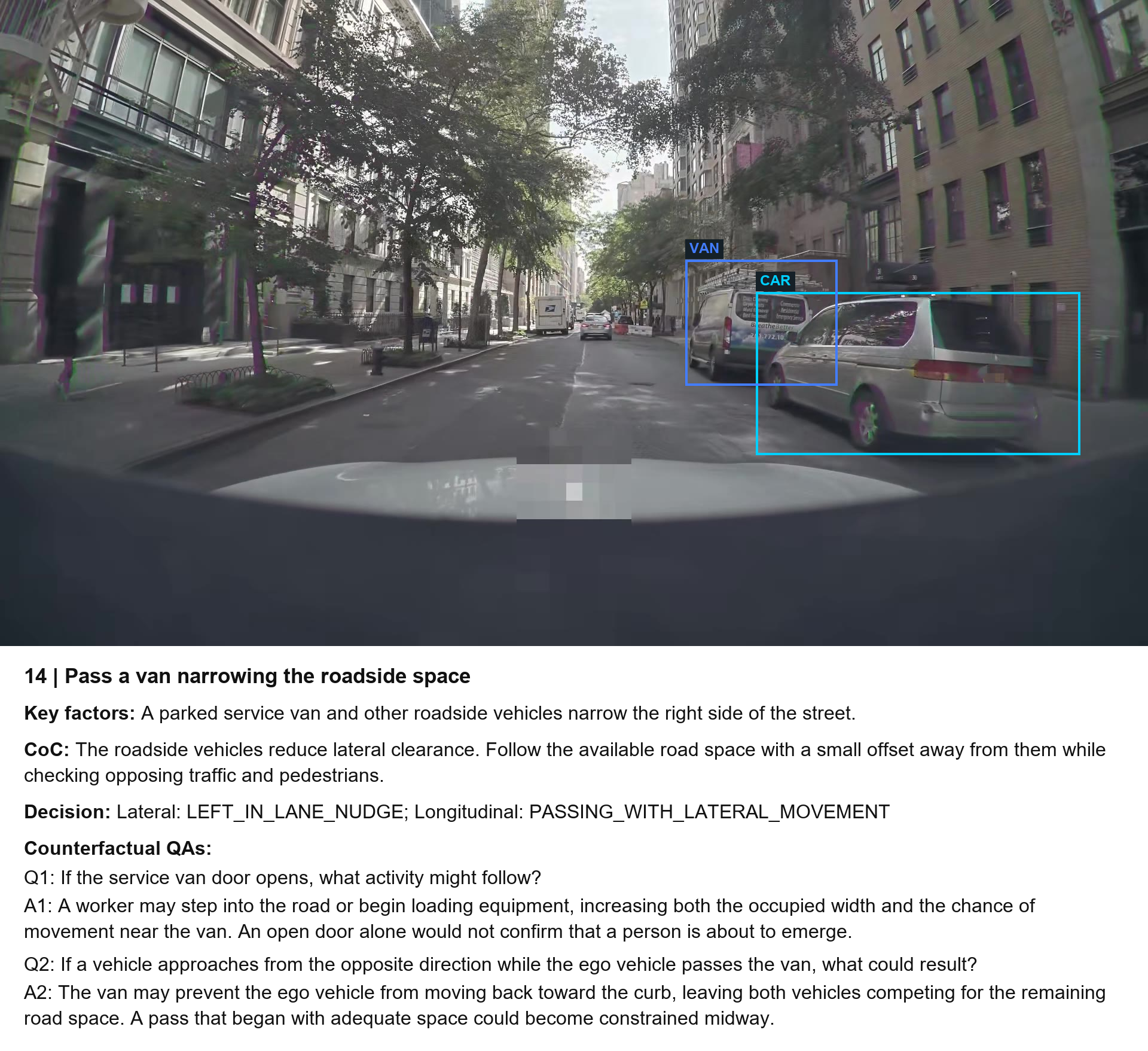}
    \end{minipage}\hfill
    \begin{minipage}{0.48\textwidth}
      \centering
      \includegraphics[width=\linewidth]{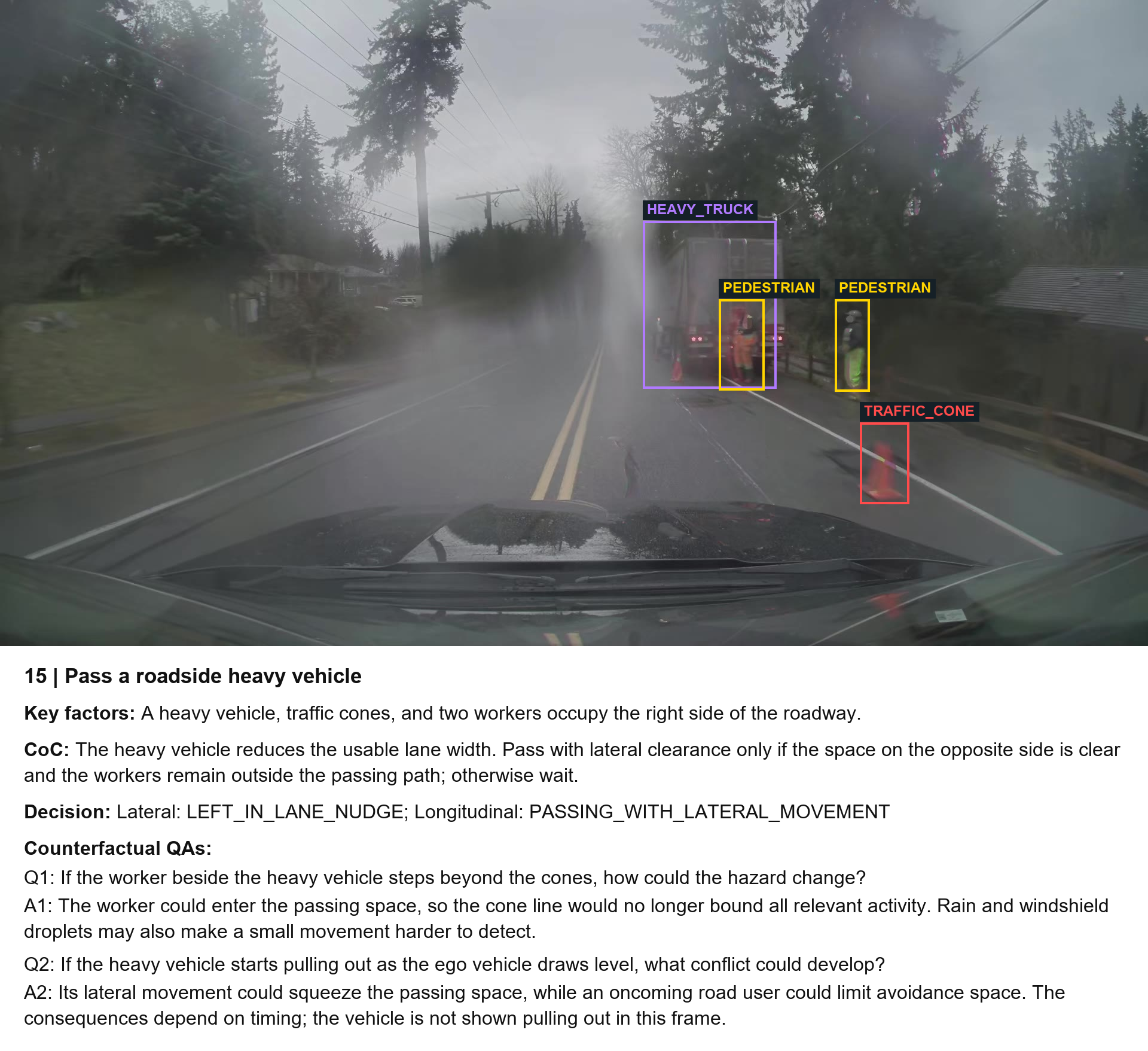}
    \end{minipage}\hfill
    \begin{minipage}{0.48\textwidth}
      \centering
      \includegraphics[width=\linewidth]{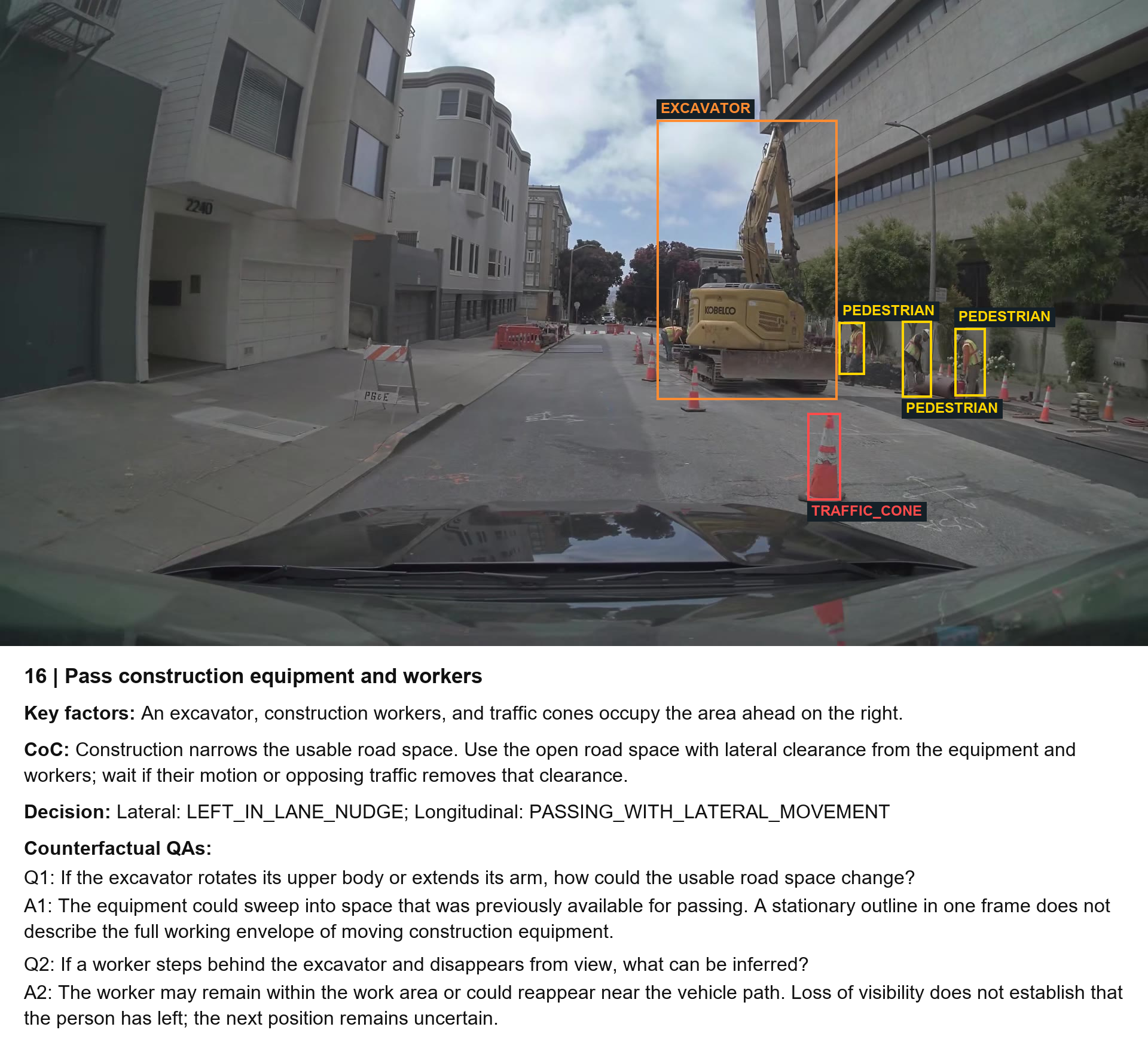}
    \end{minipage}\hfill
    \begin{minipage}{0.48\textwidth}
      \centering
      \includegraphics[width=\linewidth]{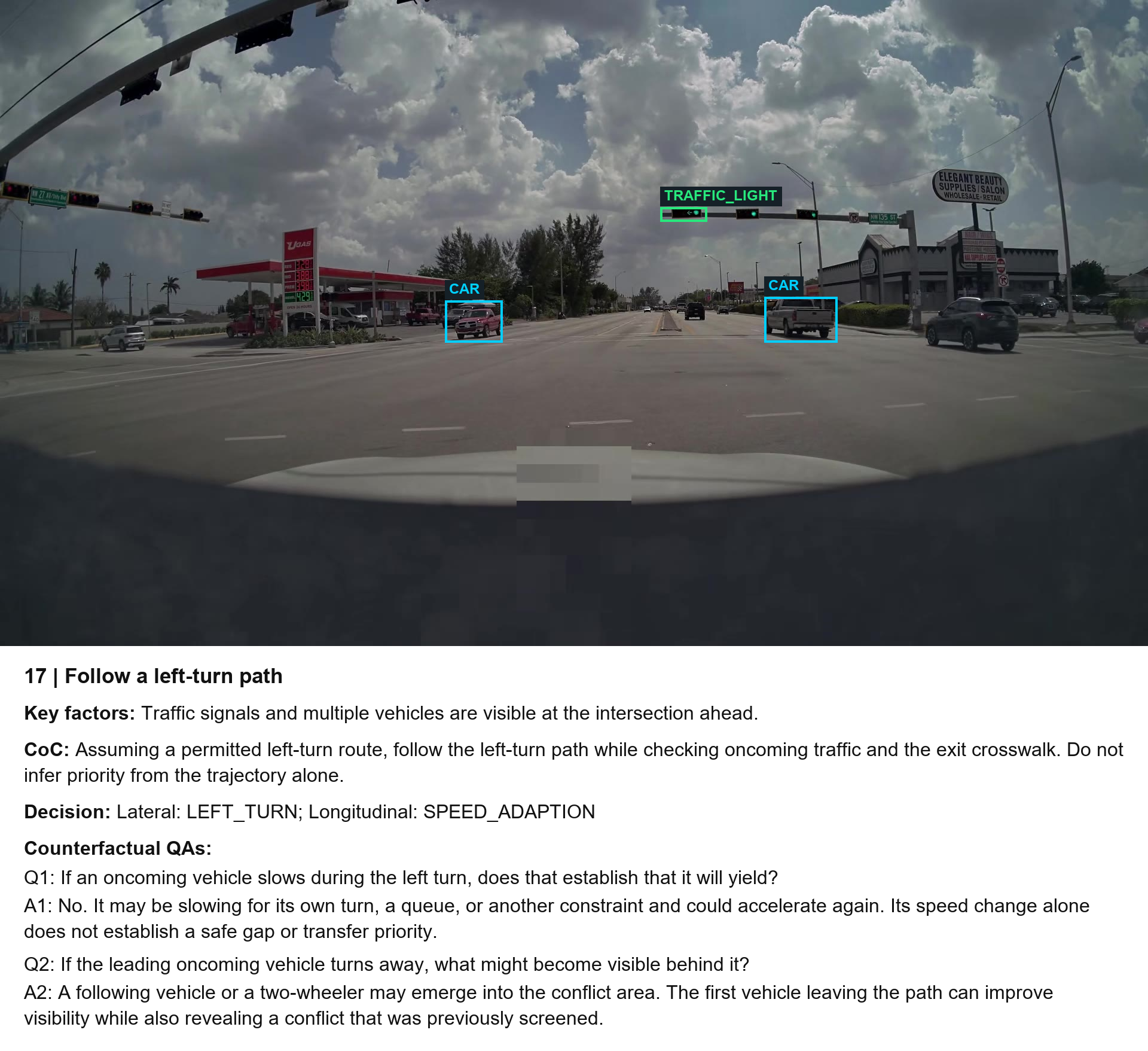}
    \end{minipage}\hfill
    \begin{minipage}{0.48\textwidth}
      \centering
      \includegraphics[width=\linewidth]{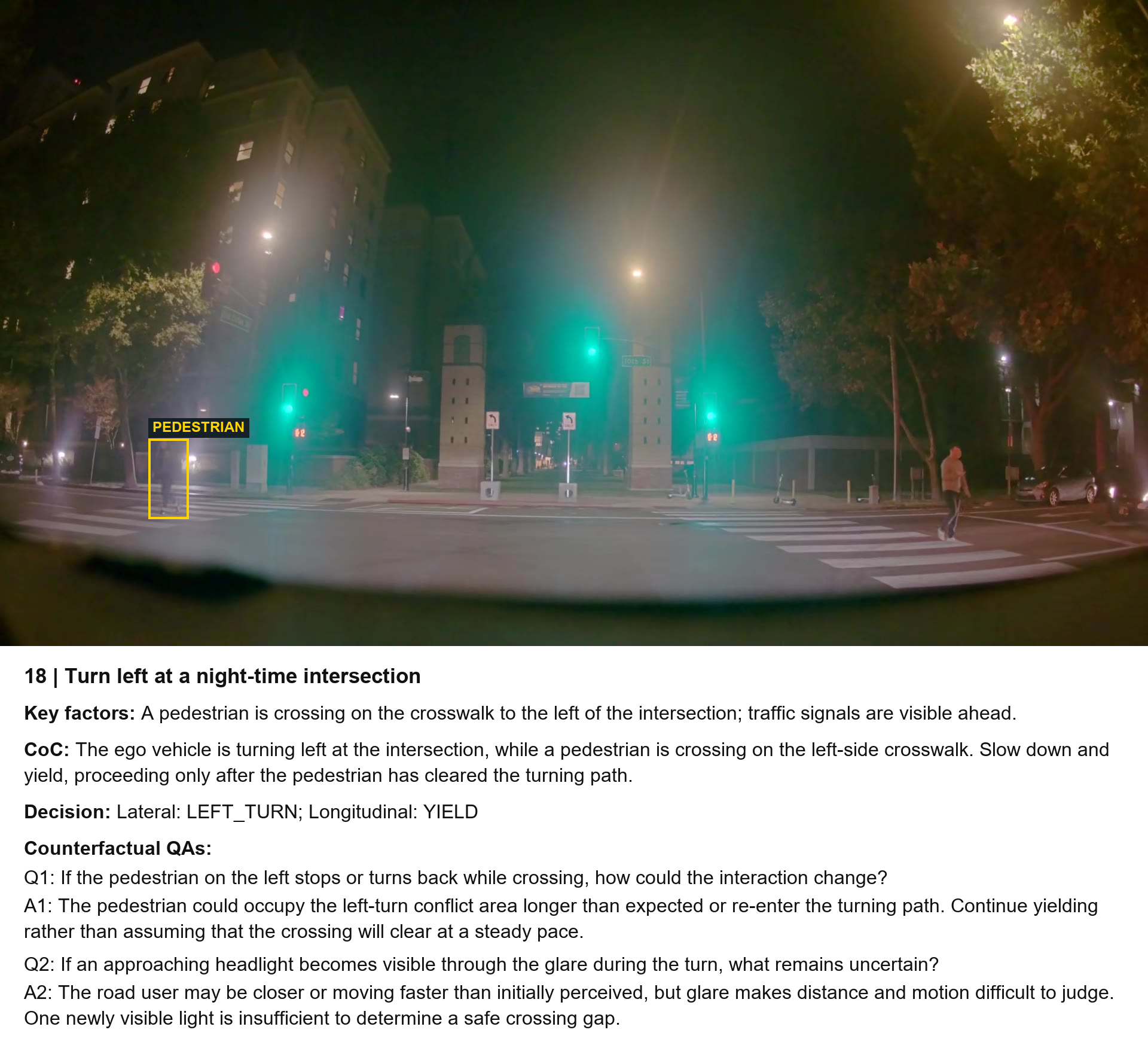}
    \end{minipage}\hfill
    \caption{Qualitative case study on 6 sign-dense scenes (13--18). Each case presents the key factors that the ego vehicle must attend to in the current frame, the CoC, the lateral and longitudinal decisions the ego vehicle should take, and the counterfactual QA pairs.}
    \label{fig:coc_case_study_3}
  \end{figure}

  \begin{figure}[htbp]
    \centering
    \begin{minipage}{0.48\textwidth}
      \centering
      \includegraphics[width=\linewidth]{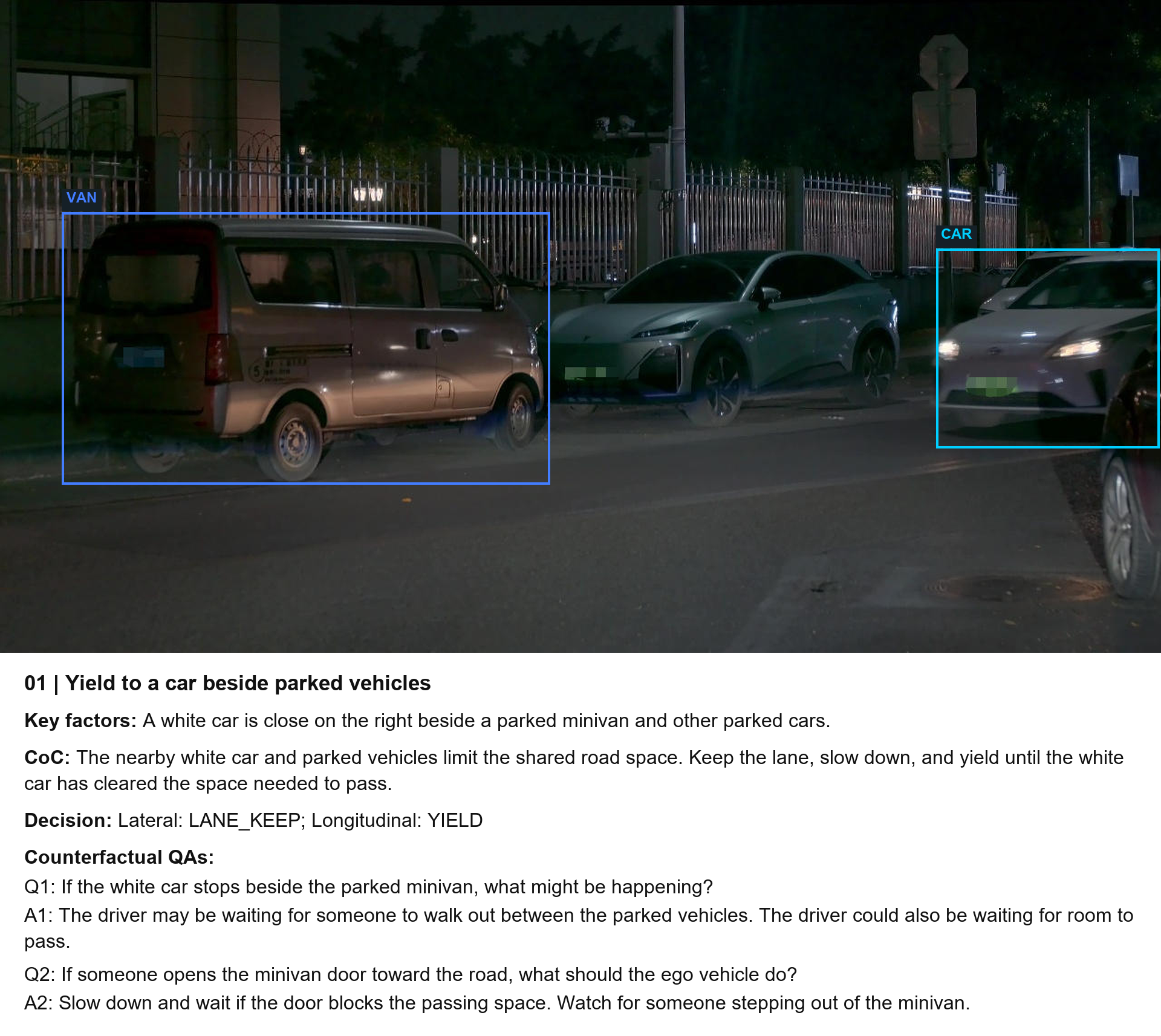}
    \end{minipage}\hfill
    \begin{minipage}{0.48\textwidth}
      \centering
      \includegraphics[width=\linewidth]{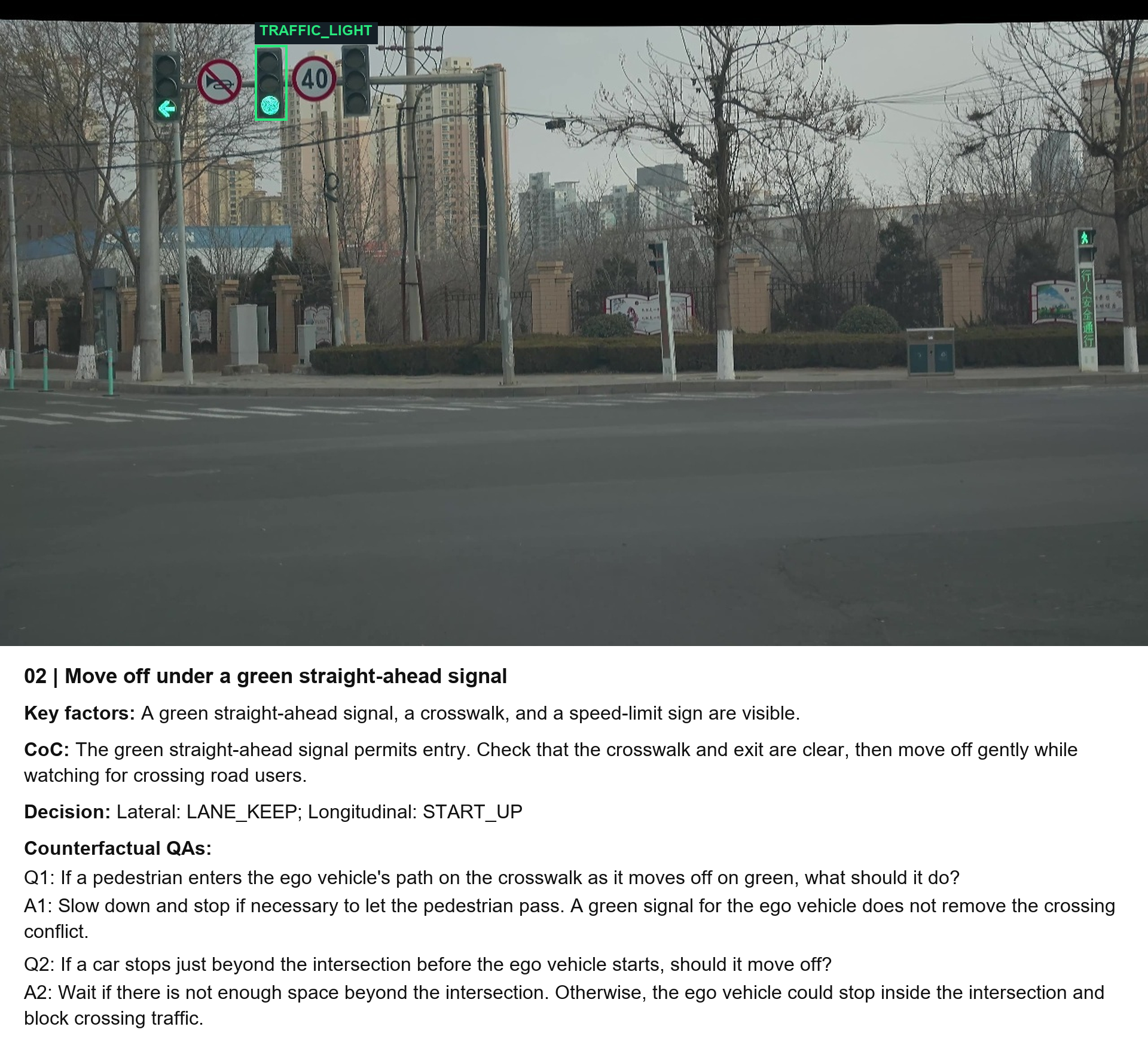}
    \end{minipage}\hfill
    \begin{minipage}{0.48\textwidth}
      \centering
      \includegraphics[width=\linewidth]{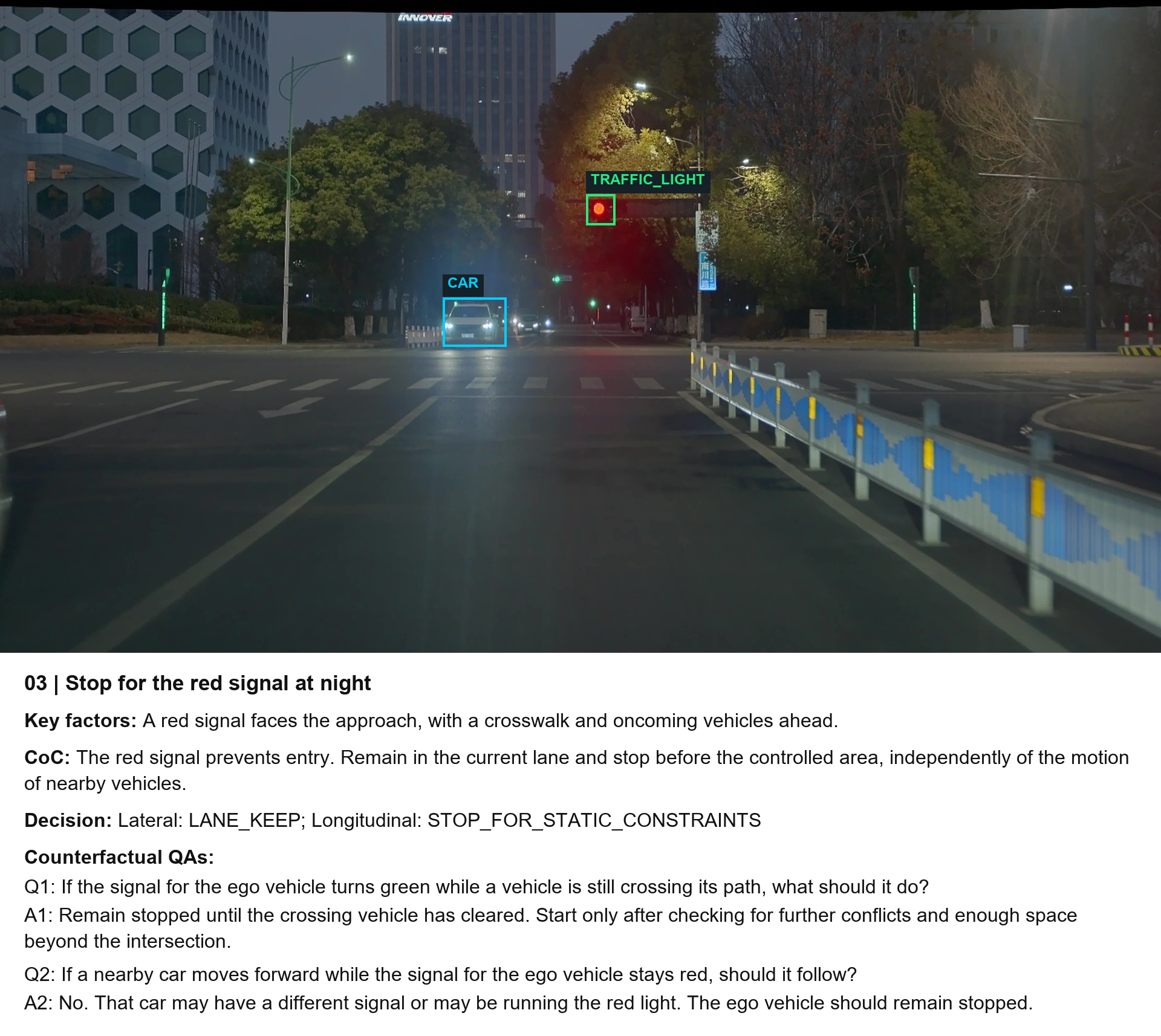}
    \end{minipage}\hfill
    \begin{minipage}{0.48\textwidth}
      \centering
      \includegraphics[width=\linewidth]{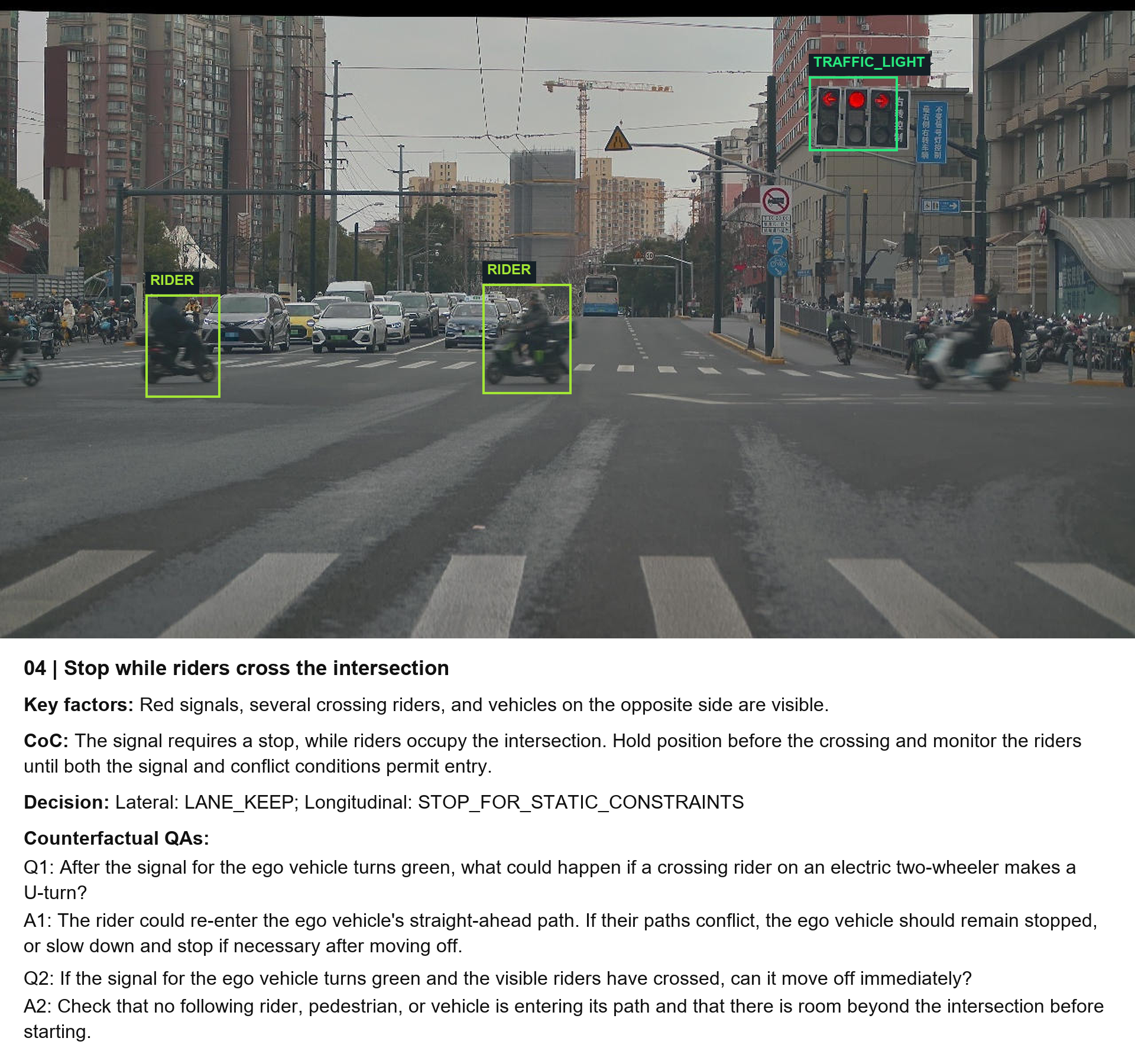}
    \end{minipage}\hfill
    \begin{minipage}{0.48\textwidth}
      \centering
      \includegraphics[width=\linewidth]{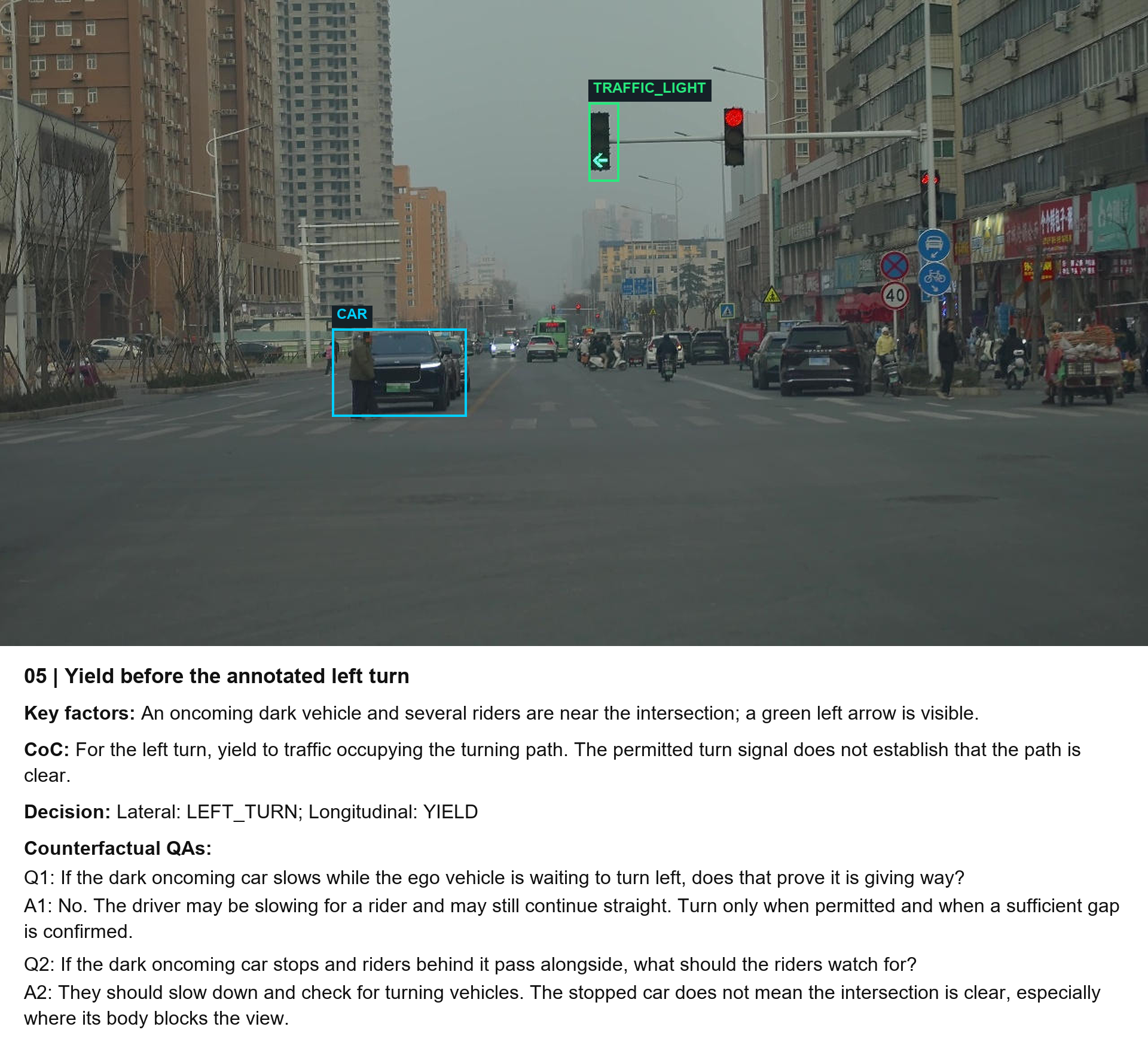}
    \end{minipage}\hfill
    \begin{minipage}{0.48\textwidth}
      \centering
      \includegraphics[width=\linewidth]{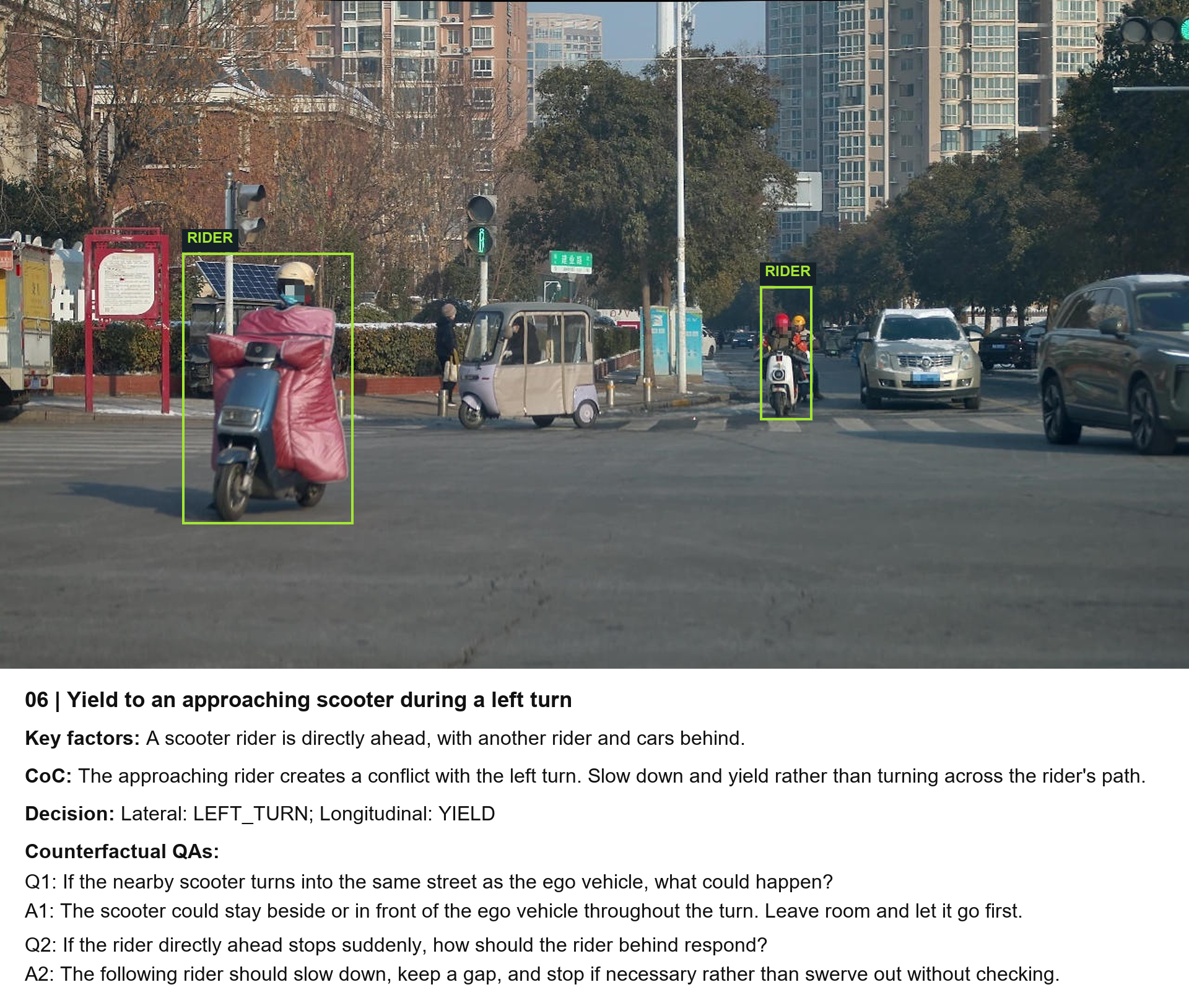} 
    \end{minipage}\hfill
    \caption{Qualitative case study on 6 sign-dense scenes (19--24). Each case presents the key factors that the ego vehicle must attend to in the current frame, the CoC, the lateral and longitudinal decisions the ego vehicle should take, and the counterfactual QA pairs.}
    \label{fig:coc_case_study_4}
  \end{figure}

  \begin{figure}[htbp]
    \centering
    \begin{minipage}{0.48\textwidth}
      \centering
      \includegraphics[width=\linewidth]{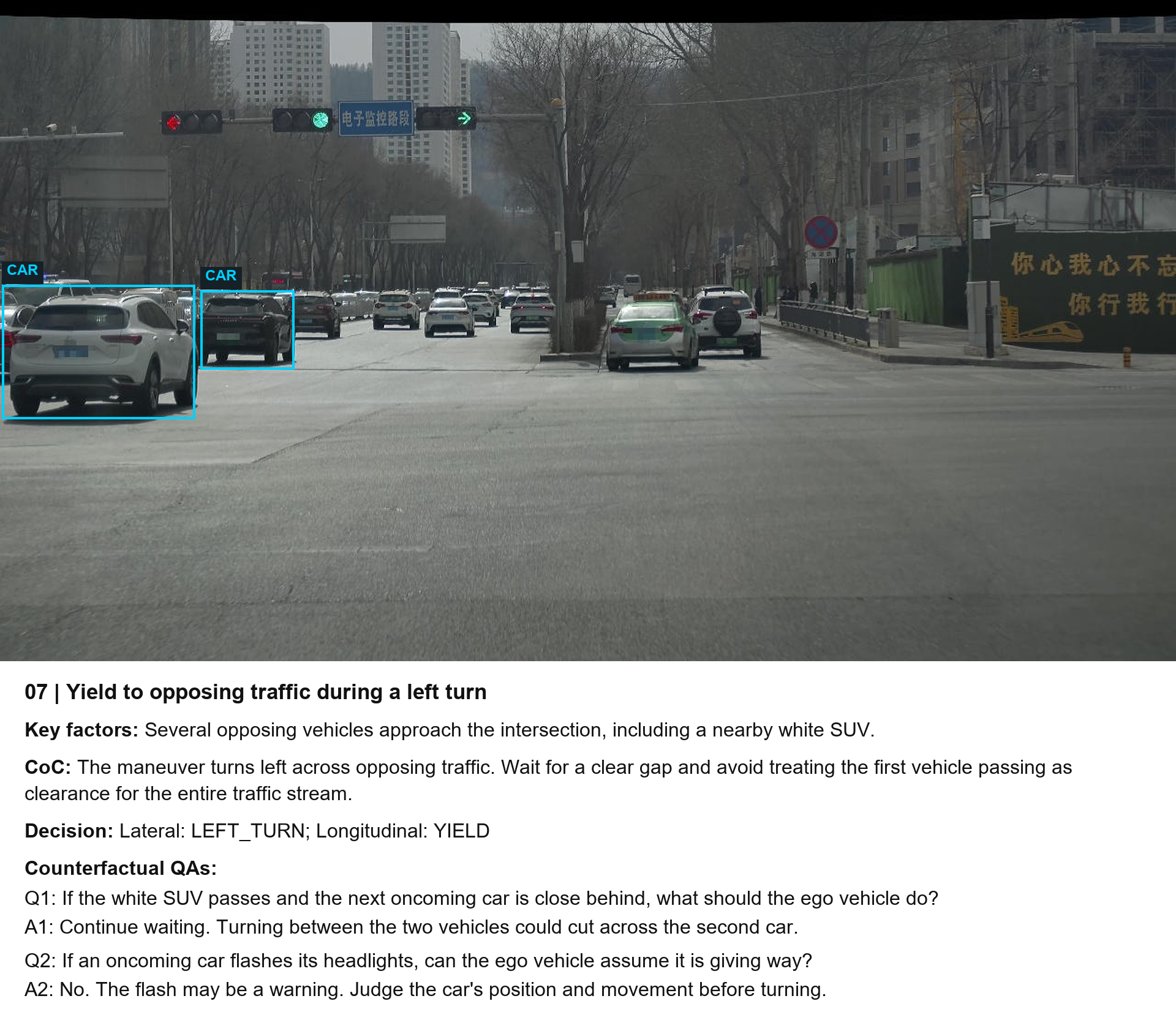}
    \end{minipage}\hfill
    \begin{minipage}{0.48\textwidth}
      \centering
      \includegraphics[width=\linewidth]{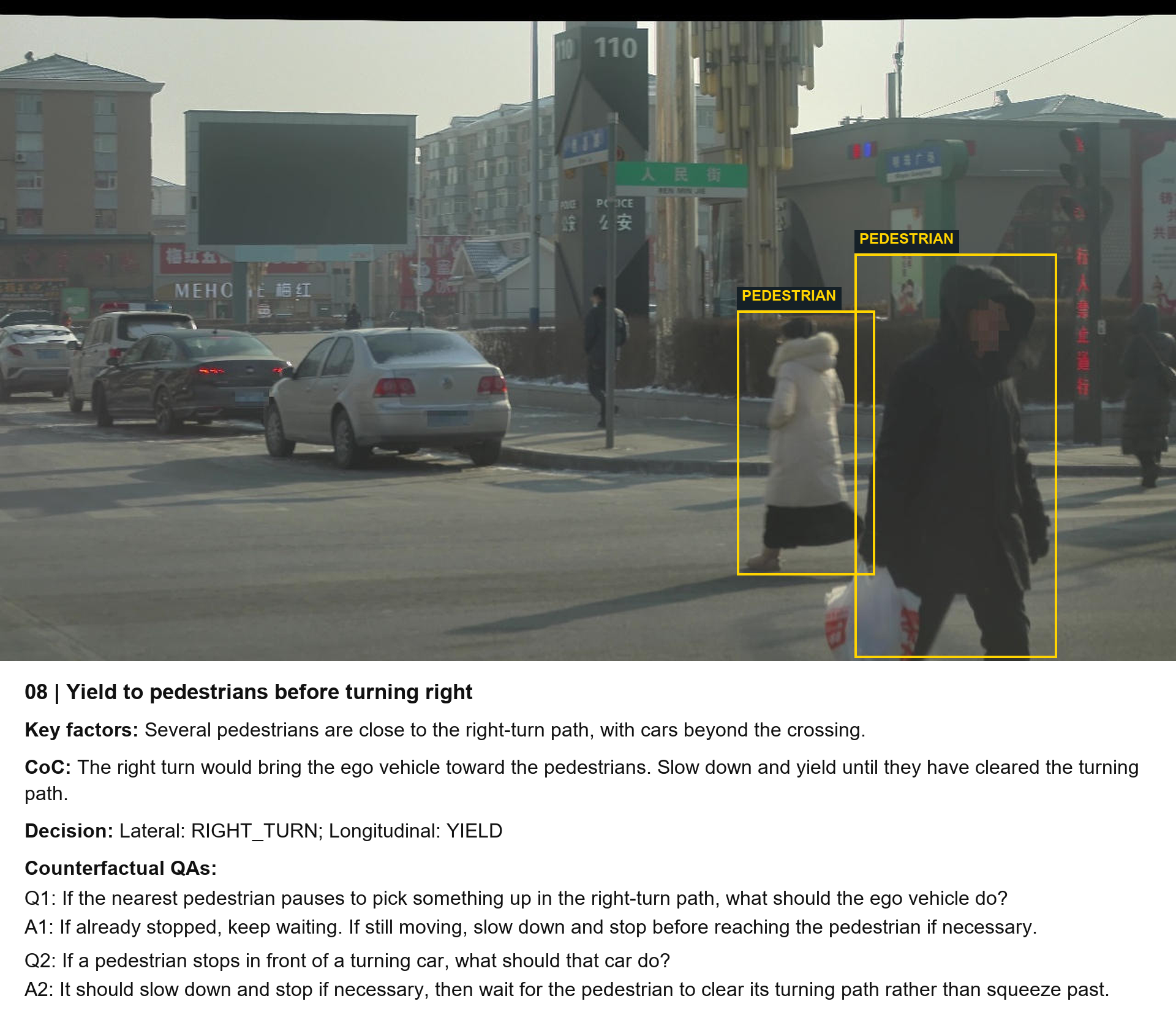}
    \end{minipage}\hfill
    \begin{minipage}{0.48\textwidth}
      \centering
      \includegraphics[width=\linewidth]{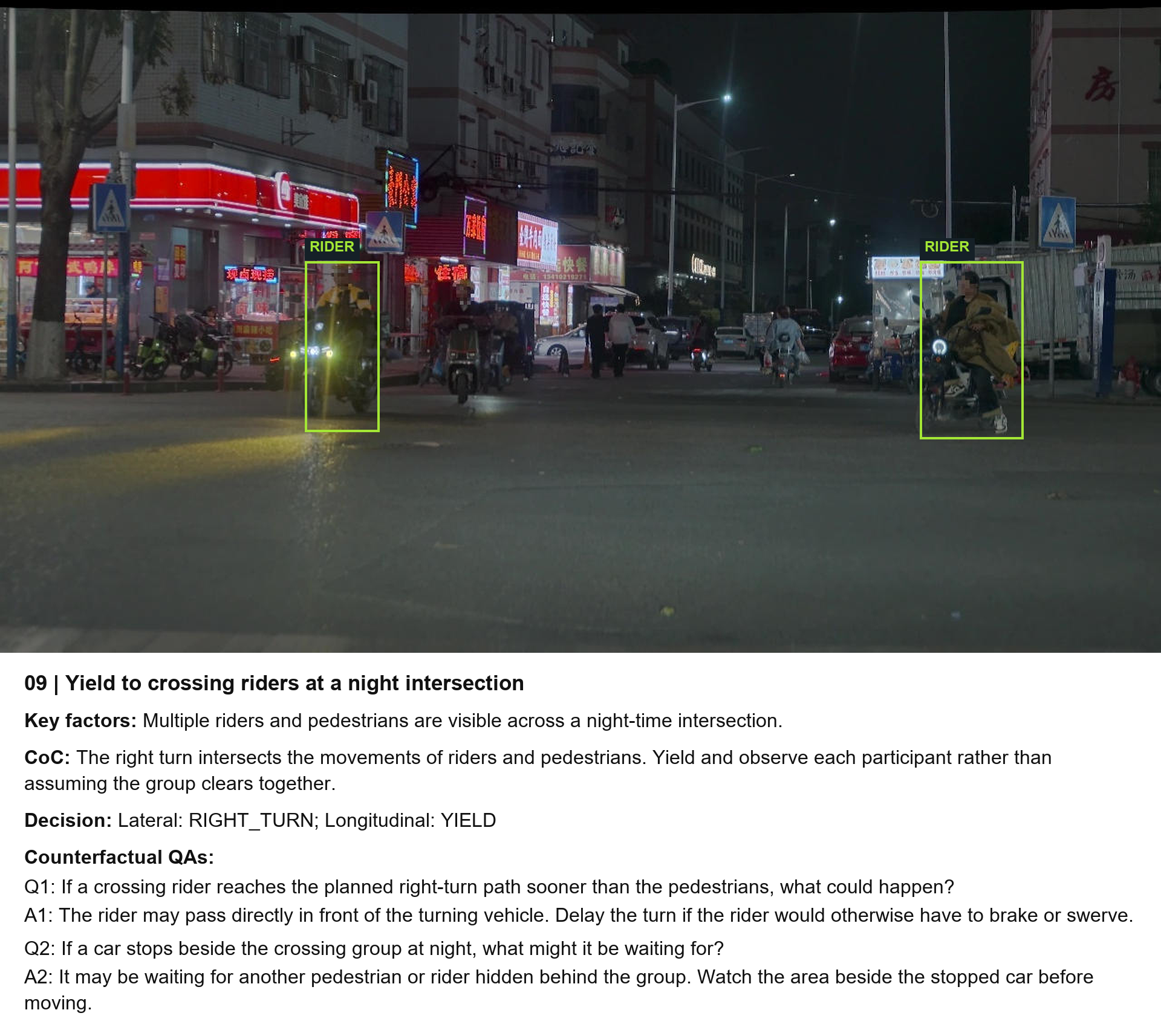}
    \end{minipage}\hfill
    \begin{minipage}{0.48\textwidth}
      \centering
      \includegraphics[width=\linewidth]{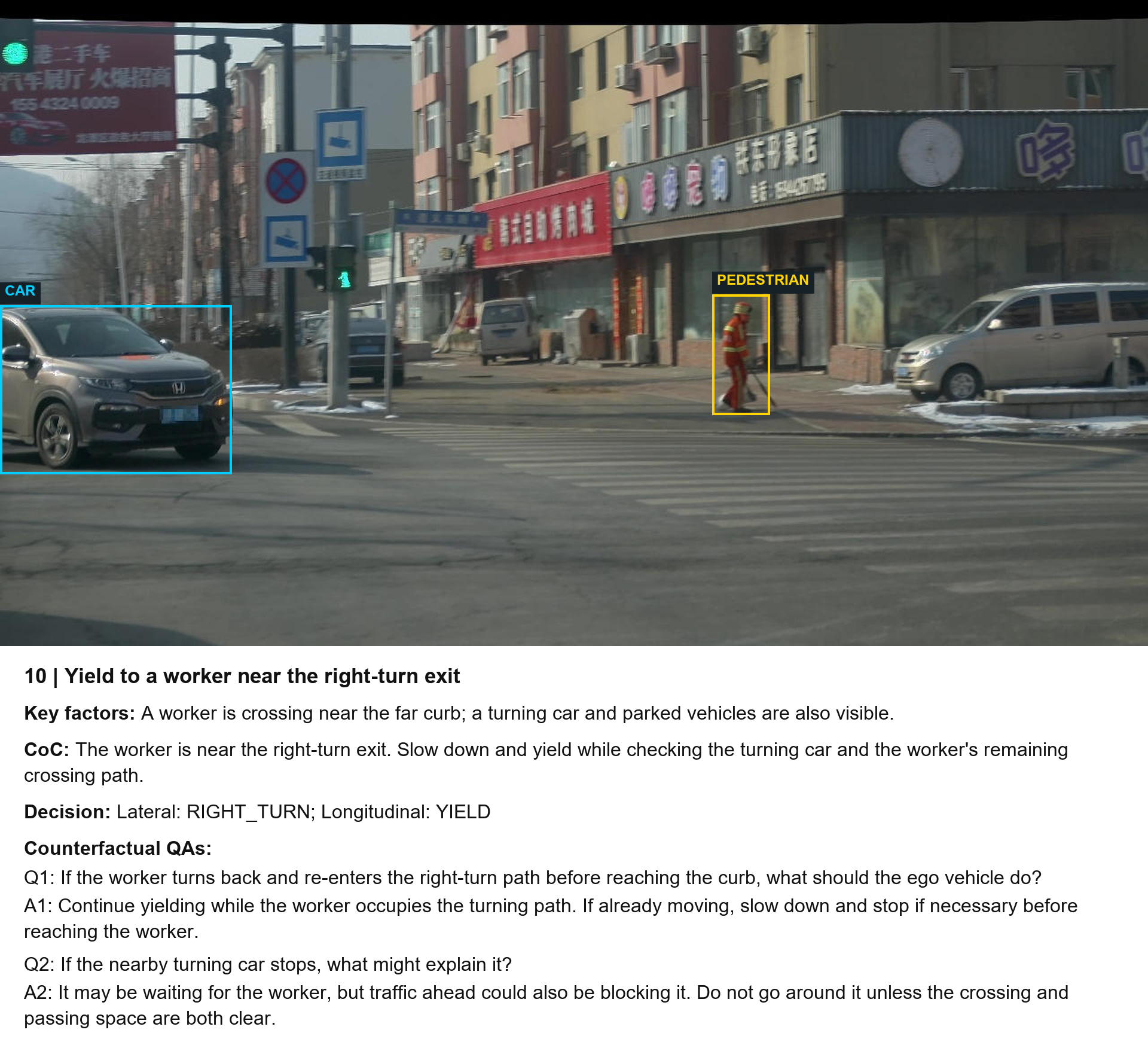}
    \end{minipage}\hfill
    \begin{minipage}{0.48\textwidth}
      \centering
      \includegraphics[width=\linewidth]{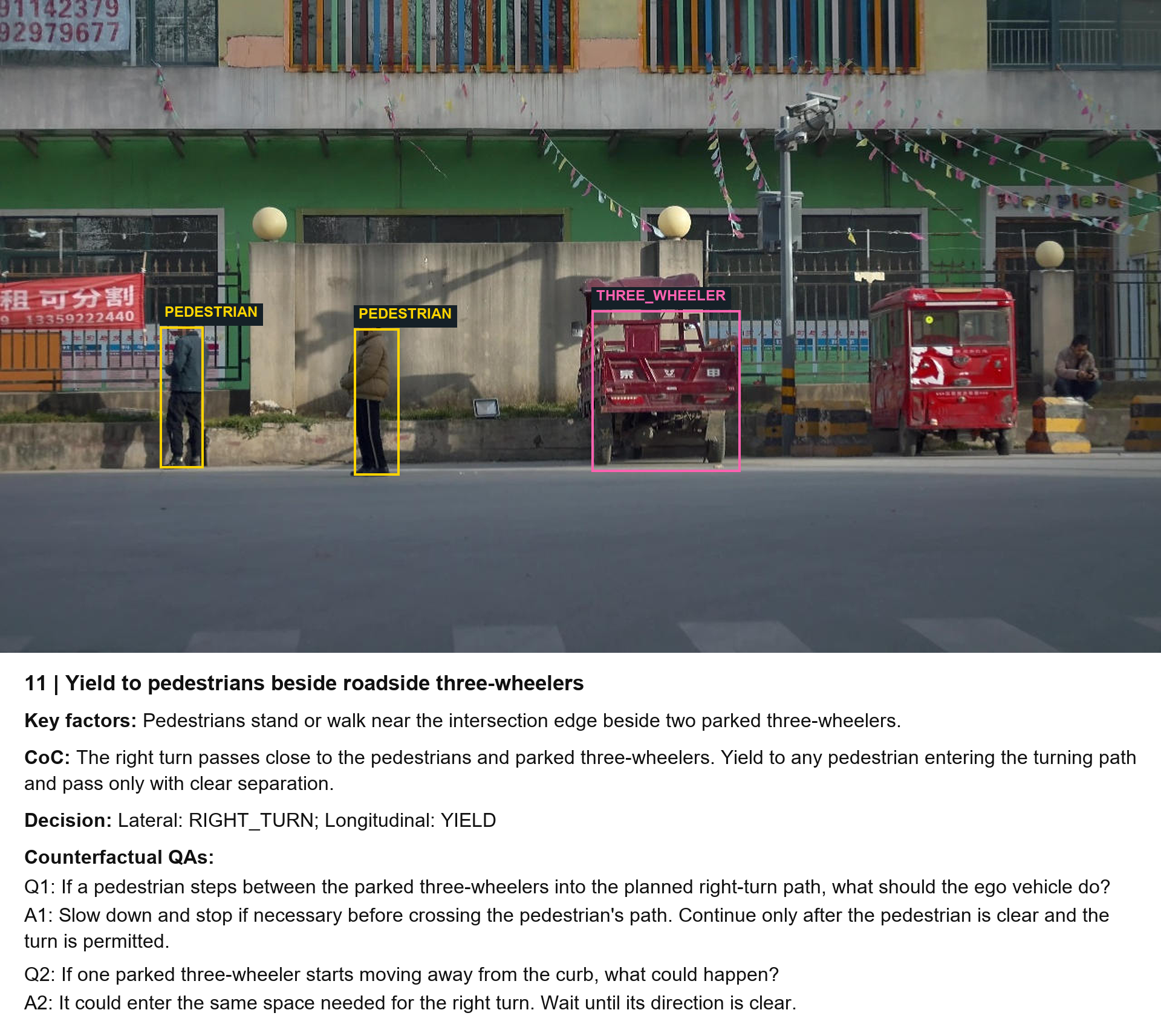}
    \end{minipage}\hfill
    \begin{minipage}{0.48\textwidth}
      \centering
      \includegraphics[width=\linewidth]{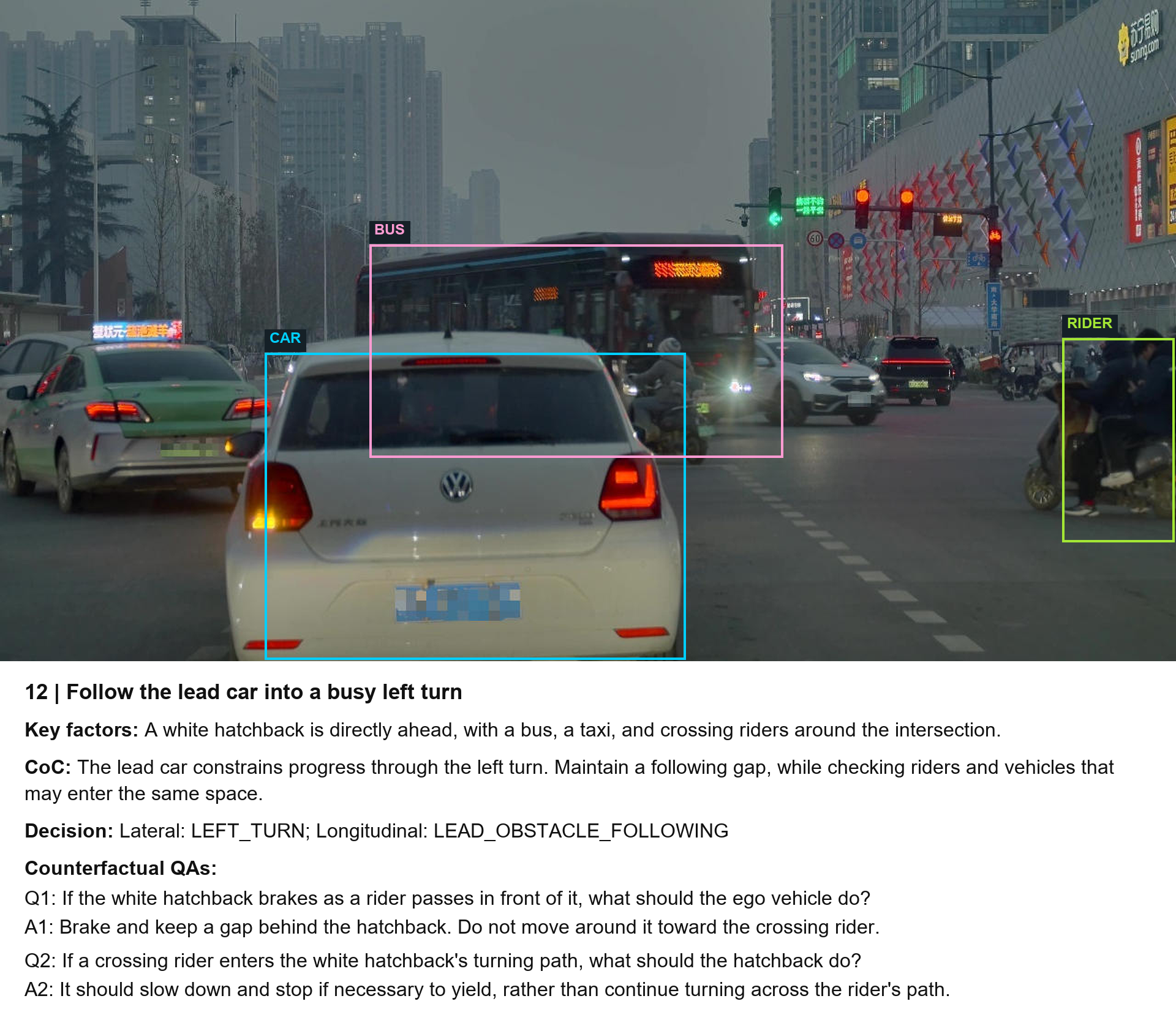} 
    \end{minipage}\hfill
    \caption{Qualitative case study on 6 sign-dense scenes (25--30). Each case presents the key factors that the ego vehicle must attend to in the current frame, the CoC, the lateral and longitudinal decisions the ego vehicle should take, and the counterfactual QA pairs.}
    \label{fig:coc_case_study_5}
  \end{figure}

  \begin{figure}[htbp]
    \centering
    \begin{minipage}{0.48\textwidth}
      \centering
      \includegraphics[width=\linewidth]{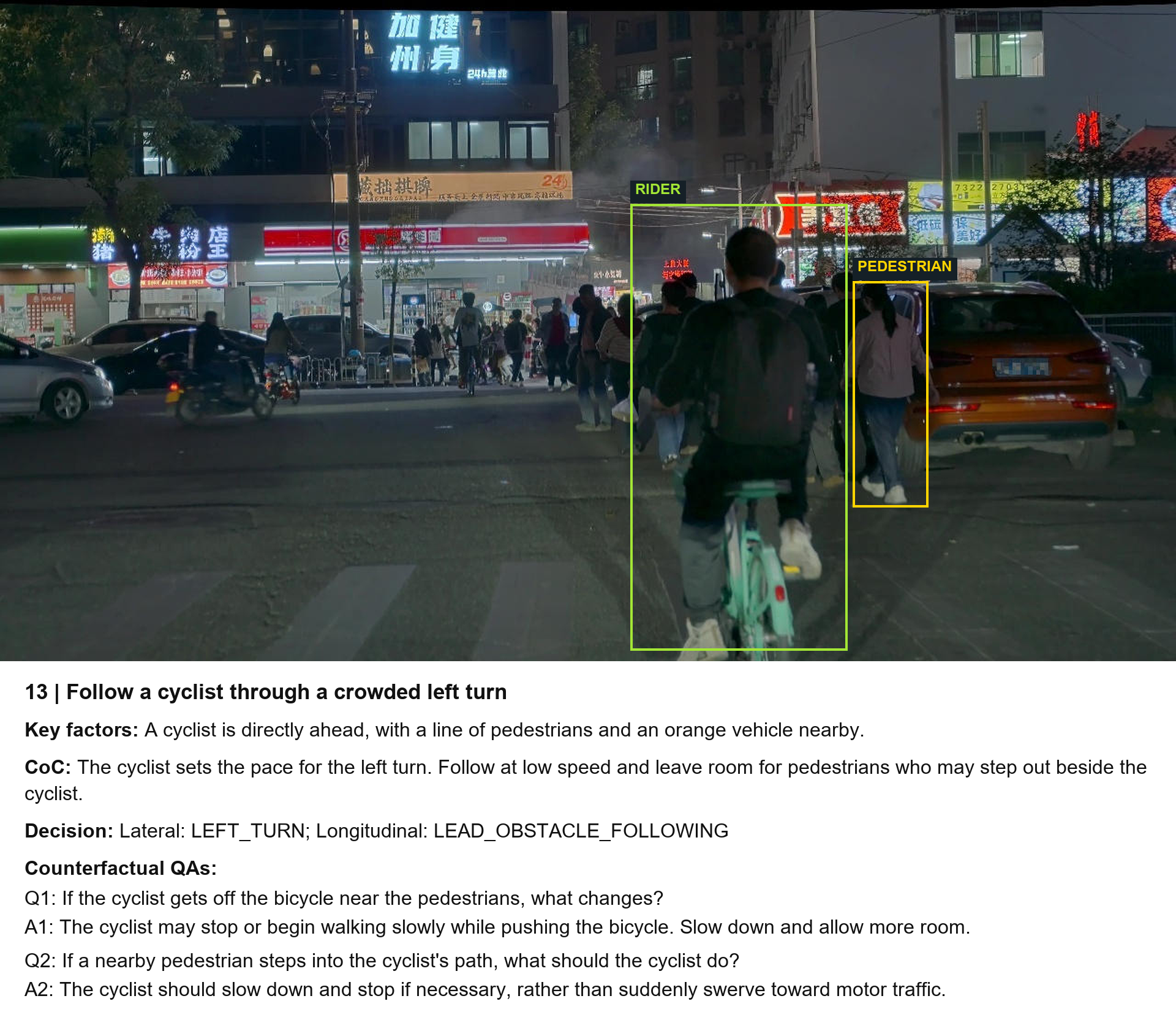}
    \end{minipage}\hfill
    \begin{minipage}{0.48\textwidth}
      \centering
      \includegraphics[width=\linewidth]{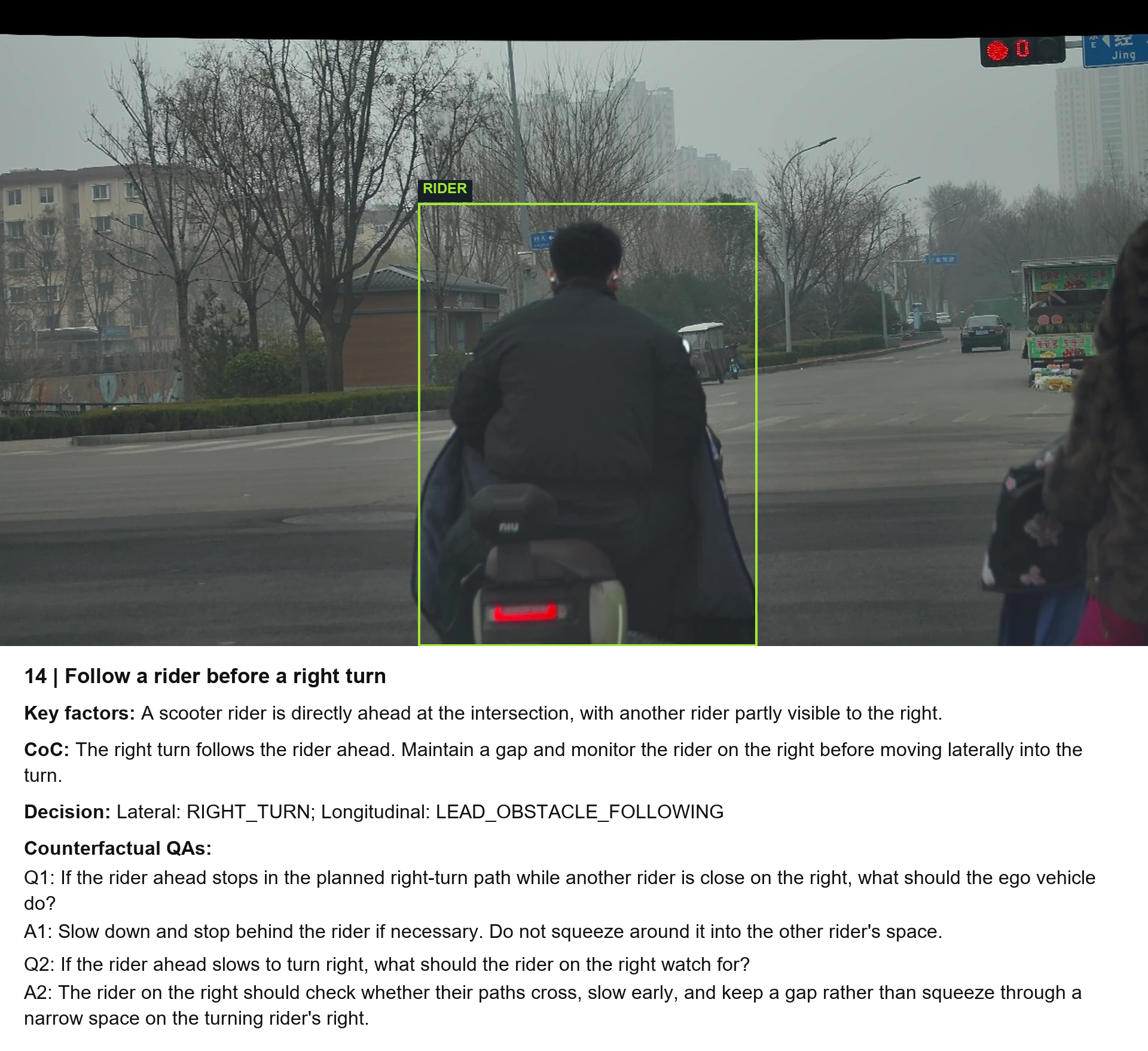}
    \end{minipage}\hfill
    \begin{minipage}{0.48\textwidth}
      \centering
      \includegraphics[width=\linewidth]{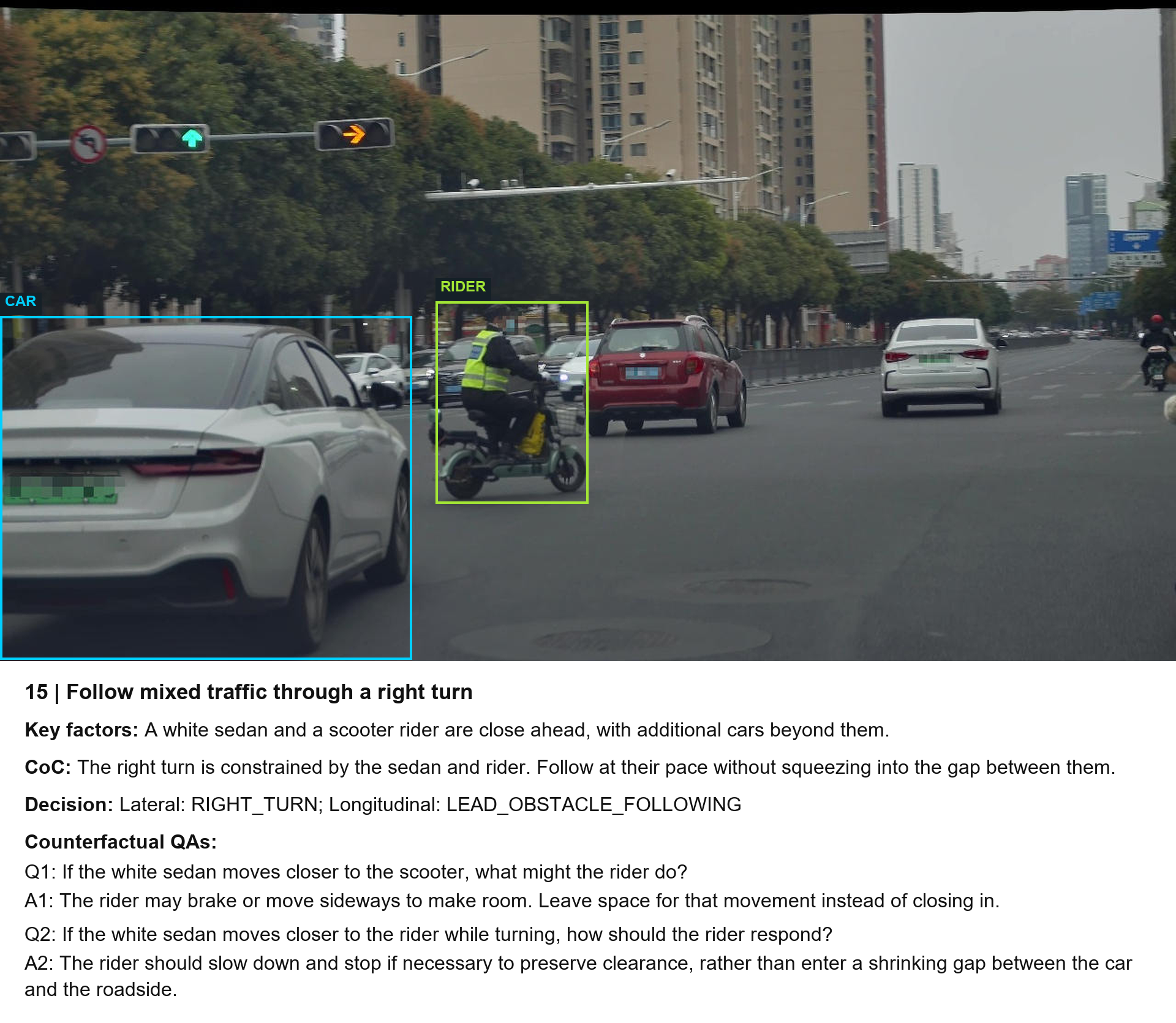}
    \end{minipage}\hfill
    \begin{minipage}{0.48\textwidth}
      \centering
      \includegraphics[width=\linewidth]{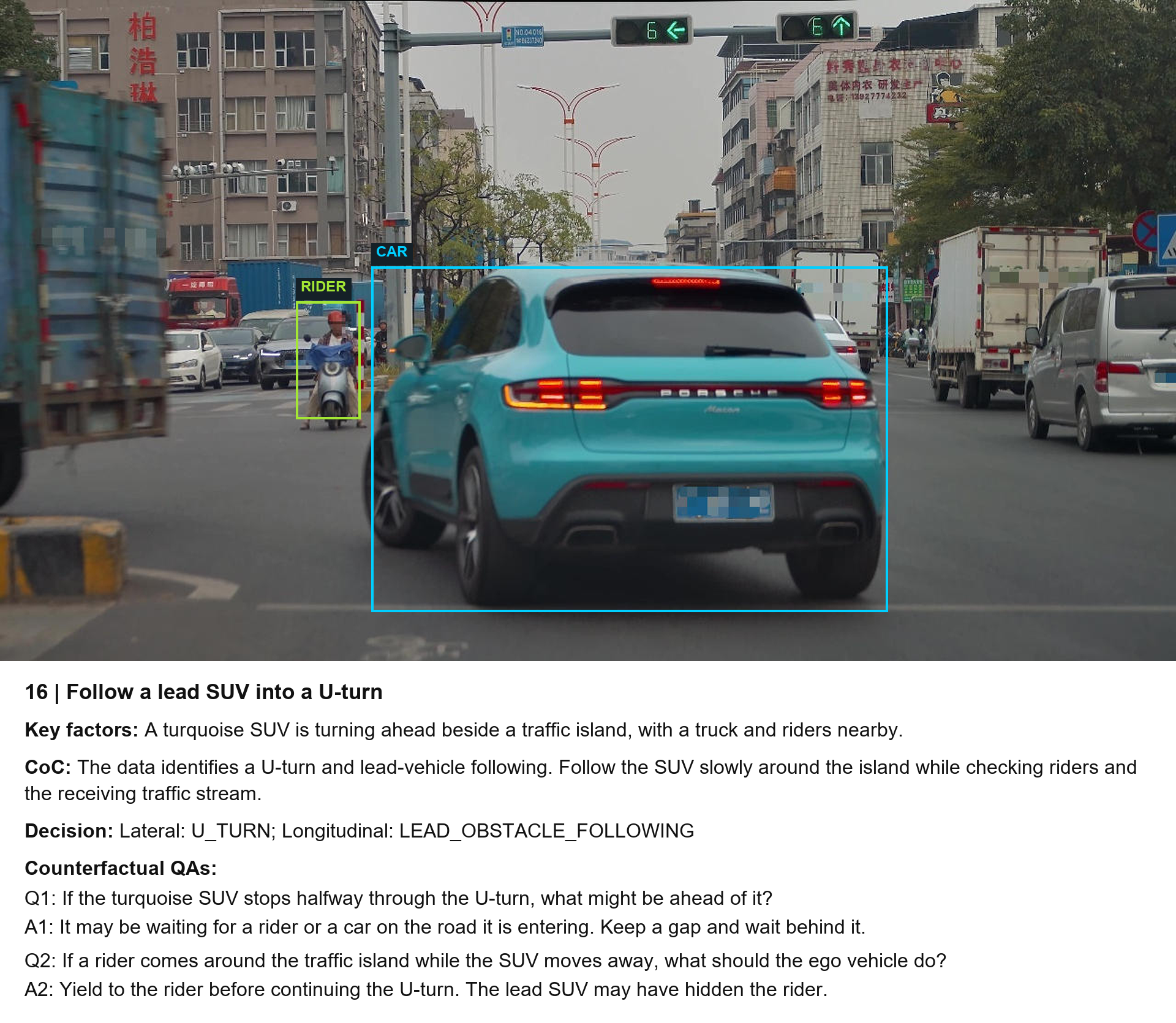}
    \end{minipage}\hfill
    \begin{minipage}{0.48\textwidth}
      \centering
      \includegraphics[width=\linewidth]{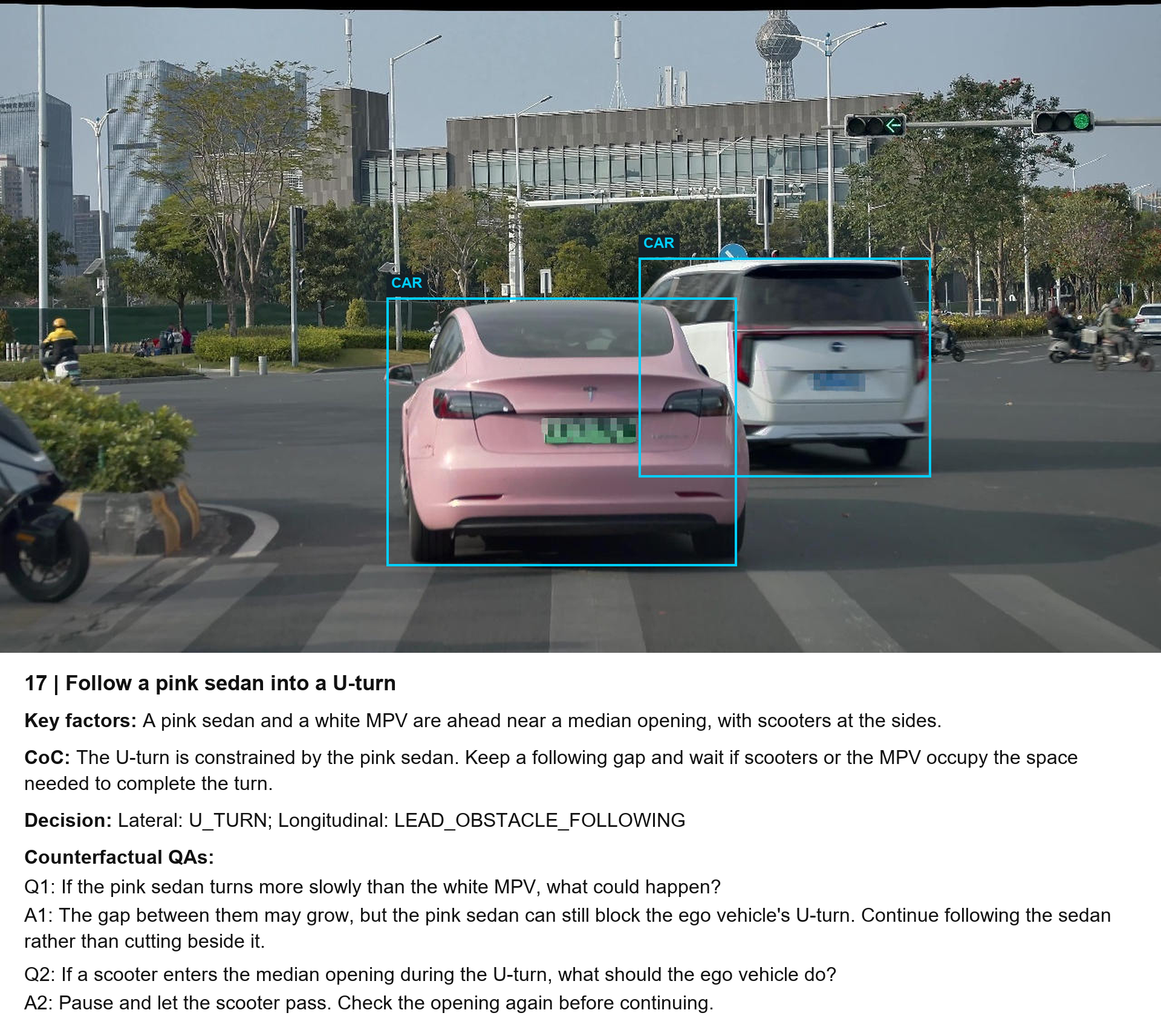}
    \end{minipage}\hfill
    \begin{minipage}{0.48\textwidth}
      \centering
      \includegraphics[width=\linewidth]{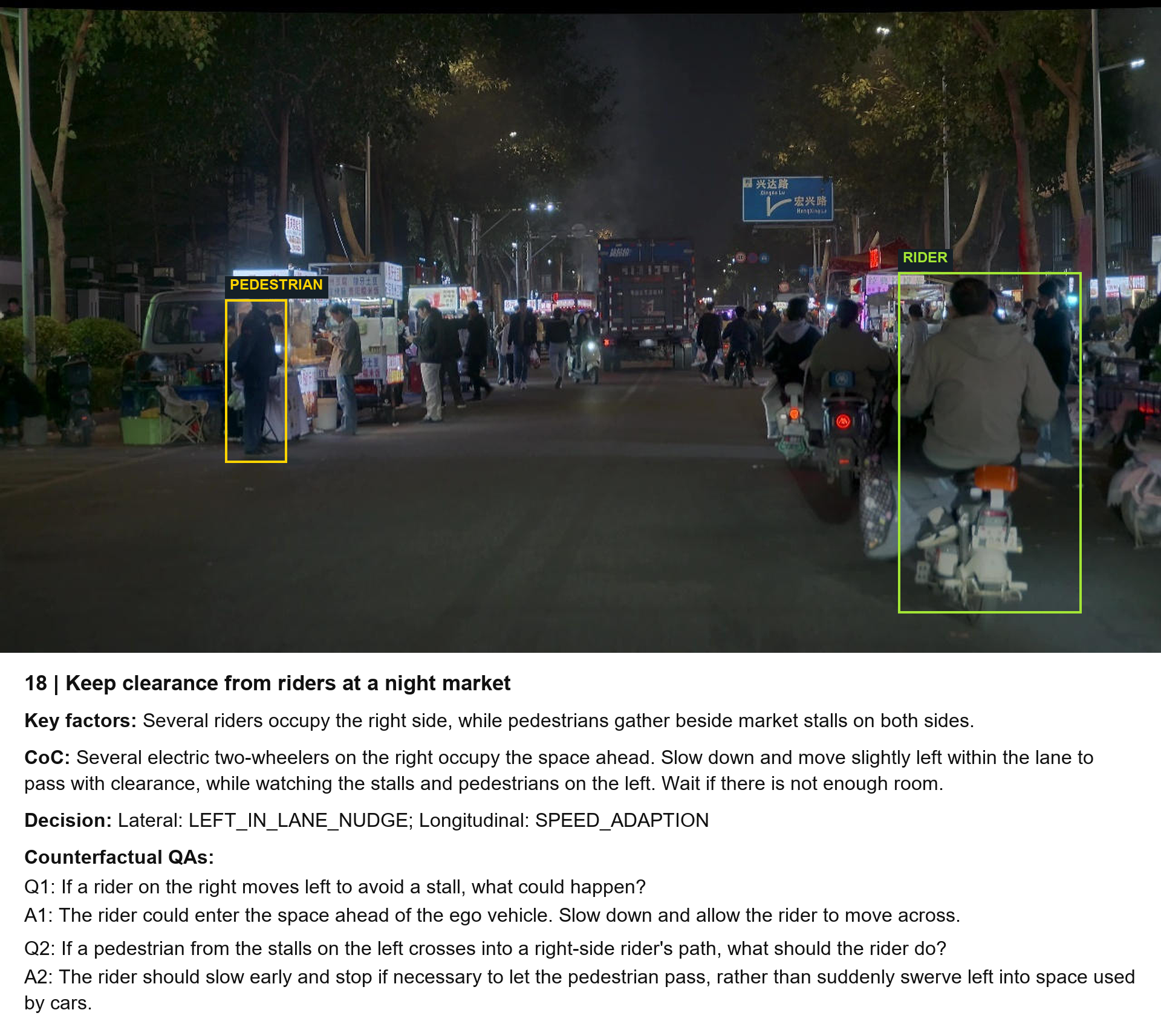} 
    \end{minipage}\hfill
    \caption{Qualitative case study on 6 sign-dense scenes (31--36). Each case presents the key factors that the ego vehicle must attend to in the current frame, the CoC, the lateral and longitudinal decisions the ego vehicle should take, and the counterfactual QA pairs.}
    \label{fig:coc_case_study_6}
  \end{figure}

  \begin{figure}[htbp]
    \centering
    \begin{minipage}{0.48\textwidth}
      \centering
      \includegraphics[width=\linewidth]{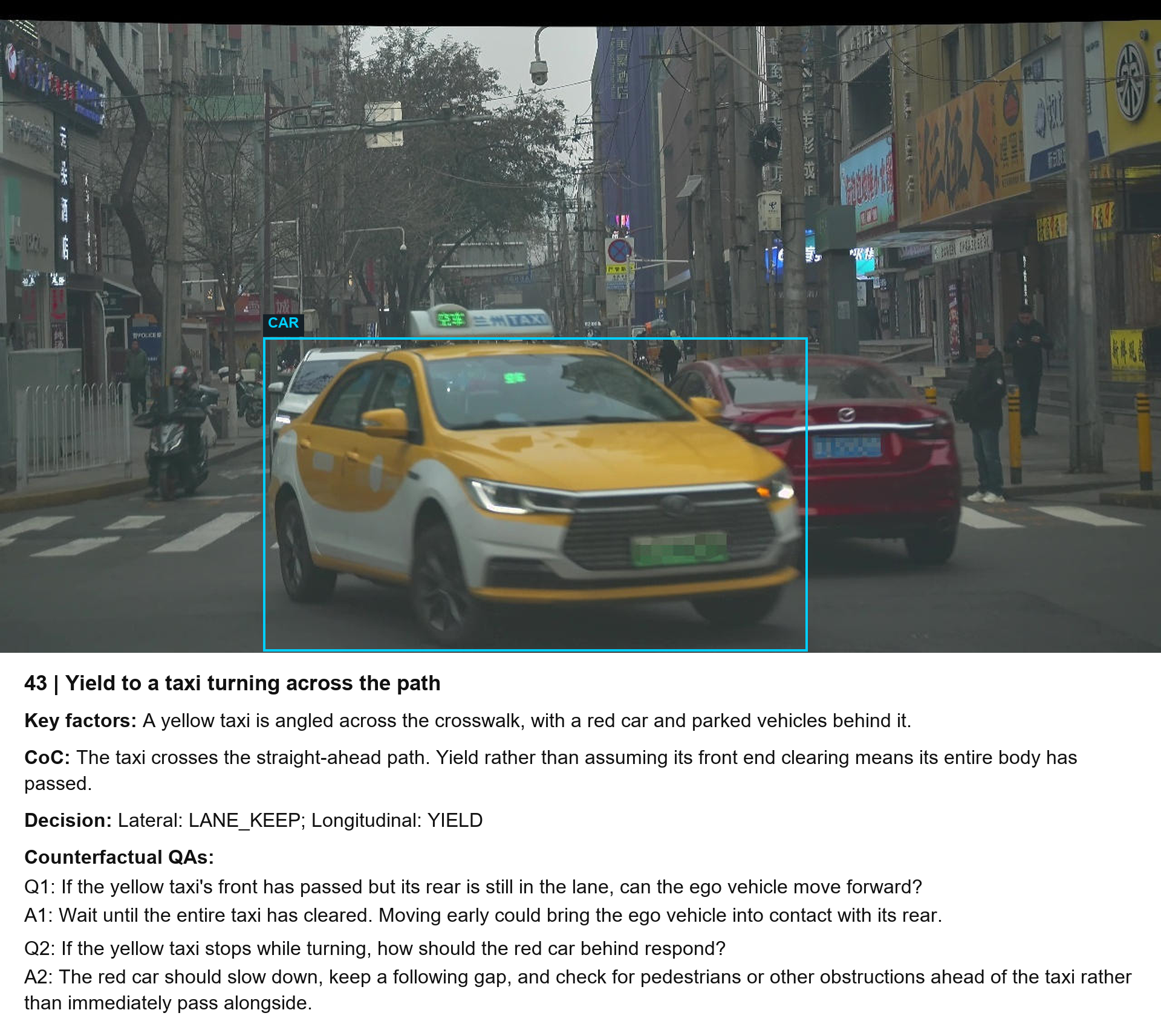}
    \end{minipage}\hfill
    \begin{minipage}{0.48\textwidth}
      \centering
      \includegraphics[width=\linewidth]{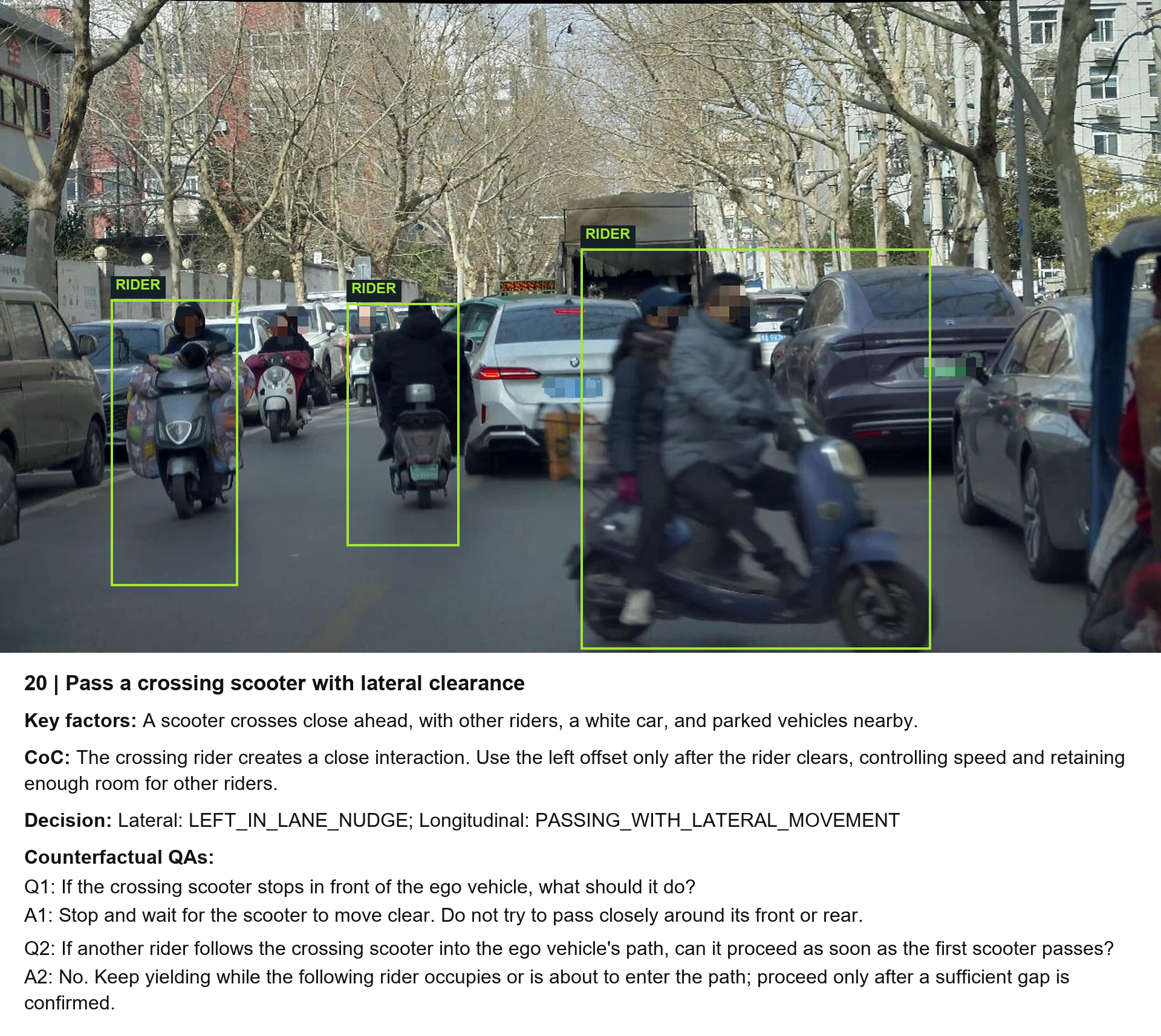}
    \end{minipage}\hfill
    \begin{minipage}{0.48\textwidth}
      \centering
      \includegraphics[width=\linewidth]{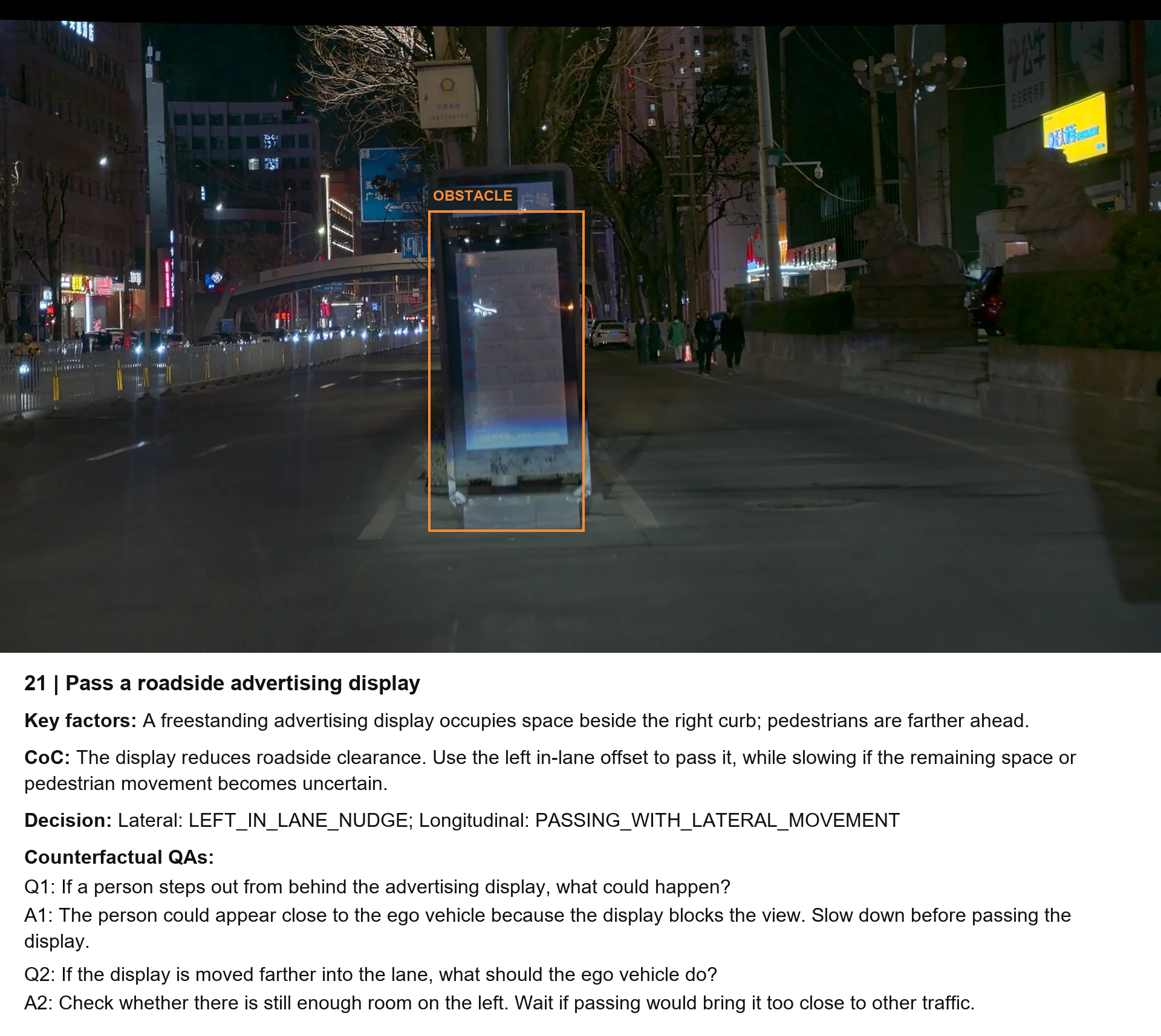}
    \end{minipage}\hfill
    \begin{minipage}{0.48\textwidth}
      \centering
      \includegraphics[width=\linewidth]{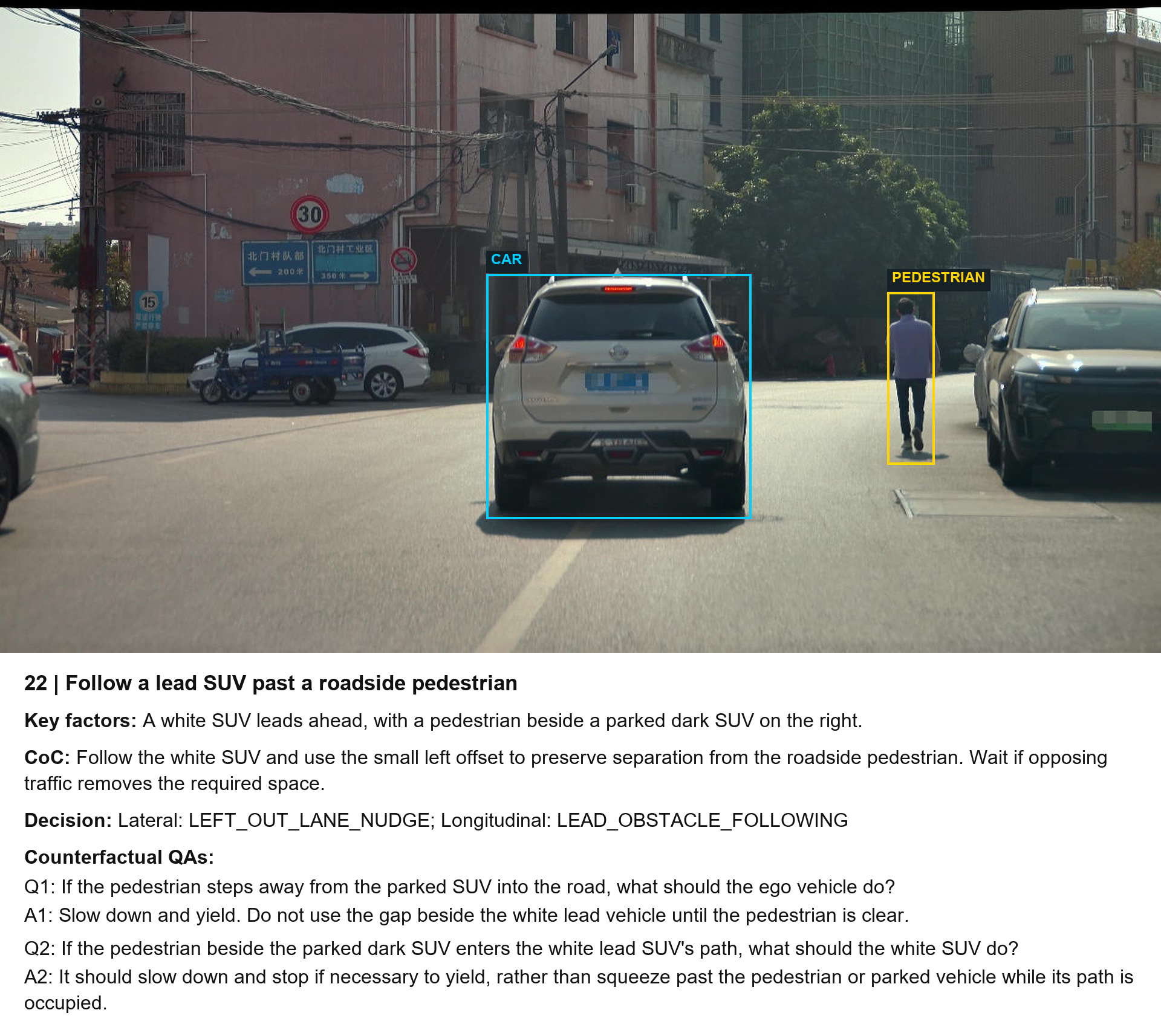}
    \end{minipage}\hfill
    \begin{minipage}{0.48\textwidth}
      \centering
      \includegraphics[width=\linewidth]{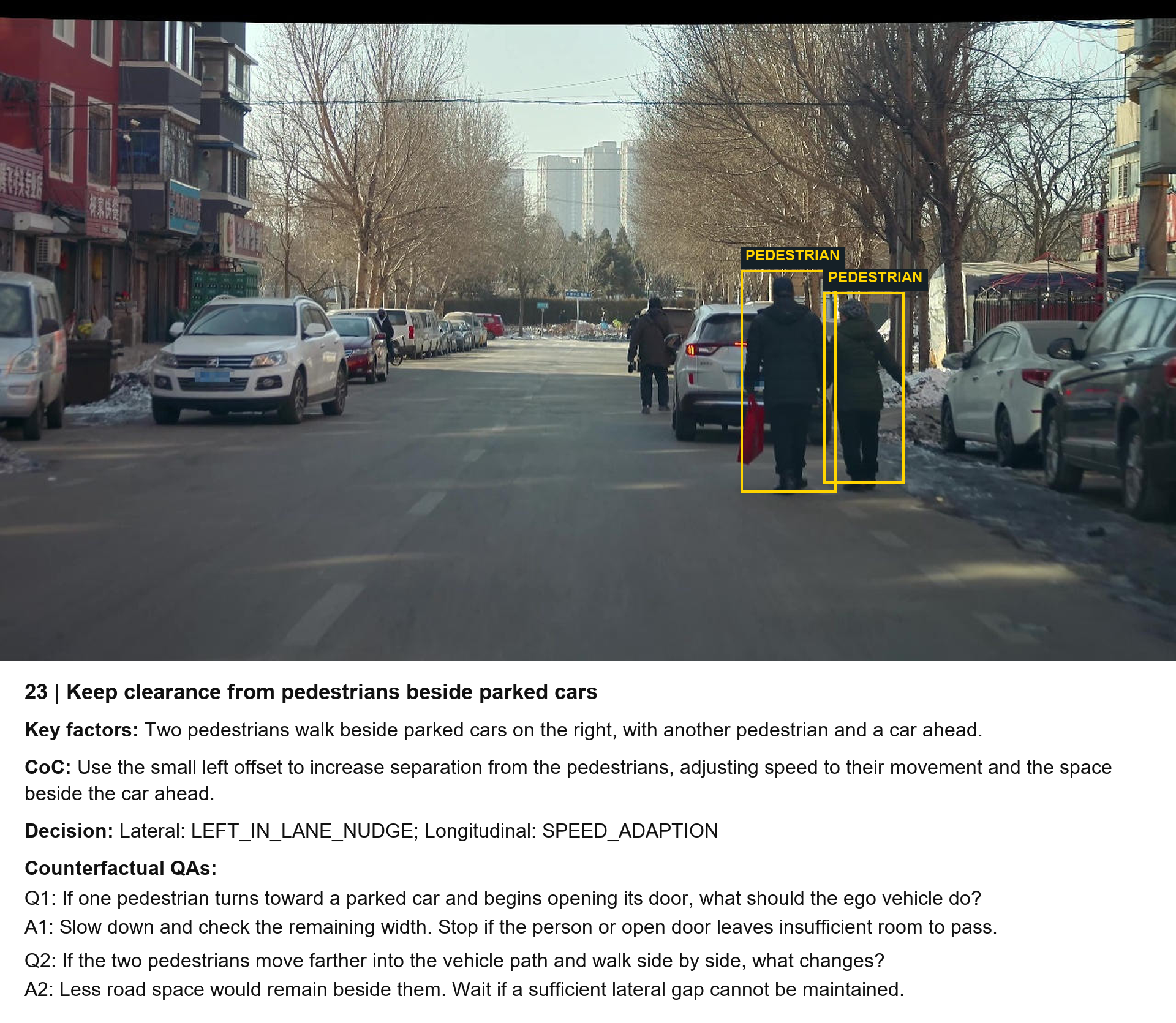}
    \end{minipage}\hfill
    \begin{minipage}{0.48\textwidth}
      \centering
      \includegraphics[width=\linewidth]{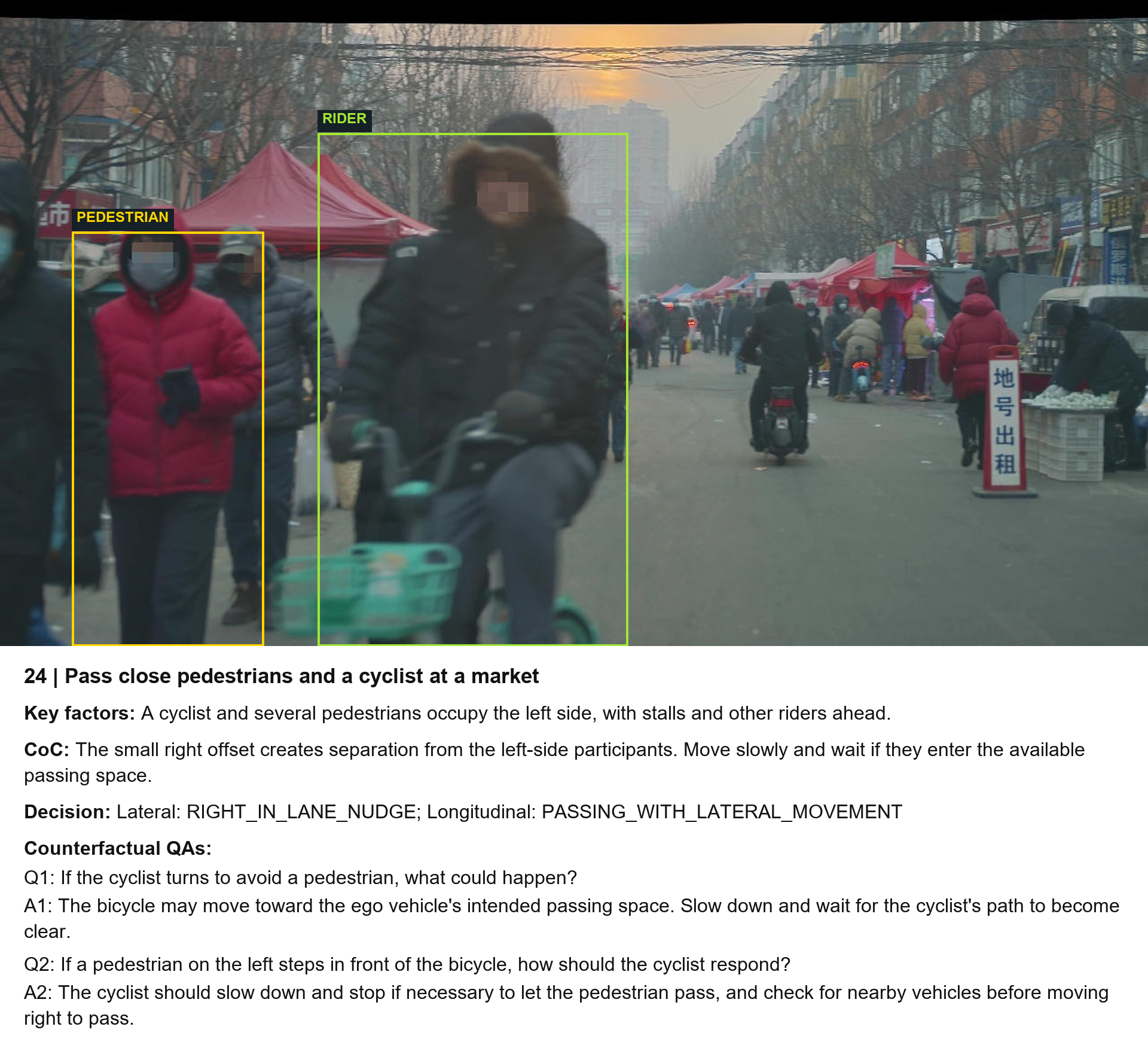} 
    \end{minipage}\hfill
    \caption{Qualitative case study on 6 sign-dense scenes (37--42). Each case presents the key factors that the ego vehicle must attend to in the current frame, the CoC, the lateral and longitudinal decisions the ego vehicle should take, and the counterfactual QA pairs.}
    \label{fig:coc_case_study_7}
  \end{figure}

  \begin{figure}[htbp]
    \centering
    \begin{minipage}{0.48\textwidth}
      \centering
      \includegraphics[width=\linewidth]{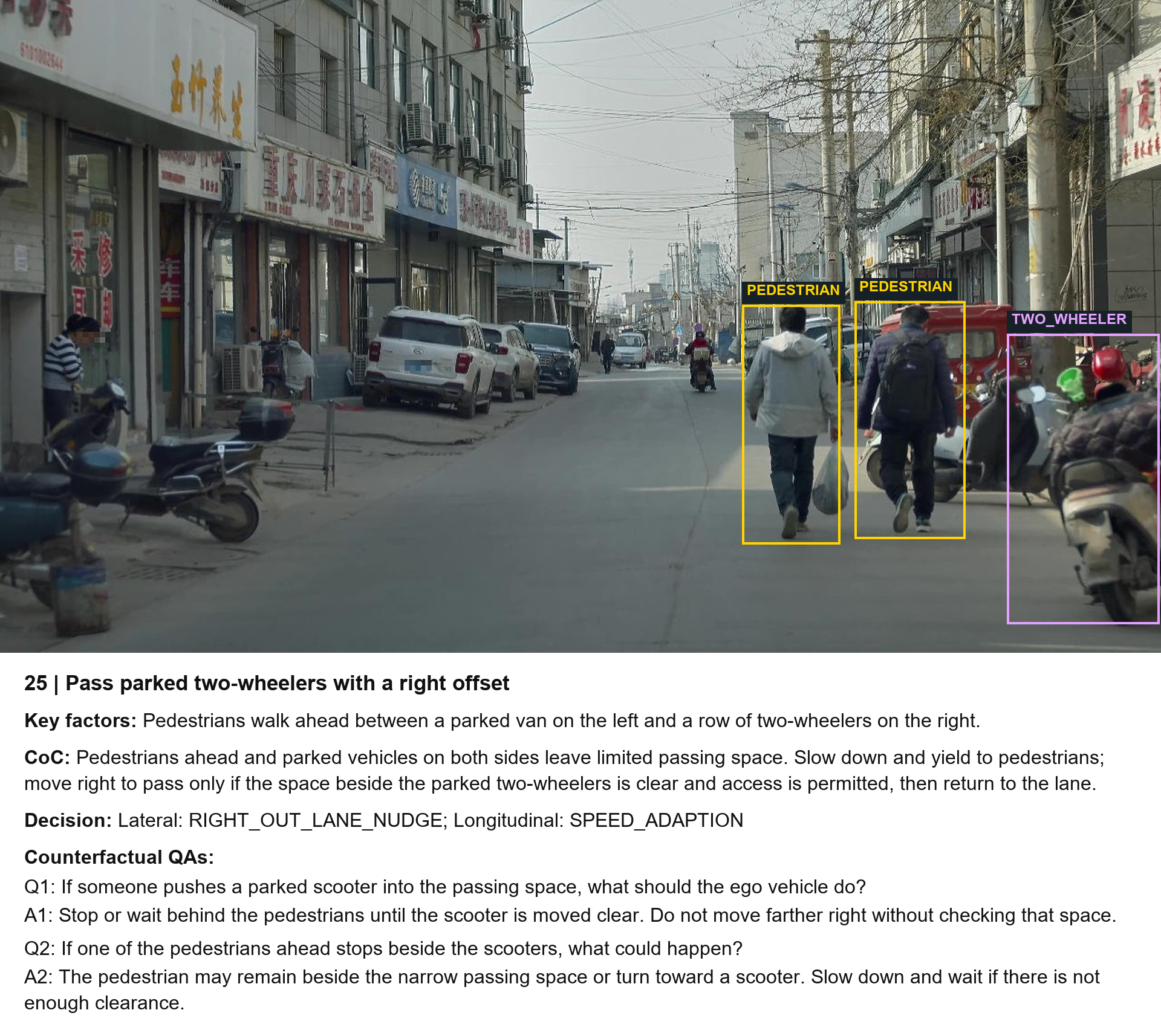}
    \end{minipage}\hfill
    \begin{minipage}{0.48\textwidth}
      \centering
      \includegraphics[width=\linewidth]{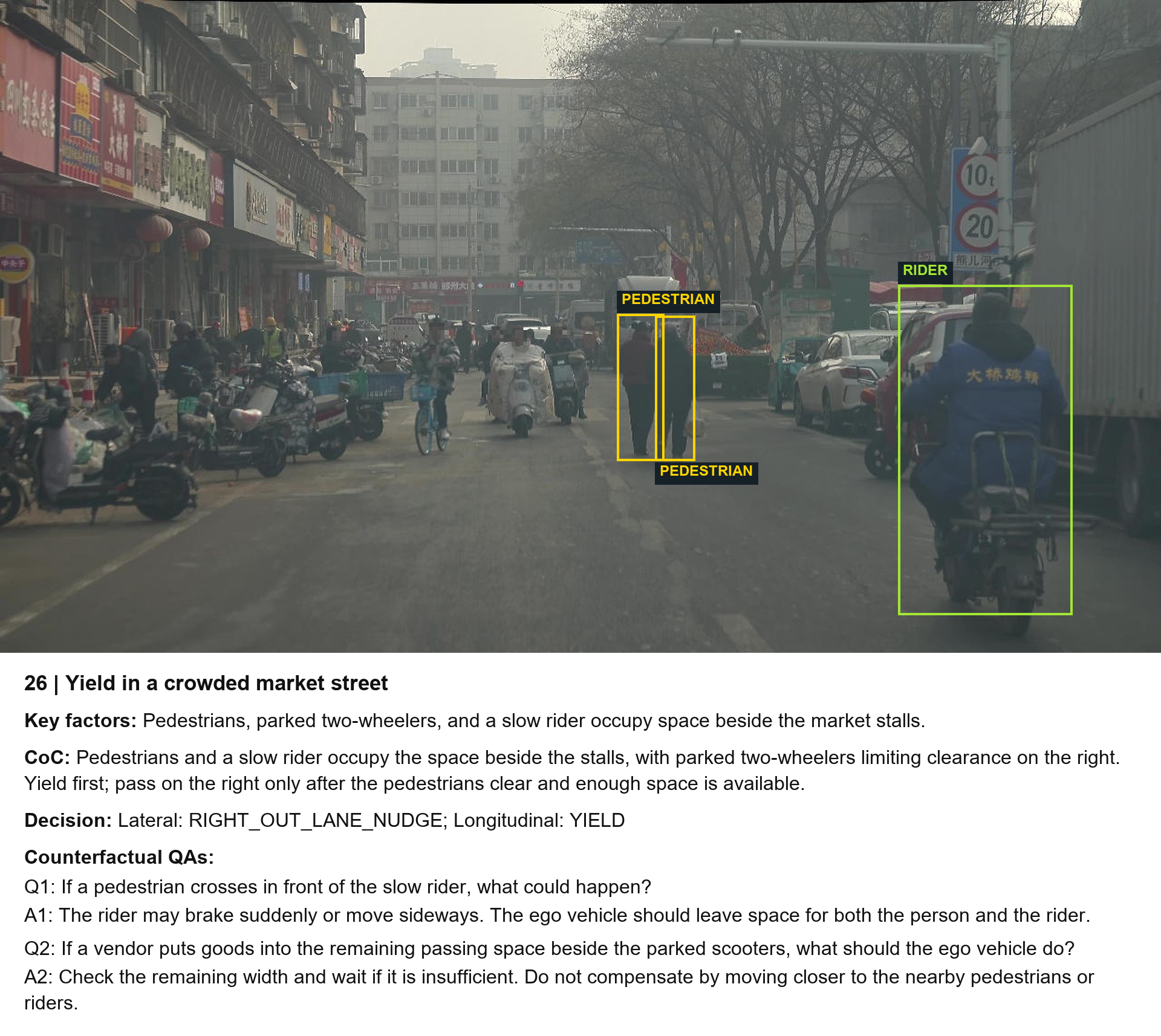}
    \end{minipage}\hfill
    \begin{minipage}{0.48\textwidth}
      \centering
      \includegraphics[width=\linewidth]{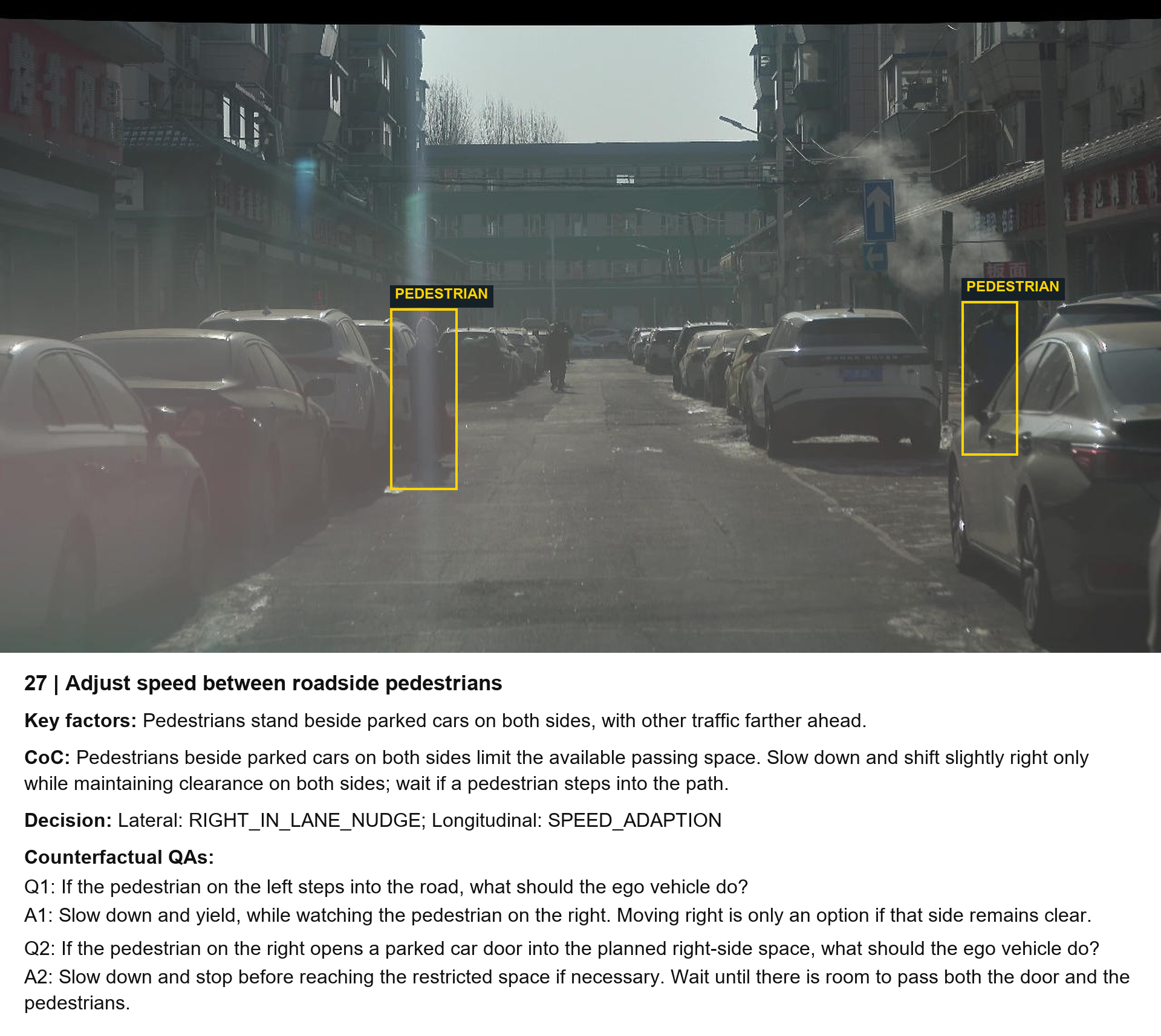}
    \end{minipage}\hfill
    \begin{minipage}{0.48\textwidth}
      \centering
      \includegraphics[width=\linewidth]{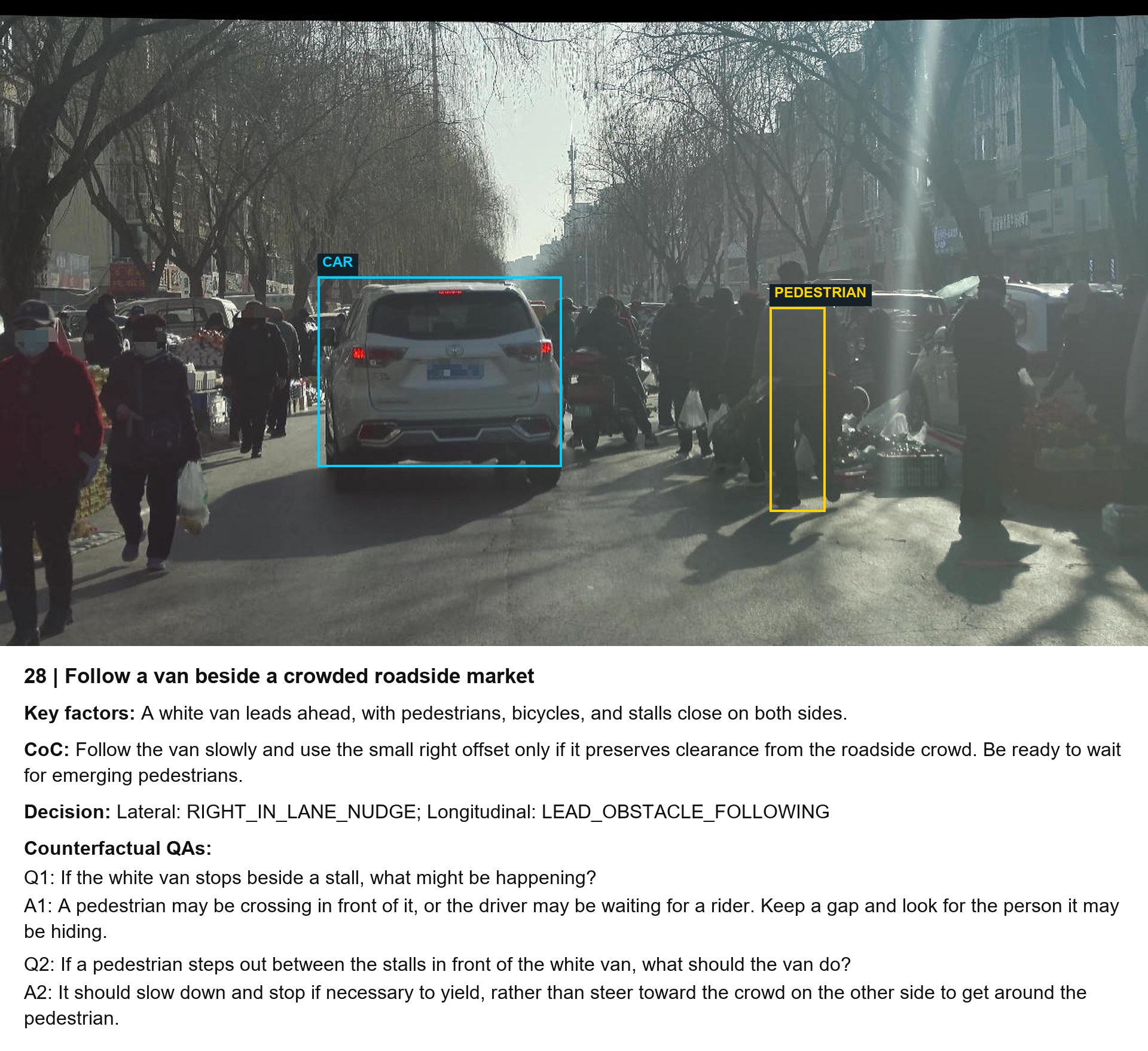}
    \end{minipage}\hfill
    \begin{minipage}{0.48\textwidth}
      \centering
      \includegraphics[width=\linewidth]{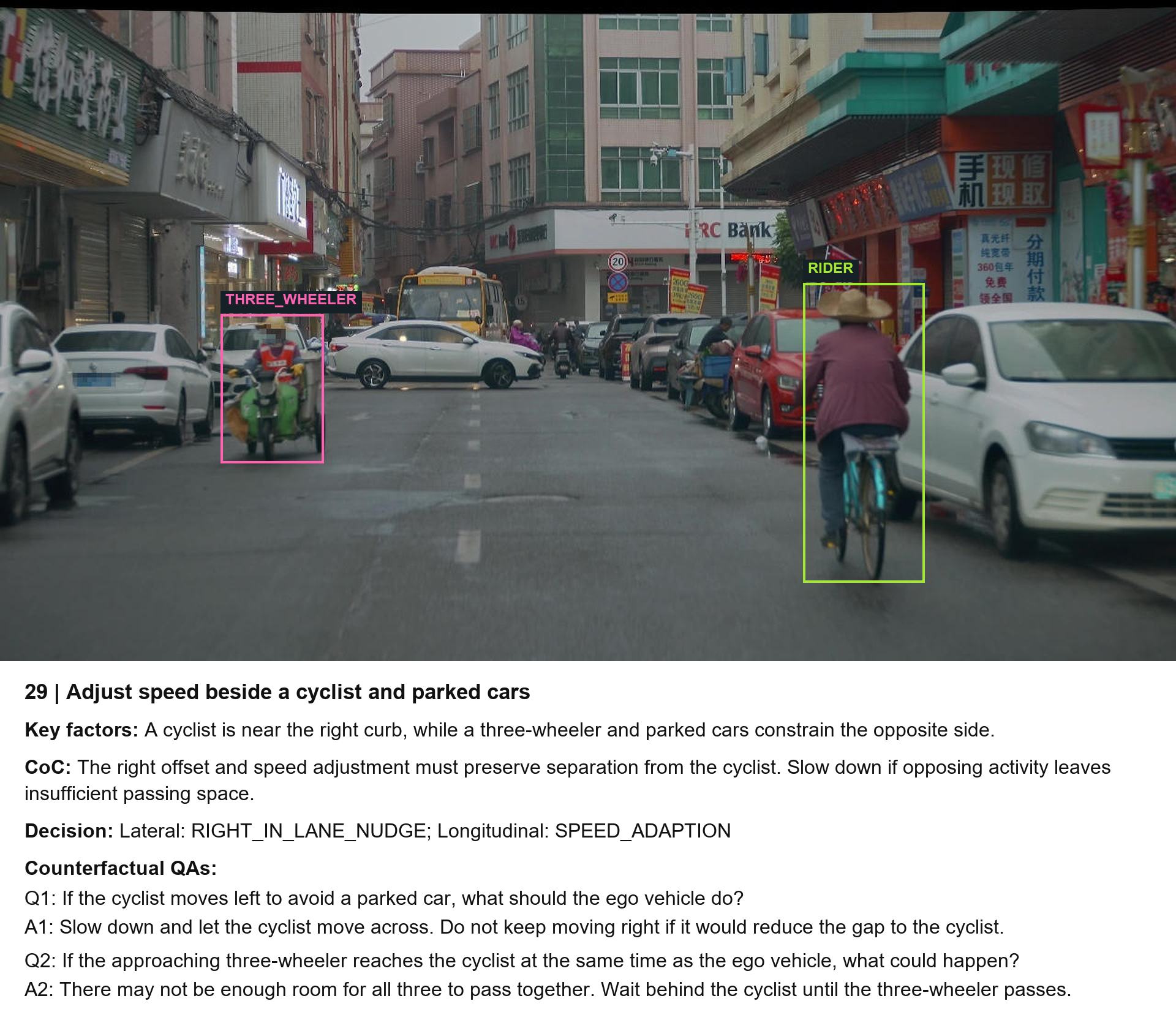}
    \end{minipage}\hfill
    \begin{minipage}{0.48\textwidth}
      \centering
      \includegraphics[width=\linewidth]{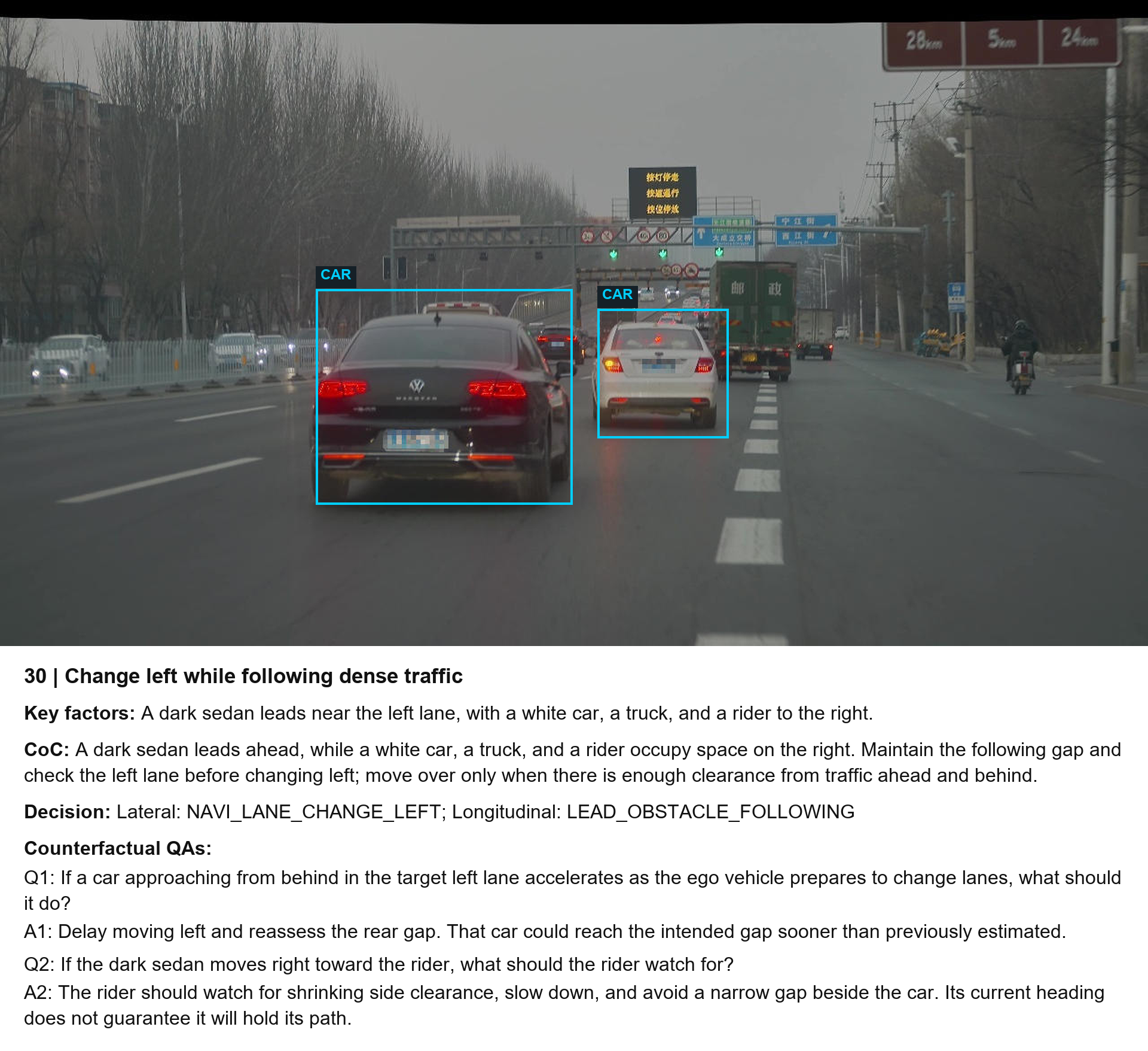} 
    \end{minipage}\hfill
    \caption{Qualitative case study on 6 sign-dense scenes (43--48). Each case presents the key factors that the ego vehicle must attend to in the current frame, the CoC, the lateral and longitudinal decisions the ego vehicle should take, and the counterfactual QA pairs.}
    \label{fig:coc_case_study_8}
  \end{figure}

  \begin{figure}[htbp]
    \centering
    \begin{minipage}{0.48\textwidth}
      \centering
      \includegraphics[width=\linewidth]{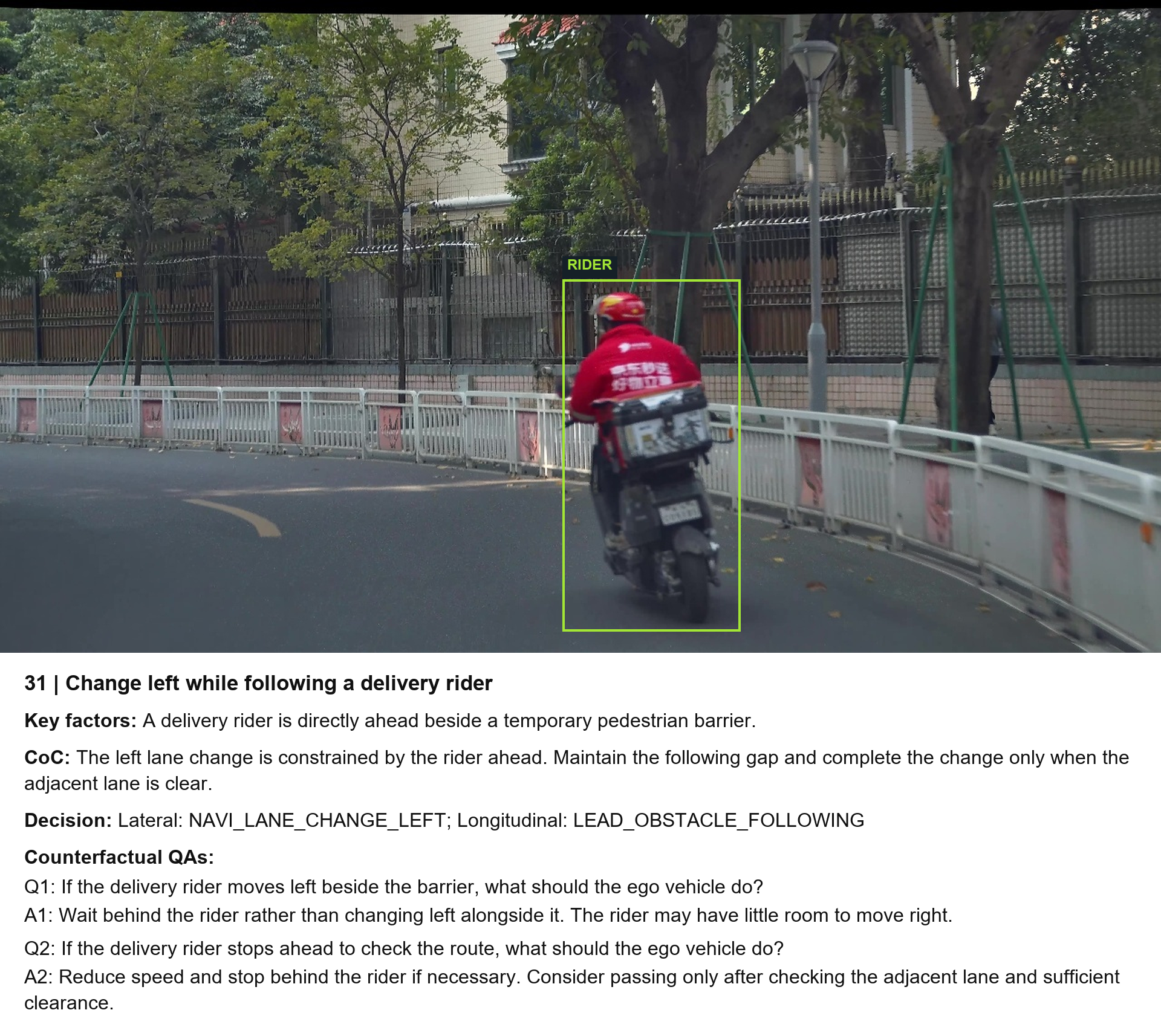}
    \end{minipage}\hfill
    \begin{minipage}{0.48\textwidth}
      \centering
      \includegraphics[width=\linewidth]{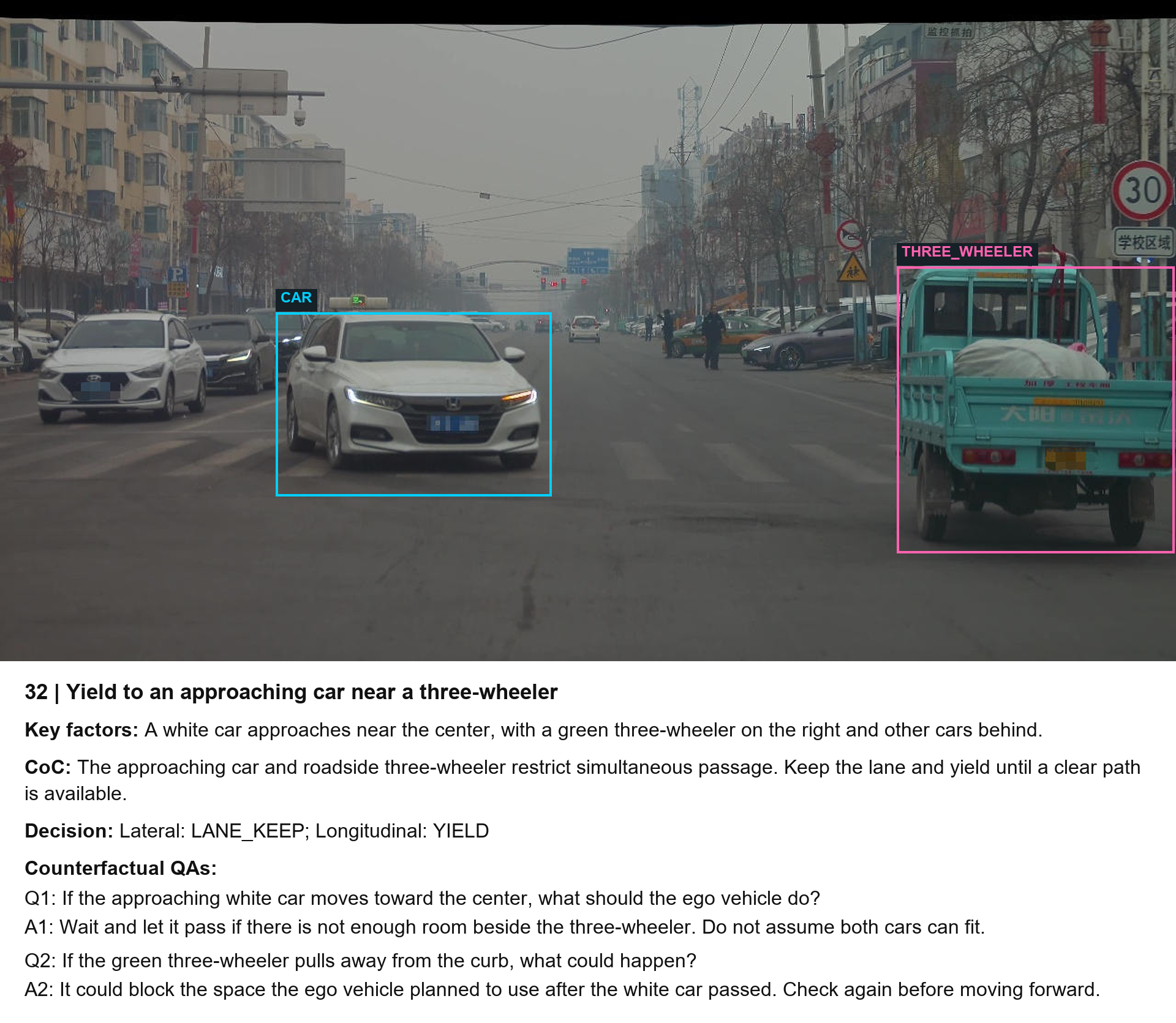}
    \end{minipage}\hfill
    \begin{minipage}{0.48\textwidth}
      \centering
      \includegraphics[width=\linewidth]{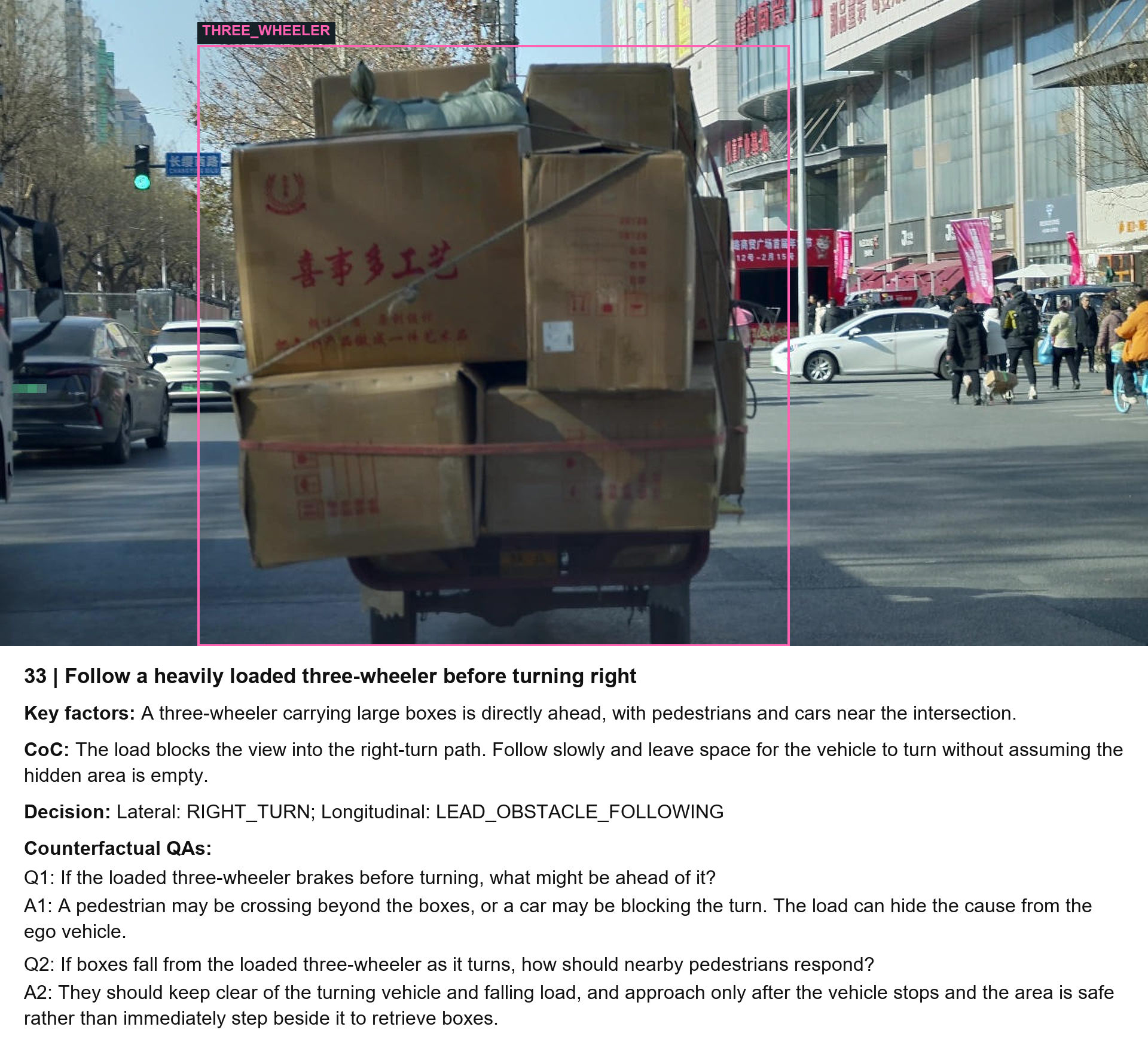}
    \end{minipage}\hfill
    \begin{minipage}{0.48\textwidth}
      \centering
      \includegraphics[width=\linewidth]{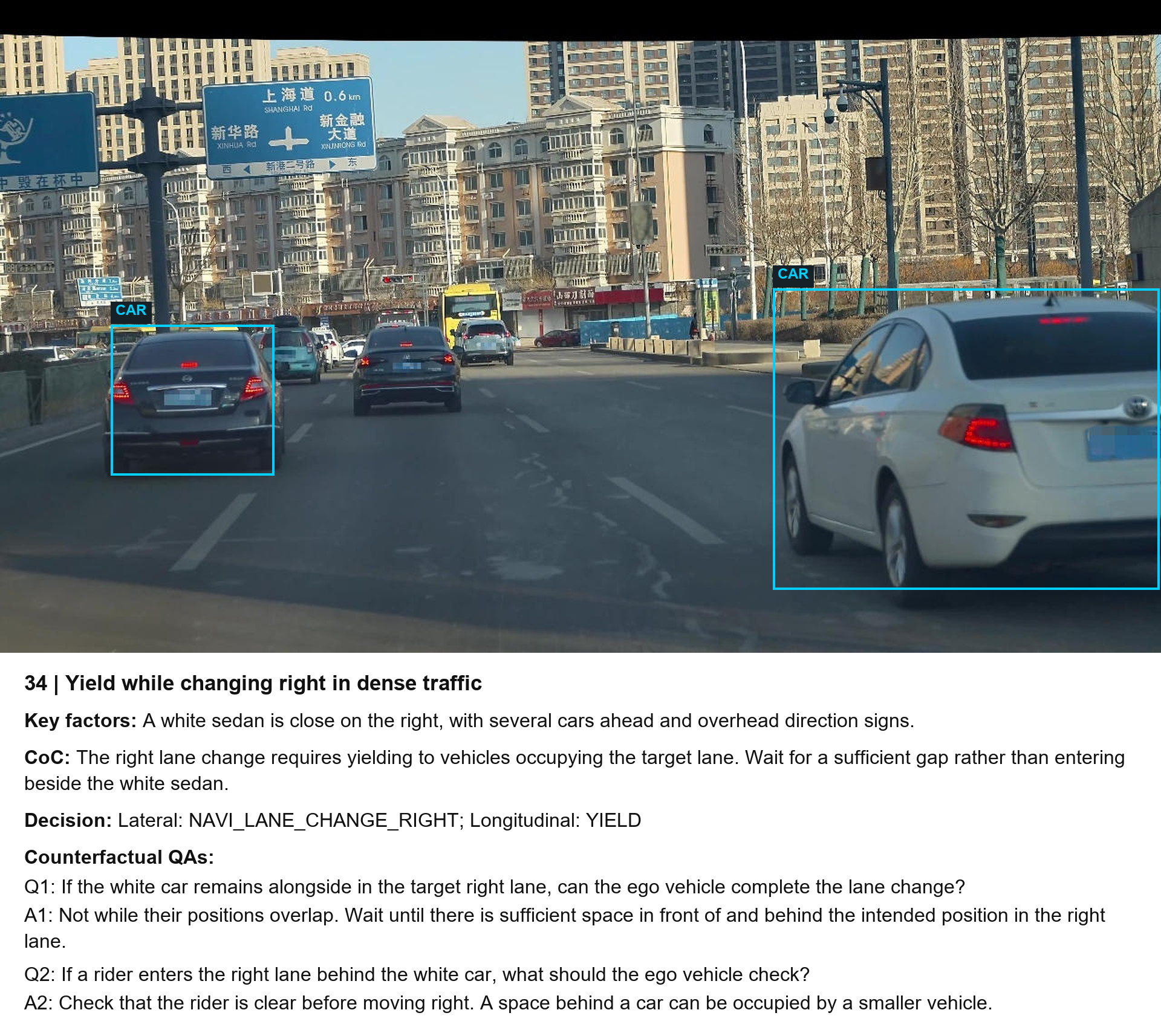}
    \end{minipage}\hfill
    \begin{minipage}{0.48\textwidth}
      \centering
      \includegraphics[width=\linewidth]{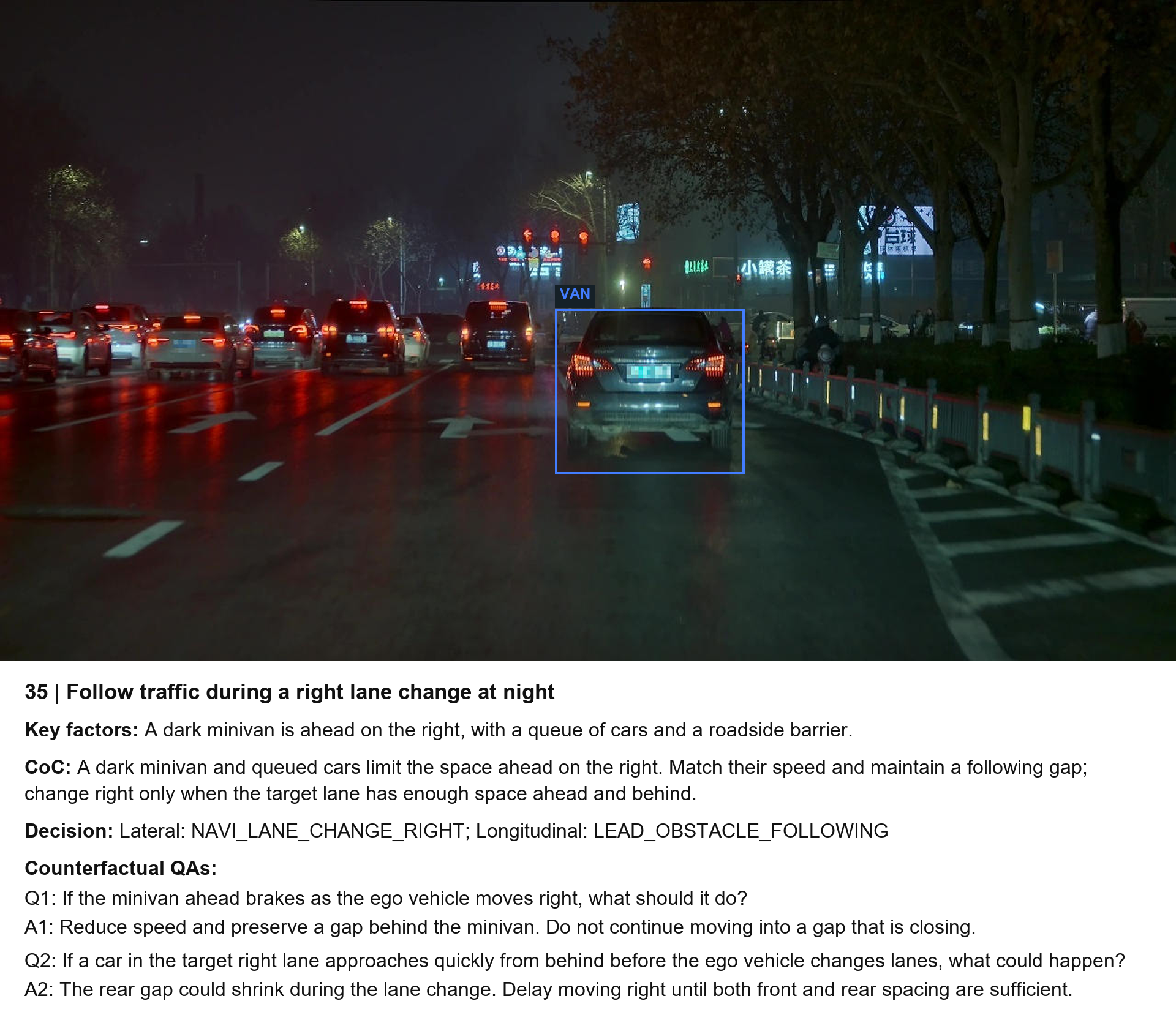}
    \end{minipage}\hfill
    \begin{minipage}{0.48\textwidth}
      \centering
      \includegraphics[width=\linewidth]{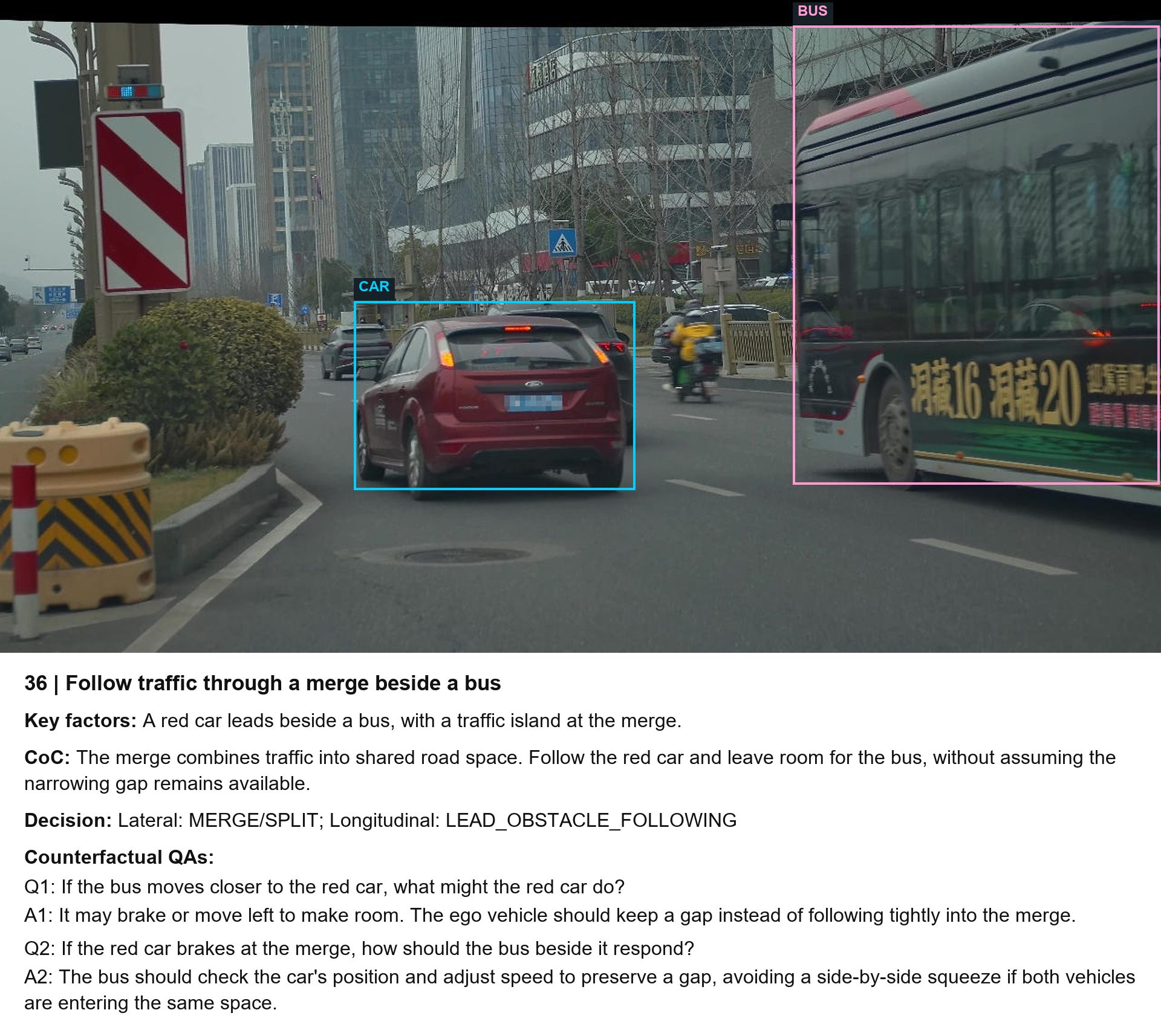} 
    \end{minipage}\hfill
    \caption{Qualitative case study on 6 sign-dense scenes (49--54). Each case presents the key factors that the ego vehicle must attend to in the current frame, the CoC, the lateral and longitudinal decisions the ego vehicle should take, and the counterfactual QA pairs.}
    \label{fig:coc_case_study_9}
  \end{figure}

  \begin{figure}[htbp]
    \centering
    \begin{minipage}{0.48\textwidth}
      \centering
      \includegraphics[width=\linewidth]{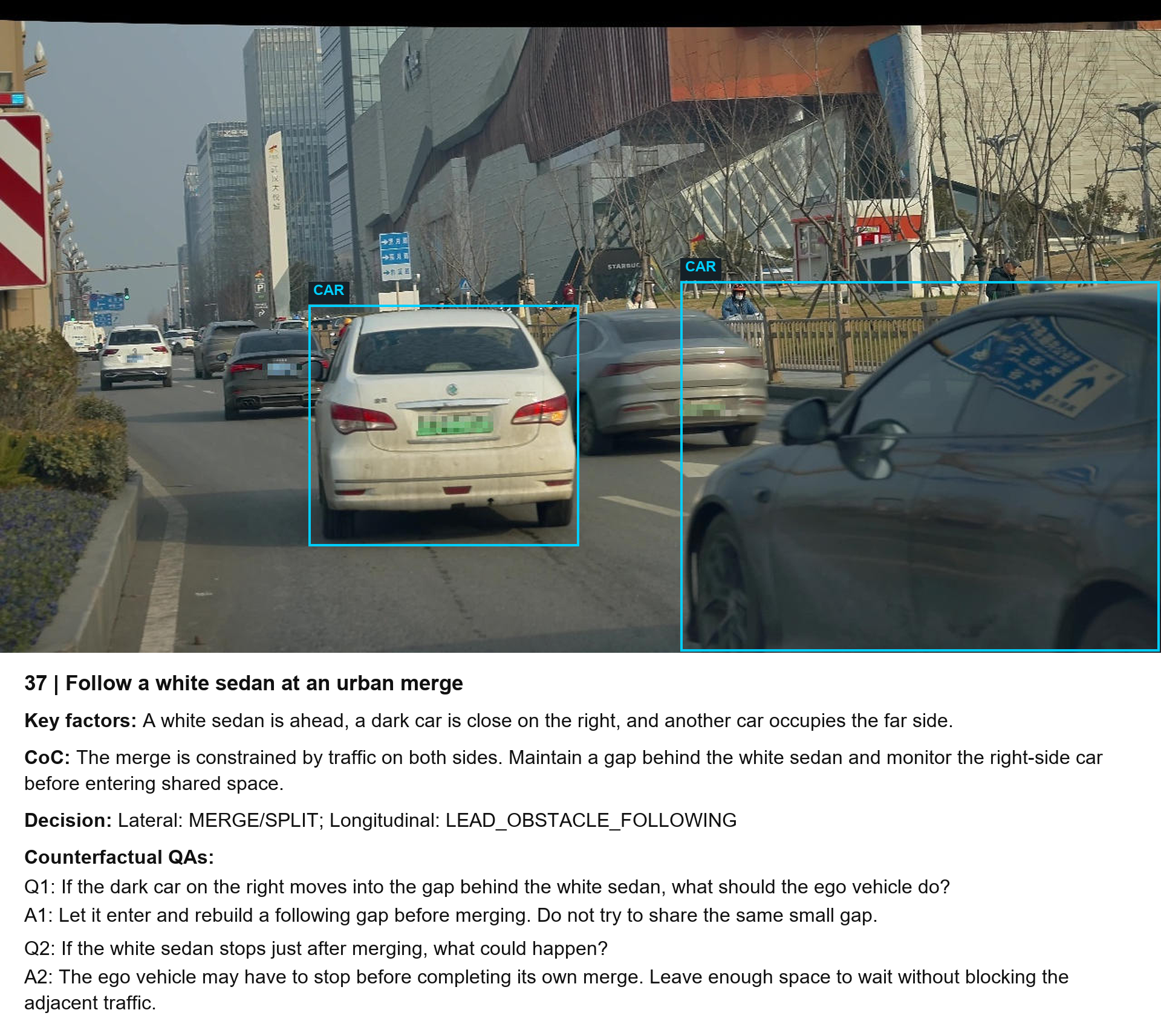}
    \end{minipage}\hfill
    \begin{minipage}{0.48\textwidth}
      \centering
      \includegraphics[width=\linewidth]{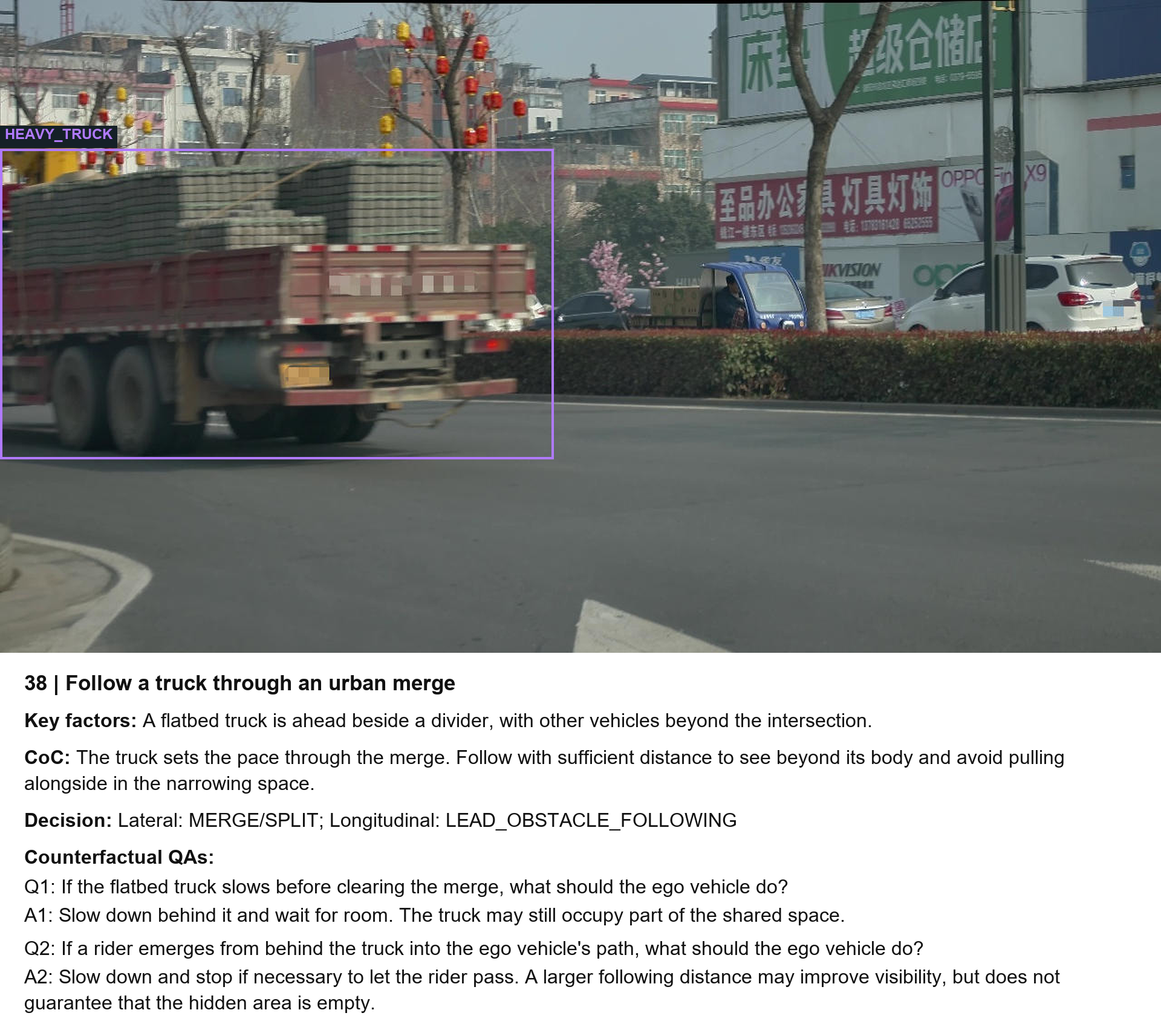}
    \end{minipage}\hfill
    \begin{minipage}{0.48\textwidth}
      \centering
      \includegraphics[width=\linewidth]{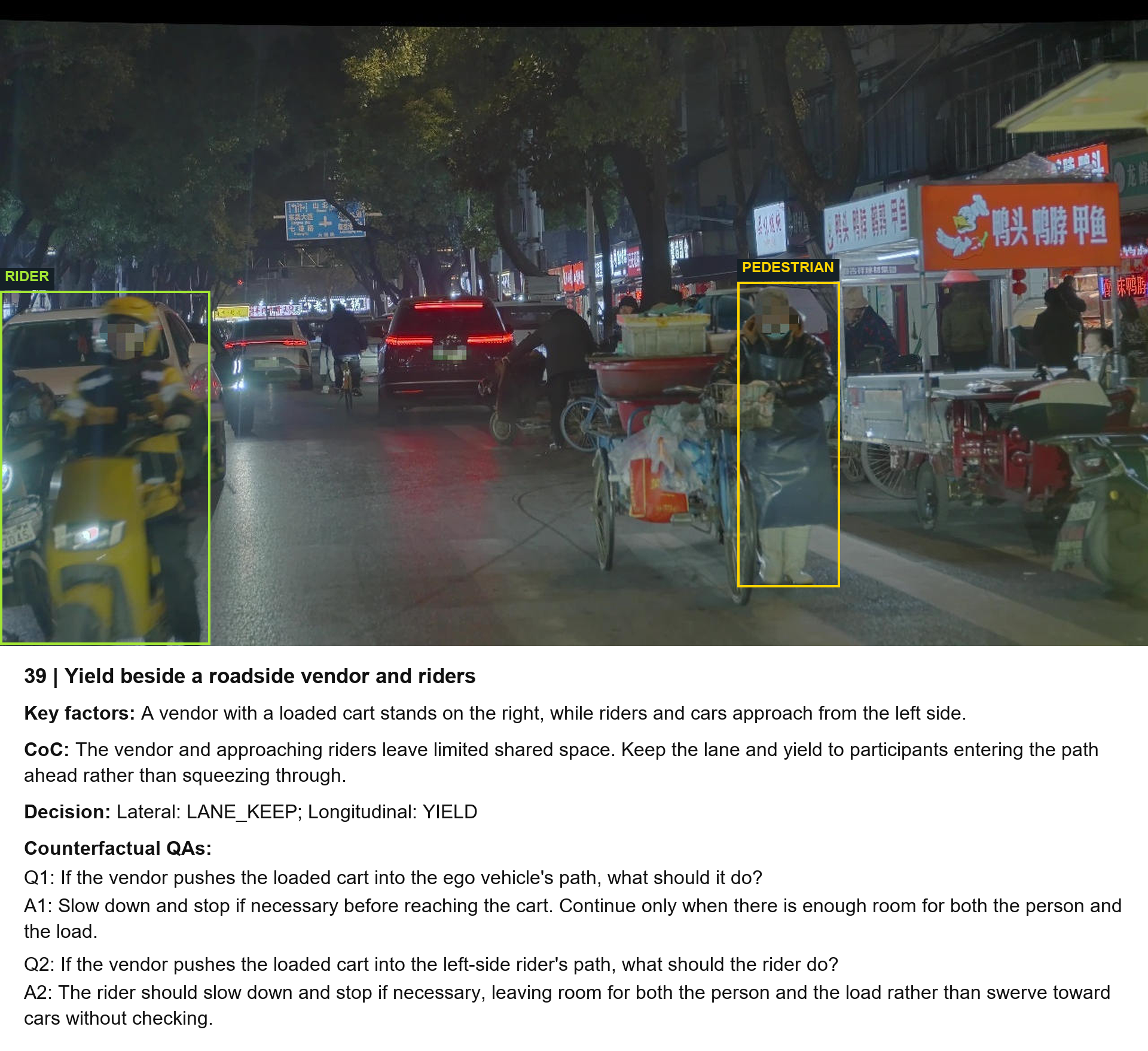}
    \end{minipage}\hfill
    \begin{minipage}{0.48\textwidth}
      \centering
      \includegraphics[width=\linewidth]{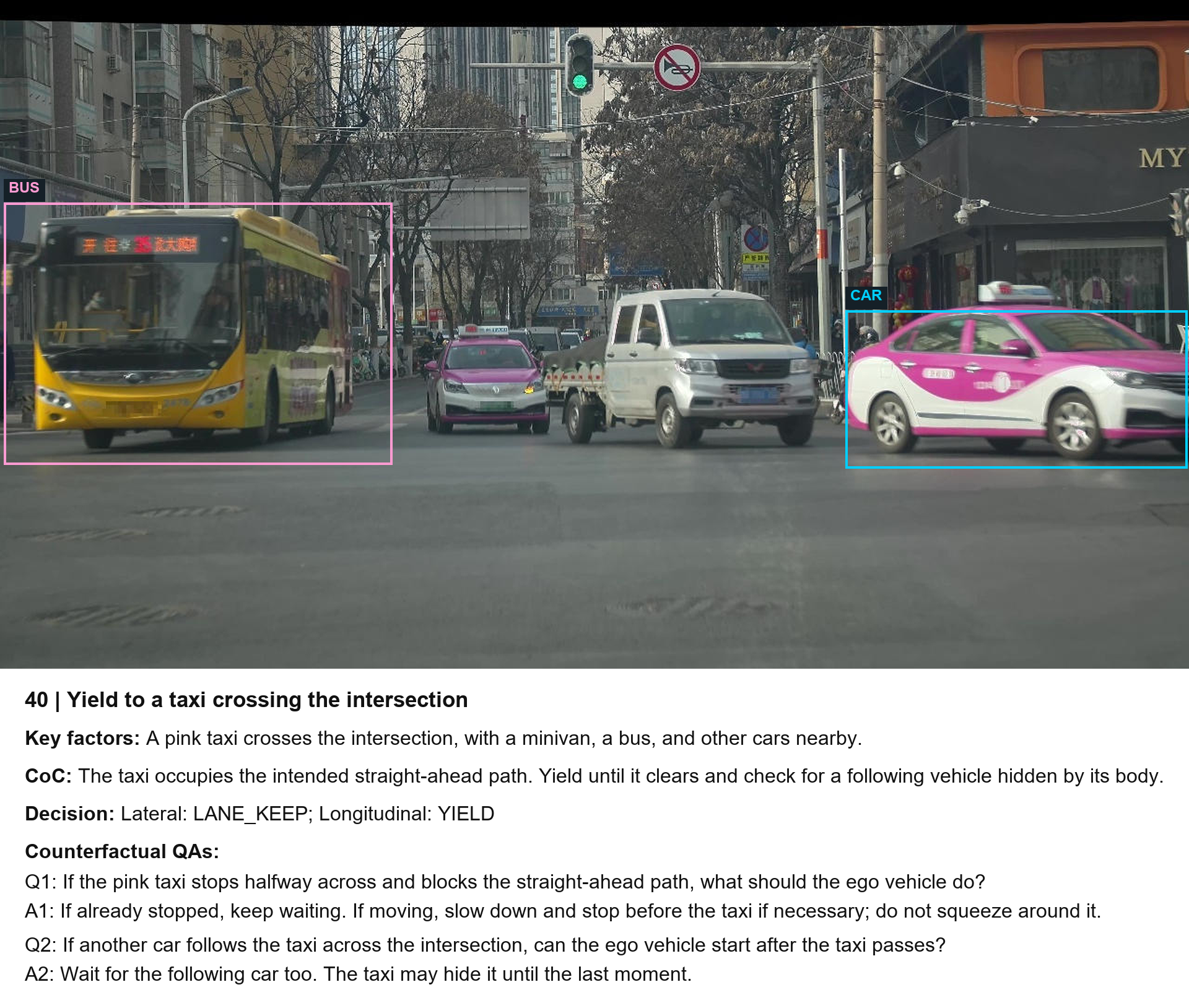}
    \end{minipage}\hfill
    \begin{minipage}{0.48\textwidth}
      \centering
      \includegraphics[width=\linewidth]{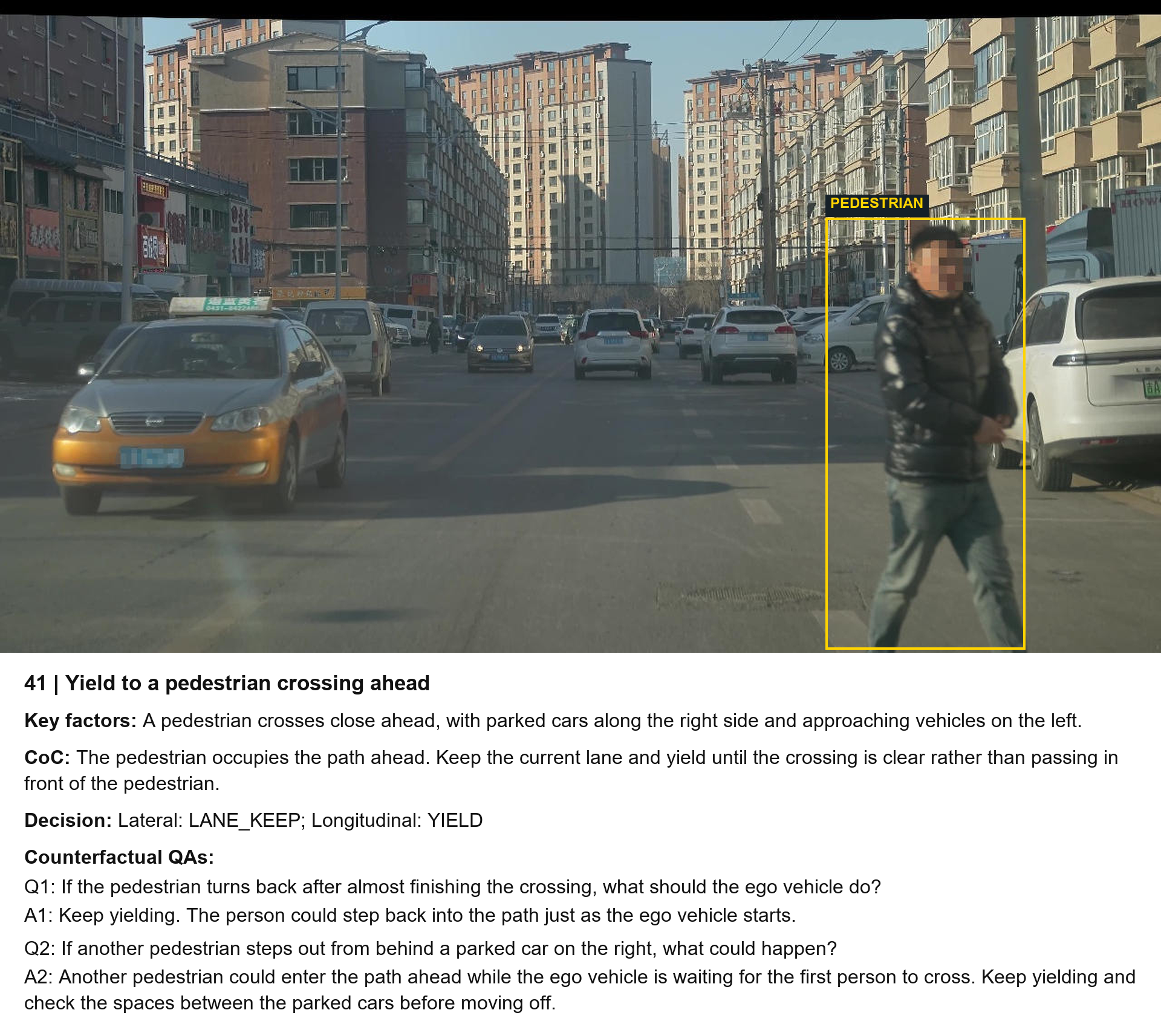}
    \end{minipage}\hfill
    \begin{minipage}{0.48\textwidth}
      \centering
      \includegraphics[width=\linewidth]{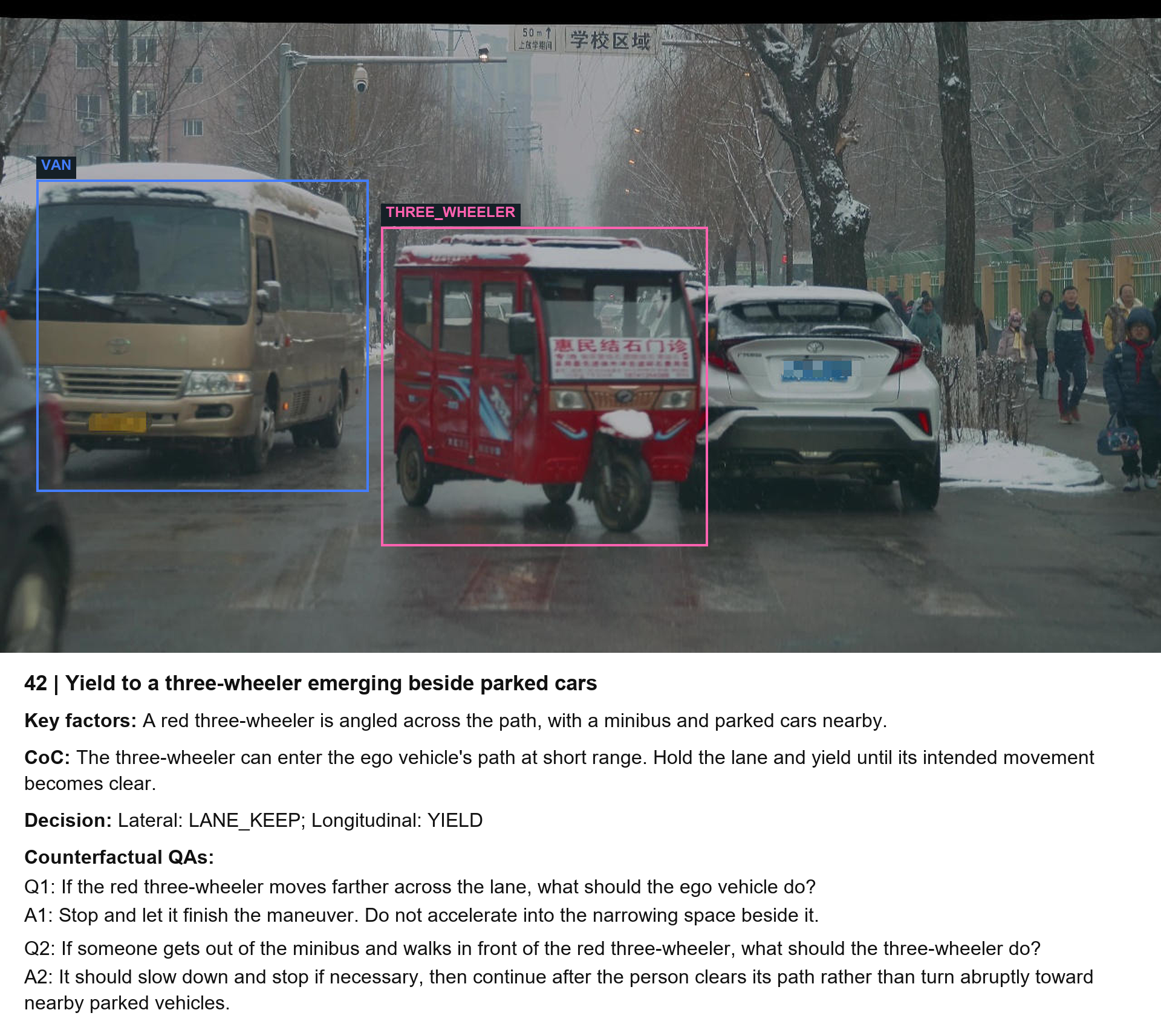} 
    \end{minipage}\hfill
    \caption{Qualitative case study on 6 sign-dense scenes (55--60). Each case presents the key factors that the ego vehicle must attend to in the current frame, the CoC, the lateral and longitudinal decisions the ego vehicle should take, and the counterfactual QA pairs.}
    \label{fig:coc_case_study_10}
  \end{figure}

To qualitatively evaluate our model's ability to understand driving scenes, we first compare the CoC outputs of our method and AR2 on six cases (Fig.~\ref{fig:coc_case_study_0}). We then present 60 representative visualized cases (Figs.~\ref{fig:coc_case_study_1}--\ref{fig:coc_case_study_10}). For each scene, we list the critical factors that the ego vehicle must attend to, the CoC, the lateral and longitudinal decisions the ego vehicle should take, and the counterfactual QA pairs.

\begin{figure}[t]
    \centering
    \includegraphics[width=\linewidth]{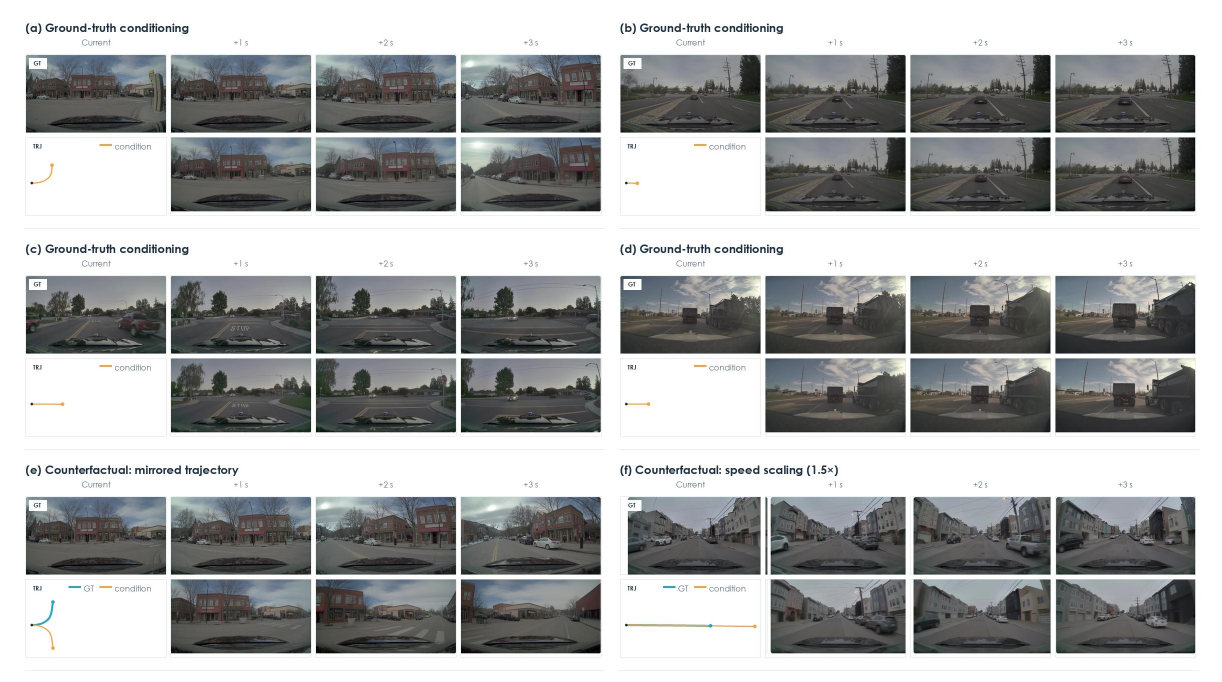}
    \caption{
        Qualitative results of trajectory-conditioned world model generation
        on the Physical AI test set.
    }
    \label{fig:physical_ai_world_model}
\end{figure}

\begin{figure}[t]
    \centering
    \includegraphics[width=\linewidth]{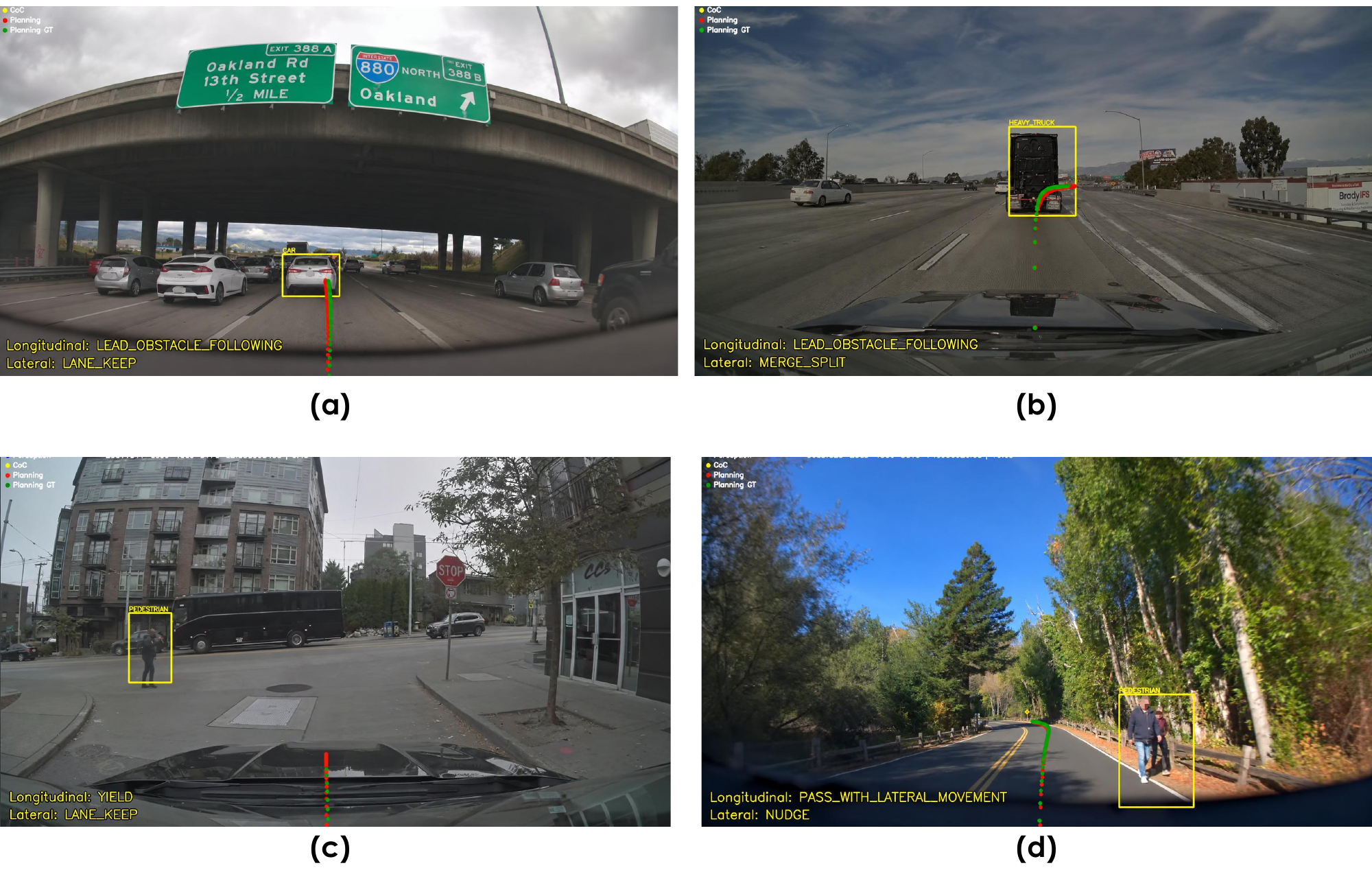}
    \caption{
         Four qualitative examples of planning results from \Model{} and the corresponding CoC boxes on the Physical AI test set.
    }
    \label{fig:coc_planning_vis}
\end{figure}

Figure~\ref{fig:physical_ai_world_model} presents qualitative results
of our world model on the Physical AI dataset.
We examine future-frame generation under both ground-truth trajectory
conditioning and counterfactual interventions.
The counterfactual examples modify the conditioning trajectory through
mirroring or speed scaling, illustrating how the generated visual
predictions respond to changes in the specified ego motion.
Figure~\ref{fig:coc_planning_vis} presents four qualitative examples of planning results from \Model{} and the corresponding CoC boxes on the PhysicalAI test set. 

%% file: sections/related_work.tex
\section{Related Work}
\label{sec:related}

\subsection{VLM and VLA for Autonomous Driving}
\label{sec:related:vlm-vla}

Vision-language models (VLMs) connect visual observations with language supervision for driving-scene understanding, building on visual instruction tuning and general-purpose multimodal language models~\cite{liu2023visual,bai2023qwenvlversatilevisionlanguagemodel,liu2024improved,tong2024cambrian}. LingoQA~\cite{marcu2024lingoqa} evaluates video-based driving question answering, while DriveLM~\cite{sima2024drivelm} organizes perception, prediction, and planning questions through graph-structured dependencies. OmniDrive~\cite{wang2024omnidrive} extends this supervision to holistic scene understanding and counterfactual reasoning. DriveGPT4~\cite{xu2024drivegpt4} further connects visual evidence and language explanations with end-to-end vehicle control. Driving-specific benchmarks~\cite{qian2024nuscenes,cao2024maplm,lu2025can,chen2025automated,hou2025drivemrp} provide complementary coverage of VQA, map understanding, and scene-level assessment. Together, these efforts establish language as an interface for expressing scene semantics and relating them to driving behavior.

Vision-language-action (VLA) models incorporate action prediction into the multimodal interface, connecting semantic reasoning with executable motion. General VLA systems~\cite{zitkovich2023rt,kim2024openvla} transfer visual-language knowledge into control and open-ended action prediction, while driving planners~\cite{hu2023planning,chitta2022transfuser,chen2024vadv2,liao2025diffusiondrive,zhang2026geoworldad,zuo2026dvgt} have progressed from sensor-fusion and vectorized planning to trajectory-generating models. Recent driving VLAs add cognitive, causal, or world-model signals: Alpamayo-R1~\cite{wang2025alpamayo} couples structured Chain-of-Causation (CoC) traces with trajectory generation, FutureSightDrive~\cite{zeng2025futuresightdrive} introduces spatio-temporal reasoning for future-aware planning, and related systems~\cite{luo2026reflect,huang2025robotron,team2026omnibrain,li2025drivevla,zhou2025autovla,lu2025uniugp,chen2026last,luo2026unleashing,he2026drivever} combine multimodal reasoning with end-to-end action learning. World-model-based driving methods~\cite{hu2023gaia,hu2024drivingworld,wang2024driving,wang2024drivedreamer} provide future-state supervision or interactive prediction. These approaches motivate treating scene understanding, reasoning, and action as interdependent capabilities. \XFM{} extends this organization to include auxiliary perception, causal decision formation, action-conditioned future prediction, and candidate evaluation within a shared latent interface. Its supervision connects discrete policy decisions to continuous trajectories and their evaluated consequences, giving these capabilities a common scene--candidate reference.

\subsection{Chain-of-Thought Reasoning: Explicit and Latent Forms}
\label{sec:related:cot}

\textbf{Explicit Chain-of-Thought}. 
Explicit Chain-of-Thought (CoT)~\cite{wei2022chain,lu2022learn,deng2024explicit,li2026think} represents intermediate reasoning as a sequence of discrete language tokens between an observation and a final prediction. Its central benefit is supervision: intermediate steps expose which evidence supports a conclusion and can improve compositional reasoning. Grounded visual explanations~\cite{kim2018textual} connect such steps to visual entities and spatial relations. For autonomous driving, an effective trace should identify relevant agents and traffic constraints, reason about their interactions and likely consequences, and use this evidence to select a maneuver. Such supervision teaches how a decision is formed, rather than only how its final trajectory is decoded.

Driving-oriented models increasingly instantiate this structure. DriveLMM-o1~\cite{ishaq2025drivelmm} learns step-by-step driving traces; FutureSightDrive~\cite{zeng2025futuresightdrive} introduces spatio-temporal reasoning to connect current observations with future outcomes; and Alpamayo-R1~\cite{wang2025alpamayo} organizes the trace as a structured Chain-of-Causation (CoC) that conditions trajectory generation. CoC gives this textual interface a driving-specific structure by linking decision-relevant evidence to the maneuver being explained or predicted. It therefore makes the evidence--decision relationship explicit while retaining the inspectable form of language supervision.

The same discrete representation creates a systems trade-off for closed-loop driving. Autoregressive decoding incurs computation for every generated token, so latency grows with trace length~\cite{adathinkdrive,lu2026xiaomi}. Language tokens describe constraints semantically but do not directly encode the continuous geometry and dynamics required for a feasible trajectory. These limitations motivate latent and implicit CoT methods, which retain intermediate computation in continuous hidden states and are discussed next.

\textbf{Latent and Implicit Chain-of-Thought}. Latent reasoning moves some or all intermediate computation from discrete text into continuous hidden states. Step-by-step internalization and continuous thought models~\cite{deng2024explicit,coconut,codi,cheng2024compressed} progressively remove or replace explicit reasoning steps during training. Other approaches~\cite{xu2025softcot,su2025token,simcot,zhu2025scaling} combine soft representations, latent--text mixtures, auxiliary supervision, or recurrent latent refinement. These methods vary in how they compress, supervise, and expose intermediate reasoning, but share the goal of reducing the cost of generating a long discrete trace.

Driving-specific methods connect latent computation to spatial, temporal, and action prediction. LaST-VLA~\cite{lastvla} performs reasoning in latent spatio-temporal space, FutureX~\cite{lin2025futurex} combines latent CoT with world modeling, and other driving models~\cite{lu2026xiaomi,ma2025dvlm,li2025discrete,ye2025dap} use compact visual or discrete action latents for controllable planning. World-model-based representation learning~\cite{hu2023gaia,hu2024drivingworld,wei2024occllama,li2025drivevla,wang2024drivedreamer} supplies complementary future-state constraints. OneVL~\cite{lu2026xiaomi} couples language reconstruction and future visual prediction to supervise compact latent representations, then uses prefill inference to avoid decoding an explicit reasoning trace. Our focus is the organization of latent computation around decision dependencies. Perception, CoC, Action, World, and Value concepts have distinct supervision targets, and block-wise autoregressive refill makes an upstream block's output available to downstream queries. This allows evidence extraction, candidate formation, consequence prediction, and evaluation to share intermediate states while retaining their different roles in the decision process.

\subsection{Adaptive Reasoning and Computation Allocation}
\label{sec:related:adaptive}

Adaptive reasoning makes inference computation depend on the scene and the expected benefit of further reasoning. AdaThinkDrive~\cite{adathinkdrive} studies how explicit CoT affects driving across different scenario complexities and learns to select between direct prediction and reasoning-conditioned prediction. AutoVLA~\cite{zhou2025autovla}, LaST-R1~\cite{chen2026last}, and cognitive or adaptive world-model planners~\cite{li2025recogdrive,wang2025adawm} extend this idea through reinforcement fine-tuning, physical latent reasoning, or adaptive future prediction. Their hybrid supervised and reinforcement-learning objectives jointly account for trajectory quality and reasoning economy. These methods establish a concrete alternative to executing the same reasoning procedure for every input: learn when intermediate reasoning improves the action enough to justify its computational cost.

Efficient VLM/VLA systems~\cite{chen2025nanovla,zhang2024minidrive,zhang2026fastddrive,lu2026xiaomi} also study token routing, block-wise execution, and reduced-cost action decoding. In parallel, closed-loop reinforcement-learning and policy-refinement methods~\cite{jiang2025irlvla,yan2026ad,liu2026r2se,song2025diver,shang2026drivedpo} optimize action quality under safety or interaction feedback. Our method connects this allocation decision to candidate verification. An initial Action--Value computation produces and evaluates a candidate.
An initial Action--Value computation generates and evaluates candidates.
The VOC head then guides the selection of task execution and
latent-token allocations for Perception, Causation, and World,
using the current inference state and remaining budget.
This adapts both task activation and the amount of latent
computation within each enabled task. The shared latent state carries information across these routes, while Value feedback links additional computation to the quality of the current candidate. Adaptive inference thus operates over the same evidence, decision, and consequence structure used by the annotation system.

%% file: sections/annotationV2.tex
\section{Unified Causal Data Annotation Loop}
\label{sec:annotation}

\subsection{Motivation, Overall Principles, and Annotation-Model Iteration}
\label{sec:annotation_motivation}

The development efficiency of a production autonomous-driving system depends not only on open-loop or closed-loop performance on fixed evaluation sets, but also on how efficiently real-vehicle risks can be converted into learnable data and then into improved driving experience. Conventional data systems often center on a single type of supervision, such as trajectories, object boxes, or lane lines. Such supervision supports perception and behavioral imitation, but it does not fully specify which scene facts affect a decision, why a candidate behavior is selected, or how the outcomes of different candidates differ.

Annotation contents differ substantially in information complexity, production cost, throughput, and degree of automation. Object geometry, road elements, and basic states can be generated at scale by rules, sensor-geometry systems, and autolabeling models. Driving relations, decision factors, causal explanations, and value judgments require more expensive model assistance or human annotation. DriveFoundry therefore uses hierarchical annotation both to organize information and to allocate data-production resources: low-cost, high-throughput supervision is used mainly for pretraining, while high-cost, high-value supervision is used for Chain-of-Causation (CoC) reasoning, planning SFT, and Value Model training.

DriveFoundry is designed to support multiple tasks and training stages. Structured annotations provide a common factual source from which Perception QA, CoC QA, decision and causal SFT data, planning trajectory data, and Value evaluation data are derived according to task requirements. World learns to predict visual observations at future horizons of 1, 2, and 3~s, conditioned on the scene history and a candidate
trajectory. Its visual targets are derived from image sequences
and linked to the corresponding scene and candidate records.

The system follows four principles. First, supervision is stratified by content and production cost. Second, automatic annotation provides scale, while human annotation supplies complex relations, causal explanations, decisions, and value judgments. Third, low-cost Scene annotation covers basic scene facts, while the evidence required by a concrete driving candidate, decision segment, or Critical Factor in Policy determines where expensive annotation and review are allocated. Fourth, human annotation, annotation models, and quality review form a loop that progressively increases annotation automation.

Annotation automation follows an iterative process. Human annotators first build high-quality seed data, which is used to train annotation models for different label types. The models then pre-label new data in batches, while humans focus on low-confidence samples, long-tail scenes, and cases in which models disagree. Reviewed results are added back to the training set and used to update the annotation models and rules. The process applies to objects, road relations, decisions, Critical Factors, and CoC. For CoC, humans initially identify Critical Factor instances, bind the associated decision labels to those factors, and annotate their attributes and roles to establish reliable supervision. As the annotation model improves, production gradually shifts toward model-based labeling, human sampling inspection, and difficult-case review.

\subsection{Unified Causation Annotation System}
\label{sec:annotation_hierarchy}

The basic data unit of DriveFoundry is a \emph{scene--candidate record}. Every record is based on recorded driving data that includes the actual ego trajectory. The scene describes environmental observations and ego state over the time interval, while the candidate in the base record denotes the executed driving behavior, including discrete decisions and a continuous trajectory. When counterfactual analysis is required, additional candidate behaviors can be introduced under the same scene and planning horizon and linked to their feasibility, outcome judgments, and value evaluations. Each record preserves label sources, confidence, and evidence references, and distinguishes observed outcomes from counterfactual outcome judgments.
The annotation system groups labels into Scene, Policy, and
Value. Scene labels describe observed entities, states, and
relations; Policy labels describe factor--decision relations
and behavior; Value labels assess individual candidates.
These annotation families supply supervision for the model's
Perception, Causation, Action, and Value tasks.
Each supervision family is stratified by content complexity, production method, and training use. Table~\ref{tab:annotation-supervision} summarizes the hierarchy and its downstream uses.

\input{tables/annotation_supervision}

Scene annotation is organized progressively from physical evidence to structured state and then to scene relations. L0 Physical / Evidence records physical evidence that is directly observed or computed. L1 Structured State organizes objects, road elements, signals, and ego state into a structured state. L2 Driving Relation describes the objective relations among these entities.

Policy annotation uses the continuous driving decision process as its organizing object and expands across physical execution, decision segments, decision instances, and counterfactual candidates. The Physical / Execution level records actual ego motion, action execution, the recorded trajectory, velocity, acceleration, and spatial relations, while referencing the corresponding Scene states. The Event level represents Decision Segments obtained by partitioning an entire clip through rule--model collaboration. Rules generate lateral and longitudinal action keyframes defined by Meta Action, while human annotators mark lateral and longitudinal semantic keyframes. Rules and large language models then jointly merge and filter these keyframes to produce event-level temporal segments, termed Decision Segments. The Decision level consists of factor-decision causal pairs within each Decision Segment. Annotation first identifies a Critical Factor instance and then binds the associated lateral or longitudinal decision label to that factor. The resulting pair constitutes a Decision Instance. A segment may contain multiple instances, which retain their own valid time ranges and are managed asynchronously across different factors and lateral or longitudinal decisions. The Counterfactual level records controlled changes to scene factors and comparisons among alternative candidate behaviors.

Annotation first extracts Critical Factor instances that affect the formation, persistence, or realization of a decision. It then annotates each factor with the associated lateral or longitudinal decision label, relation type, valid time range, source object, source state, affected aspect, and evidence source. The decision label identifies the concrete lateral or longitudinal decision that the factor affects and constitutes the decision end of the causal pair. The source object and source state identify the entity and state from which the factor arises. The evidence source references Scene labels or raw observations supporting the judgment. The relation type distinguishes \texttt{MOTIVE}, which promotes or maintains a decision, from \texttt{CONSTRAINT}, which limits its realization or feasible set. The affected aspect specifies the planning aspect influenced by the factor. The factor and its bound decision label, together with these attributes and their valid time range, are recorded as one Decision Instance. Different factors associated with the same behavior label retain separate instances; a factor associated with both lateral and longitudinal labels likewise forms separate pairs. A source object is an evidence index rather than a Decision Instance: its state and planning relevance identify the factor. The referenced Scene facts can be reused across instances.

Table~\ref{tab:appendix_decision_classification_en} summarizes
the semantic decision labels used in factor--decision pairs.
Lateral and longitudinal labels may co-occur.

\begin{table}[t]
\centering
\small
\caption{Semantic decision labels used in factor--decision annotation.}
\label{tab:appendix_decision_classification_en}
\setlength{\tabcolsep}{6pt}
\renewcommand{\arraystretch}{1.15}
\begin{tabularx}{\linewidth}{
    @{}
    >{\raggedright\arraybackslash}p{0.28\linewidth}
    >{\raggedright\arraybackslash}X
    @{}
}
\toprule
\rowcolor[HTML]{FFE0CC}
\textbf{Decision label} & \textbf{Meaning} \\
\midrule

\multicolumn{2}{@{}l}{\textbf{Lateral decisions}} \\
\addlinespace[3pt]

Lane Keep &
Remain within lane boundaries, allowing in-lane offsets
below 0.3\,m. \\

Lane Change &
Move to an adjacent lane; specify left or right. \\

Nudge &
Temporarily offset within the lane, or briefly cross a lane
line and return, to increase clearance. \\

Merge/Split &
Follow a ramp/main-road transition or a lane merge/split
at an intersection. \\

Turn &
Enter another road segment with a significant heading change;
specify left or right. \\

\midrule
\multicolumn{2}{@{}l}{\textbf{Longitudinal decisions}} \\
\addlinespace[3pt]

Gap Searching &
Adjust speed to match the target traffic flow or create
a suitable gap for a lateral maneuver. \\

Passing with Lateral Movement &
Adjust speed to pass an obstacle with a lateral maneuver,
or move ahead of an adjacent vehicle or its blind spot. \\

Overtaking &
Proceed with the ego vehicle's right of way, adjusting
speed or lateral position as needed. \\

Speed Adaptation &
Adjust speed for road geometry, topology, or visibility;
excludes blind spots caused by vehicles. \\

Yield &
Give way to a specific road user. \\

Start up &
Move from rest without interference from other road users;
excludes restarting while following another vehicle. \\

Stop for Static Constraints &
Stop and remain stopped at a required control point,
such as a stop line or traffic signal. \\

Lead Obstacle Following &
Follow the nearest same-direction road user on the ego path
at a safe time gap, including stops and restarts. \\

Set Speed Tracking &
Track the target speed when no other driving constraint applies. \\

\bottomrule
\end{tabularx}

\smallskip
\begin{minipage}{\linewidth}
\footnotesize
All decisions remain subject to safety and feasibility constraints.
\end{minipage}
\end{table}

Value annotation records binary problem indicators for safety,
comfort, efficiency, compliance, and route adherence, with
1 indicating a problem in the corresponding dimension.
Separate feasibility labels record hard-constraint violations,
while physical measures such as clearance, time margins,
smoothness, and progress support continuous quality supervision
and candidate preferences.

Labels retain their evidence, confidence, and applicability.
Uncertain or unobservable judgments are masked, and incomparable
pairs do not supply preference supervision. At inference,
hard-constraint tests are performed by the separate rule-based
Check module.
Figure~\ref{fig:annotation-overview}(a--b) summarizes the annotation structure and the shared scene--candidate record.

\par\medskip
\noindent\begin{minipage}{\linewidth}
\captionsetup{type=figure}
\centering
\includegraphics[width=\linewidth]{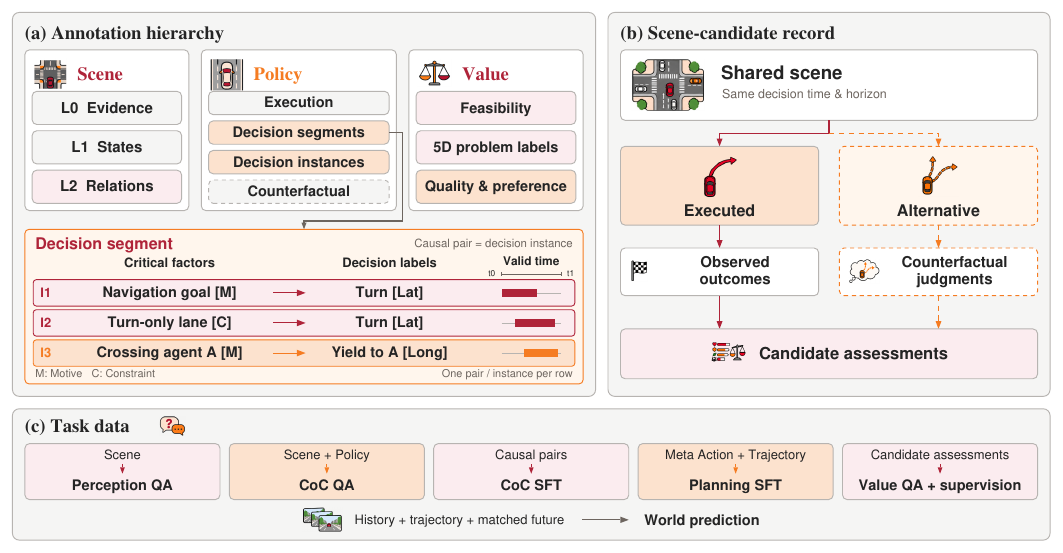}
\caption{Unified annotation hierarchy, scene--candidate records, and task-data derivation.}
\label{fig:annotation-overview}
\end{minipage}
\par\medskip

\subsection{Deriving Task Data from Structured Annotations}
\label{sec:annotation_derivation}

Structured annotations provide a common factual source for Perception, CoC, Action, and Value tasks. Perception pretraining uses hierarchical QA translated from Scene states L0--L2 labels and covers physical evidence, structured state, and objective scene relations. CoC pretraining jointly uses Scene and Policy ground truth to generate QA about scene evidence, Decision Segments, Critical Factor instance extraction, decision-label binding on each factor, and the attributes and roles of the resulting factor-decision causal pairs (Decision Instances). Because Scene and Policy are hierarchical, the derived QA retains the source level, question type, and difficulty.
Value pretraining also uses textual analyses of candidate
trajectories in the five evaluation dimensions. The continuous
Action decoder is introduced during SFT.
Figure~\ref{fig:annotation-overview}(c) maps the structured annotations to task data and separates the raw-image World-model interface.

A large language model performs linguistic translation during QA production but does not create facts. Every QA example must trace back to its source record, source fields, and annotation version. After generation, factual-consistency checks prevent the QA from introducing objects, states, relations, or future information that are absent from the structured ground truth.

During SFT, Perception learns from the corresponding scene
labels, while Causation learns factor extraction, decision-label
binding, relation roles, and grounded CoC explanations.
Action learns meta-actions and continuous trajectories.
Temporal annotations align the factors and decisions with
the corresponding trajectory segments.

World predicts visual tokens for future observations at
1, 2, and 3~s, conditioned on the scene history and a specific
candidate trajectory. For recorded behavior, the corresponding
future observations provide the supervision targets.
Alternative candidates require their own valid future targets;
the recorded future is not automatically assigned to them.
World loss is computed only on executed World paths with
valid candidate-matched targets.

Value uses the five-dimensional binary problem labels and
continuous quality supervision defined above, with invalid
or inapplicable targets masked. These signals are retained
separately during data production.

For post-training, human-refined assessments of recorded
trajectories train and calibrate the annotation model.
The calibrated annotator evaluates diverse candidates generated
by the SFT model, forming fixed scene--candidate supervision
for Value.

VOC supervision reuses these candidate assessments.
Alternative task and latent-budget expansions are executed
from the same saved SFT inference prefix, recording the current
state, proposed task and latent-budget configuration, selected
candidate, and incremental execution cost. Value learns
individual candidate quality; VOC learns the improvement in
the selected candidate relative to stopping, minus the additional
execution cost.

These rollout annotations are prepared offline before
post-training to reduce generation and annotation overhead.
Online collection is also possible. Rule rewards for Action
are computed separately during RL.

\subsection{Overall Structured-Annotation Production Pipeline}
\label{sec:annotation_pipeline}

The DriveFoundry data-production process starts from recorded driving data containing raw observations and the actual ego trajectory. After synchronization, cleaning, and segmenting, the data enter an automatic annotation line and a CoC annotation line. The two lines share raw observations, ego trajectories, and basic labels. The automatic line provides reusable Scene facts, while the CoC line organizes Policy annotations and feeds evidence gaps discovered during analysis back to the relevant annotation modules. Human annotation and evaluation run through both lines, and reviewed outputs are assembled into unified Labels and \emph{scene--candidate records}. Table~\ref{tab:annotation-production} lists the production modules and quality gates. Figure~\ref{fig:annotation-pipeline} shows their organization into parallel production lines and the exchange of Scene evidence.

\input{tables/annotation_production}

The automatic annotation line generates Scene information at scale. Autolabeling models use available sensor data and geometric auxiliary information to generate objects, lanes, stop lines, drivable areas, attributes, occlusion, signal states, and cross-frame tracks. Rule modules compute kinematic quantities, temporal boundaries, collision relations, and explicit safety constraints from the available geometry, states, and trajectories. Rule--model fusion establishes associations between objects and road elements, road topology, and temporal relations, and checks physical boundaries and cross-frame consistency. The line covers L0, L1, and L2 relations that can be generated reliably by rules and models; complex or ambiguous relations are supplemented by VLM pre-labeling and human review.

The CoC annotation line combines recorded clips with Scene
annotations to produce temporal factor--decision pairs and
grounded explanations, as detailed in
Section~\ref{sec:decision_centric_coc}. Evidence gaps trigger
supplementation and review in the automatic annotation line.
The resulting records link decisions to trajectories, observed
outcomes, and candidate assessments.

\par\medskip
\noindent\begin{minipage}{\linewidth}
\captionsetup{type=figure}
\centering
\includegraphics[width=\linewidth]{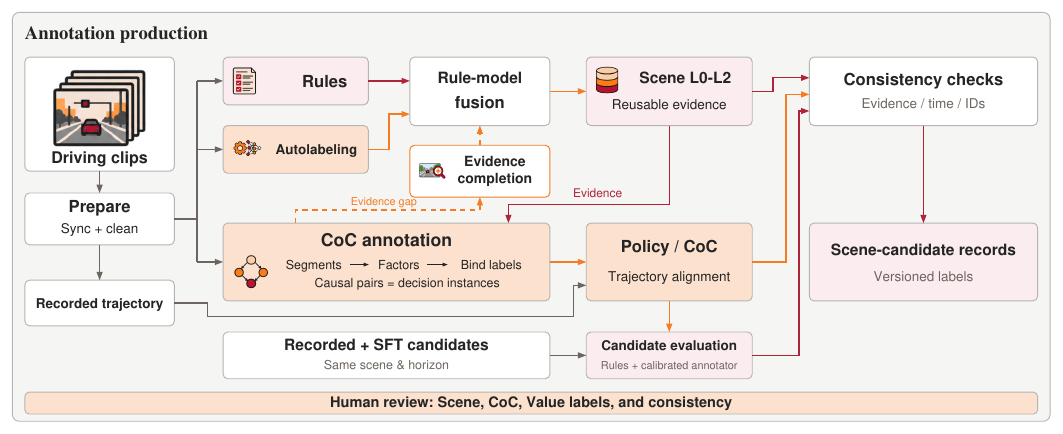}
\caption{Overall structured-annotation production pipeline.}
\label{fig:annotation-pipeline}
\end{minipage}
\par\medskip

\subsection{Decision-Centric CoC Annotation Process}
\label{sec:decision_centric_coc}

A frame-level CoC record can pair a Meta Action keyframe with
one factor and one decision label to explain the recorded
trajectory. Restricting annotation to this form leaves six
aspects of continuous decision-making under-specified:
\begin{enumerate}
\item \textbf{No temporal representation of the decision process.} A frame-level Decision Label and Critical Factor cannot describe the process from intent formation through execution, completion, or interruption, nor can they represent changes in the entry, exit, and influence of associated factors.
\item \textbf{No asynchronous lateral and longitudinal decisions.} Lateral and longitudinal decisions have different formation, persistence, and switching rhythms, which are difficult to express with a frame-level label or a joint lateral--longitudinal category.
\item \textbf{No representation of multiple objects and simultaneous decisions.} A vehicle may turn, yield to one object, and pass before another object at the same time. A single decision label loses the factors and composition of these local decisions.
\item \textbf{No planning-role semantics for Critical Factors.} The same scene factor may motivate or maintain a decision, or constrain the feasible set for realizing it. Merely recording a Critical Factor does not distinguish these roles.
\item \textbf{No counterfactual candidate analysis.} Recording only the executed trajectory prevents comparison with alternative trajectories in safety, efficiency, comfort, and rule compliance, and cannot distinguish an infeasible candidate from a feasible but inferior one.
\item \textbf{Unstable machine semantics in free-text CoC.} Free text can introduce hallucination and post-hoc rationalization and is difficult to validate as a closed set, analyze statistically, or use as stable ground truth.
\end{enumerate}

DriveFoundry represents CoC as a temporally structured set of factor-decision causal pairs grounded in Scene evidence. This representation supports concurrent lateral and longitudinal decisions and traceable causal explanations across a driving clip.

Meta Action defines the lateral and longitudinal actions used by rules to generate action keyframes over the clip. Human annotators separately mark lateral and longitudinal semantic keyframes reflecting meaningful decision boundaries. Rules and large language models jointly merge and filter the two keyframe sources, retaining the temporal anchors used to partition the clip into Decision Segments with explicit start and end times. These segments constitute the Event level, as shown in Figure~\ref{fig:coc-production}(a).

Within each Decision Segment, annotation first extracts Critical Factor instances and forms a Critical Factor List. It then binds the associated lateral or longitudinal decision label to each factor and records the applicable role or roles
(\texttt{MOTIVE}, \texttt{CONSTRAINT}, or both), valid time range, source object, source state, affected aspect, and evidence source. Each resulting factor-decision causal pair constitutes a Decision Instance at the Decision level; the decision label is the decision end of the pair rather than a separate instance. These instances are managed asynchronously according to their factors, decision dimensions, and valid time ranges, allowing multiple local decisions to coexist. Figure~\ref{fig:coc-production}(b) summarizes this factor extraction and decision-label binding. Changes in factor state, role, or bound label are recorded on the corresponding pair timeline, with object identity, state, and relation references kept consistent with Scene evidence.

Structured CoC organizes the Decision Instances within each segment, together with their temporal relations and outcomes. These explicitly typed relations can be checked for consistency. Natural-language reasoning derives a concise causal explanation from these relations for CoC SFT, analysis, and debugging. Each explanation states the relevant factors before the decisions they support, using the explicit form ``Because [factors], the ego vehicle [decisions].'' The clauses trace back to the same structured pairs and evidence available at the decision time. Trajectory alignment uses the actual ego trajectory already recorded in the driving data and associates it with the valid time ranges of the Decision Segment, Decision Instance, and Critical Factors. It describes how a decision is realized as action execution and whether the action is completed, interrupted, or switched.

Counterfactual annotation covers both changes to scene factors
and comparisons among alternative behaviors. Factor interventions
specify a physically consistent change and assess its effect on
decision feasibility or candidate preference.

For behavior comparisons, at least two candidates with alternative
semantic decisions or trajectories are evaluated under the same
scene history and planning horizon. Feasibility and mandatory
constraints are checked first. Feasible candidates are then
compared using the five-dimensional assessments and continuous
quality measures. Each judgment retains its evidence source and
confidence, including those for unexecuted candidates.

\par\medskip
\noindent\begin{minipage}{\linewidth}
\captionsetup{type=figure}
\centering
\includegraphics[width=\linewidth]{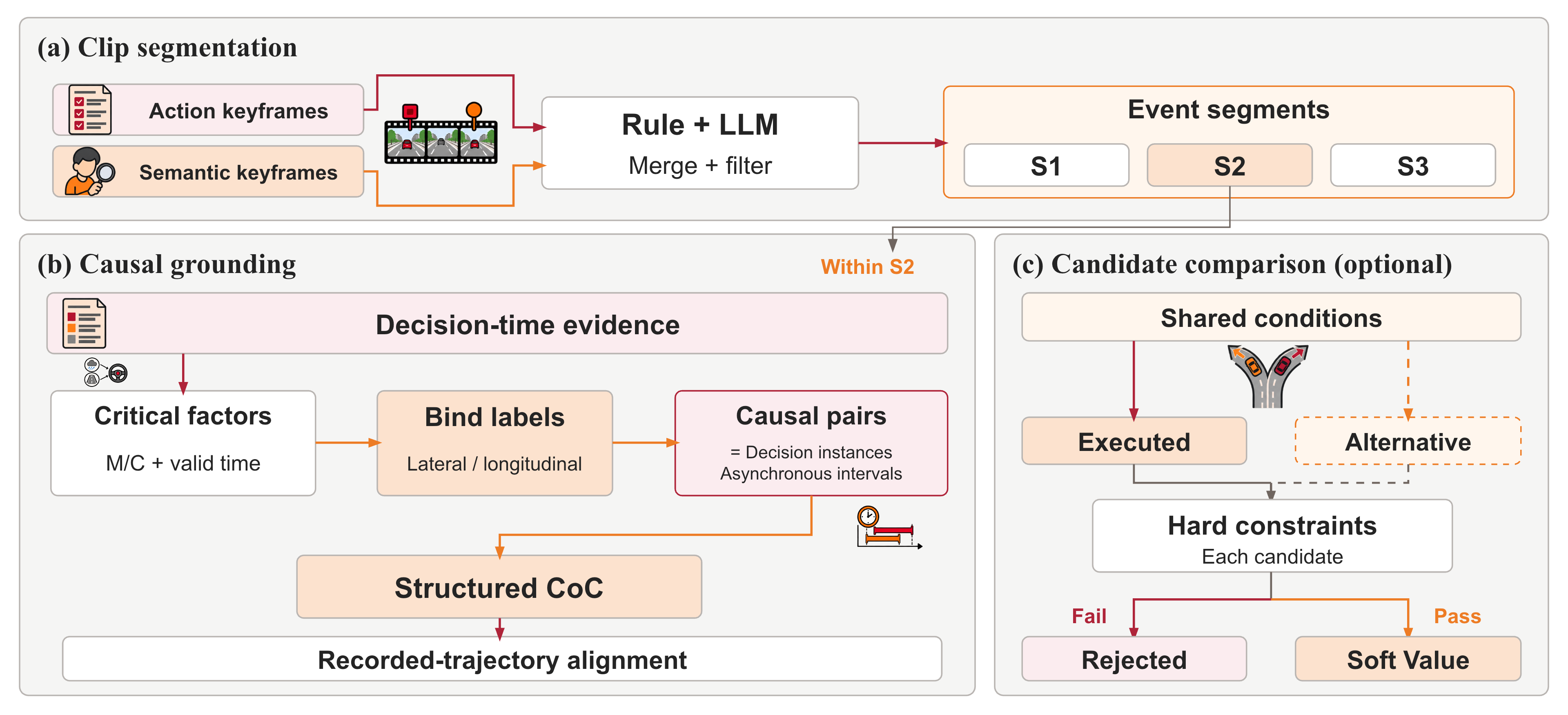}
\caption{Decision-Centric CoC: clip segmentation, factor-decision grounding, and optional candidate comparison.}
\label{fig:coc-production}
\end{minipage}
\par\medskip

Figure~\ref{fig:overview_coc_autolabel_pipeline} illustrates the temporal alignment of visual evidence,
lateral and longitudinal decisions, and CoC explanations in an
intersection scenario. The ego vehicle changes lanes to prepare
for a left turn, yields to conflicting traffic, and follows the
vehicle ahead after turning.


\subsection{Quality Control and Annotation-Model Iteration Loop}
\label{sec:annotation_quality}

Before release, the data undergo stratified quality checks. Rule labels are tested with unit tests and boundary-case regression. Autolabeling results are sampled by category and scenario and checked for cross-frame consistency. The fusion module is checked for model--rule conflict rate, topological validity, and temporal consistency. CoC annotation is checked for action and semantic keyframe coverage, rule--LLM merging and filtering, Event-level decision-segment boundaries, membership of each factor-decision causal pair as one Decision Instance within its segment, and asynchronous intervals across factors and lateral or longitudinal labels. Critical Factor instance extraction is checked for omission, wrong selection, duplicates, and object-identity consistency; attribute and role annotation is checked for accurate binding of each lateral or longitudinal decision label to its factor, agreement between the causal-pair record and Decision Instance identity and valid time range, state accuracy, MOTIVE/CONSTRAINT and affected-aspect consistency, valid time ranges, and Scene evidence references. Decision causes must be supported by evidence available at the decision time, while observed outcomes and decision evidence are recorded separately. Value annotation is checked for consistency in candidate comparison, the rate of incomparable cases, and conflicts among dimensions. QA data are checked for source-field traceability and factual correctness. Figure~\ref{fig:annotation-feedback} connects these checks to accepted-data release, issue-specific corrections, and annotation-model iteration. Its quality gate also checks the temporal CoC example in Figure~\ref{fig:overview_coc_autolabel_pipeline}: each explanation must include all factors valid in its interval, preserve the associated decisions, and place the factor clause before the decision clause.

Quality control is also the entry point for annotation-model iteration. Human-confirmed high-quality labels are added to the corresponding annotation-model training set. The model pre-labels new data in batches, and confidence, scene difficulty, and model disagreement determine which samples require human review. Low-quality rules trigger rule revision; weak model performance on particular scenes triggers difficult-sample mining and retraining; disagreement among human annotators triggers specification updates and expert adjudication; and factual errors in QA trigger updates to generation templates or translation models.

For CoC, human-confirmed Critical Factor instances, their bound decision labels, factor attributes and roles, and Structured CoC continuously train and update the annotation model. Evaluation covers factor-instance extraction accuracy and completeness, factor--label binding accuracy, consistency of pair and instance identity and timing, and the consistency of role semantics and evidence references. Weak scenes are fed into the next round of sampling and human review. As performance improves, human work shifts toward high-value samples, low-confidence review, and long-tail evaluation, forming a loop of seed annotation, batch model pre-labeling, difficult-case review, and model retraining.

Verified failures from development evaluation and fleet operation
also guide scene selection, candidate generation, and targeted
label correction. Reviewed records feed subsequent driving-model
training as well as annotation-model updates.

Derived QA, generated candidates, and scheduling records inherit
the split assignment of their source clips. Held-out test records
are excluded from driving-model and annotation-model training
and from calibration.

\par\medskip
\noindent\begin{minipage}{\linewidth}
\captionsetup{type=figure}
\centering
\includegraphics[width=\linewidth]{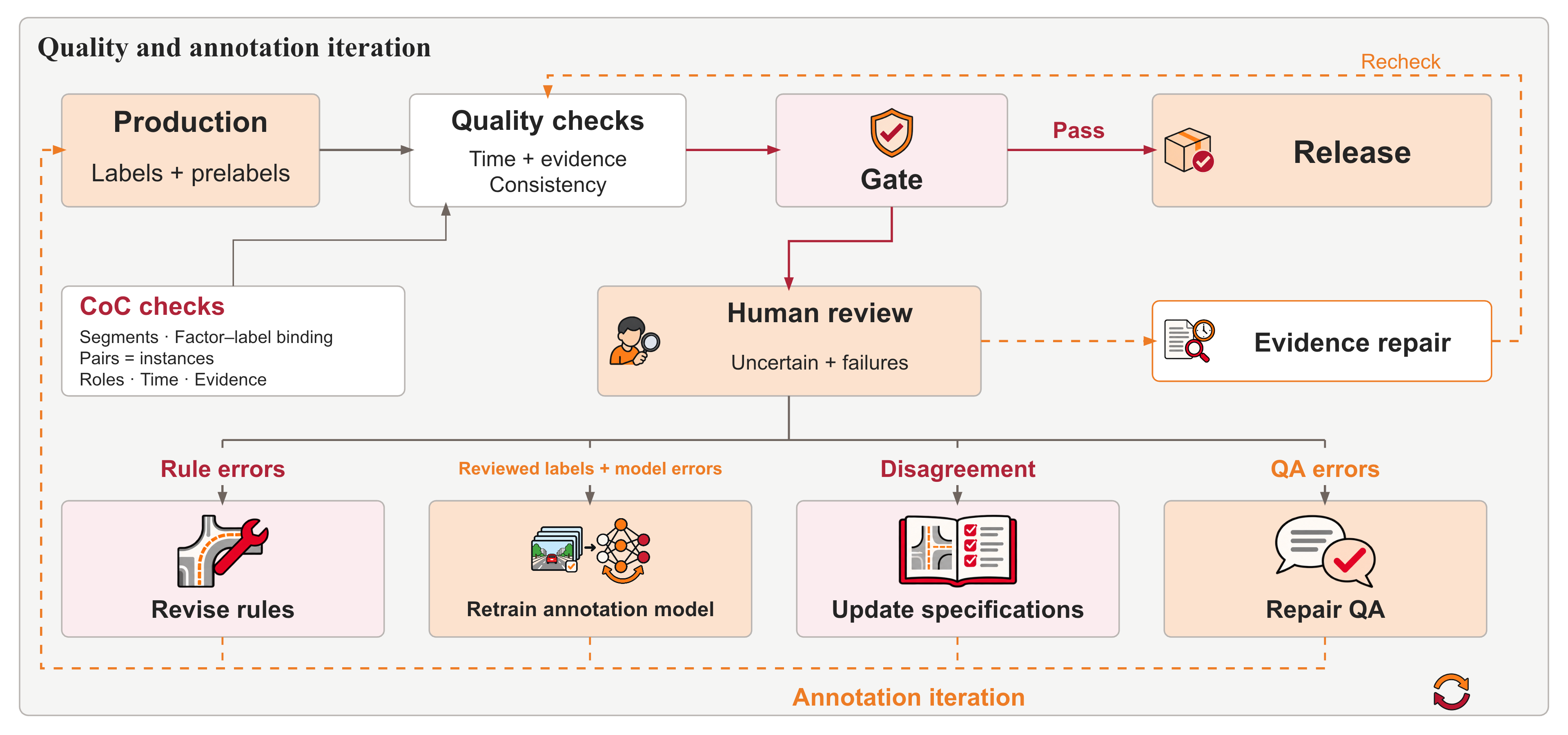}
\caption{Quality-gated data release and annotation-model iteration loop.}
\label{fig:annotation-feedback}
\end{minipage}
\par\medskip

%% file: tables/annotation_supervision.tex
\begin{table}[t]
\centering
\small
\caption{Annotation hierarchy, production methods, and training
uses in \XFM{}.}
\label{tab:annotation-supervision}
\setlength{\tabcolsep}{4pt}
\renewcommand{\arraystretch}{1.16}

\begin{tabularx}{\linewidth}{@{}
    >{\raggedright\arraybackslash}p{0.15\linewidth}
    >{\raggedright\arraybackslash}p{0.34\linewidth}
    >{\raggedright\arraybackslash}p{0.24\linewidth}
    >{\raggedright\arraybackslash}X
@{}}
\toprule
\rowcolor[HTML]{FFE0CC}
\textbf{Annotation level} &
\textbf{Content and targets} &
\textbf{Production} &
\textbf{Training use} \\
\midrule

\textbf{Scene}\newline
L0--L1 &
Objects, road geometry, signals, occupancy, and motion states;
structured labels and QA. &
Rules, sensor geometry, and autolabeling with sampling inspection. &
Perception pretraining and SFT. \\

\textbf{Scene}\newline
L2 &
Relations among objects, roads, and signals;
relation labels and grounded QA. &
Rule--model fusion and VLM pre-labeling with human verification. &
Perception and Causation training. \\

\midrule

\textbf{Policy}\newline
Execution &
Recorded trajectories, kinematics, and meta-action labels. &
Trajectory processing and rule-based labeling. &
Action SFT; temporal grounding for CoC. \\

\textbf{Policy}\newline
Event &
Decision Segments and their temporal boundaries. &
Rule--LLM merging of action keyframes and human-marked
semantic keyframes. &
Temporal alignment for Causation and Action. \\

\textbf{Policy}\newline
Decision &
Factor--decision pairs (Decision Instances), roles,
valid time ranges, and grounded CoC explanations. &
Factor extraction, decision-label binding, and human review. &
Causation pretraining and SFT. \\

\textbf{Policy}\newline
Counterfactual &
Controlled changes to scene factors and alternative candidate
behaviors; feasibility and preference comparisons. &
Controlled comparisons, candidate generation,
rule checks, and expert review. &
Causation and Value supervision. \\

\midrule

\textbf{Value} &
Five-dimensional binary problem labels and textual assessments;
continuous quality scores, feasibility labels, and preferences. &
Rules and calibrated annotators with human review. &
Value training; shared candidate assessments for VOC supervision. \\

\bottomrule
\end{tabularx}

\par\smallskip
\begin{minipage}{\linewidth}
\footnotesize
\raggedright
Feasibility labels support rule-based checks.
World supervision comes from candidate-matched future visual targets.
\end{minipage}
\end{table}

%% file: tables/annotation_production.tex
\begin{table}[t]
\centering
\small
\setlength{\tabcolsep}{4pt}
\renewcommand{\arraystretch}{1.15}
\caption{Production modules, quality checks, and feedback targets
in the \XFM{} annotation pipeline.}
\label{tab:annotation-production}

\begin{tabularx}{\linewidth}{
    @{}
    >{\raggedright\arraybackslash}p{0.17\linewidth}
    >{\raggedright\arraybackslash}X
    >{\raggedright\arraybackslash}X
    >{\raggedright\arraybackslash}p{0.18\linewidth}
    @{}
}
\toprule
\rowcolor[HTML]{FFE0CC}
\textbf{Module} &
\textbf{Main outputs} &
\textbf{Quality checks} &
\textbf{Feedback target} \\
\midrule

Rule-based &
Kinematic quantities and feasibility labels. &
Physical validity and boundary-case tests. &
Rules and thresholds. \\

\addlinespace[4pt]
Autolabeling &
Objects, attributes, and tracks (L0--L1). &
Stratified inspection, confidence, and temporal consistency. &
Labeling model and sampling policy. \\

\addlinespace[4pt]
Rule--model fusion &
Scene relations, road topology, and associations (L2). &
Rule--model conflicts, topology, and temporal consistency. &
Fusion rules and model. \\

\addlinespace[4pt]
CoC annotation &
Decision segments, factor--decision pairs, and CoC. &
Factor accuracy and coverage; label/role binding;
timing and evidence. &
CoC model and annotation guidelines. \\

\addlinespace[4pt]
Human / expert &
Semantic keyframes, reviewed relations, and candidate assessments. &
Annotator agreement, evidence support, and observability. &
Guidelines and hard cases. \\

\addlinespace[4pt]
Record release &
Versioned scene--candidate records and task datasets. &
Field completeness, provenance, and cross-module consistency. &
Data version and training configuration. \\

\bottomrule
\end{tabularx}
\end{table}